\documentclass[draftclsnofoot, compsoc, peerreview, onecolumn, 11pt]{IEEEtran}
\ifCLASSOPTIONcompsoc
  \usepackage[nocompress]{cite}
\else
  \usepackage{cite}
\fi
\ifCLASSINFOpdf
  \usepackage[pdftex]{graphicx}
\else
  \usepackage[dvips]{graphicx}
\fi
\usepackage[utf8]{inputenc} % allow utf-8 input
\usepackage[T1]{fontenc}    % use 8-bit T1 fonts
\usepackage{hyperref}       % hyperlinks
\usepackage{url}            % simple URL typesetting
\usepackage{booktabs}       % professional-quality tables
\usepackage{amsfonts}       % blackboard math symbols
\usepackage{nicefrac}       % compact symbols for 1/2, etc.
\usepackage{microtype}      % microtypography
\usepackage{amsmath}
\newtheorem{theorem}{Theorem}
\newtheorem{lemma}{Lemma}
\newtheorem{prop}{Proposition}
\newtheorem{defn}{Definition}

\newtheorem{assumption}{Assumption}
\usepackage{mathrsfs}
\usepackage{enumitem}
\usepackage{amssymb}
\usepackage{multirow}
\usepackage{graphics}
\usepackage{subfigure} 
\usepackage{wrapfig}
\usepackage{array}
\usepackage{xcolor}
\usepackage{bm}
\usepackage{float}
\newcommand{\nj}{n_{j}}

\begin{document}
\title{%Hybrid ISTA Based on Free-Form Deep Neural Networks With Convergence Guarantees
\LARGE Hybrid ISTA: Unfolding ISTA With Convergence Guarantees Using Free-Form Deep Neural Networks}

\author{
\small Ziyang Zheng, 
Wenrui Dai,~\IEEEmembership{Member,~IEEE,}
Duoduo Xue,
Chenglin Li,~\IEEEmembership{Member,~IEEE,} 
Junni Zou,~\IEEEmembership{Member,~IEEE,}
and Hongkai Xiong,~\IEEEmembership{Senior~Member,~IEEE}
\IEEEcompsocitemizethanks{
\IEEEcompsocthanksitem Z. Zheng, D. Xue, C. Li, and H. Xiong are with the Department of Electronic Engineering, Shanghai Jiao Tong University, Shanghai 200240, China. E-mail: doll001@sjtu.edu.cn, xueduoduo@sjtu.edu.cn, lcl1985@sjtu.edu.cn, xionghongkai@sjtu.edu.cn.
\IEEEcompsocthanksitem W. Dai and Z. Jou are with the Department of Computer Science and Engineering, Shanghai Jiao Tong University, Shanghai 200240, China. E-mail:daiwenrui@sjtu.edu.cn, zoujunni@sjtu.edu.cn.
\IEEEcompsocthanksitem This is a draft and the final version has been accepted by TPAMI (\textbf{DOI: 10.1109/TPAMI.2022.3172214}).}
}

\markboth{}%
{Zheng \MakeLowercase{\textit{et al.}}: %Hybrid ISTA Based on Free-Form Deep Neural Networks with Convergence Guarantees
Hybrid ISTA: Unfolding ISTA With Convergence Guarantees Using Free-Form Deep Neural Networks}


\IEEEtitleabstractindextext{%

\begin{abstract}
It is promising to solve linear inverse problems by unfolding iterative algorithms (\emph{e.g.}, iterative shrinkage thresholding algorithm (ISTA)) as deep neural networks (DNNs) with learnable parameters. However, existing ISTA-based unfolded algorithms restrict the network architectures for iterative updates with the partial weight coupling structure to guarantee convergence. In this paper, we propose hybrid ISTA to unfold ISTA with both pre-computed and learned parameters by incorporating free-form DNNs (\emph{i.e.}, DNNs with arbitrary feasible and reasonable network architectures), while ensuring theoretical convergence. We first develop HCISTA to improve the efficiency and flexibility of classical ISTA (with pre-computed parameters) without compromising the convergence rate in theory. Furthermore, the DNN-based hybrid algorithm is generalized to popular variants of learned ISTA, dubbed HLISTA, to enable a free architecture of learned parameters with a guarantee of linear convergence. To our best knowledge, this paper is the first to provide a convergence-provable framework that enables free-form DNNs in ISTA-based unfolded algorithms. This framework is general to endow arbitrary DNNs for solving linear inverse problems with convergence guarantees. Extensive experiments demonstrate that hybrid ISTA can reduce the reconstruction error with an improved convergence rate in the tasks of sparse recovery and compressive sensing.
\end{abstract}

}

\maketitle

\IEEEdisplaynontitleabstractindextext

% \fi
% \IEEEpeerreviewmaketitle
\newpage
{\tableofcontents}
\newpage

\section{Introduction}\label{sec:introduction}

\IEEEPARstart{I}{n} many practical problems, the features of most interest cannot be observed directly, but have to be inferred from samples. The simplest approximation that considers the linear relationship between the features and samples works surprisingly well in a wide range of cases. Therefore, a series of linear inverse problems have arisen to address the linear relationship in a variety of applications such as signal and image processing~\cite{mairal2010online, candes2008an}, statistical inference~\cite{tarantola2005inverse}, and optics~\cite{katz2009compressive}.  
Amongst these inverse problems, special attention has been devoted to the Lasso problem \cite{Tibshirani1996Regression} that is popular to model sparse coding \cite{mairal2010online} and compressive sensing (CS)~\cite{candes2008an}. In general, the Lasso problem is defined as:
\begin{equation}\label{lasso}
\min_{\mathbf{x}} \left\{F(\mathbf{x})\equiv f(\mathbf{x})+g(\mathbf{x})=\frac{1}{2}\|\mathbf{Ax}-\mathbf{b}\|_{2}^{2}+\lambda\|\mathbf{x}\|_{1}\right\},
\end{equation}
where $\mathbf{A}\in\mathbb{R}^{M\times N}$ is an over-complete basis matrix, $\mathbf{x}\in\mathbb{R}^{N}$ is an unknown sparse signal constrained by the $\ell_1$ norm, $\mathbf{b}\in\mathbb{R}^{M}$ is the vector of linear measurements, and $\lambda>0$ is the regularization parameter. 
%Suppose that Eq.~\eqref{lasso} is solvable, i.e., $X_{*}:=\arg\min F \neq \emptyset$, and for $x^{*}\in X_{*}$ we set $F_{*}:=F(x^{*}).$
The Lasso problem aims at recovering the $N$-dimensional signal $\mathbf{x}^{*}$ from the $M$ observed measurements $\mathbf{b}$ and the basis matrix $\mathbf{A}$. Since $M\ll N$, the recovery is ill-posed and under-determined without the $\ell_{1}$ sparsity regularization. However, if $\mathbf{x}^{*}$ is sparse enough, it can be exactly recovered with an overwhelming probability~\cite{candes2005decoding}. 

Various traditional algorithms have been developed to solve Eq.~\eqref{lasso}, such as proximal coordinate descent~\cite{tseng2001convergence, friedman2007pathwise}, least angle regression~\cite{efron2004least}, and iterative shrinkage-thresholding algorithm (ISTA)~\cite{blumensath2008iterative}. In this paper, we focus on ISTA, a proximal gradient method widely adopted to solve the Lasso problem. At the $n$th step of ISTA, $\mathbf{x}^n$ is updated to approximate the optimum $\mathbf{x}^\ast$.
%Basically, the general step of ISTA is the form of:
\begin{align}\label{e1}
\mathbf{x}^{n+1}
&=\mathop{\arg\min}_{\mathbf{x}}\left\{\lambda\|\mathbf{x}\|_{1}+\frac{1}{2t}\|\mathbf{x}-[\mathbf{x}^{n}-t\mathbf{A}^{T}(\mathbf{Ax}^{n}-\mathbf{b})]\|_{2}^{2}\right\} \nonumber\\
&=\mathcal{S}_{\lambda t}\left(\mathbf{x}^{n}-t\mathbf{A}^{T}(\mathbf{Ax}^{n}-\mathbf{b})\right)\nonumber\\
&=\mathcal{S}_{\lambda t}\left((\mathbf{I}-t\mathbf{A}^{T}\mathbf{A})\mathbf{x}^{n}+t\mathbf{A}^{T}\mathbf{b}\right)
\end{align}
Here, $t$ is the step size and $\mathcal{S}_{\lambda t}(\cdot)$ denotes the soft-thresholding operator that is defined in a component-wise way as $\mathcal{S}_{\lambda t}(x)={\rm sgn}(x)\max(0, |x|-\lambda t)$, where ${\rm sgn}(x)=|x|/x$ for non-zero $x$ and $0$ when $x=0$.
%At each update step of ISTA, $x^n$ is improved to approximate the optimal value.
Despite the concise steps, ISTA suffers from the sublinear convergence rate~\cite{doi:10.1137/080716542}. Promising alternatives to improve the convergence rate include modifying the update steps~\cite{doi:10.1137/080716542}, introducing relaxed conditions~\cite{bredies2008linear, zhang2017new} and identifying the support of $\mathbf{x}^\ast$~\cite{ndiaye2017gap, pmlr-v80-massias18a}. A large step size $t$ can be taken in Eq.~\eqref{e1} by identifying the support to rapidly approach the optimum~\cite{poon2018local}.
%One feasible solution is to improve the convergence rate by modifying the update steps~\cite{doi:10.1137/080716542}, or analyze new results under different conditions~\cite{bredies2008linear, zhang2017new}. Another popular line is to try to identify the support of $x^*$, in order to diminish the size of optimization problem~\cite{ndiaye2017gap, pmlr-v80-massias18a}. Once the support is identified, a large step size $t$ can be taken in Equation~\eqref{e1} to rapidly approach the optimum~\cite{poon2018local}.

%Recently, deep neural network (DNN) based unfolded iterative algorithms have been empirically successful in solving optimization problems. Iterative algorithms, like alternating direction of method of multipliers (ADMM)~\cite{admm2010}, approximate message passing (AMP)~\cite{amp2009} and ISTA~\cite{blumensath2008iterative} have been unfolded into DNNs~\cite{NIPS2016_6406,8550778,xie2019differentiable,7934066,metzler2017learned,10.5555/3104322.3104374}, to improve the solutions to linear inverse problems.
Recently, iterative algorithms, \emph{e.g.}, alternating direction of method of multipliers (ADMM)~\cite{admm2010}, approximate message passing (AMP)~\cite{amp2009}, and ISTA~\cite{blumensath2008iterative}, have been unfolded into deep neural networks (DNNs)~\cite{NIPS2016_6406,8550778,xie2019differentiable,7934066,metzler2017learned,10.5555/3104322.3104374} to improve the solutions to linear inverse problems. These methods attempt to construct interpretable DNNs by incorporating the framework of traditional iterative algorithms and have been empirically successful in solving optimization problems. 

Revisiting the realm of sparse coding, the pioneering work of LISTA \cite{10.5555/3104322.3104374}, a trained version of ISTA, unfolds $K$ iterations of Eq.~\eqref{e1} and substitutes $t\mathbf{A}^{T}$ and $(\mathbf{I}-t\mathbf{A}^{T}\mathbf{A})$ with the learnable parameters $\mathbf{W}_{1}^{n}$ and $\mathbf{W}_{2}^{n}$ for $n=0,1,\cdots,K$.
\begin{equation}\label{eq:lista}
\mathbf{x}^{n+1}=\mathcal{S}_{\bm\theta^{n}}\left(\mathbf{W}_{1}^{n}\mathbf{b}+\mathbf{W}_{2}^{n}\mathbf{x}^{n}\right)%,\quad n=0, 1, 2,\cdots,K
\end{equation} 
Empirical evaluations show that LISTA and its variants can converge in 10-20 iterations and substantially speed up the convergence of ISTA \cite{wang_article,wang_paper}. 
%However, LISTA is not flexible in architecture, as the operation for employing the learned matrices $\mathbf{W}_{1}^{n}$ and $\mathbf{W}_{2}^{n}$ is limited to matrix multiplication and the sizes of $\mathbf{W}_{1}^{n}$ and $\mathbf{W}_{2}^{n}$ are fixed by the dimensions of $\mathbf{b}$ and $\mathbf{x}$  (\emph{i.e.}, $M$ and $N$), respectively. 
However, LISTA is not flexible in architecture, as the sizes of $\mathbf{W}_{1}^{n}$ and $\mathbf{W}_{2}^{n}$ are fixed by the dimensions of $\mathbf{b}$ and $\mathbf{x}^n$ (\emph{i.e.}, $M$ and $N$) to allow matrix multiplication in Eq.~\eqref{eq:lista}. By contrast, ISTA-Net~\cite{zhang2018ista} introduces DNNs into the framework of ISTA to realize the learned transform that induces the sparsity of signals. Different from LISTA, ISTA-Net does not constrain the architectures of DNNs\footnote{In ISTA-Net, there are 6-7 convolutional layers of kernel size $3\times 3$ in one iterative block. %However, these DNNs can be replaced by any other DNNs with feasible and reasonable architecture while not destroying the interpretation of ISTA-Net.
However, this architecture can be replaced with arbitrary feasible and reasonable architecture without obscuring the interpretation of ISTA-Net.}, \emph{i.e.}, incorporates \emph{free-form DNNs}. The \emph{free-form DNNs} can contain arbitrary feasible and reasonable deep architectures, e.g., convolutional layers, rectified linear unit (ReLU), batch normalization~\cite{ioffe2015batch}, and residual connectivity~\cite{he2016deep}.
%These \emph{free-form DNNs}\footnote{The term \emph{free-form DNNs} means that DNNs with arbitrary feasible and reasonable architecture.} can contain heuristic deep architectures, including convolutional layers, rectified linear unit (ReLU), and batch normalization (BN)~\cite{ioffe2015batch}.

A fundamental deficiency of LISTA and ISTA-Net is the absence of convergence guarantees, as they simply unfold ISTA without theoretical analysis. Thus, a necessary condition of \emph{partial weight coupling structure} is introduced to constrain the learnable parameters $\mathbf{W}_{1}^{n}$ and $\mathbf{W}_{2}^{n}$ for $n=0,1,\cdots,K$ in LISTA for convergence guarantees~\cite{NIPS2018_8120}.
%as specified in Eq.~\eqref{pwcs}.
\begin{equation}\label{pwcs}
\mathbf{W}_{2}^{n} = \mathbf{I}-\mathbf{W}_{1}^{n}\mathbf{A}
\end{equation}
Under the necessary condition, two improved versions of LISTA (\emph{i.e.}, LISTA-CP and LISTA-CPSS) are demonstrated to achieve linear convergence in theory. Furthermore, analytic LISTA (ALISTA)~\cite{liu2018alista} and Gated LISTA~\cite{Wu2020Sparse} extend LISTA-CP and LISTA-CPSS to simplify the unfolded model and improve the reconstruction performance with a guarantee of convergence property, respectively. Recently, ELISTA~\cite{li2021learned} developed an interpretable residual structure with convergence guarantees by introducing extragradient and substituting the soft-thresholding operator with the multistage-thresholding operator. 
%\request{Recently, by introducing extragradient and replacing the soft-thresholding operator with the multistage-thresholding operator, Extragradient based LISTA (ELISTA)~\cite{li2021learned} was proposed with an interpretable residual structure and convergence guatantee.}
Although convergence guarantees have been successfully established, these ISTA-based unfolded DNNs suffer from restricted network architectures and degraded recovery performance.
%Based on the convergence property of LISTA-CP(SS), analytic LISTA (ALISTA) and Gated LISTA were proposed to simplify the complexity of model and improve the reconstruction performance, respectively~\cite{liu2018alista, Wu2020Sparse}. Although establishing the convergence analysis of ISTA-based DNN models achieved great progress, these methods suffer from limitations with respect to flexibility and restricted recovery performance.

%Table~\ref{T1} summarizes the flexibility, convergence guarantees and generality of the aforementioned ISTA-based DNNs. 
In Table~\ref{T1}, we summarize existing ISTA-based unfolded DNNs in the sense of flexibility, convergence and generality. Here, flexibility evaluates the ability to support free-from DNNs in the unfolded algorithms, whereas generality indicates whether the ISTA-based unfolded DNNs can be viewed as a general framework. For example, Gated LISTA can be applied to LISTA and its variants based on the proposed gate mechanisms. As shown in Table~\ref{T1}, existing ISTA-based DNNs cannot satisfy all these criteria, as summarized below.
%The aforementioned ISTA-based DNNs, listed in Table~\ref{T1}, are limited in flexibility, adaptability, or theoretical analysis, as follows:
\begin{itemize}[leftmargin=*]
\item LISTA-CP, LISTA-CPSS, ALISTA, Gated LISTA and ELISTA are guaranteed to achieve a linear convergence rate but constrain the learnable parameters with the partial weight coupling structure. The learned parameters are restricted by the design of unfolded algorithms.
%\request{The operations to employ the learned parameters are restricted by the design of unfolded algorithms, \emph{e.g.}, matrix multiplication derived from classical ISTA.} %These facts imply that the convergence of these methods are ensured at the cost of the efficiency of DNNs, and consequently, cannot support free-form DNNs. 
These methods ensure the convergence in theory at the cost of the efficiency of DNNs and cannot support free-form DNNs. %\request{Actually, most existing empirically efficient DNN architectures are not feasible for theoretical analysis, \emph{e.g.}, convolutional layers, ReLU and residual connectivity~\cite{he2016deep}.	}
\item LISTA and ISTA-Net simply utilize the framework of ISTA without considering its convergence guarantees. ISTA-Net introduces free-form DNNs into classical ISTA to improve the reconstruction performance, but cannot make convergence analysis in theory. 
\end{itemize}

\begin{table}[!t]
\renewcommand{\baselinestretch}{1.0}
\renewcommand{\arraystretch}{1.0}
\centering
\caption{Comparison between popular ISTA-based unfolded DNNs.}\label{T1}
\begin{tabular}{lccc}
\toprule
Methods & Flexibility & Convergence & Generality\\
%~ & Architecture & ~ & Framework \\
\midrule
LISTA \cite{10.5555/3104322.3104374} &  &  & \\
ISTA-Net \cite{zhang2018ista} & \checkmark & & \\
LISTA-CP/CPSS \cite{NIPS2018_8120} &  & \checkmark & \\
ALISTA \cite{liu2018alista} &  & \checkmark &  \\
Gated LISTA \cite{Wu2020Sparse} &  & \checkmark & \checkmark \\
ELISTA \cite{li2021learned} &  & \checkmark &  \\
\midrule
HCISTA & \checkmark & \checkmark &  \checkmark \\
HLISTA & \checkmark & \checkmark &  \checkmark \\
\bottomrule
\end{tabular}
\end{table}

It is worth mentioning that these problems are not limited to ISTA-based unfolded DNNs. We take unfolded ADMM as another example. ADMM-Net~\cite{NIPS2016_6406,8550778} reformulates the ADMM iterations as a learnable network for magnetic resonance imaging (MRI) reconstruction. ADMM-Net substitutes the soft-thresholding operator with a learned nonlinear transform but cannot be guaranteed to converge in theory. 
%Analogous to LISTA-CP and LISTA-CPSS, D-LADMM \cite{xie2019differentiable} develops an unfolded ADMM model with convergence guarantee using fixed learned parameters following the ADMM scheme. 
In analogy to LISTA-CP and LISTA-CPSS, D-LADMM~\cite{xie2019differentiable} unrolls the ADMM scheme with convergence guarantees using learnable parameters that are restricted to fixed dimension to support matrix multiplication. These facts further support that the introduction of efficient free-form DNNs is not consistent with the theoretical guarantees of convergence. %Motivated by this observation, this paper proposes hybrid ISTA to incorporate free-form DNNs with a guarantee of convergence.

It is meaningful but difficult to analyze theoretically to incorporate free-form DNNs into conventional algorithms and unfolded networks. Conventional algorithms without introducing deep learning technology and unfolded networks with restricted network architectures can be analyzed in theory, but they cannot support flexible design of network architectures and suffer from limited performance. The free-form DNNs can break through the restriction and enhance the performance. However, introducing free-form DNNs without violating the theoretical guarantees of convergence is an unsolved problem.  

%\textbf{Our Contributions:}
Motivated by this observation, in this paper, we propose novel hybrid algorithms that incorporate free-form DNNs with classical ISTA (pre-computed parameters) and LISTA (learned parameters) to simultaneously achieve the efficiency and flexibility of DNNs and ensure convergence in theory. The contributions of this paper are summarized as below.
\begin{itemize}[leftmargin=*]
\item %We develop HCISTA that integrates classical ISTA with free-form DNNs to improve efficiency and flexibility while still retaining the convergence rate. 
We develop HCISTA that integrates classical ISTA (with pre-computed parameters) with free-form DNNs to improve efficiency and flexibility with a guarantee of convergence. HCISTA is demonstrated to converge at a rate that is equivalent to ISTA in the worst case, even with untrained DNNs.
%Rigorous arguments are made to guarantee that HCISTA converges with at least a sublinear rate of $O(1/n)$, even with untrained DNNs.
	
\item We further generalize the hybrid algorithm to variants of LISTA to simultaneously free the restricted DNN architectures and achieve linear convergence in theory. The proposed HLISTA is guaranteed to achieve constrained upper bound of recovery error and linear convergence rate under mild conditions.
\item We make extensive evaluations to corroborate the theoretical results and demonstrate that the proposed hybrid ISTA can reduce the reconstruction error with an enhanced convergence rate.
\end{itemize}

To our best knowledge, this paper is the first attempt to realize ISTA-based unfolded DNNs that can support network architectures without constraints and guarantee the convergence with enhanced rates. From the perspective of classical ISTA and LISTA, incorporating free-form DNNs without obscuring the theoretical convergence offers flexibility and efficiency to these algorithms. From the perspective of DNNs, the hybrid ISTA provides an interesting direction for designing interpretable DNNs that are used for inverse problems. To be concrete, the proposed method can be viewed as a special ISTA-based connectivity that is similar to residual connectivity and provide a way to endow the empirically constructed DNNs with theoretical interpretation and convergence guarantees. We thoroughly discuss it in Section~\ref{sec5.3}.

% The proposed hybrid ISTA (\emph{i.e.}, HCISTA and HLISTA) can be viewed as a general framework that benefits almost all DNNs for solving linear inverse problems with convergence guarantees. Specifically, hybrid ISTA can be regarded as ISTA-based connectivity that is similar to residual connectivity \cite{he2016deep} for endowing the DNNs with convergence. 

The rest of this paper is organized as follows. Section~\ref{sec:2} provides a brief overview of unfolded algorithms. In Sections~\ref{sec:3} and \ref{sec:4}, we propose the hybrid algorithms that incorporate free-form DNNs into classical ISTA with pre-computed parameters and LISTA with learned parameters, respectively. Discussion on the free-form DNNs and hybrid algorithms is provided in Section~\ref{sec:5}. Section~\ref{exper_sec} demonstrates the theoretical results developed for the hybrid algorithms in sparse recovery and natural image compressive sensing. Finally, we draw the conclusion in Section~\ref{sec:7}.
%Extensive experiments corroborate the theoretical results and demonstrate that the proposed Hybrid ISTA can reduce the reconstruction error with an enhanced convergence rate.
%Note that all the proofs of lemmas and theorems in this paper are provided in the Appendix.

\section{Related Work}\label{sec:2}
In this section, we briefly overview the unfolded iterative algorithms that inherit the interpretability of classical iterative algorithms and the efficiency of DNNs. 
%\subsection{Unfolded DNNs Based on ISTA}
\subsection{Unfolding ISTA  for the Convex Lasso problem}
%In recent years, tremendous progress has been made in incorporating classical iterative algorithms with deep neural networks. The outcome, called unfolded iterative algorithms, inherits both interpretation as optimization algorithms and efficiency of DNNs. Focusing on ISTA, 
Gregor and Lecun~\cite{10.5555/3104322.3104374} first developed LISTA by unfolding $K$ iterations of classical ISTA as a novel DNN model and achieved a substantial speedup over ISTA. To establish the theoretical convergence of LISTA, LISTA-CP~\cite{NIPS2018_8120} introduced a necessary condition of partial weight coupling structure in Eq.~\eqref{pwcs} to achieve linear convergence in theory. In addition, a special thresholding operator with support selection was presented in LISTA-CPSS~\cite{NIPS2018_8120} to improve the reconstruction performance of LISTA.
%To establish the theoretical convergence of LISTA, a necessary condition as Eq.~\eqref{pwcs}, dubbed partial weight coupling structure, was developed in~\cite{NIPS2018_8120} to achieved linear convergence in theory. LISTA equipped with partial weight coupling structure is called LISTA-CP. Besides, a special thresholding operator with support selection was proposed in~\cite{NIPS2018_8120} to improve the reconstruction performance of LISTA. LISTA-CP that utilizes the special thresholding operator is named LISTA-CP(SS).
Furthermore, based on the convergence property of LISTA-CP and LISTA-CPSS, analytic LISTA (ALISTA)~\cite{liu2018alista} simplified the learned parameters in LISTA and required to train only the step sizes and thresholds. Gated mechanisms were introduced in~\cite{Wu2020Sparse} to enhance the convergence speed of LISTA-CP and LISTA-CPSS. Gain gate and overshoot gate were developed to mitigate the underestimated magnitude of code components and compensate the small step size, respectively. Recently, ELISTA~\cite{li2021learned} introduced the idea of extragradient into LISTA and designed an alternative multistage-thresholding operator for the soft-thresholding operation. The network structure trained with extragradient was interpreted as a residual structure.
%Due to the extragradient technique, ELISTA interpreted the network structure as a residual structure.

In addition to the above ISTA-based unfolded DNNs, sibling architectures of LISTA have also been investigated. In~\cite{Moreau2016Understanding}, LISTA is shown to converge sublinearly yet faster than ISTA from the perspective of matrix factorization and was reparameterized into a new factorized architecture that achieves similar acceleration gain.  %In~\cite{Moreau2016Understanding}, LISTA was reparameterized into a new factorized architecture that achieves similar acceleration gain to LISTA. \request{A proof from the perspective of matrix factorization indicates that LISTA can converge sublinearly yet faster than ISTA.} 
LISTA was interpreted as a projected gradient descent with an inaccurate projection step, which implies a trade-off between the reconstruction error and convergence rate~\cite{Giryes2018Tradeoffs}.
%The learned IHT is proven to achieve a linear convergence rate under some strong assumptions that cannot be easily extended to LISTA. 
ISTA-Net~\cite{zhang2018ista} introduced free-form DNNs into the framework of ISTA as a learned transform to induce sparsity of signals and achieved a remarkable reconstruction performance in compressive sensing. 

\subsection{Unfolding ISTA for Convolutional Sparse Coding}

Some ISTA-based DNNs focused on convolutional sparse coding (CSC) to improve the model efficiency. Learned CSC~\cite{sreter2018learned} simply utilized LISTA to solve the CSC problem and achieved a better performance than traditional methods. Simon and Elad~\cite{simon2019rethinking} suggested a feed-forward network by unfolding ISTA and connected it with traditional patch-based methods. ALISTA was also extended for CSC.

\subsection{Unfolding Algorithms for Non-convex Problems}

The Lasso problem is often employed as a convex surrogate in place of the non-convex linear inverse problems with $l_0$-norm constraint, leading to a more tractable optimization problem. However, solving $l_0$-norm based sparse approximation is still preferable in some cases, inspiring many valuable works~\cite{dai2009subspace, foucart2011hard, shen2017tight, nguyen2017linear, needell2009cosamp}. Inspired by LISTA~\cite{10.5555/3104322.3104374}, there are some unfolded algorithms that investigated unrolling non-convex algorithms with $l_0$-norm constraint~\cite{wang_paper, NIPS2016_6346}.
\emph{Deep $l_0$ Encoders}~\cite{wang_paper} formulated two iterative algorithms for the $l_0$-norm regularized problem and the $M$-sparse problem as two feed-forward neural networks. By introducing learnable parameters and recasting the hard thresholding operation as a trainable linear unit or a pooling/unpooling operation, \emph{Deep $l_0$ Encoders} were optimized in a task-driven, end-to-end manner and obtained impressive performance on the tasks of image classification and clustering. LIHT~\cite{NIPS2016_6346} directly unfolded the IHT algorithm to improve the reconstruction quality when the transformation matrix has coherent columns, as quantified by a large restricted isometry constant. 
Although the theoretical convergence of LIHT was proved under strong assumptions, the analysis cannot be extended to LISTA. 

\subsection{Unfolding Other Optimization Algorithms}
In addition to ISTA, ADMM~\cite{admm2010}, AMP~\cite{amp2009}, and the Frank-Wolfe algorithm~\cite{frank1956} have also been unfolded into DNNs. Yang \emph{et al}.~\cite{NIPS2016_6406, 8550778} recast the ADMM procedure as a learnable network, called ADMM-Net, and applied it to CS based MRI reconstruction. D-LADMM~\cite{xie2019differentiable} developed an unfolded ADMM model for solving constrained optimization and provided rigorous argument for convergence guarantees. Borgerding \emph{et al}.~\cite{7934066} presented an AMP-inspired unfolded network for solving sparse linear inverse problems and Metzler \emph{et al}.~\cite{metzler2017learned} recast the denoising-based AMP algorithms as a novel network for compressive image recovery. Inspired by the Frank-Wolfe algorithm, Liu \emph{et al}.~\cite{liu2019frank} developed an unfolded Frank-Wolfe Network for solving $\ell_{p}$-norm constrained optimization with $p\geq 1$.

\section{HCISTA: Hybrid Algorithm of Classical ISTA and Free-Form DNNs}\label{sec:3} 
%Inspired by classical ISTA with pre-computed parameters, we first propose HCISTA that utilizes free-form DNNs into the ISTA scheme to solve Equation~\eqref{lasso}. 
We first propose HCISTA that incorporates free-form DNNs into the classical ISTA with pre-computed parameters for solving the Lasso problem. We reformulate Eq.~\eqref{lasso} by adaptively determining the regularization parameters for various iterations to improve the efficiency, \emph{i.e.}, $\lambda^n$ for the $n$th iteration, $n\in\mathbb{N}$.
At the $n$th iteration, the proposed HCISTA updates $\mathbf{x}^n$ according to the following steps:
%Steps of the proposed \textbf{HCISTA} in $n$-th iteration are defined as: 
\begin{equation}\label{e4}
\begin{aligned}
&\mathbf{v}^{n}= \mathcal{S}_{\lambda^n t^{n}}(\mathbf{x}^{n}-t^{n}\nabla f(\mathbf{x}^{n})), \\
&\mathbf{u}^{n}=N_{\mathcal{W}^{n}}(\mathbf{v}^{n}), \\
&\mathbf{w}^{n}= \mathcal{S}_{\lambda^n t^{n}}(\mathbf{u}^{n}-t^{n}\nabla f(\mathbf{u}^{n})),\\
&\mathbf{x}^{n+1}= \alpha^{n} \mathbf{v}^{n}+(1-\alpha^{n})\mathbf{w}^{n},
\end{aligned}
\end{equation}
where $\nabla f(\mathbf{x}^{n})=A^T(A\mathbf{x}^n-b)$, $N_{\mathcal{W}^{n}}$ is a free-form DNN with learnable parameters $\mathcal{W}^{n}$, $\alpha^n$ is the balancing parameter that controls the convex combination of $\mathbf{v}^n$ and $\mathbf{w}^n$, and $t^{n}$ is the step size for updating $\mathbf{v}^n$ and $\mathbf{w}^n$. 
%and $\lambda^n$ denotes the regularization parameter adaptively determined for the $n$th iteration.
It is obvious that the steps of updating $\mathbf{v}^{n}$ and $\mathbf{w}^{n}$ are the same as the basic step of ISTA specified in Eq.~\eqref{e1}, and a \emph{free-form DNN} $N_{\mathcal{W}^{n}}$ without constraint on its architecture is introduced for producing $\mathbf{u}^n$ for $n\in \mathbb{N}$. 

Without explicit specification, in this section, we consider the sequence $\{\mathbf{x}^n\}_{n\in\mathbb{N}}$ iteratively generated by Eq.~\eqref{e4}, where $\Theta=\{\delta^n, t^{n}, \lambda^{n}, \alpha^{n}, \mathcal{W}^{n}\}_{n\in \mathbb{N}}$ except for $\lambda^0$ are learnable parameters to be trained. 
Following~\cite{10.5555/3104322.3104374, NIPS2018_8120, liu2018alista, Wu2020Sparse, li2021learned}, the loss function for training models and obtaining $\Theta$ with $K$ iterations is defined as
\begin{equation}\label{loss_function_general}
\min_{\Theta}\mathcal{F}(\mathbf{x}^{K}, \mathbf{x}^{*})+\mathcal{G}(\mathbf{x}^K),
\end{equation}
where $\mathbf{x}^{K}$ is the output of the $K$th iteration, $\mathbf{x}^{*}$ is the learning target, $\mathcal{F}$ is the fidelity term and $\mathcal{G}$ is the regularization term. Without extra descriptions, we denote $\mathcal{F}$ as the mean squared error (MSE) loss and $\mathcal{G}(\mathbf{x}^K)=0$ in this paper. Note that there may be slight differences on different tasks for $\mathcal{F}$.

To guarantee convergence, we constrain the ranges of $\delta^n, t^n$, $\alpha^n$ and $\lambda^n$, $n\in\mathbb{N}$. For arbitrary $n\in\mathbb{N}$, if $\|\mathbf{v}^{n}-\mathbf{x}^{n}\|_{2}\neq 0$, $\alpha^n$ satisfies that
%where $\nabla f(x)=A^{T}(Ax-b)$, $N_{\mathcal{W}^{n}}$ is the DNN with learnable parameters $\mathcal{W}^{n}$ in the $n$-th iteration, $\alpha^n$ is the parameter that controls the convex combination of $v^n$ and $w^n$ in the $n$-th iteration, and $t^{n}$ represents the step size for updating $v^n$ and $w^n$ in the $n$-th iteration. Specifically, $\alpha^n$ is chosen such that
\begin{equation}\label{add5}
\frac{\|\mathbf{u}^{n}-\mathbf{x}^{n}\|_{2}^{2}}{\|\mathbf{u}^{n}-\mathbf{x}^{n}\|_{2}^{2}+(1-2t^{n}\delta^n\|\mathbf{A}\|_2^2)\|\mathbf{v}^{n}-\mathbf{x}^{n}\|_{2}^{2}}\leq\alpha^{n} < 1,  %\\\forall n \in \mathbb{N},
\end{equation}
%if $\|v^{n}-x^{n}\|_{2}\neq 0$, 
%where $\delta^n$ is the learned parameter that takes its value in $(0.25,0.5)$. 
where $\delta^n$ takes its value in $(0.25,0.5)$ and $t^n$ is chosen such that  
\begin{equation} \label{ls4}
\frac{1}{4\delta^n\|\mathbf{A}\|_2^2}\leq t^{n}\leq \frac{1}{\|\mathbf{A}\|_2^2},~\forall n \in\mathbb{N}.
\end{equation} 
If $\|\mathbf{v}^{n}-\mathbf{x}^{n}\|_{2}=0$, $\alpha^n$ is set to 1 and $\mathbf{x}^{n+1}=\mathbf{x}^n$. 
Moreover, $\lambda^n$ is chosen such that, for arbitrary $n \in\mathbb{N}_+$,
\begin{equation} \label{ls5}
0<\lambda^n\leq \min\{\lambda^{n-1}, C_{\lambda}\|\mathbf{x}^n-\mathbf{x}^{n-1}\|_2\},
\end{equation}
where $C_{\lambda}$ and $\lambda^0$ are hyper-parameters. 
%The ranges of $\delta^n, t^n$ and $\alpha^n$ are chosen for the theoretical analysis on convergence.

%In this section, without explicit specification, we consider the sequence $\{\mathbf{x}^n\}_{n\in\mathbb{N}}$ iteratively generated by Equation~\eqref{e4}. In Equation~\eqref{e4}, $\Theta=\{\delta^n, t^{n}, \alpha^{n}, \mathcal{W}^{n}\}_{n\in \mathbb{N}}$ are learned parameters to train and the specified values  are determined after the network training phase. It is obvious that the update steps of $\mathbf{v}^{n}$ and $\mathbf{w}^{n}$ are the same as the basic step of ISTA in Equation~\eqref{e1}, and a \emph{free-form DNN} $N_{\mathcal{W}^{n}}$ without constraint on its architectures is introduced for producing $\mathbf{u}^n$ for $n\in \mathbb{N}$. 
%and we do not require any constraint on the architectures of the DNNs $\{N_{\mathcal{W}^{n}}\}_{n\in \mathbb{N}}$.
To achieve convergence analysis, we first clarify some properties of the objective function $F=f+g$ for the Lasso problem defined in Eq.~\eqref{lasso} as below.
%In the following, we will analyze the convergence behavior of HCISTA. We state some properties about the Lasso problem in Equation~\eqref{lasso} which are helpful for the convergence analysis.
\begin{prop}\label{assum_lasso}[Properties of $F$, $f$, and $g$] For the Lasso problem defined in Eq.~\eqref{lasso}, we have that
\begin{enumerate}
\item $f$ is a smooth convex function with $L$-Lipschitz continuous gradient.
\item $g$ is a continuous convex function that is possibly nonsmooth.
\item $f$ and $g$ are proper convex functions, \emph{i.e.}, $f$ and $g$ have nonempty effective domain, and never attain $-\infty$.
\item $F$ is coercive, \emph{i.e.}, $F$ is bounded from below and $F\rightarrow\infty$ if $\|\mathbf{x}\|_2\rightarrow \infty$.
\item $F$ is semi-algebraic and is a KŁ function, \emph{i.e.}, $F$ satisfies the Kurdyka-Łojasiewicz (KŁ) property~\cite{bolte2014proximal}.
\end{enumerate}
\end{prop}

%\begin{enumerate}
%\item $f(x)$ is convex and Lipschitz smooth;
%\item $g(x)$ is convex and continuous;
%\item $F(x)$ satisfies the Kurdyka-Łojasiewicz (KŁ) property and is coercive. 
%\end{enumerate}
%More details are given in the Appendix~1. Specifically, we give the definition of KŁ property as follows. 
%\begin{defn} [\cite{bolte2014proximal}]\label{defn1}
%A function $f$: $\mathbb{R}^{N}\rightarrow (-\infty, +\infty]$ is said to have the KŁ property at $\bar{u}\in {\rm dom} \partial f := \{x\in \mathbb{R}^{N}: \partial f \neq \emptyset\}$ if there exists $\eta\in(0, +\infty]$, a neighborhood $U$ of  $\bar{u}$ and a function $\varphi\in\Phi_{\eta}$, such that for all $u\in U \bigcap\{u\in\mathbb{R}^{N}:f(\bar{u})<f(u)<f(\bar{u})+\eta\}$, the following inequality holds
%\begin{equation}\label{kl}
%\varphi'(f(u)-f(\bar{u})){\rm dist}(0, \partial f(u))\geq 1,
%\end{equation}
%where ${\rm dist}(0, \partial f(u))={\rm inf}\{\|x_{*}\|: x_*\in \partial f(u)\}$, and $\Phi_{\eta}$ stands for a class of function $\varphi: [0, \eta)\rightarrow \mathbb{R}^{+}$ satisfying: (1) $\varphi$ is concave and $C^1$ on $(0, \eta)$; (2) $\varphi$ is continuous at 0, $\varphi(0)=0;$ and (3) $\varphi'(x)>0, \forall x \in (0, \eta)$.
%\end{defn}
%\request{Since all semi-algebraic functions satisfy the KŁ property~\cite{bolte2014proximal} and $F$ in Equation~\eqref{lasso} is semi-algebraic, $F$ is a KŁ function.} %Furthermore, when $F$ is a semi-algebraic function, the desingularising function $\varphi(t)$ can be in the form of
Please refer to Appendix~A.1 for detailed description of Proposition~\ref{assum_lasso}. Furthermore, the desingularising function $\varphi(t)$ for the semi-algebraic function $F$ can be in the form of
\begin{equation}\label{varphi}
\varphi(t) = \frac{C}{\theta}t^{\theta},
\end{equation} 
where $\theta\in(0,1]$ and $C$ is a positive constant~\cite{frankel2015splitting, li2015accelerated}. 

Consequently, we make convergence analysis and develop convergence rates for HCISTA. In Lemma~\ref{lemma1}, we prove the inequalities for $F(\mathbf{x}^n)$, $F(\mathbf{w}^n)$, and $F(\mathbf{v}^n)$.
\begin{lemma}\label{lemma1}
Let $\{\mathbf{x}^{n}\}_{n\in \mathbb{N}}$ be a sequence generated by Eq.~\eqref{e4} with the learnable parameters $\Theta$. We have
\begin{equation}
F(\mathbf{x}^n)-F(\mathbf{w}^n)\geq\frac{1}{2t^{n}}(\|\mathbf{w}^{n}-\mathbf{x}^{n}\|_{2}^{2} -\|\mathbf{u}^{n}-\mathbf{x}^{n}\|_{2}^{2}),
\end{equation}
and 
\begin{equation}
F(\mathbf{x}^n)-F(\mathbf{v}^n)\geq\frac{1}{2t^{n}}\|\mathbf{v}^{n}-\mathbf{x}^{n}\|_{2}^{2}.
\end{equation}
\begin{IEEEproof}
Please refer to Appendix~A.2.
\end{IEEEproof}
\end{lemma}
%We prove in Lemma~\ref{lemma1} that, for arbitrary sequence $\{\mathbf{x}^{n}\}_{n\in\mathbb{N}}$ generated by Equation~\eqref{e4} with the learnable parameters $\alpha$ and $\{t^n\}_{n\in\mathbb{N}}$ satisfying Equations~\eqref{add5} and \eqref{ls4}, the objective sequence $\{F(\mathbf{x}^n)\}_{n\in\mathbb{N}}$ converges as $n\rightarrow \infty$.

Based on Lemma~\ref{lemma1}, we prove in Theorem~\ref{theorem1} that the objective sequence $\{F(\mathbf{x}^n)\}_{n\in\mathbb{N}}$ converges for arbitrary sequence $\{\mathbf{x}^{n}\}_{n\in\mathbb{N}}$ as $n\rightarrow \infty$.
\begin{theorem}\label{theorem1}
%Let $\{\mathbf{x}^{n}\}_{n\in \mathbb{N}}$ be a sequence generated by Equation~\eqref{e4}, $\alpha$ is chosen following Equation~\eqref{add5} and $t^n$ is chosen following Equation~\eqref{ls4}.
%Let $\{\mathbf{x}^{n}\}_{n\in \mathbb{N}}$ be a sequence generated by Equation~\eqref{e4} where the learnable parameters $\alpha$ and $\{t^n\}_{n\in\mathbb{N}}$ satisfy Equations~\eqref{add5} and \eqref{ls4}, respectively. $\{\mathbf{x}^{n}\}_{n\in \mathbb{N}}$ have accumulation points and $F$ achieves the same value $F_*$ at all the accumulation points.
Let $\{\mathbf{x}^{n}\}_{n\in \mathbb{N}}$ be a sequence generated by Eq.~\eqref{e4} with the learnable parameters $\Theta$. $\{\mathbf{x}^{n}\}_{n\in \mathbb{N}}$ have accumulation points where $F$ achieves the same value $F_*$. We further have 
\begin{equation}\label{cm}
\|\mathbf{x}^{n+1}-\mathbf{x}^{n}\|_{2}^{2} \rightarrow 0,\quad n\rightarrow \infty.
\end{equation}
\begin{IEEEproof}
Please refer to Appendix~A.3.
\end{IEEEproof}
\end{theorem}

However, Theorem~\ref{theorem1} is not sufficient to ensure that $\{\mathbf{x}^{n}\}_{n\in \mathbb{N}}$ converges to the optimum. To address this problem, we bound $\{\mathbf{v}^{n}\}_{n\in \mathbb{N}}$ and $\{\mathbf{w}^{n}\}_{n\in \mathbb{N}}$ with $\{\mathbf{x}^{n}\}_{n\in \mathbb{N}}$. Let us define the index set $\mathcal{T}=\{n|n\in\mathbb{N}, \|\mathbf{v}^n-\mathbf{x}^n\|_2=0\}$ for all the iterations with the same $\mathbf{v}^n$ and $\mathbf{x}^n$. Note that $\alpha^n=1$ and $\mathbf{x}^{n+1}=\mathbf{x}^n$ for $n\in\mathcal{T}$. For $n\notin\mathcal{T}$, it is reasonable for us to suppose that $\|\mathbf{u}^{n}-\mathbf{x}^{n}\|_2=\eta^{n}\|\mathbf{v}^{n}-\mathbf{x}^{n}\|_2$ with $\eta^n\ge 0$. Thus, we constrain $\mathbf{x}^n$, $\mathbf{v}^n$, and $\mathbf{w}^n$ for $n\notin\mathcal{T}$ in Lemma~\ref{lemma2}.
%To address this problem, we prove an inequality for the sequences $\{\mathbf{v}^{n}\}_{n\in \mathbb{N}}$, $\{\mathbf{w}^{n}\}_{n\in \mathbb{N}}$, and $\{\mathbf{x}^{n}\}_{n\in \mathbb{N}}$.
\begin{lemma}\label{lemma2}
Let $\{\mathbf{x}^{n}\}_{n\in \mathbb{N}}$ be a sequence generated by Eq.~\eqref{e4} with the learnable parameters $\Theta$. For arbitrary $n\notin\mathcal{T}$, $\mathbf{v}^{n}$ and $\mathbf{w}^{n}$ are bounded by
%let $\|\mathbf{v}^{n}-\mathbf{x}^{n}\|_2\neq 0$ and $\|\mathbf{u}^{n}-\mathbf{x}^{n}\|_2=\eta^{n}\|\mathbf{v}^{n}-\mathbf{x}^{n}\|_2$, then we have 
\begin{equation}
(a-\eta^nc^n)\|\mathbf{v}^{n}-\mathbf{x}^{n}\|_2+b\|\mathbf{w}^{n}-\mathbf{x}^{n}\|_2\leq (a+b)\|\mathbf{x}^{n+1}-\mathbf{x}^n\|_2,
\end{equation}
%\begin{align}
%(a-2a\eta^n+2a\eta^n\alpha^n-2b\eta^n\alpha^n)\|\mathbf{v}^{n}-\mathbf{x}^{n}\|_2+b\|\mathbf{w}^{n}-\mathbf{x}^{n}\|_2 \nonumber\\ \leq (a+b)\|\mathbf{x}^{n+1}-\mathbf{x}^n\|_2,
%\end{align}
where $a$ and $b$ are finite positive constants and $c^n=2a(1-\alpha^n)+2b\alpha^n$.
%where $a>0$, $b>0$, and $a, b$ are upper bounded.
\begin{IEEEproof}
Please refer to Appendix~A.4.
\end{IEEEproof}
\end{lemma}

%To guarantee the convergence of HCISTA, we also make an assumption about the $\eta^n$ in Lemma~\ref{lemma1}.
To guarantee the convergence of HCISTA, we further make an assumption on the existence of the upper bound of $\eta^n=\|\mathbf{u}^n-\mathbf{x}^n\|_2/\|\mathbf{v}^n-\mathbf{x}^n\|_2$ for $n\notin\mathcal{T}$. 
\begin{assumption}\label{assum1}
%Suppose that $\{\mathbf{x}^{n}\}_{n\in \mathbb{N}}$ be a sequence generated by Equation~\eqref{e4}, $\alpha$ is chosen following Equation~\eqref{add5} and $t^n$ is chosen following Equation~\eqref{ls4}. Define a set $\mathcal{T}=\{i|i\in\mathbb{N}, \|\mathbf{v}^i-\mathbf{x}^i\|=0\}$. For $n\notin\mathcal{T}$, let $\|\mathbf{u}^{n}-\mathbf{x}^{n}\|=\eta^{n}\|\mathbf{v}^{n}-\mathbf{x}^{n}\|$, then there exists a constant $\eta_c$ such that $0\leq \eta^n\leq \eta_c$.
Let $\{\mathbf{x}^{n}\}_{n\in \mathbb{N}}$ be a sequence generated by Eq.~\eqref{e4} with the learnable parameters $\Theta$. For $n\notin\mathcal{T}$, there exists a constant $\eta_c>0$ such that $0\leq \eta^n\leq \eta_c$.
\end{assumption}

Assumption~\ref{assum1} implies that $\eta^n$ does not diverge to infinity for $n\notin T$. Actually, it is a mild condition and we demonstrate some specific $\eta_c$ for $\{\eta^n\}_{n\in \mathbb{N}}$ in the experiments in Section~\ref{6.1.2}. Under Assumption~\ref{assum1}, we demonstrate in Theorem~\ref{theorem2} that HCISTA ensures that each accumulation point of $\{\mathbf{x}_n\}_{n\in\infty}$ is an optimum of $F$. 

%Equipped with Lemmas~\ref{lemma1}, \ref{lemma2} and Assumption~\ref{assum1}, we further establish a theorem to ensure that every accumulation point is an optimal point for HCISTA.
\begin{theorem}[Convergence of HCISTA]\label{theorem2}
%Suppose that $\{x^{n}\}_{n\in \mathbb{N}}$ be a sequence generated by Equation~\eqref{e4}, $\alpha$ is chosen following Equation~\eqref{add5} and $t^n$ is chosen following Equation~\eqref{ls4}. Let $x^*$ be any accumulation point of $\{x^n\}$.
Let $\{\mathbf{x}^{n}\}_{n\in \mathbb{N}}$ be a sequence generated by Eq.~\eqref{e4} with the learnable parameters $\Theta$ and $\mathbf{x}^* \in \mathbb{R}^{N}$ be arbitrary accumulation point of $\{\mathbf{x}^n\}_{n\in\mathbb{N}}$. When Assumption~\ref{assum1} holds, $\{\mathbf{x}^{n}\}_{n\in \mathbb{N}}$ converges to the optimum $\mathbf{x}^*$ of the convex objective function $F$ for the Lasso problem. 
%If Assumption~\ref{assum1} holds, then we have that $0\in \partial F(\mathbf{x}^*)$ and $\{\mathbf{x}^{n}\}_{n\in \mathbb{N}}$ is a Cauchy sequence. As $F$ is convex for the Lasso problem, $\{x^{n}\}_{n\in \mathbb{N}}$ converges to the optimal point $x^*$ of $F$. 
\begin{IEEEproof}
Please refer to Appendix~A.5.
\end{IEEEproof}
\end{theorem}

Theorem~\ref{theorem2} guarantees that, when Assumption~\ref{assum1} holds, HCISTA definitely converges to the optimum. This fact suggests that incorporating a free-form DNN in each iteration does not obscure the convergence of classical ISTA in theory. Intuitively, HCISTA achieves a faster convergence rate than ISTA as the DNNs can easily learn complicated statistics of signals. To address this issue, we further develop the convergence rate of HCISTA. Since $F$ satisfies the KŁ property, we obtain some exciting results in Theorem~\ref{theorem3} with similar framework of \cite{frankel2015splitting} and \cite{li2015accelerated}.

\begin{theorem}[Convergence Rate of HCISTA]
\label{theorem3}
%Suppose that $\{x^{n}\}_{n\in \mathbb{N}}$ be a sequence generated by Equation~\eqref{e4}, $\alpha$ is chosen following Equation~\eqref{add5} and $t^n$ is chosen following Equation~\eqref{ls4}.  $F(x)$ is a semi-algebraic function with KŁ inequality~\eqref{kl} and the desingularising function $\varphi(t)$ can be chosen to be the form of Equation~\eqref{varphi}. If Assumption~\ref{assum1} holds,  then
Let $\{\mathbf{x}_n\}_{n\in\mathbb{N}}$ be a sequence generated by Eq.~\eqref{e4} with the learnable parameters $\Theta$ and $\varphi(t)=Ct^\theta/\theta$ be the desingularising function for $F$ defined in Eq.~\eqref{varphi}. Given $F_*$ of $F$ achieved at the accumulation points of $\{\mathbf{x}^n\}_{n\in\mathbb{N}}$, when Assumption~\ref{assum1} holds, we have
\begin{enumerate}
\item If $\theta=1$ or $T$ is an infinite set, there exists $k_1$ such that $F(\mathbf{x}^n)=F_*$ for all $n>k_1$ and HCISTA terminates in finite steps.
\item If $\theta\in[\frac{1}{2},1)$, there exists $k_2$ such that for arbitrary $n=k_2+2l,~\forall l\in\mathbb{N}_{+}$,
\begin{equation}
F(\mathbf{x}^n)-F_*\leq\left(\frac{9C^2C_{max}^2}{\|\mathbf{A}\|_2^2+9C^2C_{max}^2}\right)^{l} r^{k_2},
\end{equation}
where $r^n=F(\mathbf{x}^n)-F_*$ and
\begin{align}
C_{max}=\max\left\{[(18+9\sqrt{2})\eta_c+9+3\sqrt{2}]\|\mathbf{A}\|_2^2, 4\sqrt{N} C_{\lambda}\right\}.   
\end{align}
%and $r^n=F(\mathbf{x}^n)-F_*$.
\item If $\theta\in(0,\frac{1}{2})$, there exists $k_3$ such that for all $n>k_3$,
\begin{equation}
F(\mathbf{x}^n)-F_*\leq \left[\frac{2C}{(n-k_3)C_r(1-2\theta)}\right]^{\frac{1}{1-2\theta}},
\end{equation}
%where $F_*$ is the same function value at all the accumulation points of $\{x^n\}$, 
%$r^n=F(x^n)-F_*$, $C$ is defined in Equation~\eqref{varphi}, 
%\begin{equation}
%E=12(6\eta_c^2+5\eta_c+2),
%\end{equation}
%and 
where
\begin{equation}
C_r = \min\left\lbrace \frac{\|\mathbf{A}\|_2^2}{18C_{max}^2C}, \frac{C}{1-2\theta}(2^{\frac{2\theta-1}{2\theta-2}}-1)(r^{0})^{2\theta-1} \right\rbrace.
\end{equation}
\end{enumerate}
\begin{IEEEproof}
Please refer to Appendix~A.6.
\end{IEEEproof}
\end{theorem}

In summary, Theorem~\ref{theorem3} demonstrates that HCISTA converges in a finite number of iterations when $\theta=1$ or $T$ is an infinite set, and with a sublinear rate of at least $O(1/n)$ in the sense of the gap $F(\mathbf{x}^n)-F_*$ when $0<\theta<\frac{1}{2}$. When $\frac{1}{2}\leq\theta<1$, the sequence $\{\mathbf{x}^{2l+k_2}\}$ converges with a linear rate for arbitrary $l\in\mathbb{N}_{+}$. This result implies that the odd and even subsequences of $\{\mathbf{x}_n\}_{n\in\mathbb{N}}$ converge to $\mathbf{x}^*$ with a linear rate, respectively. In this case, the convergence rate is faster than a sublinear rate. Note that ISTA converges at a rate of $O(1/n)$. Theorem~\ref{theorem3} implies that the convergence rate of HCISTA is at least equivalent to ISTA.

It is worth mentioning that Theorems~\ref{theorem1}$-$\ref{theorem3} have no particular requirement on the incorporated DNNs $N_{\mathcal{W}^n}$, $n\in\mathbb{N}$. This fact suggests that HCISTA can also support untrained DNNs with randomly initialized parameters, though HCISTA with a trained DNN tends to achieve better performance. In Section~\ref{exper_sec}, we show that HCISTA with both trained and untrained DNNs improves the reconstruction performance of ISTA under the same number of iterations, as suggested by Theorems~\ref{theorem1}$-$\ref{theorem3}.
%which means that untrained DNNs with initialized parameters also support the theorems. Undoubtedly, HCISTA with a trained DNN can achieve a better performance than with an untrained one. Experimental results in Section~\ref{exper_sec} support this proposition and validate the correctness of theorems, as  HCISTA obtains an improved reconstruction performance in comparison to ISTA after the same number of iterations. 
%In contrast, ISTA converges at an $O(1/n)$ rate and fast ISTA (FISTA) converges at an $O(1/n^{2})$ rate \cite{doi:10.1137/080716542}. Although we prove that HCISTA achieves the same convergence rates as ISTA and FISTA, HCISTA actually obtains an improved reconstruction performance and a faster convergence rate by introducing free-form DNNs. 

We can also interpret the superior convergence rate achieved by HCISTA from the perspective of incorporated DNNs. It is well-known that DNNs  have powerful fitting capacity to approximate the distribution of signals by training over numerous paired inputs and outputs, whereas classical iterative algorithms formalize the optimization problem to exploit the knowledge of models in a principled way. Therefore, free-form DNNs can learn to fit the distribution of signals when incorporated in the ISTA algorithm. Furthermore, Theorems~\ref{theorem1}$-$\ref{theorem3} actually provide a theoretically sound paradigm for HCISTA to fuse both analytical and empirical information with a guarantee of convergence. These facts imply that HCISTA enjoys the efficiency of free-form DNNs and convergence guarantees of classical ISTA algorithm. 

\section{HLISTA: Hybrid Algorithm of LISTA and Free-Form DNNs}\label{sec:4}
All the parameters in Eq.~\eqref{e1} are predetermined for ISTA. Despite the simplicity, ISTA converges very slowly with only a sublinear rate. To improve the convergence rate, LISTA~\cite{10.5555/3104322.3104374} resembles a recurrent neural network and learns the weights in ISTA. Popular variants of LISTA, \emph{i.e.}, LISTA-CP/CPSS~\cite{NIPS2018_8120}, ALISTA~\cite{liu2018alista}, Gated LISTA~\cite{Wu2020Sparse}, and ELISTA~\cite{li2021learned}, are proved to attain a linear convergence rate.
%In update step of classical ISTA, all the parameters are predetermined, as shown in Equation~\eqref{e1}. Despite the simplicity, it converges very slowly with only a sublinear rate. To address this issue, Gregor and LeCun first proposed to learn the weights in the matrices in ISTA rather than fixing them \cite{10.5555/3104322.3104374}. This method is called LISTA and resembles a recurrent neural network. As is aforementioned, there are many variants of LISTA, \emph{e.g.}, LISTA-CP, LISTA-CPSS, and ALISTA. All the three LISTA-based models are proved to attain a linear convergence rate theoretically and empirically.
In this section, we extend the hybrid algorithm to the variants of LISTA by introducing free-form DNNs in each iteration and improve the efficiency without obscuring the convergence guarantees. 

\subsection{HLISTA-CP}
We first propose HLISTA-CP that incorporates free-form DNNs with LISTA-CP~\cite{NIPS2018_8120} to improve the efficiency and flexibility and guarantee linear convergence. At the $n$th iteration, LISTA-CP updates $\mathbf{x}^n$ by
\begin{equation}\label{LISTA}
\mathbf{x}^{n+1}= \mathcal{S}_{\theta^{n}}\left(\mathbf{x}^{n}+(\mathbf{W}^{n})^{T}(\mathbf{b}-\mathbf{Ax}^{n})\right).
\end{equation}
In Eq.~\eqref{LISTA}, the learned weights $\mathbf{W}^n$ satisfy the partial weight coupling structure shown in Eq.~\eqref{pwcs}.
The $n$th iteration of HLISTA-CP is formulated by extending HCISTA.
%The steps of the proposed \textbf{HLISTA-CP} in the $n$-th iteration are formulated as: 
\begin{equation}\label{new3}
\begin{aligned}
&\mathbf{v}^{n}=\mathcal{S}_{\theta_{1}^{n}}\left(\mathbf{x}^{n}+(\overline{\mathbf{W}}^{n})^{T}(\mathbf{b}-\mathbf{Ax}^{n})\right), \\
&\mathbf{u}^{n}=N_{\mathcal{W}^{n}}(\mathbf{v}^{n}), \\
&\mathbf{w}^{n}=\mathcal{S}_{\theta_{2}^{n}}\left(\mathbf{u}^{n}+(\widehat{\mathbf{W}}^{n})^{T}(\mathbf{b}-\mathbf{Au}^{n})\right),\\
&\mathbf{x}^{n+1}=\alpha^{n}\mathbf{v}^{n}+(1-\alpha^{n})\mathbf{w}^{n}.
\end{aligned}
\end{equation}
Here, $N_{\mathcal{W}^{n}}$ is the free-form DNN with learnable parameters $\mathcal{W}^{n}$, and $\theta_{1}\geq0$ and $\theta_{2}\geq0$ are the learned thresholds for updating $\mathbf{v}^n$ and $\mathbf{w}^n$, respectively. Eq.~\eqref{LISTA} and Eq.~\eqref{new3} show that $\mathbf{v}^{n}$ and $\mathbf{w}^{n}$ of HLISTA-CP are updated in the same manner as LISTA-CP, whereas free-form DNNs are incorporated in HLISTA-CP to improve the flexibility and efficiency. 

In this subsection, we focus on the sequence $\{\mathbf{x}^n\}_{n\in\mathbb{N}}$ iteratively generated by Eq.~\eqref{new3} with the parameters $\Theta'=\{\theta^n_{1}, \theta^n_{2}, \overline{\mathbf{W}}^{n},  \widehat{\mathbf{W}}^{n}, \mathcal{W}^{n},
\alpha^{n}\}_{n\in\mathbb{N}}$ learned in the phase of network training.
%$N_{\mathcal{W}^{n}}$ is the free-form DNN with learnable parameters $\mathcal{W}^{n}$ in the $n$-th iteration.  Similar to HCISTA, we choose $\alpha_n$ in the following way:
In analogy to HCISTA, if $\theta_{1}^{n}\neq 0$, $\alpha^n$ is selected to satisfy
\begin{equation}\label{new8}
\frac{ \theta_{2}^{n}}{ \theta_{1}^{n}+ \theta_{2}^{n}}\leq\alpha^{n} <1.
\end{equation}
%if $\theta_{1}^{n}\neq 0$. 
If $\theta_{1}^{n}=0$, we set $\alpha^n=1$. The minimum of $\alpha^{n}$ can be determined once $\mathbf{v}^{n}$ and $\mathbf{w}^{n}$ are obtained. Thus, $\alpha^{n}$ is learned under the constraint shown in Eq.~\eqref{new8}.
The learnable parameters $\Theta'$ of HLISTA-CP are updated via the network training using the same loss function as HCISTA in Eq.~\eqref{loss_function_general}, and the other HLISTA models follow the same manner.
%In Equation~\eqref{new3}, $\Theta=\{\theta^n_{1}, \theta^n_{2}, \overline{W}^{n}, \mathcal{W}^{n}, \widehat{W}^{n}, \alpha^{n}\}_{n\in \mathbb{N}}$ are parameters to train and the specified values would be determined after the network training phase.  

%Note that the basic update step of LISTA-CP is defined as:
%\begin{equation}\label{LISTA}
%	x^{n+1}= \mathcal{S}_{\theta^{n}}\left(x^{n}+(W^{n})^{T}(b-Ax^{n})\right), \forall n \in \mathbb{N},
%\end{equation}
%where the learned weights satisfy the partial weight coupling structure shown in Equation~\eqref{pwcs}. The update steps of $v^{n}$ and $w^{n}$ in HLISTA-CP are the same as Equation~\eqref{LISTA}, while the free-form DNNs are incorporated to improve the flexibility and efficiency.

%In the following, we theoretically analyze the HLISTA-CP model. To guarantee the convergence of HLISTA-CP, we first make an assumption on the “ground truth” signal $x^{*}$.
Consequently, we establish convergence analysis for HLISTA-CP. We first introduce the same assumptions on the ``ground truth" signal $\mathbf{x}^*$  as in~\cite{NIPS2018_8120} and~\cite{liu2018alista}.
\begin{assumption}[Assumption~1 in~\cite{liu2018alista}]\label{assum2}
The signal $\mathbf{x}^{*}$ is supposed to be sampled from the constrained set $\mathcal{X}(B_{\mathbf{x}}, \mathbb{S})$ defined on the support $\mathbb{S}$ of $\mathbf{x}^*$.
\begin{equation}
\mathcal{X}(B_{\mathbf{x}}, \mathbb{S})\triangleq \left\{\mathbf{x}^*||x^{*}_{i}|\leq B_\mathbf{x}, 1\le i\le N, \|\mathbf{x}^{*}\|_{0}=|\mathbb{S}|\geq 2 \right\},
\end{equation}
where $x_i^*$ is the $i$th element of $\mathbf{x}^*$, $B_{\mathbf{x}}$ is a non-negative constant, and $|\mathbb{S}|$ is the cardinality of $\mathbb{S}$. 
%where $\mathbb{S}$ is the support of $x^*$ and $B_x$ is a constant. In other words, $x^*$ is bounded and has at least two non-zero elements.
\end{assumption}

Assumption~\ref{assum2} suggests that $\mathbf{x}^*$ is bounded and sparse and has at least two non-zero elements. We also simplify the proofs by assuming zero noise as in~\cite{liu2018alista} and \cite{Wu2020Sparse}. However, in the experiments, we demonstrate that the proposed methods are also robust to noise. 
%Here, we assume zero-noise to simplify the proofs, as in \cite{liu2018alista} and \cite{Wu2020Sparse}. In the experiments, we demonstrate that our models are also robust to noise.
Since a dictionary with small mutual coherence achieves better reconstruction performance in the tasks of sparse recovery and compressive sensing~\cite{candes2008an, candes2005decoding}, we introduce the generalized mutual coherence $\mu(\mathbf{A})$ for the under-complete basis matrix $\mathbf{A}$ in Eq.~\eqref{lasso}.
%Motivated by this point, we introduce the following definition of generalized mutual coherence, which is essential for the proofs.
\begin{defn}[Definition~1 in~\cite{liu2018alista}]\label{defn2}
%The generalized mutual coherence $\mu$ of $A\in \mathbb{R}^{M\times N}$ (each column of A is normalized) is defined as
The generalized mutual coherence $\mu(\mathbf{A})$ of $\mathbf{A}\in \mathbb{R}^{M\times N}$ consisting of normalized columns $\mathbf{A}_i\in\mathbb{R}^M$, $i=1,\cdots,N$ is defined as
\begin{equation}\label{gmc}
\mu(\mathbf{A})=\inf_{\substack{\mathbf{W}\in\mathbb{R}^{M\times N} \\ (\mathbf{W}_{i})^{T}\mathbf{A}_{i}=1, 1\leq i\leq N}}\left\{ \max_{\substack{i\neq j \\ 1\leq j\leq N}} |(\mathbf{W}_{i})^{T}\mathbf{A}_{j}|\right\},
\end{equation}
%where the subscript $i$ (resp. $j$) represents the $i$ (resp.$j$)-th column.
where $\mathbf{W}_i$ is the $i$th column of $\mathbf{W}$. Moreover, we define the space $\mathcal{W}_s(\mathbf{A})$ of $\mathbf{W}$ that achieves infimum in Eq.~\eqref{gmc}.
\begin{align}
\mathcal{W}_s(\mathbf{A})=\{\mathbf{W}|&(\mathbf{W}_i)^T\mathbf{A}_i=1, 1\leq i\leq N,\nonumber\\ &\max_{1\leq i\neq j\leq N}|(\mathbf{W}_{i})^{T}\mathbf{A}_{j}|=\mu(\mathbf{A})\}
\end{align}
%\begin{multline}
%W_s(A)=\{W\in \mathbb{R}^{M\times N}: \\ W~\rm{attains~ the~infimum~given~in~}Equation\eqref{gmc}\},
%\end{multline}
%which means that 
%\begin{equation}
%(W_{i})^{T}A_{i}=1, 1\leq i\leq N, \max_{\substack{i\neq j \\ 1\leq i,j \leq N}}|(W_{i})^{T}A_{j}|=\mu(A).
%\end{equation}
\end{defn}
%Note that there exists matrices $W$ that $W \in W_s(A)$, \emph{i.e.}, $W_s(A)\neq \emptyset$, which was proven in Lemma~1 of \cite{NIPS2018_8120}.
Note that $\mathcal{W}_s(\mathbf{A})\neq\emptyset$, as proved in Lemma~1 in~\cite{NIPS2018_8120}. 

To guarantee the convergence of HLISTA-CP, the ranges of $\theta^n_{1}$, $\theta^n_{2}$, $\overline{\mathbf{W}}^{n}$, and $\widehat{\mathbf{W}}^{n}$ are further constrained in the similar manners as \cite{NIPS2018_8120, liu2018alista, Wu2020Sparse}.
For simplicity, we use $\mu$ for $\mu(\mathbf{A})$ in the rest of this paper. Given arbitrary $\mathbf{x}^*\in\mathcal{X}(B_{\mathbf{x}},\mathbb{S})$, the parameters $\{\overline{\mathbf{W}}^{n}, \widehat{\mathbf{W}}^{n}, \theta_{1}^{n}, \theta_{2}^{n}\}_{n\in\mathbb{N}}$ are determined by
\begin{equation}\label{condition}
\begin{aligned}
&\overline{\mathbf{W}}^{n}\in\mathcal{W}_s(\mathbf{A}),\  \widehat{\mathbf{W}}^{n}\in\mathcal{W}_s(\mathbf{A}), \\
&\theta_{1}^{n}=\sup_{\mathbf{x}^{*}\in\mathcal{X}(B_{\mathbf{x}},\mathbb{S})}\{\mu\|\mathbf{x}^{n}-\mathbf{x}^{*}\|_{1}\}, \\
&\theta_{2}^{n} = \sup_{\mathbf{x}^{*}\in\mathcal{X}(B_{\mathbf{x}},\mathbb{S})}\{\mu\|\mathbf{u}^{n}-\mathbf{x}^{*}\|_{1}\}+\frac{\mu(N-|\mathbb{S}|)}{|\mathbb{S}|-1}\|\mathbf{u}^n\|_{1}.
\end{aligned}
\end{equation}

Consequently, we make analysis on the convergence properties of $\{\mathbf{x}^{n}\}_{n\in\mathbb{N}}$ generated by Eq.~\eqref{new3} using the learnable parameters $\Theta'$ specified by Eq.~\eqref{new8} and Eq.~\eqref{condition}.
In Theorem~\ref{theorem4}, we develop the upper bound of recovery error for HLISTA-CP under Assumption~\ref{assum2}. 

%From Assumption~\ref{assum2} and Definition~\ref{defn2}, we establish the recovery error bound of HLISTA-CP in Theorem~\ref{theorem3}. For simplicity, we use $\mu$ to replace $\mu(A)$.
\begin{theorem} 
[Upper Bound of Recovery Error for HLISTA-CP]
\label{theorem4}
Given arbitrary signal $\mathbf{x}^*\in\mathcal{X}(B_{\mathbf{x}},\mathbb{S})$, let $\{\mathbf{x}^{n}\}_{n\in\mathbb{N}}$  be the sequence generated by Eq.~\eqref{new3} from $\mathbf{x}^0=0$ using the learnable parameters $\Theta'$ specified in Eq.~\eqref{new8} and Eq.~\eqref{condition}. If $B_\mathbf{x}>0$ and $|\mathbb{S}|<\frac{2+1/\mu}{4}$, for arbitrary $n\in \mathbb{N}$,
%Take any $\mathbf{x}^*\in\mathcal{X}(B_{\mathbf{x}},\mathbb{S})$. Define the parameters $\{\overline{\mathbf{W}}^{n}, \widehat{\mathbf{W}}^{n}, \theta_{1}^{n}, \theta_{2}^{n}\}_{n\in\mathbb{N}}$:
%\begin{align}\label{condition}
%&\overline{\mathbf{W}}^{n}\in\mathcal{W}_s(\mathbf{A}),\  \widehat{\mathbf{W}}^{n}\in\mathcal{W}_s(\mathbf{A}), \nonumber \\
%&\theta_{1}^{n}=\sup_{\mathbf{x}^{*}\in\mathcal{X}(B_{\mathbf{x}},\mathbb{S})}\{\mu\|\mathbf{x}^{n}-\mathbf{x}^{*}\|_{1}\}, \nonumber \\
%&\theta_{2}^{n} = \sup_{\mathbf{x}^{*}\in\mathcal{X}(B_{\mathbf{x}},\mathbb{S})}\{\mu\|\mathbf{u}^{n}-\mathbf{x}^{*}\|_{1}\}+\frac{\mu(N-|\mathbb{S}|)}{|\mathbb{S}|-1}\|\mathbf{u}^n\|_{1},
%\end{align}
%while the sequence $\{\mathbf{x}^{n}\}_{n\in\mathbb{N}}$ is generated by Equation~\eqref{new3} using the above parameters and $\mathbf{x}^{0}=0$. Let $\{\alpha^{n}\}$ satisfy Equation~\eqref{new8}, and Assumption~\ref{assum2} holds with $B_\mathbf{x}>0$ and $|\mathbb{S}|<\frac{2+1/\mu}{4}$. Then, we have 
% \begin{equation}
% \|\mathbf{x}^n-\mathbf{x}^*\|_2\le|\mathbb{S}|B_{\mathbf{x}}\exp{(n\log{(4\mu|\mathbb{S}|-2\mu)})}.   
% \end{equation}
\begin{equation}\label{cp}
\begin{aligned}
{\rm supp}(\mathbf{x}^n)\subset \mathbb{S},\  \|\mathbf{x}^{n}-\mathbf{x}^{*}\|_{2}\leq |\mathbb{S}|B_{\mathbf{x}}\exp\left(-cn\right),
\end{aligned}
\end{equation}
where
\begin{equation}\label{c_cp}
c=-\log\left(4\mu|\mathbb{S}|-2\mu\right)>0.
\end{equation}
\begin{IEEEproof}
Please refer to Appendix~B.1.
\end{IEEEproof}
\end{theorem}

Theorem~\ref{theorem4} indicates that there exists a sequence of parameters $\{\overline{\mathbf{W}}^{n},\widehat{\mathbf{W}}^{n},\theta_{1}^{n},\theta_{2}^{n}\}_{n\in\mathbb{N}}$ such that there is no ``false positive" in $\mathbf{x}^n$, and the recovery error vanishes at a linear convergence rate as the number of layers grows to infinity. This conclusion is similar to LISTA-CP. 
In empirical evaluations, however, HLISTA-CP obviously outperforms LISTA-CP in the sense of reconstruction performance and convergence rate. This gain comes from the efficiency of free-form DNNs without constraints on their architectures. 
%From the perspective of DNNs, it seems to be true as the learned parameters of HLISTA-CP are seemingly more than LISTA-CP with iterations of same times.
Although HLISTA-CP introduces additional learnable parameters into LISTA-CP in each iteration according to Eq.~\eqref{LISTA} and Eq.~\eqref{new3}, we show in Section~\ref{exper_sec} that HLISTA-CP is superior to LISTA-CP in reconstruction performance with fewer iterations, when it uses the same or even fewer parameters for each iteration, \emph{e.g.}, by making $\overline{\mathbf{W}}^{n}=\widehat{\mathbf{W}}^{n}$. This result corroborates the efficiency of HLISTA-CP.

\subsection{HLISTA-CPSS}
We extend the hybrid algorithm to LISTA-CPSS that introduces a special thresholding function with support selection in comparison to LISTA-CP \cite{NIPS2018_8120}. At the $n$th iteration, LISTA-CPSS updates $\mathbf{x}^n$ by 
\begin{equation}\label{LISTA-CPSS}
\mathbf{x}^{n+1}= \mathcal{S}^{p^n}_{ss, \theta^{n}}\left(\mathbf{x}^{n}+(\mathbf{W}^{n})^{T}(\mathbf{b}-\mathbf{Ax}^{n})\right).
\end{equation}
Here, we denote $\mathcal{S}^{p^n}_{ss, \theta^{n}}$ as the thresholding operator with support selection. 
\begin{equation}\label{thresholding}
(\mathcal{S}^{p^n}_{ss, \theta^{n}}(\mathbf{z}))_i=
\begin{cases}
z_i: & ~~~z_i>\theta^n, i\in S^{p^n}(\mathbf{z}), \\
z_i - \theta^n: & ~~~z_i>\theta^n, i\not\in S^{p^n}(\mathbf{z}), \\
0: & ~~~-\theta^n\leq z_i <\theta^n, \\
z_i + \theta^n: & ~~~z_i<-\theta^n, i\not\in S^{p^n}(\mathbf{z}), \\
z_i: & ~~~z_i<-\theta^n, i\in S^{p^n}(\mathbf{z}),
\end{cases}
\end{equation}
where $S^{p^n}(\mathbf{z})$ includes the elements with the largest $p^n \%$ magnitudes in vector $\mathbf{z}$:
\begin{align}\label{thresholding2}
S^{p^n}(\mathbf{z})=&\left\lbrace i_1, i_2,\cdots, i_{p^n}\Big| \right.\nonumber\\ &\quad\left.|z_{i_1}| \geq |z_{i_2}|\geq\cdots|z_{i_{p^n}}|\geq\cdots\geq|z_{i_N}| \right\rbrace .
\end{align}
According to Eq.~\eqref{thresholding} and Eq.~\eqref{thresholding2}, $\mathcal{S}^{p^n}_{ss, \theta^{n}}$ selects the $p^n$ percentage of entries with the largest magnitudes that do not pass through the threshold. 

%Based on Equation~\eqref{LISTA-CPSS}, the steps of the proposed \textbf{HLISTA-CPSS}  in the $n$-th iteration are formulated as:
HLISTA-CPSS incorporates free-form DNNs into Eq.~\eqref{LISTA-CPSS}. The $n$th iteration is formulated as
\begin{equation}\label{HLISTA-CPSS}
\begin{aligned}
&\mathbf{v}^{n}= \mathcal{S}^{p^n}_{ss, \theta_{1}^{n}}\left(\mathbf{x}^{n}+(\overline{\mathbf{W}}^{n})^{T}(\mathbf{b}-\mathbf{Ax}^{n})\right), \\
&\mathbf{u}^{n}=N_{\mathcal{W}^{n}}(\mathbf{v}^{n}), \\
&\mathbf{w}^{n}= \mathcal{S}^{p^n}_{ss, \theta_{2}^{n}}\left(\mathbf{u}^{n}+(\widehat{\mathbf{W}}^{n})^{T}(\mathbf{b}-\mathbf{Au}^{n})\right),\\
&\mathbf{x}^{n+1}= \alpha^{n}\mathbf{v}^{n}+(1-\alpha^{n})\mathbf{w}^{n}.
\end{aligned}
\end{equation}
In Eq.~\eqref{HLISTA-CPSS}, the notations are consistent with HLISTA-CP and $\alpha^n$ is constrained by Eq.~\eqref{new8}, except for the introduction of the thresholding operator. Following the same setting as \cite{NIPS2018_8120}, $p^n$ is a hyper-parameter to be manually tuned. Specifically, we determine $p^n$ for the $n$th iteration with $p^n=\min(p\cdot n, p_{\rm{max}})$, where $p$ is the positive constant and $p_{\rm{max}}$ is the maximal percentage of the support cardinality. Thus, the parameters to be trained are the same as HLISTA-CP, \emph{i.e.}, $\Theta'=\{\theta^n_{1}, \theta^n_{2}, \overline{\mathbf{W}}^{n}, \widehat{\mathbf{W}}^{n}, \mathcal{W}^{n}, \alpha^{n}\}_{n\in \mathbb{N}}$.

Equipped with Assumption~\ref{assum2} and Definition~\ref{defn2}, theoretical analysis is achieved for HLISTA-CPSS in Theorem~\ref{theorem5}.
\begin{theorem}[Upper Bound of Recovery Error for HLISTA-CPSS]\label{theorem5}
Given arbitrary $\mathbf{x}^*\in \mathcal{X}(B_{\mathbf{x}}, \mathbb{S})$, consider the sequence $\{\mathbf{x}^{n}\}_{n\in\mathbb{N}}$ generated by Eq.~\eqref{HLISTA-CPSS} using the parameters $\{\overline{\mathbf{W}}^{n}, \widehat{\mathbf{W}}^{n}, \theta_{1}^{n}, \theta_{2}^{n}\}_{n\in\mathbb{N}}$ defined in Eq.~\eqref{condition} and $\mathbf{x}^{0}=0$. Let 
\begin{equation}
\begin{aligned}
&\Psi_{\mathbf{x}^n}=\lbrace i|i \in \mathbb{S}, x^n_{i}\neq 0, i\in S^{p^n}(\mathbf{x}^n)\rbrace,\\
&|\Psi_{*^n}|=\min \{|\Psi_{\mathbf{v}^n}|, |\Psi_{\mathbf{w}^n}|\},
\end{aligned}
\end{equation}
and $\{\alpha^{n}\}_{n\in\mathbb{N}}$ be constrained by Eq.~\eqref{new8}. Under Assumption~\ref{assum2} that $B_\mathbf{x}>0$ and $|\mathbb{S}|<(2+2\min_{n}\{|\Psi_{*^n}|\}+1/\mu)/4$, we have for arbitrary $n\in\mathbb{N}$,
\begin{equation}\label{cpss}
\begin{aligned}
{\rm supp}(\mathbf{x}^n)\subset \mathbb{S},\ \|\mathbf{x}^{n}-\mathbf{x}^{*}\|_{2}\leq|\mathbb{S}|B_\mathbf{x} \exp\left(-\sum_{k=0}^{n-1}c_{ss}^{k}\right),
\end{aligned}    
\end{equation}
where
\begin{equation}
c_{ss}^{k}=-\log\left[ 4\mu|\mathbb{S}|-2\mu-2\mu|\Psi_{*^k}|\right] >0.
\end{equation}
Furthermore, we have $c_{ss}^k\geq c$, where $c$ is defined in Eq.~\eqref{c_cp} in Theorem~\ref{theorem4}.
\begin{IEEEproof}
Please refer to Appendix~B.2.
\end{IEEEproof}
\end{theorem}

Similar to Theorem~\ref{theorem4} for HLISTA-CP, Theorem~\ref{theorem5} implies that there exists a sequence of parameters such that there is no ``false positive" in $\mathbf{x}^n$ and $\mathbf{x}^n$ converges to $\mathbf{x}^*$ at a linear rate. Moreover, the upper bound in Eq.~\eqref{cpss} guarantees that, with the same number of iterations, HLISTA-CPSS is at least no worse than HLISTA-CP. This conclusion implies that HLISTA-CPSS achieves a wider upper bound of $|\mathbb{S}|$ and a sharper upper bound of recovery error, when compared with HLISTA-CP. Experimental results also validate that HLISTA-CPSS is superior to HLISTA-CP in sparse recovery. Moreover, HLISTA-CPSS is shown to outperform LISTA-CPSS in terms of reconstruction performance, even with fewer learned parameters and numbers of iterations. 

\subsection{HALISTA}
We further extend the hybrid algorithm to ALISTA~\cite{liu2018alista}. Based on LISTA-CP in Eq.~\eqref{LISTA}, ALISTA decomposes the weights $\mathbf{W}^n$ into the product of a scalar $\gamma^n$ and a matrix $\mathbf{W}\in\mathcal{W}_s(\mathbf{A})$ independent of the layer index $n$, \emph{i.e.},
\begin{equation}\label{ALISTA}
\mathbf{W}^n = \gamma^n\mathbf{W}.
\end{equation}
%where $\mathbf{W}\in\mathcal{W}_s(\mathbf{A})$.  

HALISTA formulates each step by introducing Eq.~\eqref{ALISTA} into Eq.~\eqref{new3} and constraining $\alpha^n$ with Eq.~\eqref{new8}. 
Similar to HLISTA-CP, we develop in Theorem~\ref{theorem6} the upper bound of recovery error for HALISTA based on Assumption~\ref{assum2} and Definition~\ref{defn2}. Given arbitrary $\mathbf{x}^* \in \mathcal{X}(B_{\mathbf{x}}, \mathbb{S})$ and $\mathbf{W}\in \mathcal{W}_s(\mathbf{A})$, the parameters $\{\overline{\mathbf{W}}^{n}, \widehat{\mathbf{W}}^{n}, \theta_{1}^{n}, \theta_{2}^{n}\}_{n\in\mathbb{N}}$ are determined by
\begin{align}\label{eq34}
%\begin{aligned}
&\overline{\mathbf{W}}^{n} = \gamma_1^n \mathbf{W}, ~\widehat{\mathbf{W}}^{n} = \gamma_2^n \mathbf{W}, \nonumber\\
&\theta_{1}^{n}=\gamma_1^n\sup_{\mathbf{x}^{*}\in\mathcal{X}(B_{\mathbf{x}}, \mathbb{S})}\{\mu\|\mathbf{x}^{n}-\mathbf{x}^{*}\|_{1}\},\\
&\theta_{2}^{n} =\gamma_2^n \sup_{\mathbf{x}^{*}\in\mathcal{X}(B_{\mathbf{x}}, \mathbb{S})}\{\mu\|\mathbf{u}^{n}-\mathbf{x}^{*}\|_{1}\}+\frac{\gamma_2^n\mu(N-|\mathbb{S}|)}{|\mathbb{S}|-1}\|\mathbf{u}^n\|_{1}.\nonumber
%\end{aligned}
\end{align}
where $\gamma_1^n$ takes its value in $(0, 2/(1+4\mu|\mathbb{S}|-2\mu))$ and $\gamma_2^n\equiv1$.
\begin{theorem}[Upper Bound of Recovery Error for HALISTA]\label{theorem6}
%Take any $\mathbf{x}^* \in \mathcal{X}(B_{\mathbf{x}}, \mathbb{S})$, any $\mathbf{W}\in W_s(A)$, and any sequences $\gamma_1^n\in(0, 2/(1+4\mu|\mathbb{S}|-2\mu))$, $\gamma_2^n\equiv1$. Define the parameters $\{\overline{\mathbf{W}}^{n}, \widehat{\mathbf{W}}^{n}, \theta_{1}^{n}, \theta_{2}^{n}\}^{\infty}_{n=0}$:
%\begin{align}\label{eq34}
%&\overline{\mathbf{W}}^{n} = \gamma_1^n \mathbf{W}, ~\widehat{\mathbf{W}}^{n} = \gamma_2^n \mathbf{W}, \nonumber \\
%&\theta_{1}^{n}=\gamma_1^n\sup_{\mathbf{x}^{*}\in\mathcal{X}(B_{\mathbf{x}}, \mathbb{S})}\{\mu\|\mathbf{x}^{n}-\mathbf{x}^{*}\|_{1}\}, \nonumber \\
%&\theta_{2}^{n} =\gamma_2^n \sup_{\mathbf{x}^{*}\in\mathcal{X}(B_{\mathbf{x}}, \mathbb{S})}\{\mu\|\mathbf{u}^{n}-\mathbf{x}^{*}\|_{1}\}+\frac{\gamma_2^n\mu(N-|\mathbb{S}|)}{|\mathbb{S}|-1}\|\mathbf{u}^n\|_{1}.
%\end{align}
Let $\{\mathbf{x}^{n}\}_{n\in\mathbb{N}}$ be the sequence generated by Eq.~\eqref{new3} using the parameters $\{\overline{\mathbf{W}}^{n}, \widehat{\mathbf{W}}^{n}, \theta_{1}^{n}, \theta_{2}^{n}\}_{n\in\mathbb{N}}$ determined by Eq.~\eqref{eq34}, $\{\alpha^{n}\}_{n\in\mathbb{N}}$ constrained by Eq.~\eqref{new8}, and $\mathbf{x}^{0}=0$. Under Assumption~\ref{assum2} that $B_\mathbf{x}>0$ and $|\mathbb{S}|<(2+1/\mu)/4$, we have for arbitrary $n\in\mathbb{N}$,
\begin{equation}\label{a}
\begin{aligned}
{\rm supp}(\mathbf{x}^n)\subset \mathbb{S},\ \|\mathbf{x}^{n}-\mathbf{x}^{*}\|_{2}\leq |\mathbb{S}|B_\mathbf{x} \exp\left(-\sum_{k=0}^{n-1}c_{a}^{k}\right),
\end{aligned}
\end{equation}
where
\begin{align}\label{cka}
c^k_a&=-\log\left[ 2\mu\gamma_1^k(2|\mathbb{S}|-1)+\left|1-\gamma_1^k\right|+\gamma_1^k\left|1-\gamma_2^k\right|/\gamma_2^k\right]\nonumber\\
&>0.
\end{align}
\begin{IEEEproof}
Please refer to Appendix~B.3.
\end{IEEEproof}
\end{theorem}

In Theorem~\ref{theorem6}, the scalar $\gamma_2^n$ is set to 1 to guarantee $c_a^k>0$. Please refer to Appendix~B.3 for details. When $\gamma_2^n=1$, Eq.~\eqref{eq34} and Eq.~\eqref{cka} can be further simplified.  Similar to HLISTA-CP, we obtain that there exists a sequence of parameters such that the recovery error converges to zero in a linear rate with $\gamma_1^n=1$. 
Although $\gamma_1^n=1$ gives the optimal upper bound in theory for infinite iterations, it is not the optimal choice for finite $n$. In practice, we construct the model using finite number of iterations and learn $\{\gamma_1^n\}_{n\in\mathbb{N}}$ that locate in the interval of $(0, 2/(1+4\mu|\mathbb{S}|-2\mu))$.

Note that HALISTA  requires fewer learnable parameters than HLISTA-CP, \emph{i.e.}, $\Theta''=\{\theta^n_{1}, \theta^n_{2}, \gamma^n_1,$ $\alpha^{n}, \mathcal{W}^n\}_{n\in \mathbb{N}}$. HALISTA actually achieves a comparable recovery performance with a more lightweight framework and fewer learnable parameters in comparison to HLISTA-CP, as demonstrated in the experiments. Besides, we show in Section~\ref{6.1.3} that HALISTA achieves an improved performance than ALISTA due to the introduction of free-form DNNs.

%\subsection{HCISTA with Adaptive Step Sizes and  Regularization  Parameter}

\subsection{HGLISTA}
\label{sec_HGLISTA}
We also extend the hybrid algorithm to Gated LISTA~\cite{Wu2020Sparse}, dubbed HGLISTA. Gated LISTA introduces two gate mechanisms, \emph{i.e.}, \emph{gain gates} and \emph{overshoot gates}, to improve the reconstruction performance from two perspectives. 
%Proposition~1 in~\cite{Wu2020Sparse} revealed that the components of $\mathbf{x}^n$ must be smaller than or at most equal to those of the $\mathbf{x}^*$ in the existing variants of LISTA, \emph{i.e.}, $|x^n_j|\leq|x^*_j|, ~\forall j\in [1, N]$. The reason is that the thresholds $\{\theta^n\}_{n\in\mathbb{N}}$ in Eq.~\eqref{LISTA} and Eq.~\eqref{LISTA-CPSS} are required to be large enough to eliminate all ``false positive'' in the support of $\{\mathbf{x}^n\}_{n\in\mathbb{N}}$, but sometimes are too large to suppress $\mathbf{x}^n$. To address this issue, 
\emph{Gain gates} are designed to enlarge $\mathbf{x}^n$ to improve the performance, and the step in the $n$th iteration is when introduced to LISTA-CP.
\begin{align}\label{e1_4.4}
\mathbf{x}^{n+1}= \mathcal{S}_{\theta^{n}}(&\mathbf{x}^{n}\odot g_t(\mathbf{x}^{n}, \mathbf{b}|\Lambda_g^n)\nonumber\\
&+(\mathbf{W}^{n})^{T}(\mathbf{b}-\mathbf{A}\mathbf{x}^{n}\odot g_t(\mathbf{x}^{n}, \mathbf{b}|\Lambda_g^n))),
\end{align}
where the gate function $g_t(\cdot,\cdot|\Lambda_g^n)$ outputs an $N$-dimensions vector using a set of learnable parameters $\Lambda_g^n$ in the $n$th iteration, and $\odot$ represents element-wise multiplication of two vectors.
Besides, \emph{overshoot gates} adjust the output of current iteration based on previous outputs. 
When overshoot gates are introduced to LISTA-CP, the $n$th iteration is
% Specifically, a scalar $\eta$ is first introduced  in~\cite{Wu2020Sparse} to adjust the output and improve classical ISTA. %\cite{Wu2020Sparse} first improved classical ISTA as follows.
% \begin{align}
% \mathbf{\bar{x}}^{n+1}&=\mathcal{S}_{\lambda t}\left(\mathbf{x}^{n}-t\mathbf{A}^{T}(\mathbf{Ax}^{n}-\mathbf{b})\right), \nonumber\\
% \mathbf{x}^{n+1}&=\eta\mathbf{\bar{x}}^{n+1} +(1-\eta) \mathbf{x}^{n},
% \end{align}
% %where $\eta$ is a scalar for adjusting the output and is set to 1 in classical ISTA. 
% Proposition~2 in~\cite{Wu2020Sparse} reveals that $\eta>1$ is a better choice than classical ISTA ($\eta=1$). Then this conclusion is empirically extended to LISTA-CP and the scalar $\eta$ is replaced by the overshoot gate function $o_t(\cdot,\cdot|\Lambda_o^n)$ that produces a vector output using a set of learnable parameters $\Lambda_o^n$ in the $n$th iteration for better reconstruction performance. 
\begin{align}\label{e2_4.4}
\mathbf{\bar{x}}^{n+1}&= \mathcal{S}_{\theta^{n}}\left(\mathbf{x}^{n}+(\mathbf{W}^{n})^{T}(\mathbf{b}-\mathbf{Ax}^{n})\right), \nonumber \\
\mathbf{x}^{n+1} &= o_t(\mathbf{x}^{n}, \mathbf{b}|\Lambda_o^n)\odot\mathbf{\bar{x}}^{n+1} +(1-o_t(\mathbf{x}^{n}, \mathbf{b}|\Lambda_o^n)) \odot \mathbf{x}^{n},
\end{align}
where $o_t(\cdot,\cdot|\Lambda_o^n)$ represents the overshoot gate function that produces a vector output using a set of learnable parameters $\Lambda_o^n$ in the $n$th iteration.

In this section, by incorporating free-form DNNs into Gated ISTA with gain gates, we formulate the $n$th iteration of HGLISTA as \begin{equation}\label{HGLISTA}
\begin{aligned}
&\mathbf{v}^{n}=\mathcal{S}_{\theta_{1}^{n}}\left(\Delta_{g^n}\mathbf{x}^{n}+(\overline{\mathbf{W}}^{n})^{T}(\mathbf{b}-\mathbf{A}\Delta_{g^n}\mathbf{x}^{n})\right), \\
&\mathbf{u}^{n}=N_{\mathcal{W}^{n}}(\mathbf{v}^{n}), \\
&\mathbf{w}^{n}=\mathcal{S}_{\theta_{2}^{n}}\left(\Delta_{g^n}\mathbf{u}^{n}+(\widehat{\mathbf{W}}^{n})^{T}(\mathbf{b}-\mathbf{A}\Delta_{g^n}\mathbf{u}^{n})\right),\\
&\mathbf{x}^{n+1}=\alpha^{n}\mathbf{v}^{n}+(1-\alpha^{n})\mathbf{w}^{n},
\end{aligned}
\end{equation}
where $\Delta_{g^n}\mathbf{x}^{n}= g_t(\mathbf{v}^{n-1}, \mathbf{w}^{n-1}, \mathbf{b}|\Lambda_{g}^n)\odot\mathbf{x}^{n}$, and $\Delta_{g^n}\mathbf{u}^{n}= g_t(\mathbf{v}^{n-1}, \mathbf{w}^{n-1}, \mathbf{b}|\Lambda_{g}^n)\odot\mathbf{u}^{n}$. We do not utilize the gain gates in the first iteration to generate $\mathbf{x}^{1}$. Thus, the first iteration is the same as HLISTA-CP, and the parameters are determined as Eq.~\eqref{condition}. The parameters to be trained with gain gates are $\Theta'''=\{\theta^n_{1}, \theta^n_{2}, \overline{\mathbf{W}}^{n}, \widehat{\mathbf{W}}^{n}, \mathcal{W}^{n}, \alpha^{n}\}_{n\in \mathbb{N}}\cup \{\Lambda_g^n\}_{n\in \mathbb{N}_+}$. 
%Thus, the parameters to be trained with gain gates are as follows, \emph{i.e.}, $\Theta'''=\{\theta^n_{1}, \theta^n_{2}, \overline{\mathbf{W}}^{n}, \widehat{\mathbf{W}}^{n}, \mathcal{W}^{n}, \alpha^{n}\}_{n\in \mathbb{N}}\cup \{\Lambda_g^n\}_{n\in \mathbb{N}_+}$. 
%If the overshoot gates are also introduced, $\{\Lambda_o^n\}_{n\in \mathbb{N}_+}$ need to be added to the learnable parameters.
% The reason is that we need to adjust the gain gates using the information generated in the last iteration.

Subsequently, we find the range of the $i$th element of $g_t(\mathbf{v}^{n},\mathbf{w}^{n},  \mathbf{b}|\Lambda_{g}^{n+1})$ for $i\in(\mathbb{S}\cap\mathrm{supp}(\mathbf{v}^n))\cup(\mathbb{S}\cap\mathrm{supp}(\mathbf{w}^n))$ to guarantee the convergence, where $\mathrm{supp}(\mathbf{v}^n)$ and $\mathrm{supp}(\mathbf{w}^n)$ represent the support of $\mathbf{v}^n$ and $\mathbf{w}^n$, respectively. 
Similar to Gated LISTA, we define
\begin{equation}
g_t(\mathbf{v}^{n},\mathbf{w}^{n},  \mathbf{b}|\Lambda_{g}^{n+1})_i = 1 + \kappa_t(\mathbf{v}^{n},\mathbf{w}^{n},  \mathbf{b}|\Lambda_{g}^{n+1})_i.
\end{equation}
%and
%\begin{align}
%   &\theta_{max}^{n}= \max\{\theta_1^{n},\theta_2^{n}\},~ \theta_{min}^{n}= \min\{\theta_1^{n},\theta_2^{n}\},\nonumber\\
%   &\Xi_{i}^{n}=\max\{|v_i^n|, |w_i^n|\},~
%   \Upsilon_{i}^{n}= \min\{|v_i^n|, |w_i^n|\},
%\end{align}
The range of $\kappa_t(\mathbf{v}^{n}, \mathbf{w}^{n}, \mathbf{b}|\Lambda_{g}^{n+1})_i$ is specified as
\begin{equation}\label{kappa_cond}
    \frac{(1-\varrho^{n})\theta_{max}^{n}}{\Upsilon_{i}^{n}}\leq\kappa_t(\mathbf{v}^{n}, \mathbf{w}^{n}, \mathbf{b}|\Lambda_{g}^{n+1})_i \leq \frac{(1+\varrho^{n})\theta_{min}^{n}}{\Xi_{i}^{n}},
\end{equation}
where $\theta_{max}^{n}= \max\{\theta_1^{n},\theta_2^{n}\}$, $\theta_{min}^{n}= \min\{\theta_1^{n},\theta_2^{n}\}$, $\Xi_{i}^{n}=\max\{|v_i^n|, |w_i^n|\}$, $\Upsilon_{i}^{n}= \min\{|v_i^n|, |w_i^n|\}$, and $\varrho^{n}$ is a constant satisfying that 
\begin{equation}\label{varrho}
\sup_{i\in\mathbb{Q}}\left\{\frac{\theta_{max}^{n}\Xi_{i}^{n}-\theta_{min}^{n}\Upsilon_{i}^{n}}{\theta_{max}^{n}\Xi_{i}^{n}+\theta_{min}^{n}\Upsilon_{i}^{n}}\right\}\leq \varrho^{n} \leq 1,
\end{equation}
where $\mathbb{Q}=(\mathbb{S}\cap\mathrm{supp}(\mathbf{v}^n))\cup(\mathbb{S}\cap\mathrm{supp}(\mathbf{w}^n))$. 
%The bound for $\varrho^{n}$ is to guarantee that Eq.~\eqref{kappa_cond} holds.
Given arbitrary $\mathbf{x}^* \in\mathcal{X}(B_{\mathbf{x}}, \mathbb{S})$ and the gain gate function $g_t$, the parameters $\{\overline{\mathbf{W}}^{n}, \widehat{\mathbf{W}}^{n}, \theta_{1}^{n}, \theta_{2}^{n}\}_{n\in\mathbb{N}}$ are determined by
\begin{align}\label{eqn:hglista}
&\overline{\mathbf{W}}^{n}\in\mathcal{W}_s(\mathbf{A}),\  \widehat{\mathbf{W}}^{n}\in\mathcal{W}_s(\mathbf{A}), \nonumber\\
&\theta_{1}^{n}=\sup_{\mathbf{x}^{*}}\{\mu\|\Delta_{g^n}\mathbf{x}^{n}-\mathbf{x}^{*}\|_{1}\}, \\
&\theta_{2}^{n} = \sup_{\mathbf{x}^{*}}\{\mu\|\Delta_{g^n}\mathbf{u}^{n}-\mathbf{x}^{*}\|_{1}\}+\frac{\mu(N-|\mathbb{S}|)}{|\mathbb{S}|-1}\|\Delta_{g^n}\mathbf{u}^{n}\|_{1}.\nonumber
\end{align}
\begin{theorem}[Upper Bound of Recovery Error for HGLISTA]\label{theorem7}
Let $\{\mathbf{x}^{n}\}_{n\in\mathbb{N}}$ be the sequence generated by Eq.~\eqref{HGLISTA} using the parameters $\{\overline{\mathbf{W}}^{n}, \widehat{\mathbf{W}}^{n}, \theta_{1}^{n}, \theta_{2}^{n}\}_{n\in\mathbb{N}_+}$ determined by Eq.~\eqref{eqn:hglista},
%\begin{align}
%&\overline{\mathbf{W}}^{n}\in\mathcal{W}_s(\mathbf{A}),\  \widehat{\mathbf{W}}^{n}\in\mathcal{W}_s(\mathbf{A}), \nonumber\\
%&\theta_{1}^{n}=\sup_{\mathbf{x}^{*}}\{\mu\|\Delta_{g^n}\mathbf{x}^{n}-\mathbf{x}^{*}\|_{1}\}, \\
%&\theta_{2}^{n} = \sup_{\mathbf{x}^{*}}\{\mu\|\Delta_{g^n}\mathbf{u}^{n}-\mathbf{x}^{*}\|_{1}\}+\frac{\mu(N-|\mathbb{S}|)}{|\mathbb{S}|-1}\|\Delta_{g^n}\mathbf{u}^{n}\|_{1},\nonumber
%\end{align}
$\{\alpha^{n}\}_{n\in\mathbb{N}}$ constrained by Eq.~\eqref{new8}, and $\mathbf{x}^{0}=0$. Under Assumption~\ref{assum2} that $B_\mathbf{x}>0$ and $|\mathbb{S}|<(2+1/\mu)/4$, we have for arbitrary $n\in\mathbb{N}$ and $n\geq2$,
\begin{equation}
\begin{aligned}
{\rm supp}(\mathbf{x}^n)\subset \mathbb{S},\ \|\mathbf{x}^{n}-\mathbf{x}^{*}\|_{2}\leq |\mathbb{S}|B_{\mathbf{x}}
\exp{\left(-\sum_{k=0}^{n-2}c^k_{g}-c\right)},
\end{aligned}
\end{equation}
where
\begin{align}
c^k_{g}&=-\log\left(4\mu|\mathbb{S}|-2\mu-2(1-\varrho^{k})\mu s_\mathbf{*}^k\right)>0,
\end{align}
$s_{*}^k = \min\{|\mathrm{supp}(\mathbf{v}^k)|, |\mathrm{supp}(\mathbf{w}^k)|\}$, $\varrho^{k}$ is defined in Eq.~\eqref{varrho}, and $c$ is defined in Eq.~\eqref{c_cp}. When $n=1$, one can refer to the conclusions of HLISTA-CP.
\begin{IEEEproof}
Please refer to Appendix~B.4.
\end{IEEEproof}
\end{theorem}

As overshoot gates are developed empirically (see Appendix~C.1 and \cite{Wu2020Sparse} for more details), we evaluate the performance of HGLISTA with overshoot gates in experiments but do not make theoretical analysis. HGLISTA with both gain and overshoot gates is formulated as 
\begin{align}\label{HGLISTA_both}
&\mathbf{\bar{v}}^{n}=\mathcal{S}_{\theta_{1}^{n}}\left(\Delta_{g^n}\mathbf{x}^{n}+(\overline{\mathbf{W}}^{n})^{T}(\mathbf{b}-\mathbf{A}\Delta_{g^n}\mathbf{x}^{n})\right), \nonumber\\
&\mathbf{v}^{n} = o_t(\mathbf{x}^{n}, \mathbf{b}|\Lambda_{o1}^n)\odot\mathbf{\bar{v}}^{n} +(1-o_t(\mathbf{x}^{n}, \mathbf{b}|\Lambda_{o1}^n)) \odot \mathbf{x}^{n},\nonumber\\
&\mathbf{u}^{n}=N_{\mathcal{W}^{n}}(\mathbf{v}^{n}), \nonumber\\
&\mathbf{\bar{w}}^{n}=\mathcal{S}_{\theta_{2}^{n}}\left(\Delta_{g^n}\mathbf{u}^{n}+(\widehat{\mathbf{W}}^{n})^{T}(\mathbf{b}-\mathbf{A}\Delta_{g^n}\mathbf{u}^{n})\right),\nonumber\\
&\mathbf{w}^{n} = o_t(\mathbf{u}^{n}, \mathbf{b}|\Lambda_{o2}^n)\odot\mathbf{\bar{w}}^{n} +(1-o_t(\mathbf{u}^{n}, \mathbf{b}|\Lambda_{o2}^n)) \odot \mathbf{u}^{n},\nonumber\\
&\mathbf{x}^{n+1}=\alpha^{n}\mathbf{v}^{n}+(1-\alpha^{n})\mathbf{w}^{n}.
\end{align}
In Section~\ref{exper_sec}, $\varrho^{n}$ is initialized as 1 such that the range of $\kappa_t(\mathbf{v}^{n}, \mathbf{w}^{n}, \mathbf{b}|\Lambda_{g}^{n+1})_i$ is extremely similar to that of Gated LISTA (see Eq.~(12) and Eq.~(13) in~\cite{Wu2020Sparse}). Thus, we adopt the gain and overshoot gate functions proposed in~\cite{Wu2020Sparse}. In consideration of the analysis and experimental results in~\cite{Wu2020Sparse}, we utilize the piece-wise linear function and the inverse proportional function as gain gate functions and the sigmoid-based function as overshoot gate function. The formulations are as follows for $n\in\mathbb{N}_{+}$,
\begin{equation}
\begin{aligned}
&\kappa_t(\mathbf{v}^{n-1}, \mathbf{w}^{n-1}, \mathbf{b}|\Lambda_{g}^{n})=\xi_1^n \theta_{min}^{n}\mathrm{ReLU}(1-\mathrm{ReLU}(\xi_2^n\mathbf{\Xi}^{n})), \\
&\kappa_t(\mathbf{v}^{n-1}, \mathbf{w}^{n-1}, \mathbf{b}|\Lambda_{g}^{n})=\xi_1^n \theta_{min}^{n}/ (\xi_2^n\mathbf{\Xi}^{n}+0.001),\\
&o_t(\mathbf{x}^{n}, \mathbf{b}|\Lambda_{o1}^n)=1+a_{01}\sigma(W_{o1}\mathbf{x}^{n}+U_{o1}\mathbf{b})\odot|(\overline{\mathbf{W}}^{n})^{T}\mathbf{b}|,\\
&o_t(\mathbf{u}^{n}, \mathbf{b}|\Lambda_{o2}^n)=1+a_{02}\sigma(W_{o2}\mathbf{u}^{n}+U_{o2}\mathbf{b})\odot|(\widehat{\mathbf{W}}^{n})^{T}\mathbf{b}|,
\end{aligned}
\end{equation}
where $\{\xi_1^n, \xi_2^n, a^n_{01}, a^n_{02}\}_{n\in\mathbb{N}_+}$ are learnable parameters, $W_{o1}, W_{o2}\in \mathbb{R}^{N\times N}$ and $U_{o1}, U_{o2}\in \mathbb{R}^{N\times M}$ are learnable matrices independent of index $n$, $\sigma(\cdot)$ is the sigmoid function, $\mathbf{\Xi}^{n}=\max\{|\mathbf{v}^n|, |\mathbf{w}^n|\}$, and the first and second functions are piece-wise linear and inverse proportional functions, respectively. Following the empirical studies in~\cite{Wu2020Sparse}, we adopt the piece-wise linear function in earlier iterations and use the inverse proportional function better in latter iterations. 
Please refer to \cite{Wu2020Sparse} for more details. 

In Appendix~C.1, we further elaborate the difference between the proposed hybrid algorithm and the gate mechanisms, as the gain gates and overshoot gates seem to be similar to the free-form DNNs and the balancing parameter $\alpha^n$, respectively. We thoroughly distinguish the two methods from the perspectives of motivation, formulations, and theoretical analysis. 

\subsection{HELISTA}
Finally, we extend the hybrid algorithm to ELISTA~\cite{li2021learned} that leverages extragradient in LISTA models. 
%The extragradient method was first developed in~\cite{korpelevich1976extragradient} for variational inequality problems, and was first utilized in optimization problems in~\cite{nguyen2018extragradient} to improve some first-order descent methods. 
Extragradient~\cite{korpelevich1976extragradient} was first utilized in optimization problems in~\cite{nguyen2018extragradient} to improve first-order descent methods. In the $n$th iteration of extragradient method, an intermediate result $\mathbf{x}^{n+\frac{1}{2}}$ is first obtained via the update of $\mathbf{x}^{n}$ with the gradient at $\mathbf{x}^{n}$, then $\mathbf{x}^{n+1}$ is obtained via another update of $\mathbf{x}^{n}$ with the gradient at $\mathbf{x}^{n+\frac{1}{2}}$. ELISTA presents a multistage-thresholding operator $M_{\theta, \widehat{\theta}}$ to substitute the soft-thresholding operator $S_{\theta}$ in ALISTA.
%ELISTA introduces this idea to ALISTA and presents a multistage-thresholding operator $M_{\theta, \widehat{\theta}}$ to replace soft-thresholding operator $S_{\theta}$, where $M_{\theta, \widehat{\theta}}$ is defined as
\begin{equation}
  M_{\theta, \widehat{\theta}}(x)=\left\{
  \begin{aligned}
  0 & , & 0\leq|x|<\theta, \\
  \frac{\widehat{\theta}}{\widehat{\theta}-\theta}{\rm sgn}(x)(|x|-\theta) & , & \theta\leq|x|<\widehat{\theta}, \\
  x & , & |x|\geq\widehat{\theta}.
  \end{aligned}
  \right.
\end{equation}
The $n$th iteration of ELISTA is formulated as follows.
\begin{equation}
\begin{aligned}
  &\mathbf{x}^{n+\frac{1}{2}}=M_{\theta_1^n, \widehat{\theta}_1^n}\left(\mathbf{x}^{n}+\gamma_1^n(\mathbf{W})^{T}(\mathbf{b}-\mathbf{Ax}^{n})\right), \\
  &\mathbf{x}^{n+1}=M_{\theta_2^n, \widehat{\theta}_2^n}\left(\mathbf{x}^{n}+\gamma_2^n(\mathbf{W})^{T}(\mathbf{b}-\mathbf{Ax}^{n+\frac{1}{2}})\right).
\end{aligned}
\end{equation}

HELISTA incorporates free-form DNNs into Eq.~\eqref{HELISTA}. The $n$th iteration is formulated as
\begin{equation}\label{HELISTA}
\begin{aligned}
\mathbf{v}^{n}&=M_{\theta_{1}^{n}, \hat{\theta}_{1}^{n}}\left(\mathbf{x}^{n}+\gamma^n_1(\mathbf{W})^{T}(\mathbf{b}-\mathbf{Ax}^{n})\right), \\
\mathbf{v}^{n+\frac{1}{2}}&=M_{\theta_{2}^{n}, \hat{\theta}_{2}^{n}}\left(\mathbf{x}^{n}+\gamma^n_2(\mathbf{W})^{T}(\mathbf{b}-\mathbf{Av}^{n})\right), \\
\mathbf{u}^{n}&=N_{\mathcal{W}^{n}}(\mathbf{v}^{n+\frac{1}{2}}), \\
\mathbf{w}^{n}&=M_{\theta_{3}^{n}, \hat{\theta}_{3}^{n}}\left(\mathbf{u}^{n}+\gamma^n_3(\mathbf{W})^{T}(\mathbf{b}-\mathbf{A}\mathbf{u}^{n})\right),\\
\mathbf{w}^{n+\frac{1}{2}}&=M_{\theta_{4}^{n}, \hat{\theta}_{4}^{n}}\left(\mathbf{u}^{n}+\gamma^n_4(\mathbf{W})^{T}(\mathbf{b}-\mathbf{A}\mathbf{w}^{n})\right),\\
\mathbf{x}^{n+1}&=\alpha^{n}\mathbf{v}^{n+\frac{1}{2}}+(1-\alpha^{n})\mathbf{w}^{n+\frac{1}{2}}.
\end{aligned}
\end{equation}
Given arbitrary $\mathbf{x}^* \in \mathcal{X}(B_{\mathbf{x}}, \mathbb{S})$ and $\mathbf{W}\in \mathcal{W}_s(\textbf{A})$, the parameters $\{ \theta_{l}^{n}, \hat{\theta}_{l}^{n}\}_{n\in\mathbb{N}}$ for $l=1,2,3,4$ are determined by
\begin{align}\label{HELISTA_theta}
&\theta_{1}^{n}=\gamma_1^n\sup_{\mathbf{x}^{*}}\{\mu\|\mathbf{x}^{n}-\mathbf{x}^{*}\|_{1}\}, \theta_{2}^{n}=\gamma_2^n\sup_{\mathbf{x}^{*}}\{\mu\|\mathbf{v}^{n}-\mathbf{x}^{*}\|_{1}\},\nonumber\\
&\theta_{3}^{n} = \gamma_3^n\sup_{\mathbf{x}^{*}}\{\mu\|\mathbf{u}^{n}-\mathbf{x}^{*}\|_{1}\},\nonumber\\
&\theta_{4}^{n} = \gamma_4^n\sup_{\mathbf{x}^{*}}\{\mu\|\mathbf{w}^{n}-\mathbf{x}^{*}\|_{1}\}+\frac{\gamma_4^n\mu(N-|\mathbb{S}|)}{|\mathbb{S}|-1}\|\mathbf{w}^{n}\|_{1} \nonumber\\
&\qquad~+\frac{\gamma_3^n\gamma_4^n\mu(N-|\mathbb{S}|)}{|\mathbb{S}|-1}\|\mathbf{u}^{n}\|_{1},\nonumber\\
&\hat{\theta}_{l}^{n}=(1+1/\epsilon^n_l)\theta_{l}^{n},~l=1,2,3,4,
\end{align}
where $\epsilon^n_l,~l=1,2,3,4,$ are learnable parameters and greater than 0. $\alpha^n$ is selected to satisfy when $\theta_1^n\not=0$ and  $\theta_2^n\not=0$
\begin{equation}\label{HELISTA_alpha}
\frac{\gamma_4^n\theta_3^n+\theta_4^n}{\gamma_2^n\theta_1^n+\theta_2^n+\gamma_4^n\theta_3^n+\theta_4^n}\leq \alpha^n<1,
\end{equation}
and $\alpha^n=1$ when  $\theta_1^n=0$ and  $\theta_2^n=0$.
Note that ELISTA adopts the similar setting as ALISTA that the weights $\mathbf{W}^n$ are decomposed  into the product of a scalar $\gamma^n$ and a matrix $\mathbf{W}$ obtained by network training, thus the learnable parameters are $\Theta''''=\{\theta^n_{l}, \epsilon^n_l, \gamma^n_l, \mathcal{W}^{n}, \alpha^{n}\}_{n\in \mathbb{N}}\cup\mathbf{W}$ for $l=1,2,3,4$.

\begin{theorem}[Upper Bound of Recovery Error for HELISTA]\label{theorem8}
Let $\{\mathbf{x}^{n}\}_{n\in\mathbb{N}}$ be the sequence generated by Eq.~\eqref{HELISTA} using the parameters $\mathbf{W}\in \mathcal{W}_s(\textbf{A})$, $\{\theta_{l}^{n}, \hat{\theta}_{l}^{n}\}_{n\in\mathbb{N}}$ for $l=1,2,3,4$ determined by Eq.~\eqref{HELISTA_theta}, $\{\alpha^{n}\}_{n\in\mathbb{N}}$ constrained by Eq.~\eqref{HELISTA_alpha}, and $\mathbf{x}^{0}=0$. Under Assumption~\ref{assum2} that $B_\mathbf{x}>0$, we have for arbitrary $n\in\mathbb{N}$,
\begin{equation}
\begin{aligned}
{\rm supp}(\mathbf{x}^n)\subset \mathbb{S}, \  \|\mathbf{x}^{n}-\mathbf{x}^{*}\|_{2}\leq |\mathbb{S}|B_\mathbf{x} \exp\left(-\sum_{k=0}^{n-1}c_{e}^{k}\right),
\end{aligned}
\end{equation}
where
\begin{equation}
\begin{aligned}
c_e^k=&-\log\Bigg[\gamma_1^k\gamma_2^k\Big(1+\mathcal{Q}^k_*\mu+\frac{|1-\gamma_1^k|}{\gamma_1^k}\Big)\\
&\cdot\Big(2\mathcal{Q}^k_*\mu+\frac{\left|1-\gamma_4^k+\gamma_3^k\gamma_4^k\right|}{\gamma_3^k}\Big)+|1-\gamma_2^k+\gamma_1^k\gamma_2^k|\Bigg],
\end{aligned}
\end{equation}
and $\mathcal{Q}^k_*$ is related to $\{\epsilon^n_l\}_{l=1,2,3,4}$, $|\mathbb{S}|$, and the support of $\mathbf{v}^k$ and $\mathbf{w}^k$. One can refer to Appendix~B.5.2.5) for definition of $\mathcal{Q}^k_*$ and Appendix~B.5.2.6) for detailed discussion on $c_e^k$.
\begin{IEEEproof}
Please refer to Appendix~B.5.
\end{IEEEproof}
\end{theorem}

To guarantee that $c_e^k>0$, we thoroughly discuss the values of $\gamma^k_1, \gamma^k_2, \gamma^k_3, \gamma^k_4$ in Appendix~B.5.2.6. In our experiments, we constrain the ranges $0<\gamma^k_1<1, \gamma^k_3>1$, and $0<\gamma^k_4<1$ following the discussion.

\subsection{Relations to HCISTA}\label{sec:4.6}
In Section~\ref{sec:3}, the step size $t^n$ and regularization parameter $\lambda^n$ for HCISTA are constrained by Eq.~\eqref{ls4} and Eq.~\eqref{ls5}, respectively. However, if we relax the constraints and freely apply adaptive $t^n$ and $\lambda^n$, HCISTA can be viewed as a special case of HLISTA-CP. In fact, when we set $\overline{\mathbf{W}}^{n}=\widehat{\mathbf{W}}^n=t^n \mathbf{A}^T$ and $\theta_1^n=\theta_2^n=\lambda^n t^n$ in HLISTA-CP, HCISTA and HLISTA-CP are the same in each iteration according to Eq.~\eqref{e4} and Eq.~\eqref{new3} except for different lower bounds of $\alpha^n$ specified in Eq.~\eqref{add5} and Eq.~\eqref{new8}. In this case, $\alpha^n=1/2$ for HLISTA-CP, as the constraint on $t^n$ is removed.  %\request{$\alpha^n$ is chosen as Eq.~\eqref{new8} such that $\alpha^n=1/2$}.
Thus, HCISTA with adaptive $t^n$ and $\lambda^n$ and a proper $\mathbf{A}$ can attain a linear convergence rate when parameters $t^n \mathbf{A}^T$ and $\lambda^n t^n$ satisfies Eq.~\eqref{condition} and $\alpha^n$ satisfy Eq.~\eqref{new8}. In Section~\ref{6.1.2}, we further evaluate HCISTA-F that enables free $t^n$ and $\lambda^n$ in HCISTA.
%Thus, we further evaluate HCISTA with free $t^n$ and $\lambda^n$, dubbed HCISTA-F, in the experiments, as elaborated in Section~\ref{exper_sec}.

\section{Discussion}\label{sec:5}

\begin{figure}[!t]
\renewcommand{\baselinestretch}{1.0}
\centering
\includegraphics[width=4.5in]{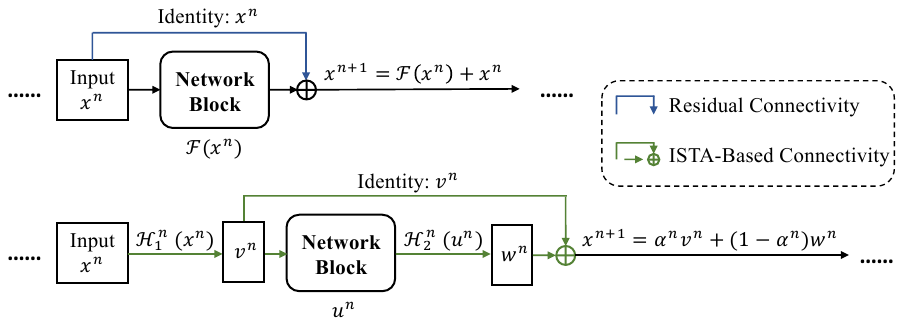}
\caption{Illustration of hybrid ISTA models. Seven models are proposed from the perspective of classical ISTA and learned ISTA.}\label{fig0}
\end{figure}

In this section, we further clarify the free-from DNNs and interpret the properties of hybrid ISTA in the sense of flexibility, convergence, and generality, as summarized in Table~\ref{T1}. Figure~\ref{fig0} illustrates the proposed hybrid ISTA models by unfolding classical ISTA with pre-computed parameters and learned ISTA.

\begin{figure}[!t]
\renewcommand{\baselinestretch}{1.0}
\centering
\includegraphics[width=0.9\textwidth]{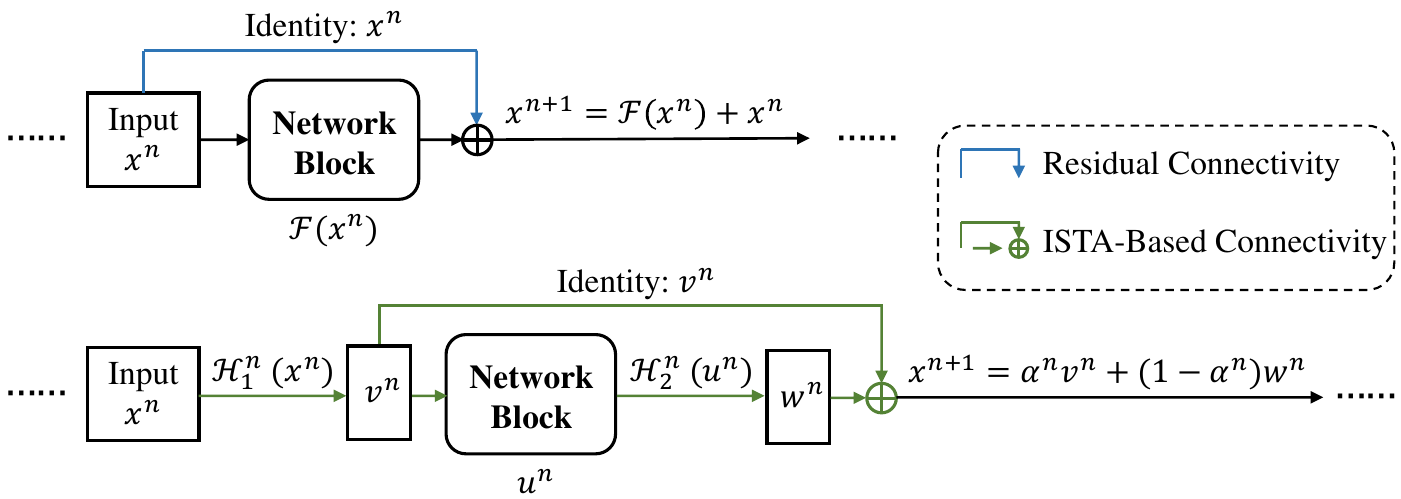}
\caption{Comparison between residual connectivity and proposed ISTA-based connectivity. $\mathcal{H}_{1}^{n}$ and $\mathcal{H}_{2}^{n}$ correspond to the counterparts in Eq.~\eqref{e4}, Eq.~\eqref{new3}, Eq.~\eqref{HLISTA-CPSS} or Eq.~\eqref{HGLISTA}, respectively. Eq.~\eqref{HELISTA} for HELISTA can be regarded as a similar but more complicated connectivity.}\label{fig1}
\end{figure}

\subsection{Flexibility of Hybrid ISTA: Free-form DNNs}
As mentioned above, the term \emph{free-form DNNs} means DNNs with any feasible and reasonable architecture in practice. For example, the architecture is not feasible, if the input and output dimensions of the DNNs are not compatible with the dimension of original signal $\mathbf{x}^*$. The architecture would not be reasonable, if the DNNs contain successive pooling operators, as it is meaningless to practical tasks.
We thoroughly discuss the inserted DNNs in Appendix~C.2.

Despite the slight requirement on the DNN architectures, the proposed hybrid ISTA can support a wide range of feasible components, including fully connected layers, convolutional layers, pooling operations like average pooling and max-pooling, normalization like batch normalization~\cite{ioffe2015batch} and layer normalization~\cite{ba2016layer}, residual connectivity~\cite{he2016deep} and dense connectivity~\cite{huang2017densely}, Transformer~\cite{dosovitskiy2020image}, non-linear activation functions like ReLU and Sigmoid. Although any feasible and reasonable combination of these components can be utilized in the proposed methods, we find in the experiments that proper DNN architectures for specified tasks can effectively boost the performance. For example, convolutional neural networks (CNNs) are particularly effective in processing natural images. 

Moreover, it is efficient to find the best architectures of the inserted DNNs by neural architecture search (NAS)~\cite{zoph2016neural} under the specified computational complexity, which is the specific advantage of our methods in comparison to classical ISTA, LISTA and the variants of LISTA. 

\subsection{Convergence of Hybrid ISTA}
\label{discussion_convergence}
There is a difference in convergence between HCISTA and HLISTA. The learnable parameters $\Theta$ of HCISTA can be randomly selected from the explicitly specified ranges to achieve convergence, even if the model is not trained. This fact suggests that convergence is independent of network training. Therefore, training with sufficient samples only finds proper parameters to improve the performance.
%As the ranges of the learnable parameters $\Theta$ of HCISTA are explicit, we can randomly choose them in the ranges to attain the convergence, even if the model is untrained. Training with sufficient samples is only to choose the proper parameters to improve the performance, \emph{i.e.}, the convergence is independent of network training. 
However, similar to LISTA and its variants, HLISTA models ensure convergence in theory when the parameters such as $\{\overline{\mathbf{W}}^{n}, \widehat{\mathbf{W}}^{n}, \theta_{1}^{n}, \theta_{2}^{n}\}_{n\in\mathbb{N}}$ are ideal but cannot be obtained directly. Thus, we prove that there exists a sequence of parameters that makes HLISTA models converge linearly.
We demonstrate in the experiments that convergence depends on network training. This fact implies that we can learn the parameters in a data-driven manner to achieve the linear convergence rate.
%Fortunately, these parameters leading to linear convergence can be obtained by data-driven learning and the experiments validate this proposition, \emph{i.e.}, the convergence depends on network training. 

\subsection{Generality of Hybrid ISTA}
\label{sec5.3}

We analyze the generality of hybrid ISTA from the views of iterative algorithms and deep neural networks, respectively.

From the perspective of classical ISTA and LISTA, we successfully generalize the hybrid scheme to these algorithms, and consequently introduce free-form DNNs to offer flexibility and efficiency with a guarantee of convergence in theory. The term `flexibility' refers to the procedure of the proposed algorithms, \emph{i.e.}, adopting different DNN architectures results in  different procedures of algorithms. The term `efficiency' refers to the reconstruction performance with the same iterations, \emph{i.e.}, various free-form DNNs bring about tremendous potential for the improvement of reconstruction performance in comparison to the classical ISTA without deep learning technology and LISTA with restricted network architectures. Extensive experiments show that hybrid ISTA can achieve superior performance in comparison to the baselines with much fewer parameters, \emph{e.g.}, HLISTA-CP and HLISTA-CPSS reduce NMSE by about 8 dB and 18 dB on the task of sparse recovery with only 6.5\% learnable parameters when compared with corresponding baselines in Section~\ref{6.1.3}. This validates that the restricted network architectures of LISTA limit the performance and hybrid ISTA can introduce efficient DNNs without violating the theoretical convergence.

From the perspective of DNNs, the hybrid ISTA provides an interesting direction for designing interpretable DNNs for inverse problems. Existing DNNs for inverse problems, \emph{e.g.}, ReconNet \cite{Kulkarni2016ReconNet}, Dr2-Net \cite{Yao2019DR}, and ISTA-Net \cite{zhang2018ista}, can be viewed as a specific version of the free-form DNNs and adopted in hybrid ISTA models. In this case, the proposed method can be viewed as a special ISTA-based connectivity that is similar to residual connectivity~\cite{he2016deep}, as illustrated in \figurename~\ref{fig1}. 
Residual connectivity transfers cross-layer information for improving performance, and can be employed in almost all DNNs. Similarly, ISTA-based connectivity transfers cross-layer information for theoretical guarantees of convergence. 
While those existing DNNs cannot be analyzed in theory, the proposed method provides a way to endow the empirically constructed DNNs with theoretical interpretation and convergence guarantees. Experiments show that hybrid ISTA can further improve the reconstruction performance of these existing DNNs in Section~\ref{sec:6.2.2}.

\subsection{Comparisons between Hybrid ISTA models}
As shown in \figurename~\ref{fig0}, the proposed hybrid ISTA consists of seven different models based on six different baselines, \emph{i.e.}, five for HLISTA, one for HCISTA, and one to bridge HLISTA and HCISTA. The main difference between HCISTA and HLISTA is the convergence, as discussed in Section~\ref{discussion_convergence}. To validate the generality of our framework, we develop five HLISTA models by extending the proposed hybrid framework to five variants of LISTA. 
%For various variants of LISTA, we extend the proposed hybrid framework to five variants to validate the generality of our method. 
It is worth mentioning that the characteristics of the variants of LISTA are reserved in the corresponding HLISTA models, \emph{e.g.}, partial weight coupling structure in LISTA-CP, support selection in LISTA-CPSS, pre-computed $\mathbf{W}$ in ALISTA, gate mechanisms in Gated LISTA and extragradient in ELISTA. Thus, the differences among those HLISTA models are mainly originated from the various technologies adopted in the baselines.

\section{Experiments} \label{exper_sec}
In this section, we perform experiments to validate our theoretical results and evaluate the reconstruction performance in the tasks of sparse recovery and compressive sensing. Following the same setting as existing unfolded iterative algorithms, all the proposed hybrid ISTA models are treated as specially structured neural networks constructed by unfolding $K$ iterations. The parameters (\emph{e.g.}, $\Theta$ for HCISTA, $\Theta'$ for HLISTA-CP/CPSS, $\Theta''$ for HALISTA, $\Theta'''$ for HGLISTA, and $\Theta''''$ for HELISTA) are learned during the phase of network training. For all the models evaluated in this section, we adopt the same stage-wise training strategy as~\cite{NIPS2018_8120, liu2018alista, Wu2020Sparse} using Adam optimizer~\cite{kingma2014adam} with learning rate decaying. Refer to Appendix~E in~\cite{NIPS2018_8120} for a detailed description about the training strategy. All the experiments are implemented using Tensorflow%~\cite{abadi2016tensorflow} 
on a workstation with an Intel Xeon E5-2603 CPU and a GTX 1080Ti GPU.

\begin{table}[!t]
\renewcommand{\baselinestretch}{1.0}
\renewcommand{\arraystretch}{1.0}
\centering
\caption{Summary of learnable parameters of different models. The postfixes `T' and `U' represent the tied and untied models, respectively. }\label{T2}
\begin{tabular}{ll}
\toprule
Models & Learnable Parameters\\
\midrule
ISTA~\cite{blumensath2008iterative} & None \\
HCISTA-UnT & None \\
HCISTA &  $\{\delta^n, t^n, \alpha^n\}_{n=0}^{K}$, $\mathcal{W}$, $\{\lambda^n\}_{n=1}^{K}$ \\
HCISTA-F &  $\{t^n, \alpha^n\}_{n=0}^{K}$, $\mathcal{W}$, $\{\lambda^n\}_{n=1}^{K}$ \\
\hline%\midrule
LISTA-CP-T/CPSS-T~\cite{NIPS2018_8120} & $\{\theta^n\}_{n=0}^{K}$, $\mathbf{W}$ \\
LISTA-CP-U/CPSS-U~\cite{NIPS2018_8120} & $\{\theta^n, \mathbf{W}^n\}_{n=0}^{K}$ \\
HLISTA-CP/CPSS & $\{\theta^n_{1}, \theta^n_{2}, \alpha^{n}\}_{n=0}^{K}$, $\mathbf{W}, \mathcal{W}$\\
\hline%\midrule
ALISTA~\cite{liu2018alista} & $\{\theta^n, \gamma^n\}_{n=0}^{K}$ \\
HALISTA &  $\{\theta^n_{1}, \theta^n_{2}, \alpha^{n}, \gamma_1^n\}_{n=0}^{K}$, $\mathcal{W}$ \\
\midrule
Gated LISTA-T~\cite{Wu2020Sparse} & $\{\theta^n, \Lambda_g^n, \Lambda_{o}^n\}_{n=0}^{K}$, $\mathbf{W}$ \\
Gated LISTA-U~\cite{Wu2020Sparse} & $\{\theta^n, \Lambda_g^n, \Lambda_{o}^n, \mathbf{W}^n\}_{n=0}^{K}$ \\
\multirow{2}*{HGLISTA} &   $\{\theta^n_{1}, \theta^n_{2}, \alpha^{n}\}_{n=0}^{K}$, $\mathbf{W}, \mathcal{W},$ \\
~ & $\{\Lambda_g^n, \Lambda_{o1}^n, \Lambda_{o2}^n\}_{n=1}^{K}$\\
\hline%\midrule
 ELISTA~\cite{li2021learned} &  $\{\theta^n_{l}, \hat{\theta}^n_l, \gamma^n_l\}_{n=0}^{K}$ for $l=1,2$, $\mathbf{W}$ \\
 \multirow{2}*{HELISTA} &   $\{\theta^n_{l}, \epsilon^n_l, \gamma^n_l, \alpha^{n}\}_{n=0}^{K}$ for $l=1,2,3,4,$ \\
~ &  $\mathbf{W}, \mathcal{W}$\\
\bottomrule
\end{tabular}
\end{table}

Table~\ref{T2} elaborates the learnable parameters of different models. For LISTA-CP, LISTA-CPSS and Gated LISTA, we introduce postfixes `T' and `U' to denote two modes of learnable parameters. The postfix `T' denotes the tied model~\cite{liu2018alista} where the matrices $\{\mathbf{W}^n\}_{n=0}^{K}$ in Eq.~\eqref{LISTA} or Eq.~\eqref{LISTA-CPSS} are tied over all the $K$ iterations, \emph{i.e.}, $\mathbf{W}^n = \mathbf{W}^m$ for arbitrary $n,m\in\mathbb{N}$ that are not greater than $K$. Thus, we use $\mathbf{W}$ to represent the weights in the tied model. The postfix `U' stands for the untied model that does not share the weights across different iterations. 
%The postfix `T' denotes tied model that was first proposed in~\cite{liu2018alista}, which means that the matrix $\mathbf{W}^n$ in~\eqref{LISTA} or \eqref{LISTA-CPSS} are tied over all the iterations, \emph{i.e.}, $\mathbf{W}^n = \mathbf{W}^m$ for arbitrary $n,m\in\mathbb{N}$. Thus, we use $\mathbf{W}$ to represent the value. The postfix `U' represents untied model that no weights are shared.
In the experiments, we limit the number of learnable parameters in the proposed hybrid ISTA models. For example, as shown in Table~\ref{T3}, the proposed HLISTA-CP/CPSS have approximately equivalent learnable parameters to LISTA-CP-T/CPSS-T and require significantly less parameters than LISTA-CP-U/CPSS-U. Specifically, we reuse the weights of incorporated DNNs $\{\mathcal{W}^n\}_{n=0}^{K}$ for all the $K$ iterations and denote them by $\mathcal{W}$, \emph{i.e.}, $\mathcal{W}^n=\mathcal{W}^m=\mathcal{W}$ for arbitrary $n,m\in\mathbb{N}$ with $0\leq n,m\leq K$. For HLISTA-CP, HLISTA-CPSS and HGLISTA, the weights $\{\overline{\mathbf{W}}^n\}_{n=0}^{K}$ and $\{\widehat{\mathbf{W}}^n\}_{n=0}^{K}$ share the same values $\mathbf{W}$ in analogy to the tied model for LISTA-CP and LISTA-CPSS. 
%To show the superiority of all the proposed hybrid ISTA models, we limit the number of learned parameters. For those large-size weights such as $\overline{\mathbf{W}}^n$, $\widehat{\mathbf{W}}^n$, and $\mathcal{W}^n$, we set $\overline{\mathbf{W}}^n = \widehat{\mathbf{W}}^n$ and reuse the weights in different iterations, which is analogous to the tied model. Then, we have for arbitrary $n,m\in\mathbb{N}$ that satisfy $0\leq n,m\leq K$,
For arbitrary $n,m\in\mathbb{N}$ that satisfy $0\leq n,m\leq K$,
\begin{equation}
\overline{\mathbf{W}}^n=\widehat{\mathbf{W}}^n=\overline{\mathbf{W}}^{m}=\widehat{\mathbf{W}}^{m}=\mathbf{W}. %\mathcal{W}^n=\mathcal{W}^m.
\end{equation}
%Thus, we denote $\{\overline{\mathbf{W}}^n\}_{n=0}^{K}$ and $\{\widehat{\mathbf{W}}^n\}_{n=0}^{K}$ by $\mathbf{W}$ and $\{\mathcal{W}^n\}_{n=0}^{K}$ by $\mathcal{W}$.
% Thus, we use $W$ to represent the value. 
% The postfix `U' means untied model.
% The postfix `US' denotes untied model with $\overline{W}^n=\widehat{W}^n$ (resp. $\gamma_1^n=\gamma_2^n$) in each iteration, but they do not share weights over all the iterations. Thus, we denote the weights by $W^n$ (resp. $\gamma^n$). The postfix `UD' denotes untied model that no weights are shared, which means that there are the most learned parameters. 

Here, Gated LISTA and HGLISTA are constructed with both gain and overshoot gates.
For HCISTA, the postfixes `UnT' and `F' represent the untrained model with random step sizes for validating Theorems~\ref{theorem1}$-$\ref{theorem3} and the trained model without constraining the step sizes $\{t^n\}_{n=0}^{K}$ and regularization parameters $\{\lambda^n\}_{n=0}^{K}$ in Section~\ref{sec:4.6}, respectively.

\begin{figure*}[!t]
\renewcommand{\baselinestretch}{1.0}
\centering
\subfigure[Summary of NMSEs]{
\includegraphics[width=0.30\textwidth]{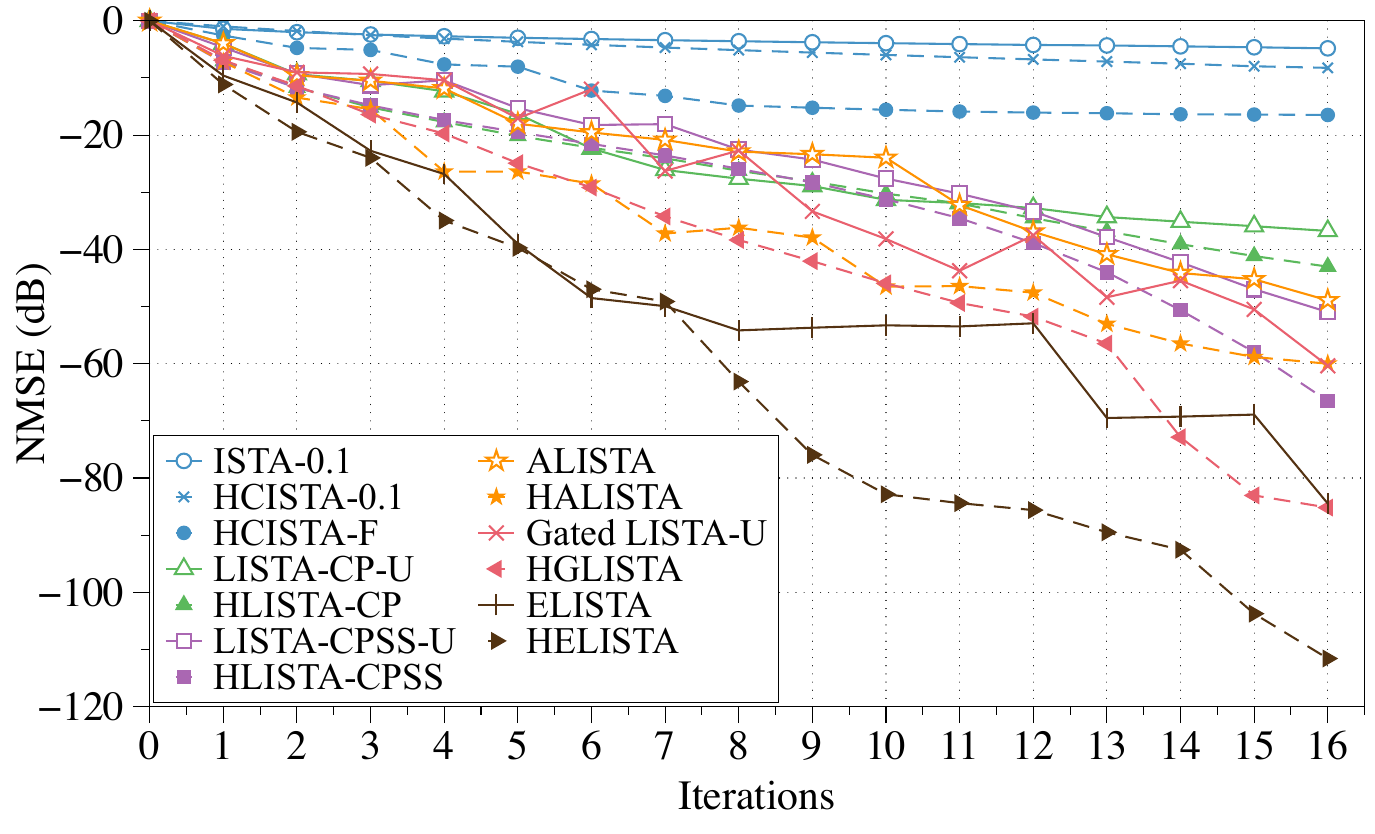}\label{fig2a}}
\quad
\subfigure[ISTA vs. HCISTA-UnT (600 iterations)]{
\includegraphics[width=0.30\textwidth]{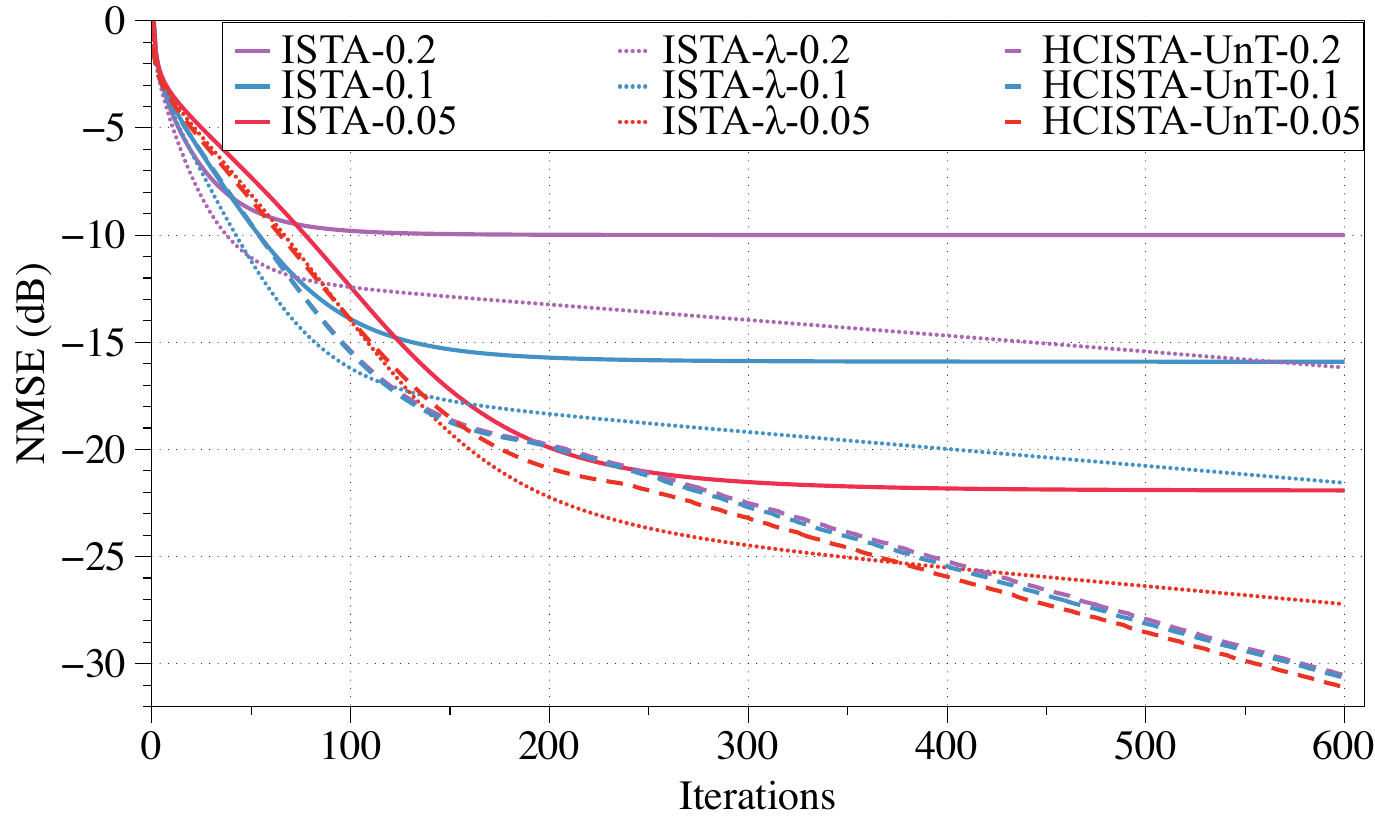}\label{fig2b}}
\quad
\subfigure[FISTA $\&$ ADMM vs. HCISTA-UnT]{
\includegraphics[width=0.30\textwidth]{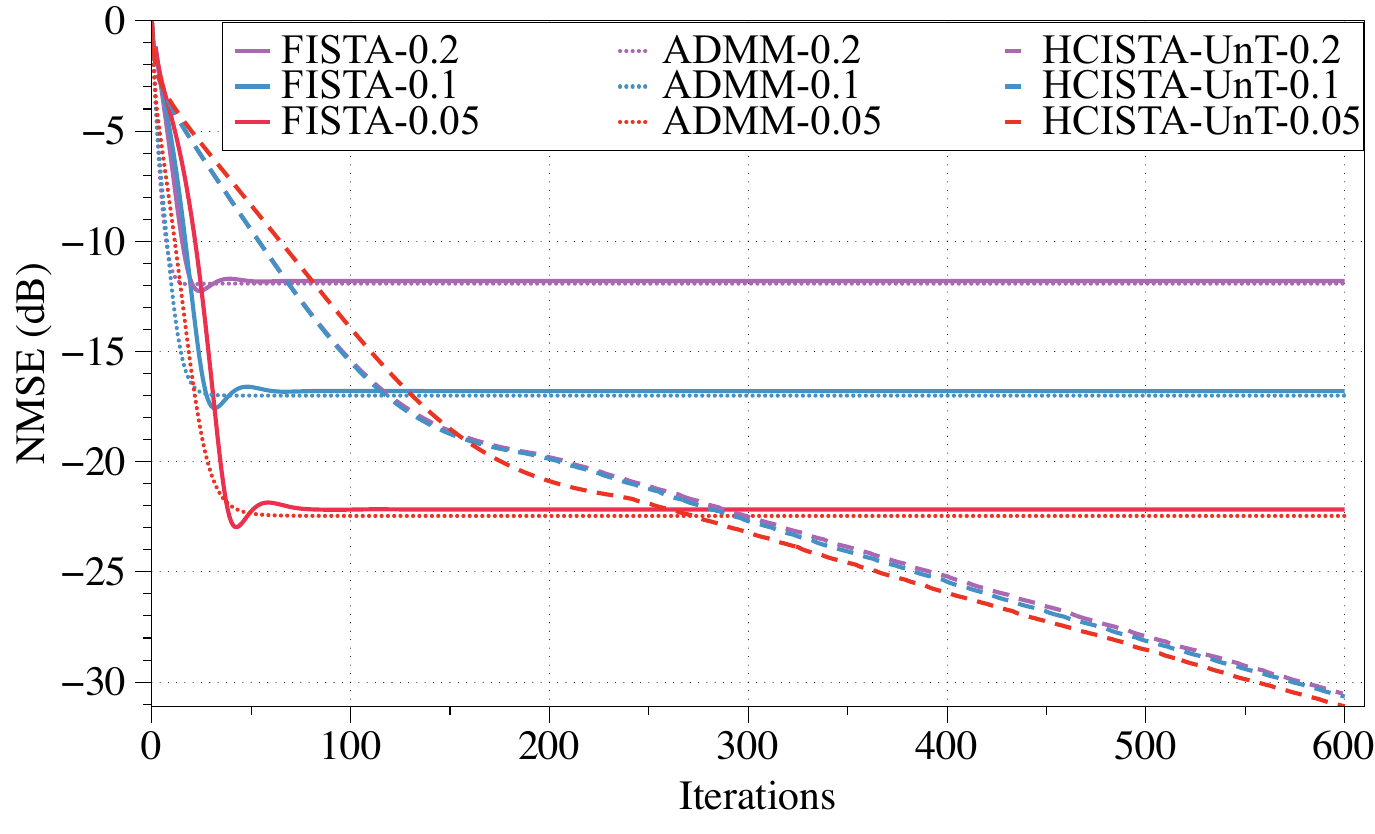}\label{fig2b_add}} \\
\subfigure[ISTA vs. HCISTA]{
\includegraphics[width=0.30\textwidth]{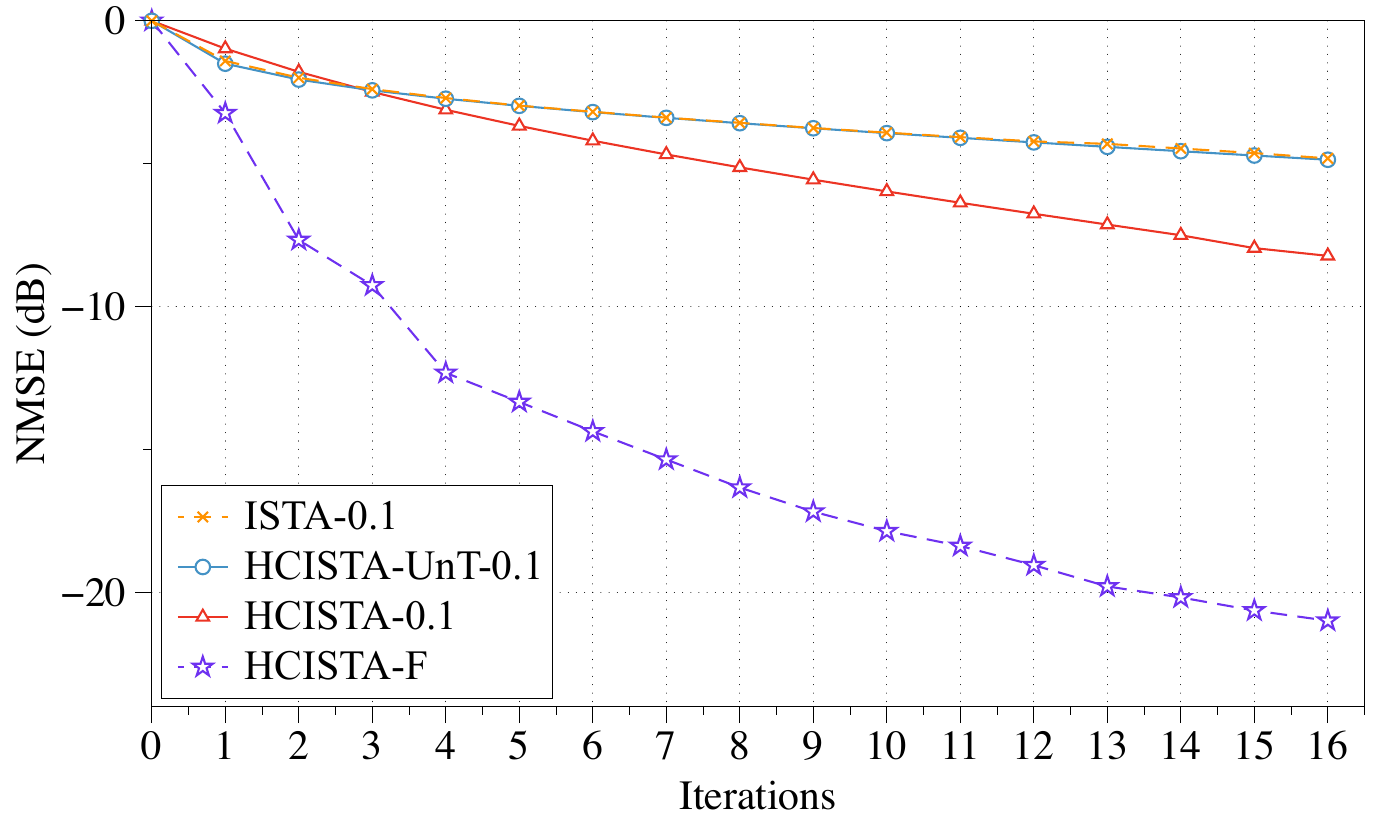}\label{fig2c}}
\quad
\subfigure[LISTA-CP vs. HLISTA-CP]{
\includegraphics[width=0.30\textwidth]{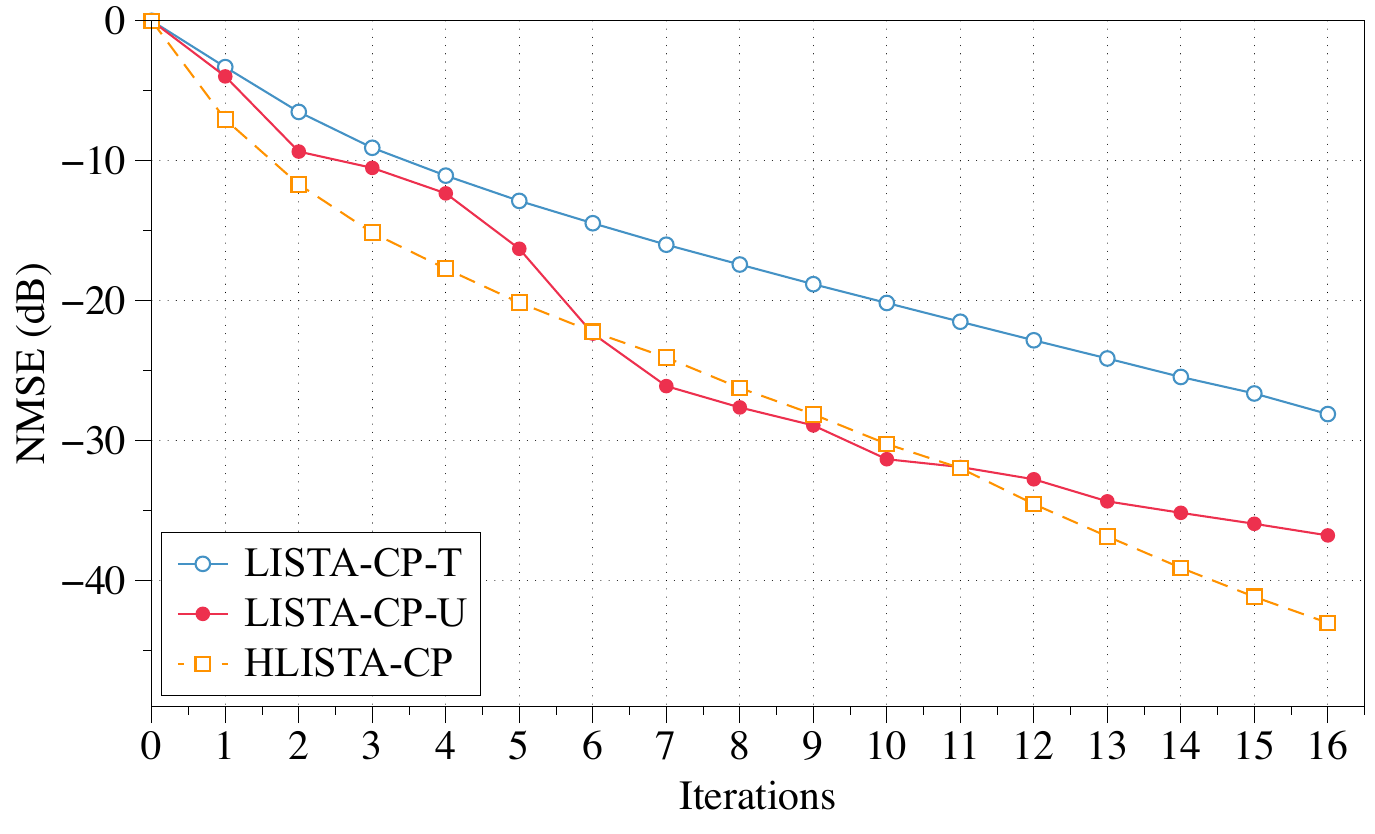}\label{fig2d}}
\quad
\subfigure[LISTA-CPSS vs. HLISTA-CPSS]{
\includegraphics[width=0.30\textwidth]{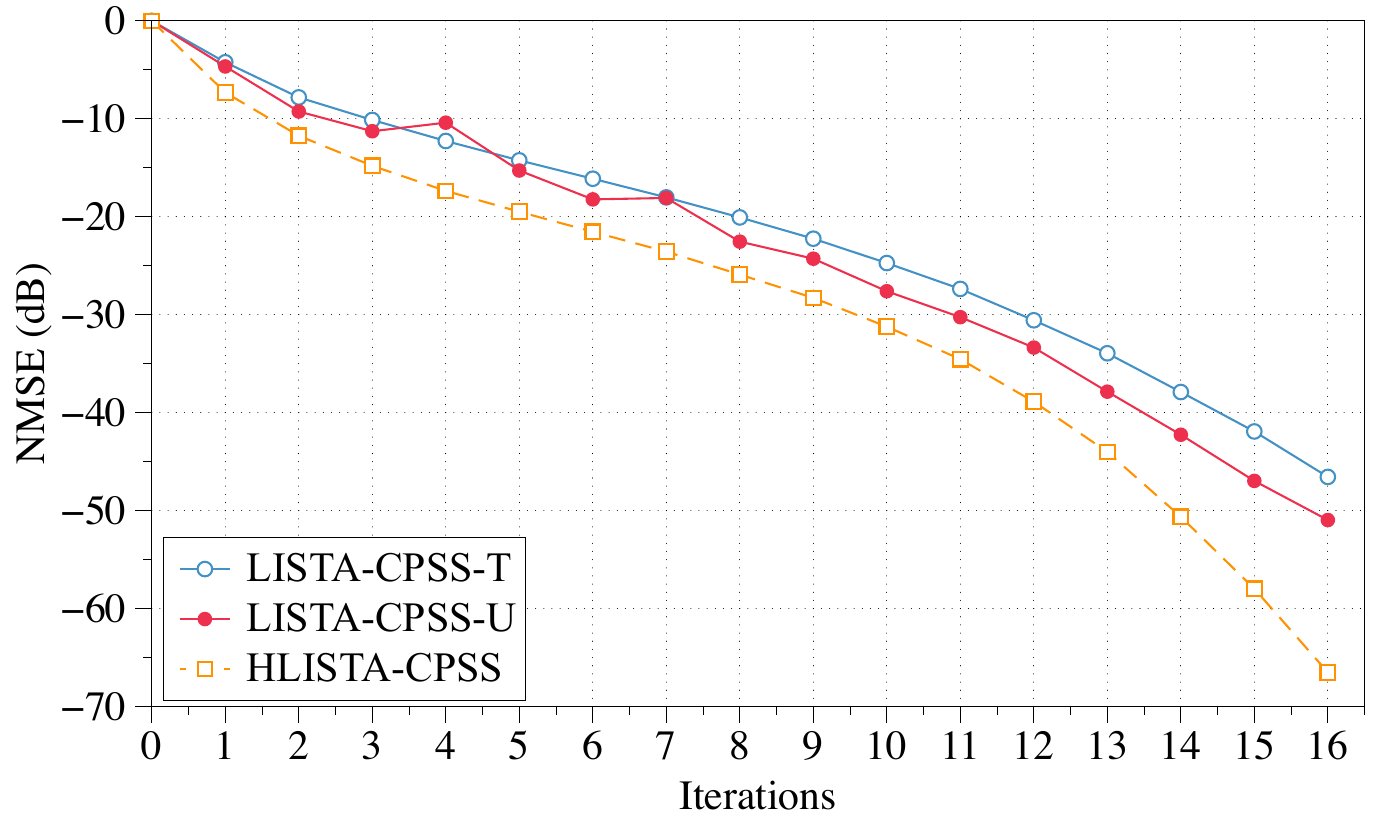}\label{fig2e}}\\
\subfigure[ALISTA vs. HALISTA]{
\includegraphics[width=0.30\textwidth]{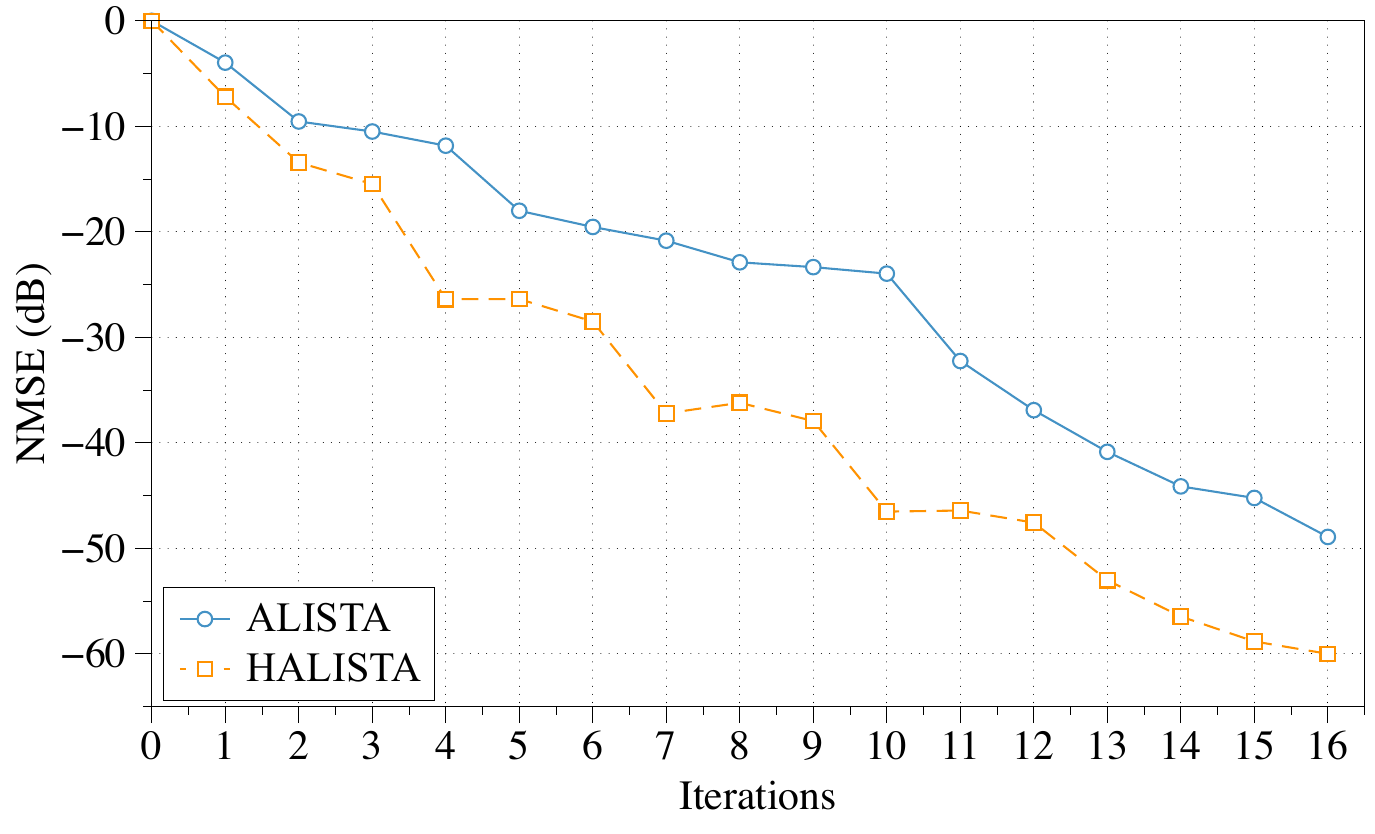}\label{fig2f}}
\quad
\subfigure[Gated LISTA vs. HGLISTA]{
\includegraphics[width=0.30\textwidth]{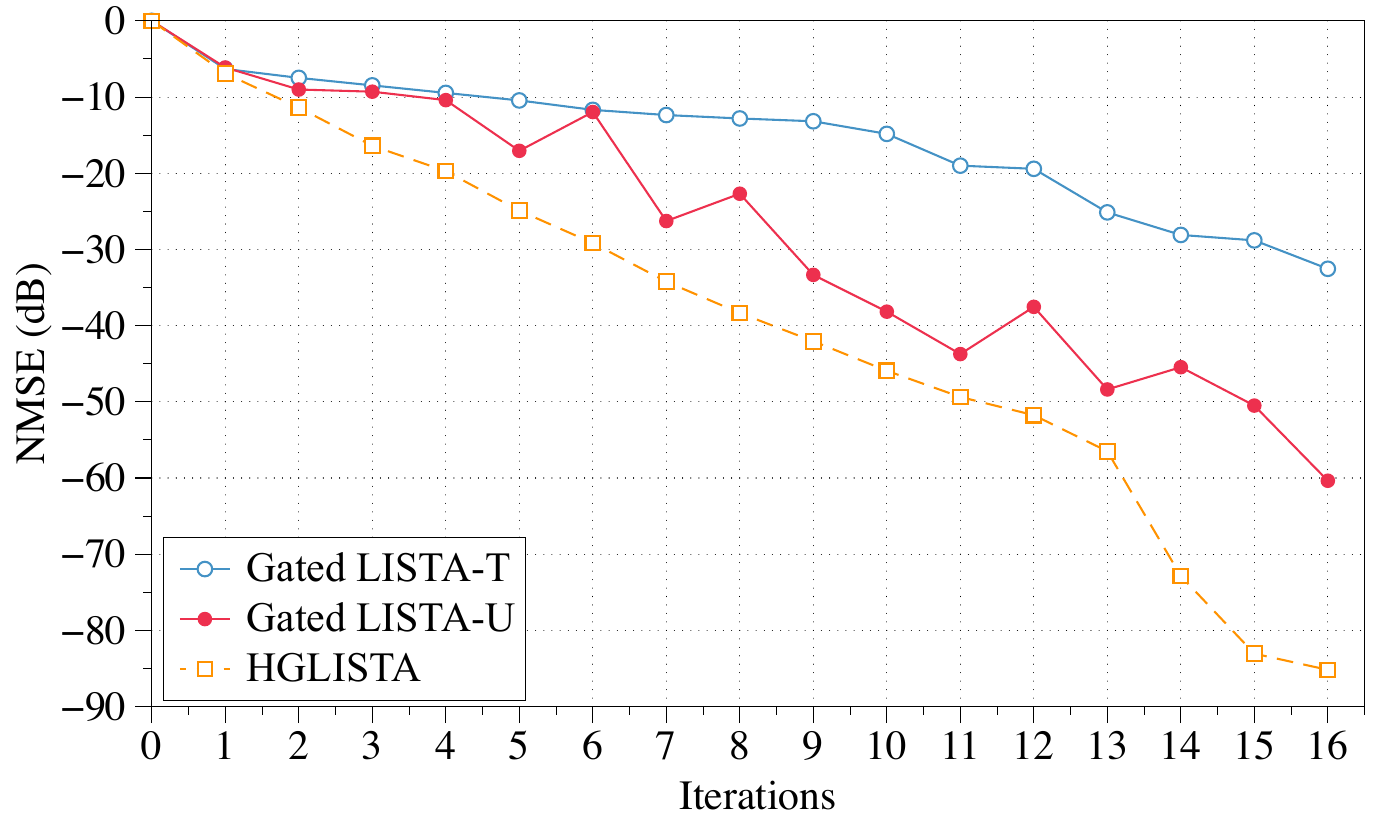}\label{fig2g}}
\quad
\subfigure[ELISTA vs. HELISTA]{
\includegraphics[width=0.30\textwidth]{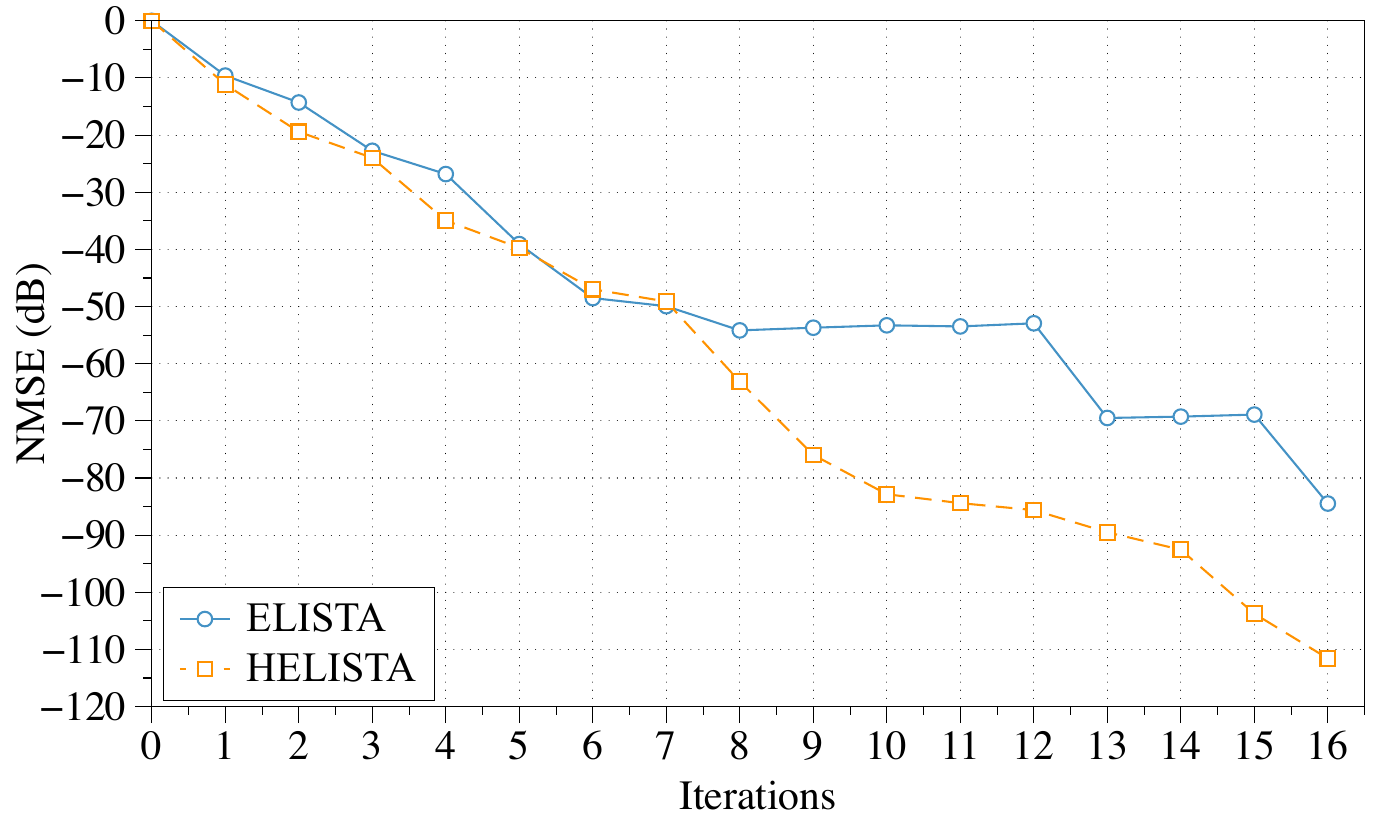}\label{fig2h}}
\caption{NMSEs with respect to iterations obtained by  ISTA~\cite{blumensath2008iterative}, FISTA~\cite{doi:10.1137/080716542},
ADMM~\cite{admm2010}, LISTA-CP/CPSS~\cite{NIPS2018_8120}, ALISTA~\cite{liu2018alista}, Gated LISTA~\cite{Wu2020Sparse}, ELISTA~\cite{li2021learned}, and the proposed hybrid ISTA models HCISTA, HCISTA-UnT, HLISTA-CP/CPSS, HALISTA, HGLISTA and HELISTA on the test set of 1000 samples randomly generated from ${\rm Ber}(0.1)\circ\mathcal{N}(0,1)$. ISTA with fixed $\lambda=0.05$, $0.1$, and $0.2$ and HCISTA and ISTA-$\lambda$ with $\lambda^0=0.05$, $0.1$, and $0.2$ and adaptive $\lambda^n$, $n=1,\cdots,K$ are evaluated. 
%(a): A summary NMSE of models.  (b): A comparison between ISTA and untrained HCISTA with different $\lambda$'s in 600 iterations. (c): A comparison between ISTA, HCISTA-UnT, HCISTA and HCISTA-F. (d), (e), (f): The specific comparisons between the baselines and the corresponding hybrid models.
}\label{fig2}
\end{figure*}

\subsection{Numerical Simulations on Sparse Recovery}\label{sec:6.1}
We follow the same experimental setting as \cite{NIPS2018_8120,liu2018alista,Wu2020Sparse, li2021learned}. Specifically, we choose $M=250$ and $N=500$. The entries of $\mathbf{A}$ are sampled from i.i.d. Gaussian distributions, namely $A_{i,j}\sim\mathcal{N}(0, 1/M)$. The columns of $\mathbf{A}$ are normalized to have the unit $\ell_{2}$-norm. $\mathbf{A}$ is fixed and shared by all the methods. To generate the sparse signal $\mathbf{x}^{*}$, we utilize a Bernoulli sampling operator (with probability 0.1) ${\rm Ber}(0.1)$ to randomly select values from the standard Gaussian distribution, \emph{i.e.}, $\mathbf{x}^{*}={\rm Ber}(0.1)\circ\mathcal{N}(0,1)$. A test set of 1000 samples is generated by fixing $\mathbf{x}^{*}$ in our simulations. When $K$ iterations are unrolled, we train the learnable parameters by minimizing the squared loss between the output of $K$th iteration $\mathbf{x}^K$ and $\mathbf{x}^*$.
%When the models are constructed by unfolding $K$ iterations, we train the learnable parameters by minimizing the squared loss between the output of $K$th iteration $\mathbf{x}^K$ and $\mathbf{x}^*$.
\begin{equation}\label{loss_function}
\min_{\Theta}\mathbb{E}[\|\mathbf{x}^{K}-\mathbf{x}^{*}\|_{2}^{2}].
\end{equation}

The number of iterations $K$ is set to 16 for all the networks. %We compare the models listed in Table~\ref{T2}.
% HLISTA-Slim means that we force $u^{n}=v^{n}$ in HLISTA and thus there are no free-form DNNs. 
We first introduce DNNs with simple architectures.
For the $K$ iterations, the DNNs incorporated in HCISTA and HLISTA share the same architecture of three one-dimensional convolutional layers with the sizes of 9$\times$1$\times $16, 9$\times$16$\times$16, and 9$\times$16$\times$1 (Kernel\_size$\times$In\_channel$\times$Out\_channel) and ReLU following the first two convolutional layers, \emph{i.e.}, the architecture of \texttt{Conv-ReLU-Conv-ReLU-Conv}. We denote this architecture by ${\rm CvRL}_3$.
%ReLU is utilized after the first two layers in each iteration. 
For $n=0,\cdots,K$, $\mathbf{v}^{n}$ and $\mathbf{u}^{n}$ are connected with a shortcut. Specifically, we have
\begin{equation}\label{un_sc}
\mathbf{u}^{n}=N_{\mathcal{W}^{n}}(\mathbf{v}^{n})=\mathbf{v}^{n}+{\rm CvRL}_3(\mathbf{v}^{n}),
\end{equation}
%where ${\rm CvRL}_3$ denotes the aforementioned convolution and ReLU operations. 
The weights of DNNs are initialized as orthogonal matrices. The results are evaluated in terms of normalized mean squared error (NMSE) in dB.
\begin{equation}
{\rm NMSE} = 10 \log_{10}(\|\mathbf{x}^{K}-\mathbf{x}^{*}\|^{2}/\|\mathbf{x}^{*}\|^{2})
\end{equation}

In Appendix~D, we further evaluate complicated DNNs and effective tricks in the task of sparse recovery, including dense connectivity~\cite{huang2017densely}, Vision Transformer~\cite{dosovitskiy2020image}, U-net~\cite{ronneberger2015u}, fully-connected layers, average pooling, batch and layer normalization. Comparisons on sparse recovery with additional conventional algorithms and unfolded models in the same experimental setups can be found in~\cite{NIPS2018_8120,liu2018alista,Wu2020Sparse, li2021learned}.
%Some complicated DNN architectures and effective tricks are adopted in the experiments including dense connectivity, Vision Transformer, U-net~\cite{ronneberger2015u}, fully-connected layers, average pooling, batch and layer normalization, and so on. Due to the limited pages, we only show results with simple DNNs here and one can refer to Appendix~D for the results with complicated DNNs. In addition, one can refer to~\cite{NIPS2018_8120,liu2018alista,Wu2020Sparse, li2021learned} for more comparisons on sparse recovery with other conventional algorithms and unfolded models in the same experimental setups.

\begin{table}[!t]
\renewcommand{\baselinestretch}{1.0}
\renewcommand{\arraystretch}{1.0}
\centering
\caption{Comparison of number of learnable parameters in sparse recovery experiments.}\label{T3}
\begin{tabular}{ll}
\toprule
Models & Number of Learnable Parameters\\
\midrule
ISTA-$0.1$~\cite{blumensath2008iterative} & None \\
HCISTA-$0.1$ &  $3\times16+2592+15=2655$ \\
HCISTA-F &  $2\times16+2592+15=2639$ \\
\hline%\midrule
LISTA-CP-T/CPSS-T~\cite{NIPS2018_8120} & $125000+16=125016$ \\
LISTA-CP-U/CPSS-U~\cite{NIPS2018_8120} & $(125000+1)\times 16=2000016$ \\
HLISTA-CP/CPSS & $3\times16+125000+2592=127640$\\
\hline%\midrule
ALISTA~\cite{liu2018alista} & $2\times 16 = 32$ \\
HALISTA &  $4\times 16 + 2592=2656$ \\
\hline%\midrule
Gated LISTA-T~\cite{Wu2020Sparse} &  $4\times 16 + 375000 + 125000=500064$ \\
Gated LISTA-U~\cite{Wu2020Sparse} & $125004\times 16 + 375000=2375064$ \\
 \multirow{2}*{HGLISTA} &   $3\times 16 +125000 +2592$ \\
~ &  $+375000\times 2+4\times 15=877700$\\
\hline%\midrule
ELISTA~\cite{li2021learned} &  $6\times 16+125000=125096$ \\
HELISTA &   $16\times 16+125000+2592=127848$ \\
\bottomrule
\end{tabular}
\end{table}
\subsubsection{Summary Results of Baselines and Hybrid ISTA}\label{6.1.1}
%We summarily report the test-set NMSE of baselines and hybrid ISTA models, as in \figurename~\ref{fig2a}. 
\figurename~\ref{fig2a} reports the NMSEs on the test set by baselines and hybrid ISTA models. Here, the untied models LISTA-CP-U/CPSS-U and Gated LISTA-U are considered, as they are superior to LISTA-CP-T/CPSS-T and Gated LISTA-T. We adopt $\lambda=0.1$ for ISTA,  $\lambda^0=0.1$ for HCISTA, and $p=0.7$ and $p_{\rm{max}}=13$ for LISTA-CPSS and HLISTA-CPSS. 
As discussed in Section~\ref{sec_HGLISTA}, we adopt the same inverse proportional function in the last six iterations and piece-wise linear function in the rest as gain gate functions, and sigmoid-based function in all sixteen iterations as overshoot gate function for Gated LISTA and HGLISTA. 
As shown in \figurename~\ref{fig2a}, all the proposed hybrid ISTA models outperform the corresponding baselines with evident gaps.

Table~\ref{T3} compares the numbers of learnable parameters for the methods listed in \figurename~\ref{fig2}. 
The numbers of parameters of free-form DNNs, $\mathbf{W}^n$, piece-wise linear function, inverse proportional function and sigmoid-based function are 2592, 125000, 2, 2, and 375001\footnote{Here, 375000 parameters are shared across all the iterations.}, respectively. 
%Note that the number of parameters of free-form DNNs , $\mathbf{W}^n$, piece-wise linear function, inverse proportional function and sigmoid-based function are $9\times 16 \times 18=2592$, $250\times 500=125000$, 2, 2 and $500\times 500+250\times 500+1=375001$\footnote{Here 375000 parameters are shared across all the iterations.}, respectively. 
Limited by the architecture of ISTA, ALISTA   and ELISTA, we cannot build the corresponding hybrid models with fewer parameters. However, the comparison between LISTA-CP-U/CPSS-U (resp. Gated LISTA-U) and HLISTA-CP/CPSS (resp. HGLISTA) suffices to corroborate the superiority of the hybrid ISTA. As shown in \figurename~\ref{fig2a}, HLISTA-CP and HLISTA-CPSS (resp. HGLISTA) outperform the corresponding baselines and require about 15 (resp. 2) times less learnable parameters. We provide more detailed comparisons in the following subsections.

\subsubsection{HCISTA vs. ISTA $\&$ FISTA $\&$ ADMM}\label{6.1.2}
We first corroborate Theorems~\ref{theorem1}-\ref{theorem3} by considering HCISTA-UnT that incorporates untrained DNNs. \figurename~\ref{fig2b} evaluates ISTA and HCISTA-UnT for up to 600 iterations under the hyper-parameters $\lambda^0=0.2$, $0.1$, and $0.05$. In HCISTA-UnT, $\mathcal{W}$ are initialized as orthogonal matrices and the initial values of $t^{n}$ and $\delta^n$ are randomly taken within the ranges of $[1/(4\delta^n\|\mathbf{A}\|_2^2),1/\|\mathbf{A}\|_2^2]$ (Eq.~\eqref{ls4}) and $(0.25, 0.5)$, respectively.
%We first evaluate ISTA and HCISTA-UnT (HCISTA with untrained DNNs) to corroborate Theorems~\ref{theorem1}-\ref{theorem3}. $\lambda^0=$0.2, 0.1, and 0.05, as shown in \figurename~\ref{fig2b}. Here, HCISTA-UnT is constructed with orthogonally initialized $\mathcal{W}$, and randomly initialized $t^{n}$ and $\delta^n$ within Eq.~\eqref{ls4} and $(0.25, 0.5)$, respectively. 
Besides, we set $\lambda^n=0.999*\min\{\lambda^{n-1}, \|\mathbf{x}^n-\mathbf{x}^{n-1}\|_2\}$ and randomly select $\alpha^n$ from the uniform distribution with the bound specified by Eq.~\eqref{add5}. For fair comparison, we also apply the same adaptive setting of $\{\lambda^n\}_{n\in\mathbb{N}}$ in ISTA and denote it by ISTA-$\lambda$. 
As shown in \figurename~\ref{fig2b}, the NMSE curves of ISTA seem to stagnate after hundreds of iterations, as they adopt fixed $\lambda$'s. ISTA-$\lambda$ and HLISTA-UnT go straight down to the optimum and the trends seem to continue after 600 iterations. 
Though ISTA-$\lambda$ performs better than HLISTA-UnT during the first dozens to hundreds of iterations (about 70, 130, 375 iterations for $\lambda^0=0.2, 0.1, 0.05$), HLISTA-UnT converges faster as iteration number goes to infinity.
Moreover, the choice of $\lambda$ has dramatic influences on the reconstruction performance of ISTA and ISTA-$\lambda$. As shown in \figurename~\ref{fig2b}, a large $\lambda$ leads to faster convergence in the initial stage but a less accurate solution.
%A large $\lambda$ leads to faster convergence in the initial stage but a less accurate solution, the curves in \figurename~\ref{fig2b} validate the proposition. Though it is not very distinct, \figurename~\ref{fig2b} shows that HCISTA-UnT inherits \request{this property} slightly.
In \figurename~\ref{fig2b_add}, we compare HCISTA-UnT with two conventional algorithms for the Lasso problems, FISTA and ADMM. Though FISTA and ADMM achieve superior performance during the first dozens to hundreds of iterations (about 70, 110, 270 iterations for $\lambda^0=0.2, 0.1, 0.05$), HLISTA-UnT performs better as iteration number goes to infinity.

Since it is time-consuming to train a HCISTA model with 600 iterations, we construct HCISTA, HCISTA-UnT, and ISTA for 16 iterations and compare their NMSEs. \figurename~\ref{fig2c} shows that HCISTA is superior to HCISTA-UnT and ISTA. Thus, it is reasonable to infer that HCISTA can still yield lower NMSE than HCISTA-UnT and ISTA after 600 iterations. The improvements of reconstruction performance originate from the trained DNNs with learnable $\lambda^n$, $t^n$, $\delta^n$ and $\alpha^n$.
%It is worth mentioning that the NMSE curves of HCISTA-UnT and HCISTA with different $\lambda^0$'s (experiments are conducted with $\lambda^0=0.2, 0.1, 0.05$) are extremely similar within 16 iterations, thus the performance with $\lambda^0=0.1$ is instructive. 
Note that the NMSE curves of HCISTA-UnT and HCISTA under $\lambda^0=0.2$, $0.1$, and $0.05$ are extremely similar within 16 iterations. Thus, the performance with $\lambda^0=0.1$ is instructive. Therefore, \figurename~\ref{fig2b} and \figurename~\ref{fig2c} support our main results that the convergence rate of HCISTA is at least equivalent to ISTA, even with untrained DNNs, as stated in Theorems~\ref{theorem1}$-$\ref{theorem3}.
%Therefore, \figurename~\ref{fig2b} and \figurename~\ref{fig2c} support our main results stated in Theorem~\ref{theorem1}$-$\ref{theorem3}, \emph{i.e.}, the convergence rate of HCISTA is at least equivalent to ISTA, even with untrained DNNs.

We further compare HCISTA with HCISTA-F to explore the influence of the constraints of step size and regularization parameter. \figurename~\ref{fig2c} shows that the NMSE achieved by HCISTA-F decreases fast as the iteration number $n$ grows and is much smaller than that by HCISTA. Moreover, by comparing \figurename~\ref{fig2c} and \figurename~\ref{fig2d}, we find that HCISTA-F is close to LISTA-CP in the recovery performance. This fact implies that HCISTA-F actually approximates or attains a linear convergence rate. 
%From Theorem~\ref{theorem4}, there are some factors that affects the recovery error and convergence rate.
In comparison to HLISTA-CP, however, HCISTA-F is worse in NMSE. As discussed in Section~\ref{sec:4.6}, although $\mathbf{A}$ is generated from a Gaussian distribution and the columns of $\mathbf{A}$ are normalized to have the unit $\ell_2$ norm, $t^n\mathbf{A}^T\in\mathcal{W}_s(\mathbf{A})$ cannot be always guaranteed for HCISTA-F. Besides, $\alpha^n$ is fixed to $0.5$ in HCISTA-F and is possibly not optimal for the hybrid models. 
%thus lost the flexibility and may not be the best value for the model. These lead to the worse NMSE of HCISTA-F in comparison to HLISTA-CP in \figurename~\ref{fig2d}.

\begin{figure}[!t]
\renewcommand{\baselinestretch}{1.0}
\centering
\includegraphics[width=0.7\textwidth]{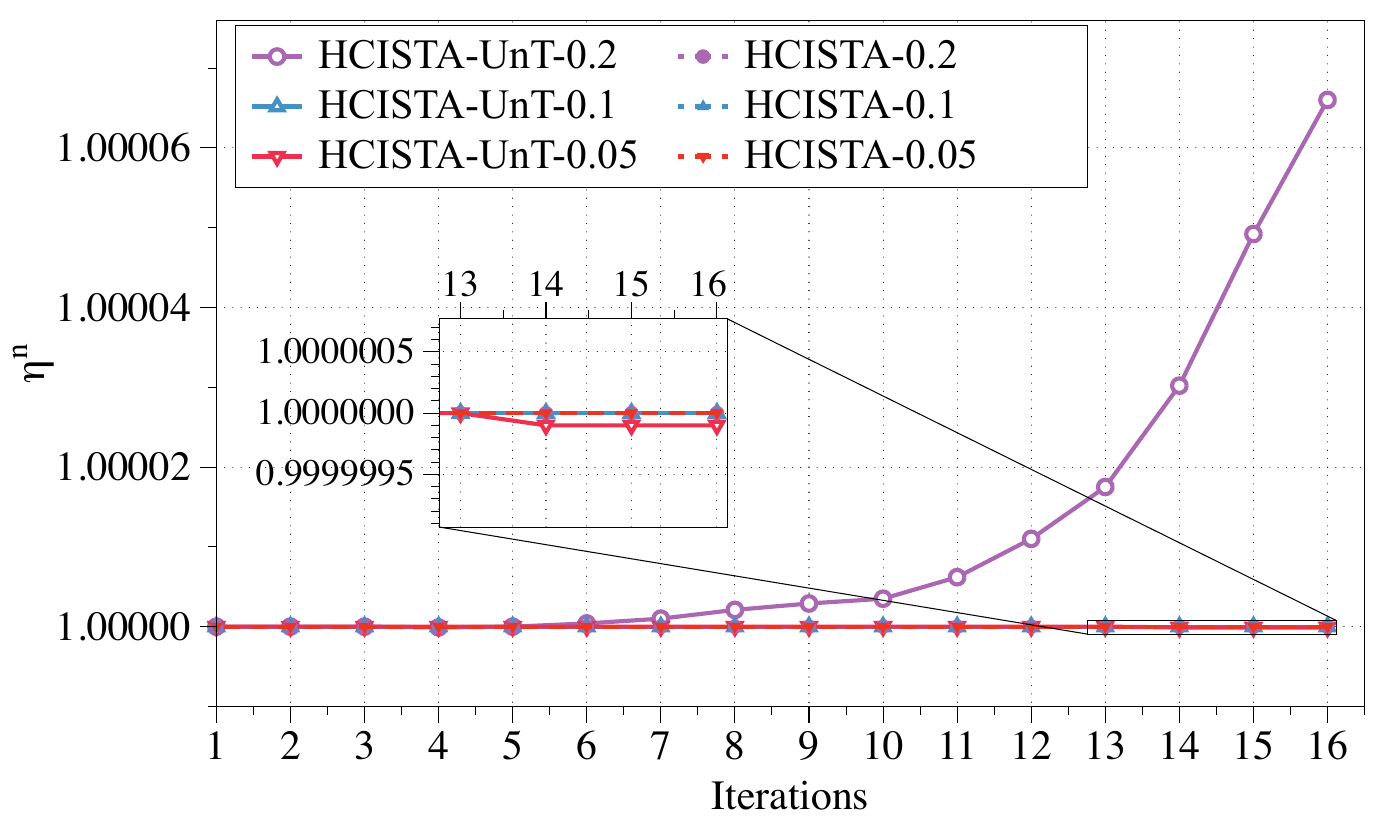}
\caption{The values of $\eta^n$, $n=1,\cdots,16$ defined in Assumption~\ref{assum1} for HCISTA with trained and untrained DNNs under $\lambda=0.05$, $0.1$, and $0.2$.}\label{fig3}
\end{figure}

Furthermore, we demonstrate that Assumption~\ref{assum1} can be easily satisfied in our experiments.
\figurename~\ref{fig3} plots the values of $\eta^n$ for the 16 iterations in HCISTA and HCISTA-UnT. For both trained and untrained DNNs, $\eta^n \approx 1$ for $n=0,\cdots,16$ and could reach $1$ in some cases (\emph{e.g.}, $\lambda^0=0.1$).  We further clarify Assumption~\ref{assum1} with more results in Appendix~D.
%We also find that in some cases $\eta^n=1$ during 16 iterations.
%\request{Limited by the measuring accuracy, these may be caused by errors that are less than $10^{-7}$.} %This result implies that Assumption~\ref{assum1} can be easily satisfied in our experiments.

\subsubsection{HLISTA vs. Variants of LISTA}\label{6.1.3}
We further compare LISTA-CP, LISTA-CPSS, ALISTA, Gated LISTA and ELISTA with tied and untied weights with the corresponding hybrid unfolded models. Note that, in ALISTA and HALISTA, $\mathbf{W}$ is pre-computed by solving the optimization problem~\cite{liu2018alista}:
\begin{equation}
\begin{aligned}
& \mathbf{W}\in \mathop{\arg\min}_{\mathbf{W}\in\mathbb{R}^{M\times N}}\|\mathbf{W}^T \mathbf{A}\|_F^2, \\
& \text{s.t.}~(\mathbf{W}_n)^T \mathbf{A}_n = 1,~\forall n =1, 2,\cdots, N,
\end{aligned}
\end{equation}
where $\mathbf{W}_n$ represents the $n$th column of $\mathbf{W}$.

\begin{figure}[!t]
\renewcommand{\baselinestretch}{1.0}
\centering
\includegraphics[width=0.7\textwidth]{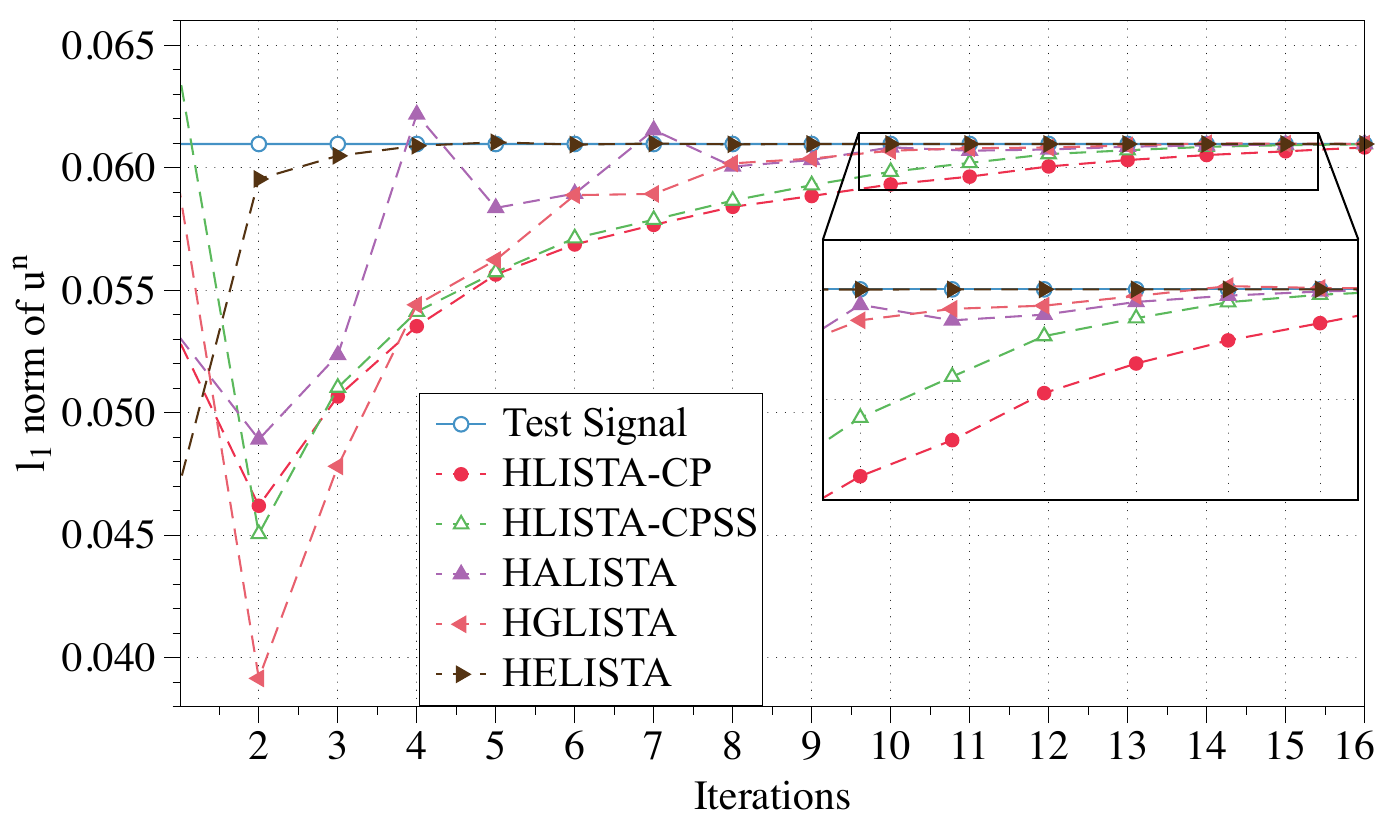}
\caption{Average $\ell_1$ norms of the test signal $\mathbf{x}^*$ and $\mathbf{u}^n$, $n=1,\cdots,16$ obtained by HLISTA-CP/CPSS, HALISTA, HGLISTA and HELISTA.
%The blue curve represents the average $\ell_1$ norm of the test signal $\mathbf{x}^*$.
}\label{fig4}
\end{figure}

%\figurename~\ref{fig2d}, \figurename~\ref{fig2e}, \figurename~\ref{fig2f}, {\color{blue}\figurename~\ref{fig2g} and \figurename~\ref{fig2h}} show that all the hybrid models yield lower NMSEs than the corresponding baselines. 
\figurename~\ref{fig2d}, \figurename~\ref{fig2e}, \figurename~\ref{fig2f}, \figurename~\ref{fig2g}, and \figurename~\ref{fig2h} show that all the hybrid models yield lower NMSEs than the corresponding baselines. For LISTA-CP/CPSS and Gated LISTA, the untied models perform better than the tied models in reconstruction but require much more learnable parameters. According to Table~\ref{T3}, HLISTA-CP and HLISTA-CPSS reduce NMSE by about 8 dB and 18 dB with 6.5\% learnable parameters when compared with LISTA-CP-U and LISTA-CPSS-U, and reduce NMSE by more than 15 dB and 20 dB using 103\% learnable parameters in comparison to LISTA-CP-T and LISTA-CP-SS. 
HGLISTA reduces NMSE by about 25 dB with 37\% learnable parameters compared with Gated LISTA-U and obtains an enormous improvement of NMSE, 53 dB, using 175\% learnable parameters in comparison to Gated LISTA-T. HELISTA reduces NMSE by about 27 dB with only 102\% learnable parameters compared with ELISTA.
%As discussed in Section~\ref{6.1.1}, although the parameters of HLISTA-CP(SS) are more than that of LISTA-CP(SS)-U, the comparison between HLISTA-CP(SS) and LISTA-CP(SS)-T implies that the proposed models obtain better performance with fewer parameters. For example, when trained for 13 iterations, HLISTA-CP achieves comparable NMSE to LISTA-CP-U trained for 16 iterations with a reduction of more than 90\% learnable parameters.
%For example, the NMSE obtain by HLISTA-CP after 13 iterations is comparable to the NMSE by LISTA-CP-U after 16 iterations, while the number of parameters of the latter is about 16 times larger than the former. 

We also evaluate the average $\ell_1$ norms of $\{\mathbf{u}^{n}\}_{n=1}^{K}$ in \figurename~\ref{fig4} to observe the relationship with the choice of $\theta_2^n$. From Eq.~\eqref{condition} and Eq.~\eqref{eq34}, the choice of $\theta_2^n$ depends on $\|\mathbf{u}^n\|_1$ and $\|\mathbf{u}^n-\mathbf{x}^*\|_1$. 
\figurename~\ref{fig4} shows that $\|\mathbf{u}^{n}\|_{1}$ tends to approximate $\|\mathbf{x}^*\|_{1}$ as $n$ grows. $\|\mathbf{u}^{n}\|_{1}$ is not stable in the first few iterations but rapidly converges to $\|\mathbf{x}^*\|_{1}$ for large $n$ (\emph{i.e.}, $n\geq 8$) as $\mathbf{x}^n\rightarrow \mathbf{x}^*$. This fact implies that $\theta_2^n$ decreases with the growth of $n$. \figurename~\ref{fig6b} validates this observation.
%Thus, we can easily deduce that $\theta_2^n$ decreases as $n$ increases and \figurename~\ref{fig6b} validates this proposition.

\begin{figure*}[!t]
\renewcommand{\baselinestretch}{1.0}
\centering
\subfigure[Proportion of True and False Positives]{\includegraphics[width=0.30\textwidth]{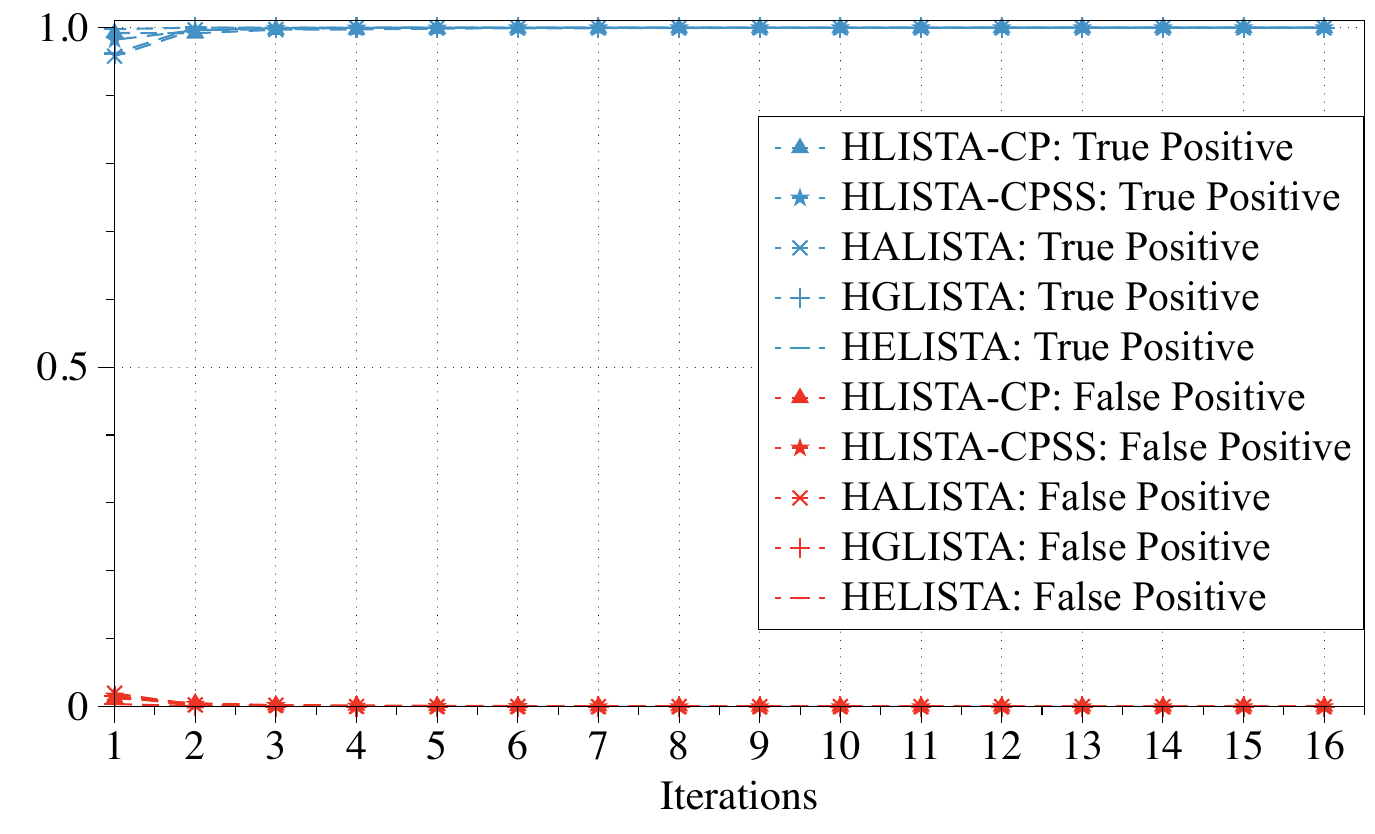}\label{fig5a}}
\quad
\subfigure[Proportion of false positives]{\includegraphics[width=0.30\textwidth]{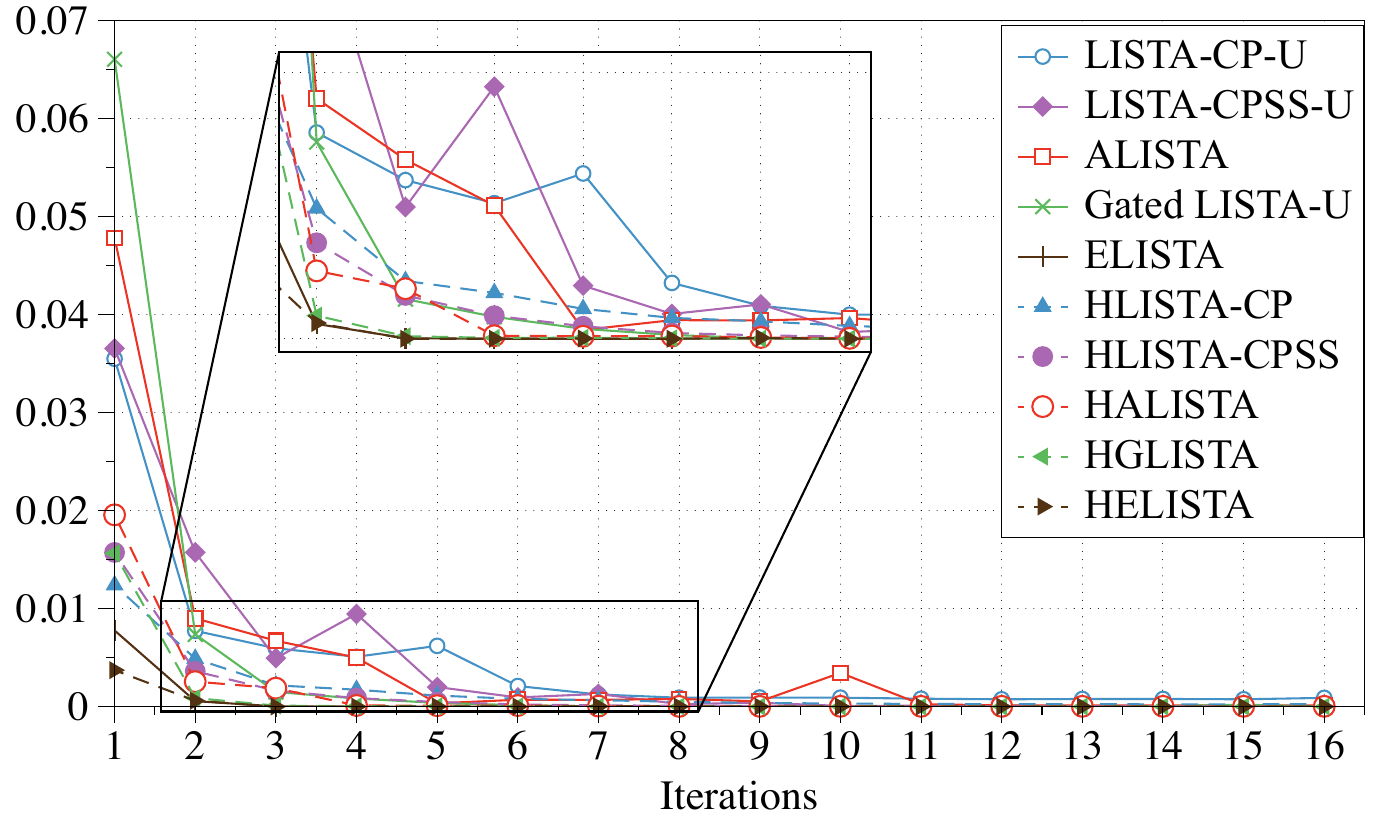}\label{fig5b}}
\quad
\subfigure[Proportion of true positives]{\includegraphics[width=0.30\textwidth]{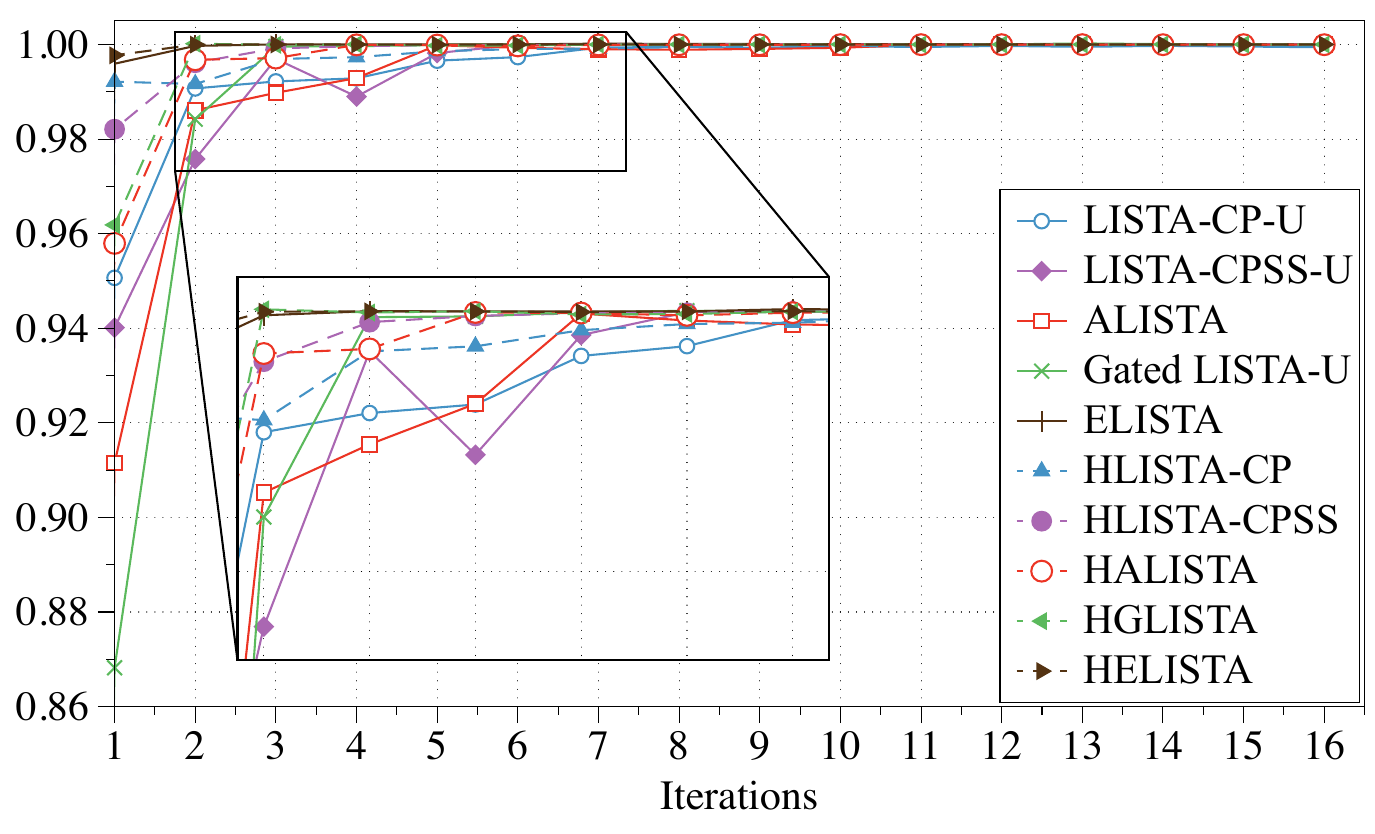}\label{fig5c}}
%\caption{Proportion of false positives vs. true positives in $\mathbf{x}^n$. The ``true positives'' curve draws the values of $\mathbb{E}\{\|\mathbf{x}^n_{\mathbb{S}}\|_{2}^{2}/\|\mathbf{x}^n\|_{2}^{2}\}$ with regard to $n$, whereas the ``false positives'' curve for $\mathbb{E}\{\|\mathbf{x}^n_{\mathbb{S}^c}\|_{2}^{2}/\|\mathbf{x}^n\|_{2}^{2}\}$.}\label{fig5}
\caption{Proportions of false positives and true positives in $\mathbf{x}^n$ obtained by LISTA-CP-U/CPSS-U~\cite{NIPS2018_8120}, ALISTA~\cite{liu2018alista}, Gated LISTA~\cite{Wu2020Sparse}, ELISTA~\cite{li2021learned}, and the proposed HLISTA models. The ``true positives'' curve draws the values of $\mathbb{E}[\|\mathbf{x}^n_{\mathbb{S}}\|_{2}^{2}/\|\mathbf{x}^n\|_{2}^{2}]$ with regard to $n$, whereas the ``false positives'' curve for $\mathbb{E}[\|\mathbf{x}^n_{\mathbb{S}^c}\|_{2}^{2}/\|\mathbf{x}^n\|_{2}^{2}]$.}\label{fig5}
\end{figure*}

%To justify there are no false positives in $\mathbf{x}^n$ (\emph{i.e.}, ${\rm support}(\mathbf{x}^n)\subset \mathbb{S}$) in Theorems~\ref{theorem4}, \ref{theorem5} and \ref{theorem6}, we report the average magnitude of the false positives and true positives in $\mathbf{x}^n$ in \figurename~\ref{fig5}. 
To justify that ${\rm supp}(\mathbf{x}^n)\subset \mathbb{S}$ in Theorems~\ref{theorem4}-\ref{theorem6}, we report the average magnitude of false positives and true positives in $\mathbf{x}^n$ in \figurename~\ref{fig5}. We adopt the same criterion as ALISTA~\cite{liu2018alista}. For $n=1,\cdots,K$, the ``true positives'' curve draws the values of $\mathbb{E}[\|\mathbf{x}^n_{\mathbb{S}}\|_{2}^{2}/\|\mathbf{x}^n\|_{2}^{2}]$ and the ``false positives'' from $\mathbb{E}[\|\mathbf{x}^n_{\mathbb{S}^c}\|_{2}^{2}/\|\mathbf{x}^n\|_{2}^{2}]$. Here, $\mathbf{x}^n_{\mathbb{S}}$ represents ${\rm supp}(\mathbf{x}^n)$ and $\mathbb{S}^c$ is the absolute complement of $\mathbb{S}$. \figurename~\ref{fig5a} shows that false positives take up a small proportion in the positives in the proposed hybrid ISTA models. This result meets with ${\rm supp}(\mathbf{x}^n)\subset \mathbb{S}$. We further explore the proportion from a refined perspective. As shown in \figurename~\ref{fig5b} (resp. \figurename~\ref{fig5c}), the proposed hybrid models attain a smaller (resp. larger) proportion of false (resp. true) positives than the corresponding baselines at the first fewer iterations, which suggests that they can achieve more accurate reconstruction.

%Based on the above discussions, the main results stated in Theorems~\ref{theorem4}, \ref{theorem5} and \ref{theorem6} are justified.

\subsubsection{Learned Thresholds and Balancing Parameter}
\begin{figure*}[!t]
\renewcommand{\baselinestretch}{1.0}
\centering
\subfigure[$\lambda^n t^n$ for HCISTA and $\theta_1^n$ for HLISTA]{
\includegraphics[width=0.30\textwidth]{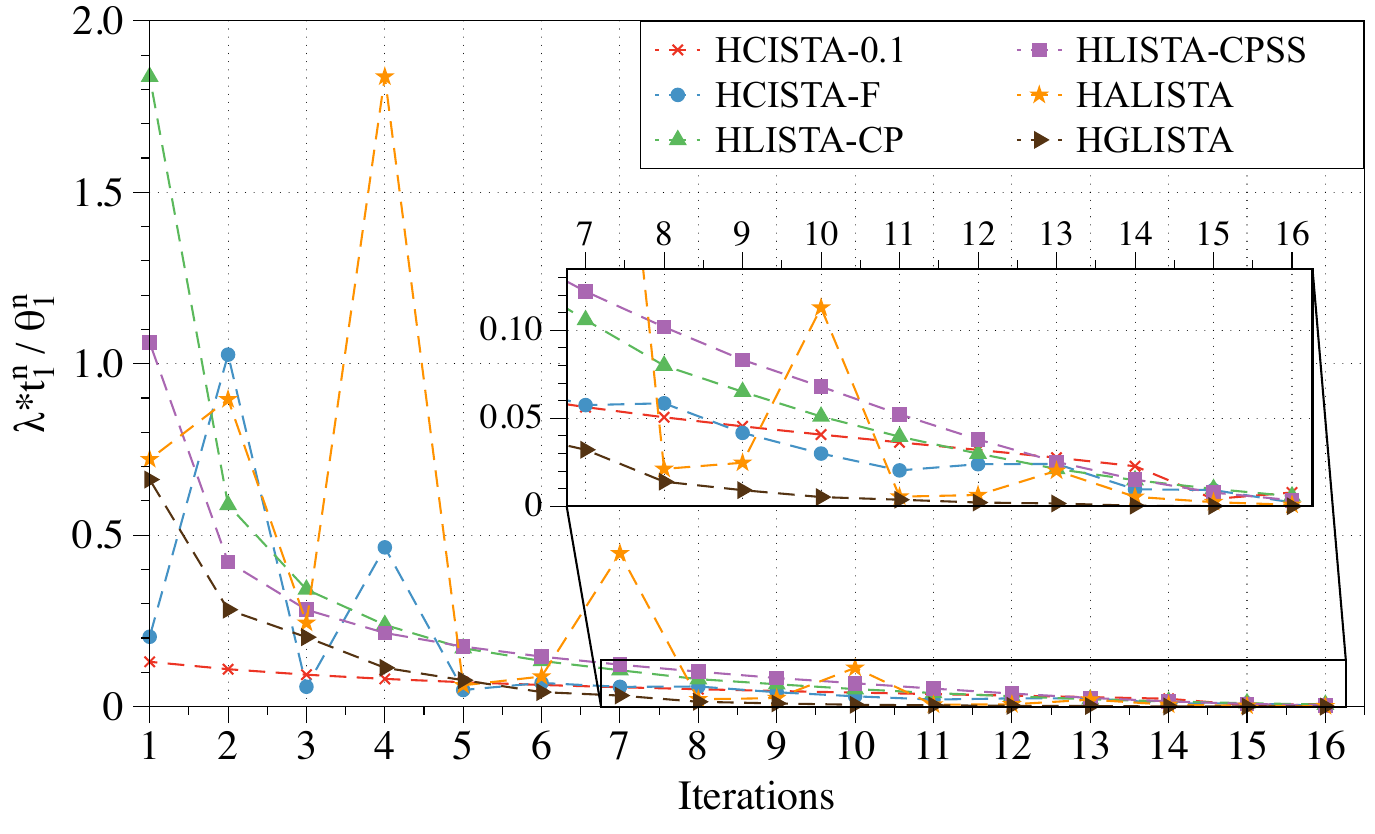}\label{fig6a}}
\quad
\subfigure[$\lambda^n t^n$ for HCISTA and $\theta_2^n$ for HLISTA]{
\includegraphics[width=0.30\textwidth]{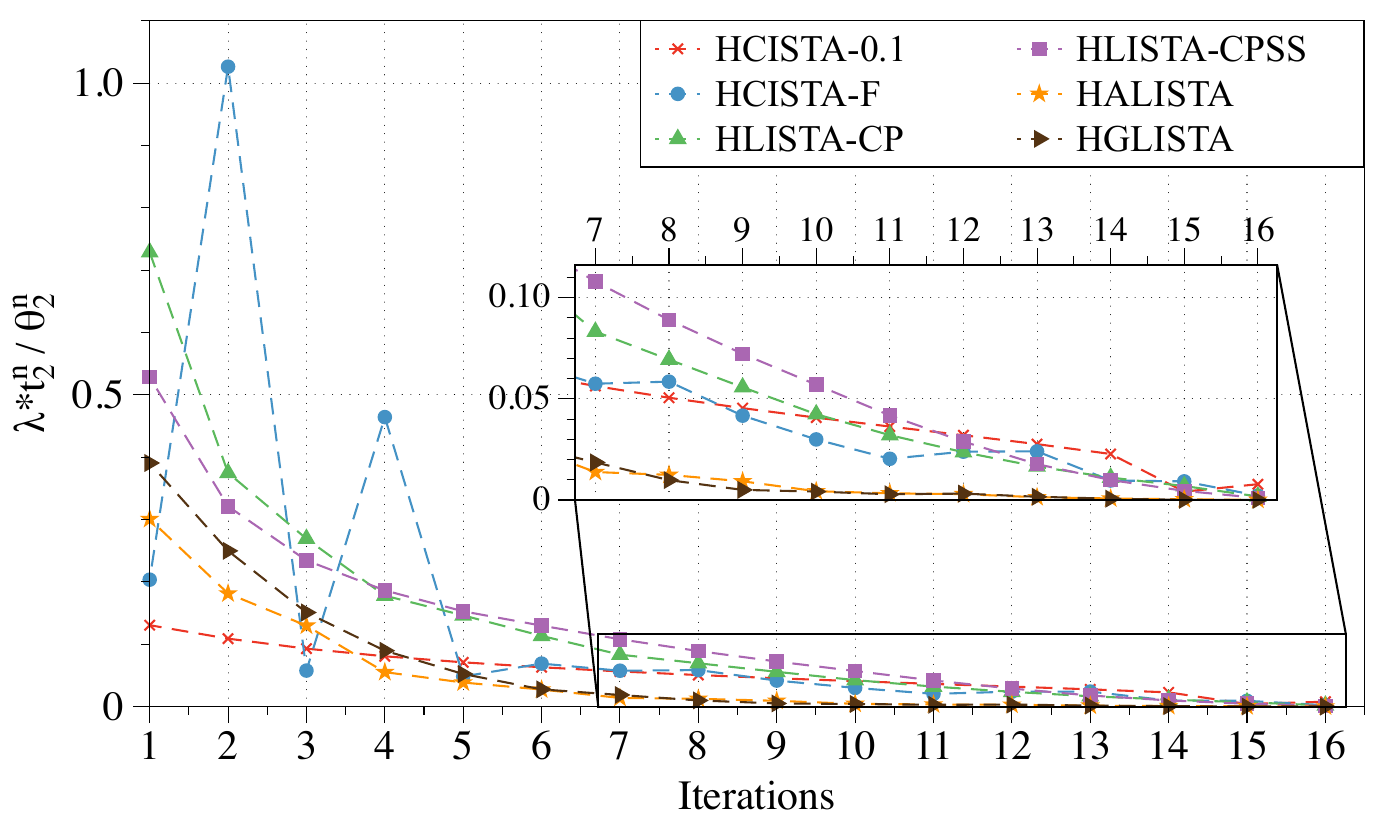}\label{fig6b}}
\quad
\subfigure[$\alpha^n$ for hybrid ISTA]{
\includegraphics[width=0.30\textwidth]{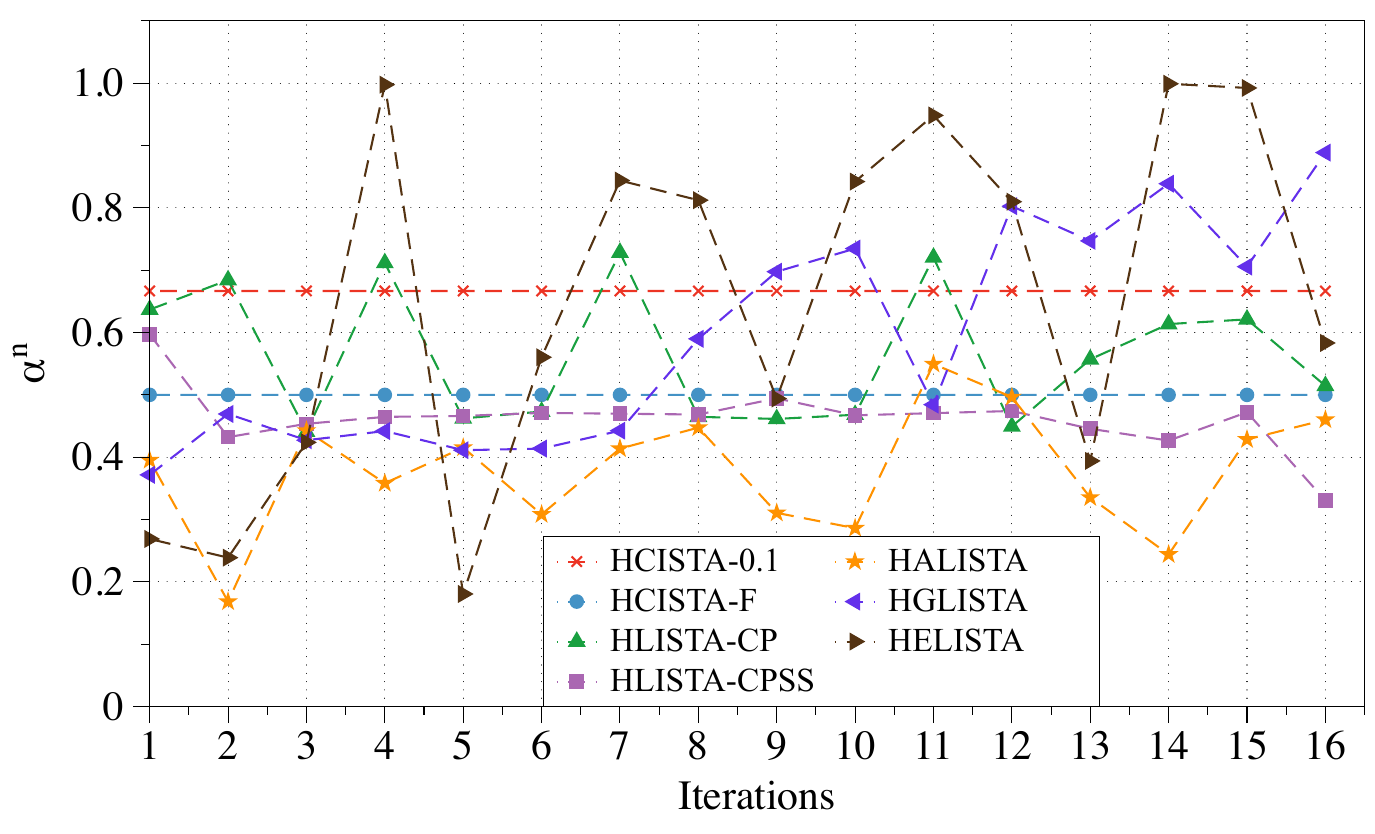}
\label{fig6c}}
\caption{The learned thresholds (\emph{i.e.}, $\lambda^n t^n$ for HCISTA and $\theta_1^n$ and $\theta_2^n$ for HLISTA) and $\alpha^n$ for hybrid ISTA models. %Specifically, $\lambda^n t^n$ is utilized to represent the thresholds for HCISTA, and $\theta_1^n$ and $\theta_2^n$ are for HLISTA.
}\label{fig6}
\end{figure*}

\figurename~\ref{fig6a} and \figurename~\ref{fig6b} demonstrate the learned thresholds, \emph{i.e.}, $\{\lambda^nt^n\}_{n=0}^{K}$ for HCISTA and $\{\theta_1^n\}_{n=0}^{K}$ and $\{\theta_2^n\}_{n=0}^{K}$ for HLISTA. 
In view of the special case of $\theta$'s for HELISTA, we do not show them here. 

%The learned thresholds are illustrated in \figurename~\ref{fig6a} and \ref{fig6b}. We utilize $\lambda^n t^n$ for HCISTA and $\theta_1^n, \theta_2^n$ for HLISTA. 
The thresholds tend to converge to 0 from a relatively large value as the iteration number increases. The downward trend of HCISTA with constrained $\lambda^n$ is smooth, whereas that of HCISTA-F with adaptive $\lambda^n$ fluctuate in the first few iterations. This fact implies that HCISTA-F learns adaptive steps to boost the reconstruction performance. For all the HLISTA models, the thresholds rapidly vanish with the growth of $n$, as $\|\mathbf{x}^n\|_1$ and $\|\mathbf{u}^n\|_1$ approximate $\|\mathbf{x}^*\|_1$ for large $n$. This fact implies that HLISTA models can achieve linear convergence based on the sequences of thresholds $\{\theta_1^n\}_{n=1}^K$ and  $\{\theta_2^n\}_{n=1}^K$ learned according to Eq.~\eqref{condition} or Eq.~\eqref{eq34}. \figurename~\ref{fig6c} demonstrates the values of $\{\alpha^n\}_{n=1}^K$ locate within the range of $[0.4, 0.6]$ in most cases. This fact implies that the output of DNN-based update step $\mathbf{w}^n$ has an approximately equivalent contribution to the outputs $\mathbf{x}^{n+1}$ compared with the original update step $\mathbf{v}^n$.
%For all the HLISTA models, because $\|\mathbf{x}^n\|_1$ and $\|\mathbf{u}^n\|_1$ approximate $\|\mathbf{x}^*\|_1$ as $n$ increases, the thresholds fast converge to 0 as $n$ grows. This fact implies that the learned thresholds $\{\theta_1^n\}_{n=1}^K$ and  $\{\theta_2^n\}_{n=1}^K$ follow Eq.~\eqref{condition} or Eq.~\eqref{eq34}, \emph{i.e.}, the sequence of parameters leading to linear convergence for HLISTA models can be obtained by data-driven learning.

%The values of  $\alpha^n$ are demonstrated in \figurename~\ref{fig6c}. One can see that $\alpha^n$ are within the range of [0.4, 0.6] in most cases, which implies that the output of DNN-based update step $\mathbf{w}^n$ has an approximately equivalent contribution to the results $\mathbf{x}^{n+1}$ compared with the original update step $\mathbf{v}^n$.

\begin{figure*}[!t]
\renewcommand{\baselinestretch}{1.0}
\centering
\subfigure[SNR = 20 dB]{
\includegraphics[width=0.30\textwidth]{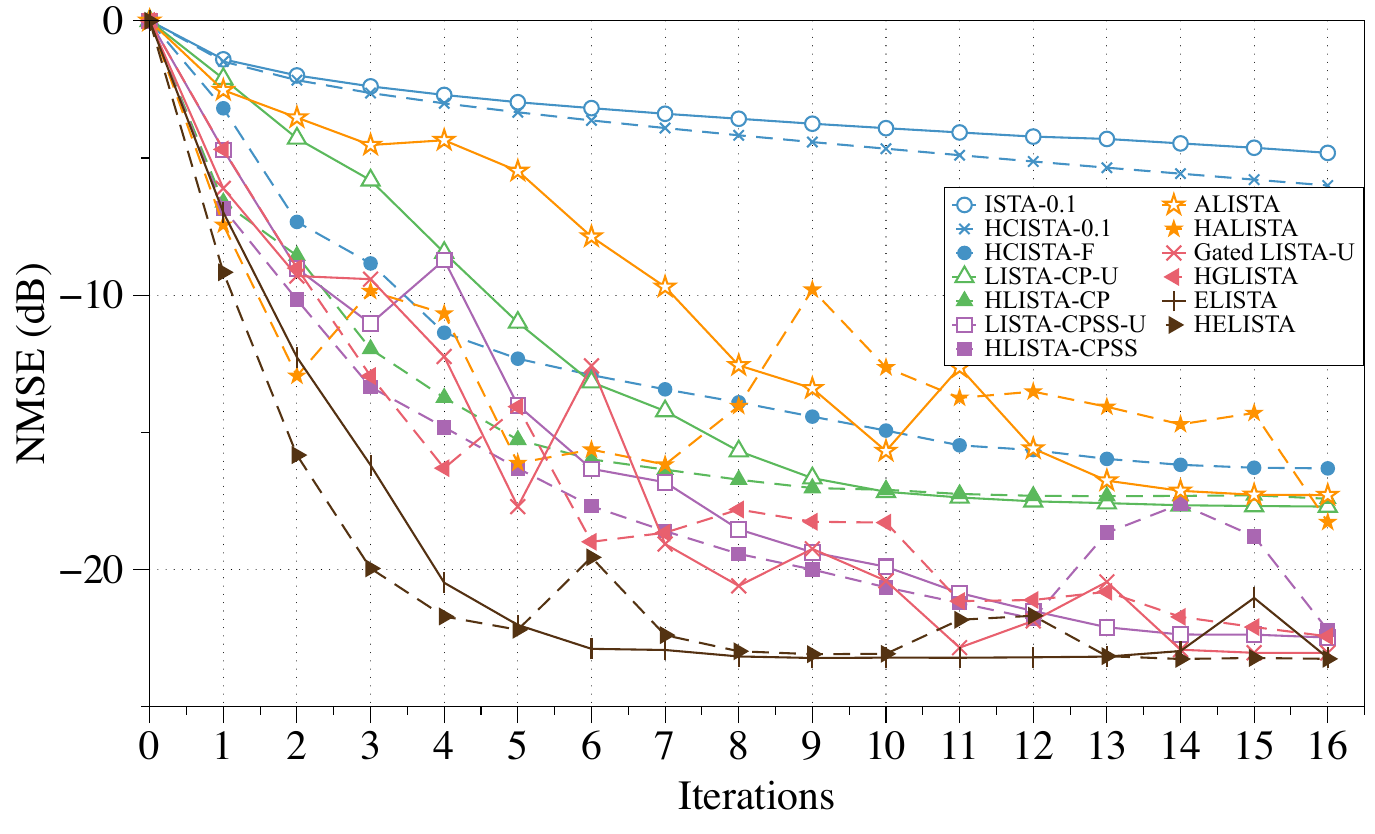}\label{fig7a}}
\quad
\subfigure[SNR = 30 dB]{
\includegraphics[width=0.30\textwidth]{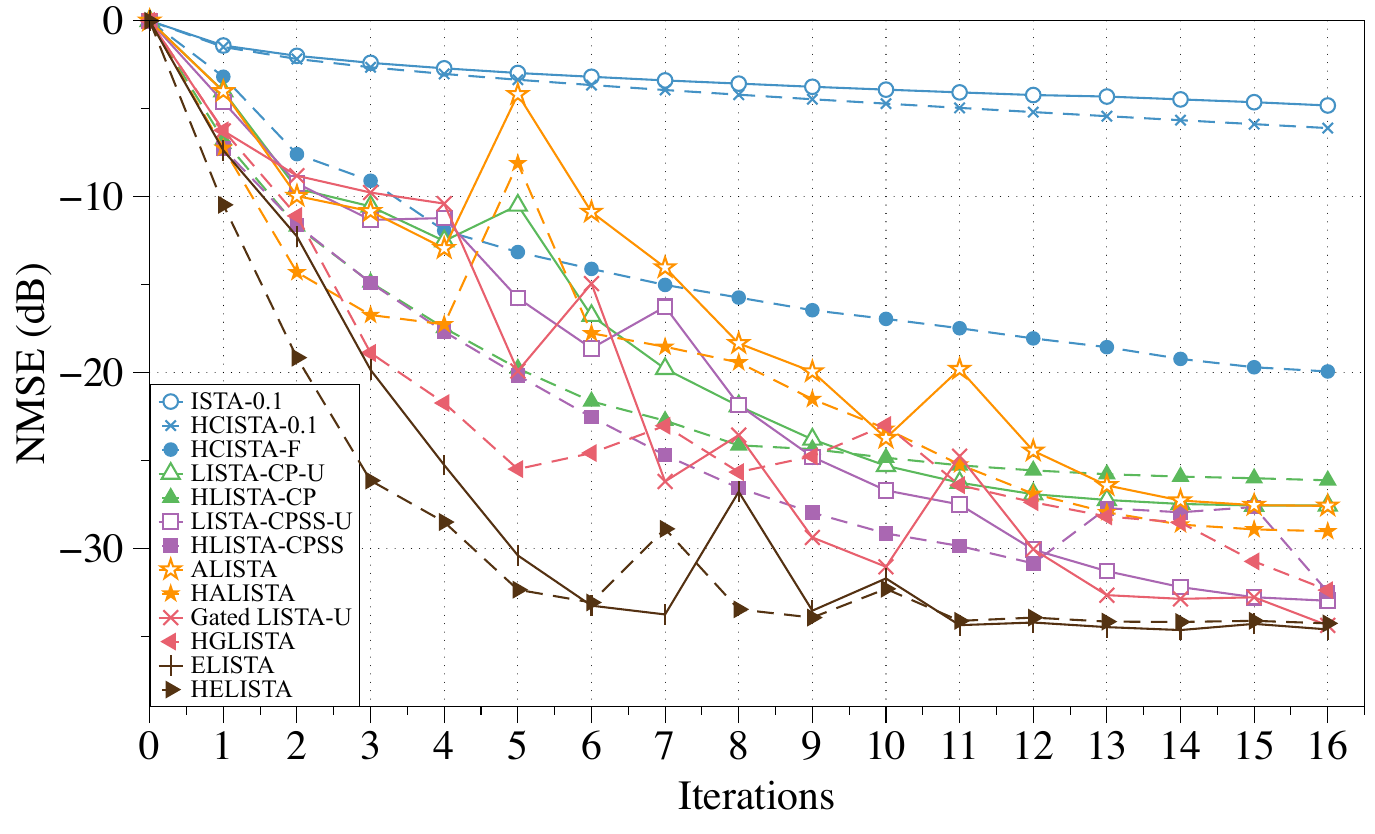}\label{fig7b}}
\quad
\subfigure[SNR = 40 dB]{
\includegraphics[width=0.30\textwidth]{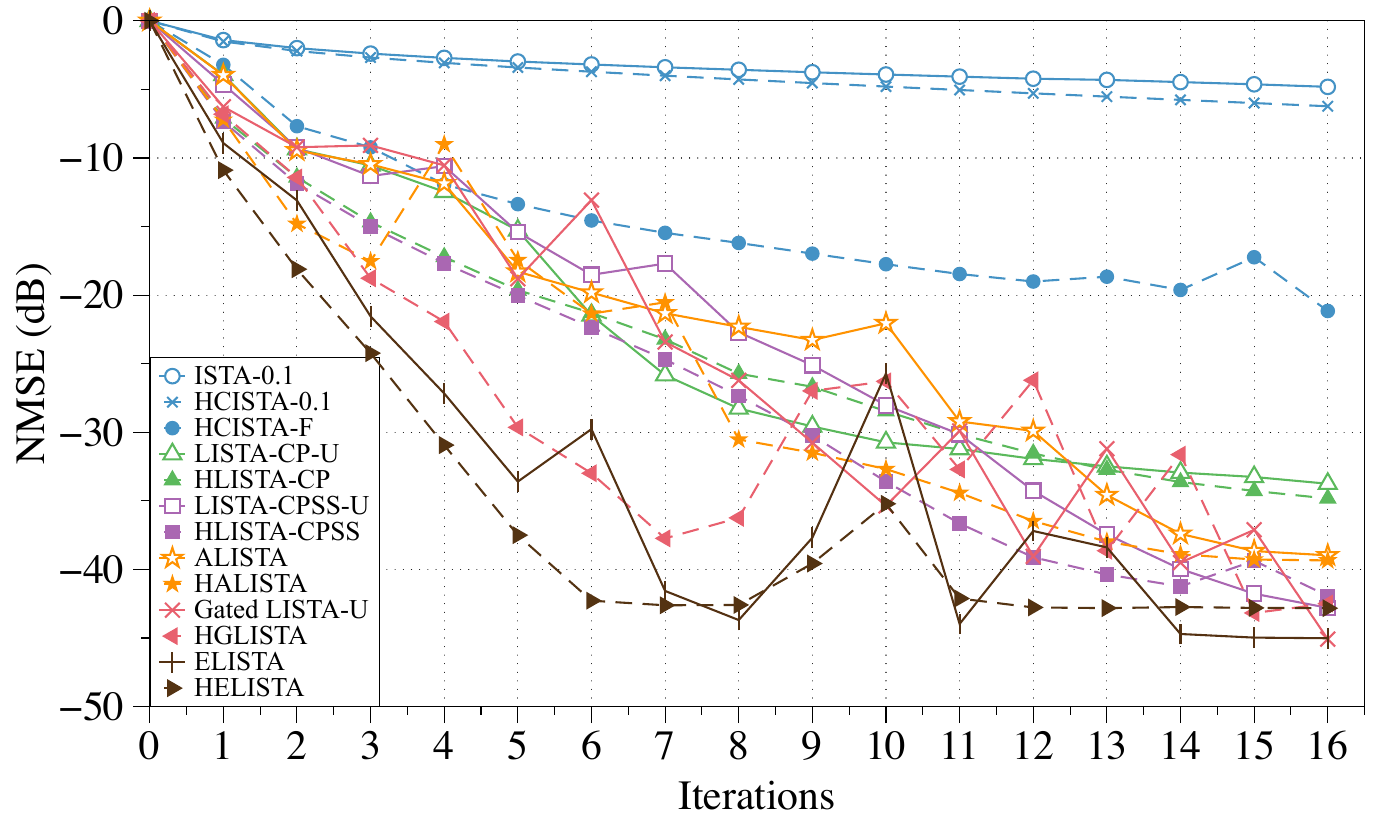}\label{fig7c}}\\
\caption{NMSEs obtained by ISTA~\cite{blumensath2008iterative}, LISTA-CP-U/CPSS-U~\cite{NIPS2018_8120}, ALISTA~\cite{liu2018alista},  Gated LISTA~\cite{Wu2020Sparse}, ELISTA~\cite{li2021learned}, and the proposed HCISTA models HCISTA-0.1 and HCISTA-F and HLISTA models HLISTA-CP/CPSS, HALISTA,  HGLISTA, and HELISTA for noisy cases under SNRs of 20, 30, 40 dB.}\label{fig:noise}
\end{figure*}

\begin{figure*}[!t]
\subfigure[$\mathcal{K}=5$]{
\includegraphics[width=0.30\textwidth]{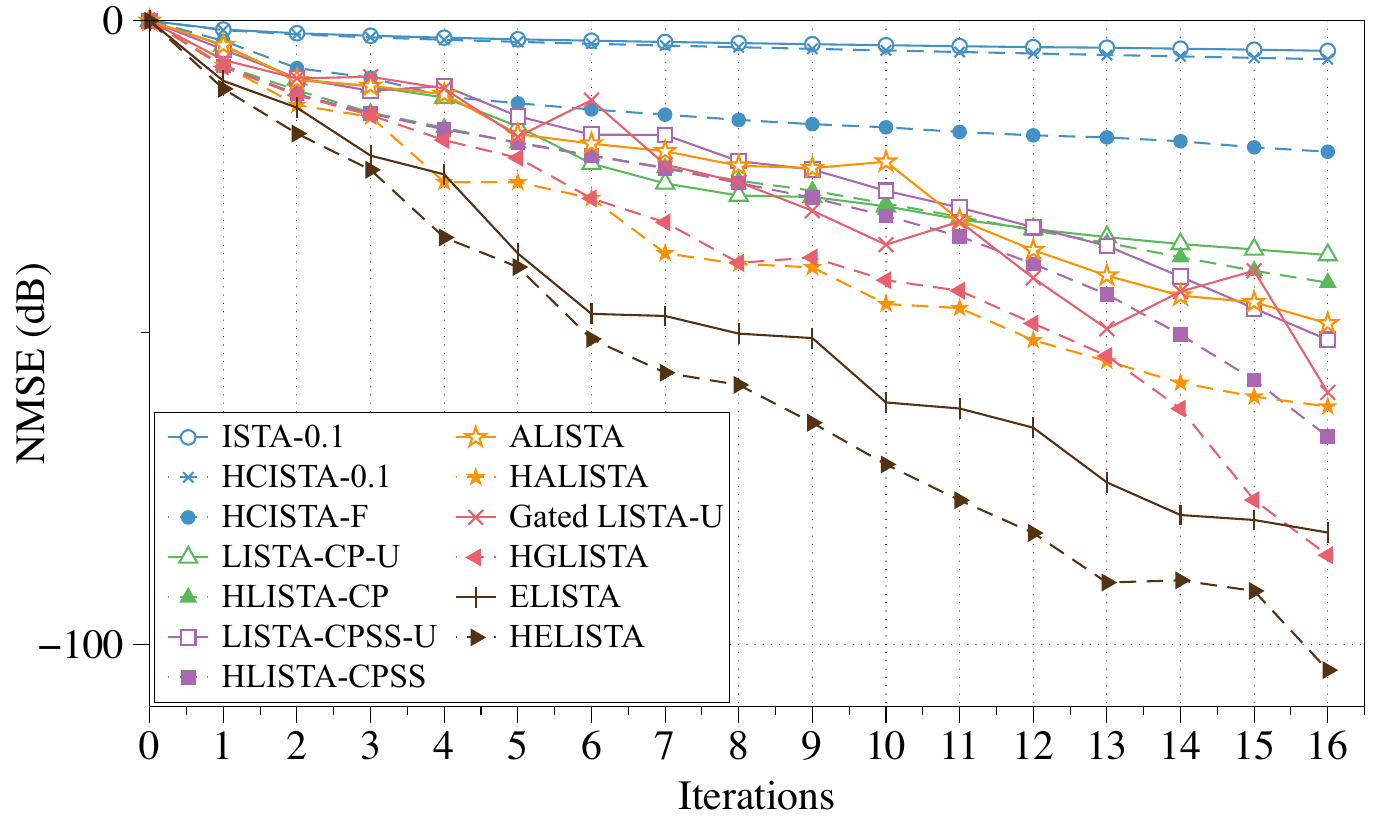}\label{fig7d}}
\quad
\subfigure[$\mathcal{K}=30$]{
\includegraphics[width=0.30\textwidth]{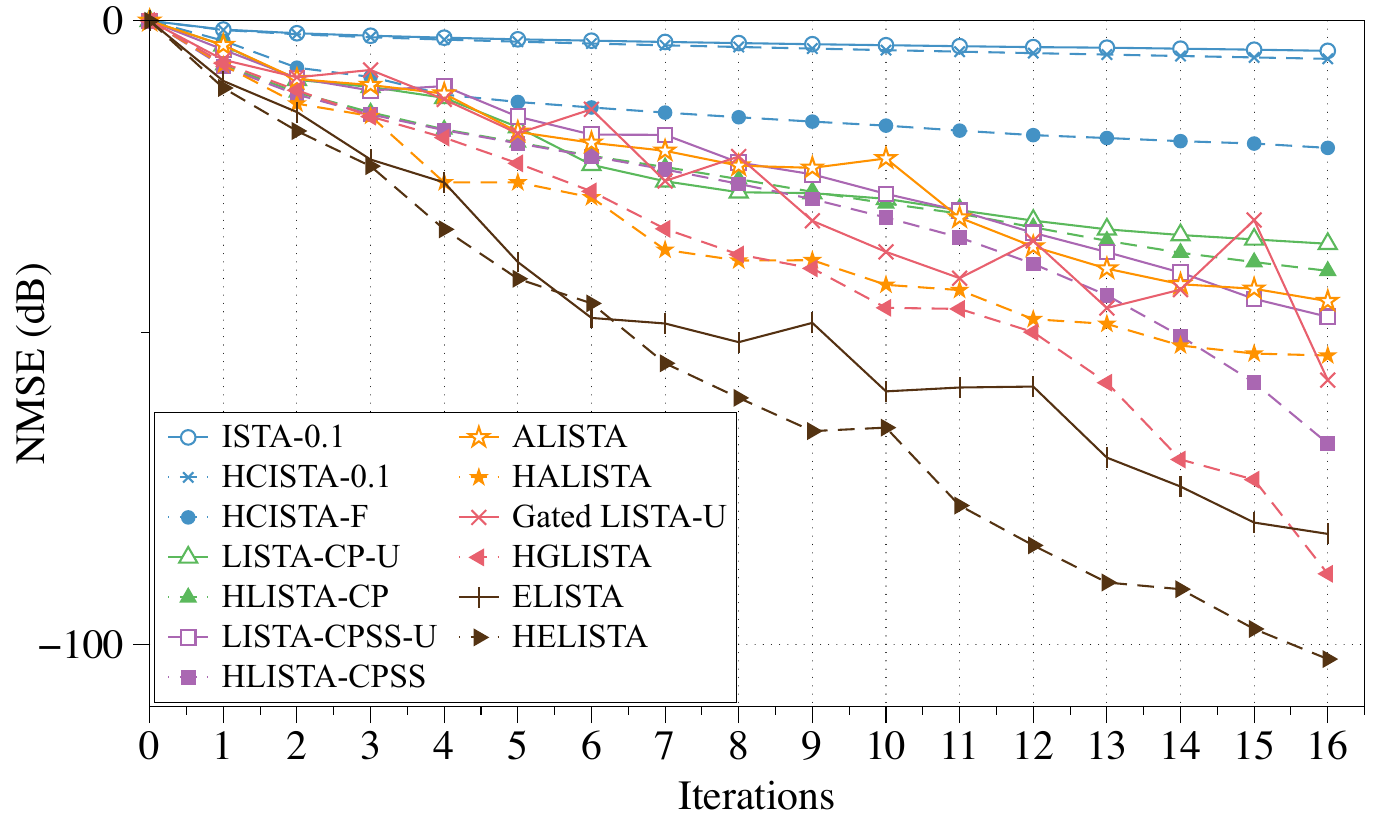}		\label{fig7e}}
\quad
\subfigure[$\mathcal{K}=50$]{
\includegraphics[width=0.30\textwidth]{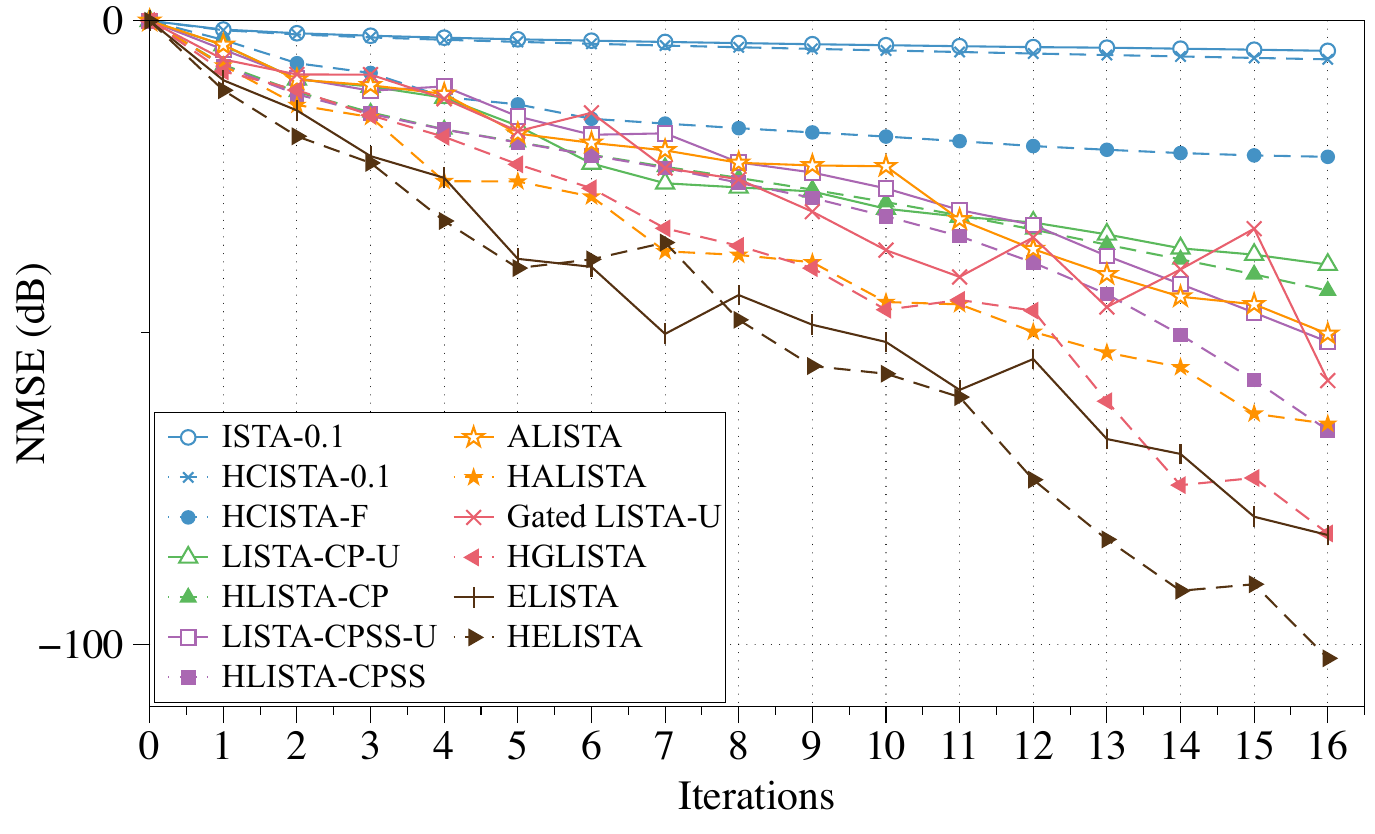}\label{fig7f}}
\caption{NMSEs obtained by ISTA~\cite{blumensath2008iterative}, LISTA-CP-U/CPSS-U~\cite{NIPS2018_8120}, ALISTA~\cite{liu2018alista},  Gated LISTA~\cite{Wu2020Sparse}, ELISTA~\cite{li2021learned} and the proposed HCISTA models HCISTA-0.1 and HCISTA-F and HLISTA models HLISTA-CP/CPSS, HALISTA, HGLISTA, and HELISTA for ill-conditioned matrices $\mathbf{A}$ with the condition number $\mathcal{K}$ of 5, 30, and 50.}\label{fig:ill}
%\caption{A summary NMSE in noisy cases or with ill-conditioned matrices $\mathbf{A}$. We evaluate the performance under SNRs of 20, 30, and 40 dB, and with ill-conditioned matrices $\mathbf{A}$ of condition number $\mathcal{K}$ of 5, 30, and 50.}
\end{figure*}

\subsubsection{Noisy Cases and Ill-conditioned Basis Matrix}
%We report the recovery in noisy cases under the SNRs of 20, 30, and 40 dB, as shown in \figurename~\ref{fig7a}, \ref{fig7b}, and \ref{fig7c}. The impact of noise is trivial for HCISTA-0.1, which implies the robustness.
%\figurename~\ref{fig7a}, \figurename~\ref{fig7b}, and \figurename~\ref{fig7c} report the recovery in noisy cases under the SNRs of 20, 30, and 40 dB, respectively. 
\figurename~\ref{fig:noise} reports the recovery performance in noisy cases under the SNRs of 20, 30, and 40 dB. HCISTA-0.1 (\emph{i.e.}, HCISTA with $\lambda^0=0.1$) is robust and trivially affected by the noise. HLISTA models are superior for most $n<K$ and achieve similar recovery performance for $n=K$, when compared with the corresponding baselines. Their NMSEs converge to a stationary level related to the level of noise (\emph{i.e.}, NMSEs decreasing with the growth of SNR). This fact suggests that HLISTA models inherit the robustness of LISTA models and perform better in the intermediate stages of reconstruction.
%Besides, the proposed HLISTA models show a superior performance within most interior iterations and achieve similar performance with the correpsonding baselines after the final iteration. The NMSE converges to a stationary level related with the noise-level. This means that HLISTA models inherit the robustness of the LISTA models and performs better in the intermediate reconstruction stage.

%Moreover, we evaluate the models with ill-conditioned matrices $\mathbf{A}$ of condition numbers $\mathcal{K}$ of 5, 30, and 50, as demonstrated in \figurename~\ref{fig7d}, \ref{fig7e}, and \ref{fig7f}. Moreover, we consider the case that $\mathbf{A}$ is ill-conditioned. In \figurename~\ref{fig7d}, \figurename~\ref{fig7e}, and \figurename~\ref{fig7f}, we evaluate the models with under the condition numbers $\mathcal{K}$ of 5, 30, and 50 for $\mathbf{A}$, respectively.
Moreover, we demonstrate that hybrid ISTA models are robust to ill-conditioned $\mathbf{A}$. The condition number $\mathcal{K}$ of $\mathbf{A}$ is set to $5$, $30$, and $50$ for evaluations. \figurename~\ref{fig:ill} shows that HCISTA models can retain the recovery performance, whereas HLISTA models outperform the corresponding baselines with and are stable to $\mathbf{A}$ with varying $\mathcal{K}$.
%Similarly, HCISTA models are robust to ill-conditioned $\mathbf{A}$, and can retain the reconstruction performance. HCISTA models can retain the reconstruction performance. For HLISTA modes, the results show that they outperform the corresponding baselines in recovery performance with an equivalent stability, which corroborates the robustness to ill-conditioned $\mathbf{A}$.

\subsection{Natural Image Compressive Sensing}
\subsubsection{Comparison with Classical ISTA and LISTA}
\label{sec:6.2.1}
We further evaluate the hybrid ISTA with $K=16$ iterations in the task of natural image CS under various measurement rates (MRs). For fair comparison, we utilize the \emph{BSD500}  dataset~\cite{martin2001database} for all DNN-based models, as in LISTA-CP and LISTA-CPSS~\cite{NIPS2018_8120}. \emph{BSD500} is divided into a training set of 400 images, a validation set of 50 images, and a test set of 50 images. We randomly extract 10000 image patches with size $16\times 16$ from each image for training. For each image patch, the means of its pixels is removed. To induce the sparsity from the natural images, we prepare a dictionary $\mathbf{D}\in \mathbb{R}^{256\times 512}$ learned from the training set using the online dictionary learning method~\cite{mairal2009online}. A measurement matrix $\mathbf{\Phi}\in \mathbb{R}^{M\times 256}$ is generated in the same manner as in Section~\ref{sec:6.1} and the MR is $M/256$.
The reconstruction performance is evaluated in terms of the average Peak Signal-to-Noise Ratio (PSNR) and Structural Similarity Index Metric (SSIM) over the test images. In addition to the test set composed of 50 images from \emph{BSD500}, we also adopt the widely used \emph{Set11} dataset for test~\cite{Kulkarni2016ReconNet,zhang2018ista}. Each test image is first divided into non-overlapped image patches with size $16\times 16$, then recovered and aggregated as a whole image. Note that the image patches may not be exactly sparse even though we introduce the dictionary $\mathbf{D}$, as Assumption~\ref{assum2} is not strictly satisfied. The primary goal of the CS experiments is to show that the proposed hybrid models are robust to the deviations and practically useful in non-ideal conditions.
%Note that even though we utilize a dictionary $\mathbf{D}$, the natural image patches $f$ may not be exactly sparse under $\mathbf{D}$ and then Assumption~\ref{assum2} no longer strictly holds. The primary goal of the CS experiments is to show that the proposed hybrid models are robust to the deviations and practically useful in non-ideal conditions. 

\begin{table*}[!t]
\renewcommand{\baselinestretch}{0.8}
\renewcommand{\arraystretch}{0.8}
\setlength{\abovecaptionskip}{-2pt}
\centering
\caption{Comparison of average PSNR (dB) | SSIM on \emph{Set11} and \emph{BSD500} (50 images for test) obtained at the measurement rates (MRs) of 0.04, 0.10, 0.25 and 0.50.}\label{T4}
\begin{tabular}{lccccc}
\toprule
Results for \emph{Set11} & MR=0.50 & MR=0.25 & MR=0.10 & MR=0.04 \\
\midrule
ISTA (Convergence)~\cite{blumensath2008iterative} & 27.63 | 0.8372 & 20.16 | 0.5874 & 17.29 | 0.4048 & 9.301 | 0.1045 \\
FISTA (Convergence)~\cite{doi:10.1137/080716542} & 27.78 | 0.8404 & 22.19 | 0.6570 &  17.62 | 0.4141 & 13.59 | 0.2366 \\
ADMM (Convergence)~\cite{admm2010} &  27.64 | 0.8376 & 21.91 | 0.6474 & 16.89 | 0.3886 &  12.55 | 0.2113 \\
HCISTA & 32.45 | 0.9251 & 27.83 | 0.8415 & 23.51 | 0.6962 & 21.40 | 0.5981 \\
HCISTA-F  & \textbf{35.03} | \textbf{0.9516} & \textbf{30.03} | \textbf{0.8878} & \textbf{24.80} | \textbf{0.7526} & \textbf{21.75} | \textbf{0.6240} \\
\midrule
LISTA-CP-U~\cite{NIPS2018_8120} & 34.53 | 0.9479 & 29.69 | 0.8832 & 24.91 | 0.7585 & 21.84 | 0.6290 \\
HLISTA-CP & \textbf{35.48} | \textbf{0.9550} & \textbf{30.25} | \textbf{0.8920} & \textbf{25.12} | \textbf{0.7651} & \textbf{21.85} | \textbf{0.6298} \\
\midrule
LISTA-CPSS-U~\cite{NIPS2018_8120} & 34.50 | 0.9477 & 29.65 | 0.8828 & 24.79 | 0.7548 & 21.80 | 0.6262 \\
HLISTA-CPSS & \textbf{35.40} | \textbf{0.9544} & \textbf{30.25} | \textbf{0.8923} & \textbf{24.85} | \textbf{0.7557} & \textbf{21.82} | \textbf{0.6271} \\
\midrule
ALISTA~\cite{liu2018alista} & 33.53 | 0.9388 & 28.76 | 0.8656 & 23.03 | 0.6818 & 20.98 | 0.5837 \\
HALISTA & \textbf{34.45} | \textbf{0.9470} & \textbf{29.89} | \textbf{0.8858} & \textbf{24.10} | \textbf{0.7243} & \textbf{21.54} | \textbf{0.6124} \\
\midrule
Gated LISTA-U~\cite{Wu2020Sparse} & 34.82 | 0.9500 &  30.04 | 0.8902 & 25.15 | 0.7659 & 21.60 | 0.6269 \\
HGLISTA & \textbf{36.77} | \textbf{0.9628} & \textbf{31.48} | \textbf{0.9111} & \textbf{25.21} | \textbf{0.7676} & \textbf{21.92} | \textbf{0.6333} \\
\midrule
ELISTA-U~\cite{li2021learned} & 33.02 | 0.9302 & 27.41 | 0.8339 & 23.27 | 0.6946 & 21.04 | 0.5869 \\
HELISTA & \textbf{36.46} |  \textbf{0.9611} & \textbf{31.40} | \textbf{0.9091} &  \textbf{25.51} | \textbf{0.7778} & \textbf{21.86} | \textbf{0.6287} \\
\bottomrule
\\
\toprule
Results for \emph{BSD500} & MR=0.50 & MR=0.25 & MR=0.10 & MR=0.04 \\
\midrule
ISTA (Convergence)~\cite{blumensath2008iterative} & 26.56 | 0.7770 & 22.46 | 0.5993 & 17.82 | 0.3874 & 9.661 | 0.1247 \\
FISTA (Convergence)~\cite{doi:10.1137/080716542} & 26.57 | 0.7771 & 22.73 | 0.6064 & 19.14 | 0.4282 & 15.90 | 0.3007 \\
ADMM (Convergence)~\cite{admm2010} &  26.50 | 0.7758 & 22.55 | 0.6022 &  18.58 | 0.4125 &  14.74 | 0.2721 \\
HCISTA & 30.41 | 0.8928 & 26.64 | 0.7812 & 23.58 | 0.6324 & 21.95 | 0.5456 \\
HCISTA-F  & \textbf{31.90} | \textbf{0.9157} & \textbf{27.75} | \textbf{0.8069} & \textbf{24.37} | \textbf{0.6615} & \textbf{22.24} | \textbf{0.5629} \\
\midrule
LISTA-CP-U~\cite{NIPS2018_8120} & 31.44 | 0.9092 & 27.50 | 0.8013 & 24.34 | 0.6608 & 22.30 | 0.5629 \\
HLISTA-CP & \textbf{32.06} | \textbf{0.9178} & \textbf{27.81} | \textbf{0.8095} & \textbf{24.50} | \textbf{0.6652} & \textbf{22.36} | \textbf{0.5652} \\
\midrule
LISTA-CPSS-U~\cite{NIPS2018_8120} & 31.42 | 0.9088 & 27.47 | 0.8007 & 24.26 | 0.6587 & 22.24 | 0.5614 \\
HLISTA-CPSS & \textbf{31.98} | \textbf{0.9168} & \textbf{27.80} | \textbf{0.8097} & \textbf{24.35} | \textbf{0.6618} & \textbf{22.25} | \textbf{0.5628} \\
\midrule
ALISTA~\cite{liu2018alista} & 30.91 | 0.9010 & 27.10 | 0.7915 & 23.25 | 0.6265 & 21.73 | 0.5448 \\
HALISTA & \textbf{31.48} | \textbf{0.9094} & \textbf{27.65} | \textbf{0.8053} & \textbf{23.95} | \textbf{0.6476} & \textbf{22.11} | \textbf{0.5583} \\
\midrule
Gated LISTA-U~\cite{Wu2020Sparse} & 31.64 | 0.9121 & 27.70 | 0.8068 & 24.49 | 0.6646 &  22.06 | 0.5566 \\
 HGLISTA &  \textbf{32.74} | \textbf{0.9260} & \textbf{28.36} | \textbf{0.8233} &  \textbf{24.51} | \textbf{0.6662} & \textbf{22.33} | \textbf{0.5664} \\
\midrule
ELISTA~\cite{li2021learned} &  30.47 | 0.8900 & 26.33 | 0.7738 &  23.38 | 0.6354 & 21.77 | 0.5479\\
 HELISTA & \textbf{32.59} | \textbf{0.9240} &  \textbf{28.33} | \textbf{0.8217} &  \textbf{24.68} | \textbf{0.6722} & \textbf{22.29} | \textbf{0.5644} \\
\bottomrule
\end{tabular}
\end{table*}

We first introduce simple DNNs into hybrid ISTA models. DNNs incorporated in HCISTA and HLISTA for CS are slightly different from those for sparse recovery, as the input images are two-dimensional signals.
%There is a slight difference of DNNs adopted in HCISTA and HLISTA for CS compared with sparse recovery, as the input images are two-dimensional signals. 
The DNNs incorporated in the $K$ iterations consist of three convolutional layers with the sizes of 9$\times$9$\times$1$\times$16, 9$\times$9$\times$16$\times$16, and 9$\times$9$\times$16$\times$1 (Kernel\_height$\times$Kernel\_width$\times$
In\_channel$\times$Out\_channel) and ReLU following the first two convolutional layers, \emph{i.e.}, the architecture of \texttt{Conv-ReLU-Conv-ReLU-Conv}. 
%The DNNs consist of three convolution layers with the sizes of 9$\times$9$\times$1$\times$16, 9$\times$9$\times$16$\times$16, and 9$\times$9$\times$16$\times$1 (Kernel\_height$\times$Kernel\_width$\times$In\_channel$\times$Out\_channel) and ReLU is utilized after the first two layers in each iteration. 
A residual connectivity is adopt to link $\mathbf{v}^{n}$ and $\mathbf{u}^{n}$ for $n=0,\cdots,K$. As we apply two-dimensional convolution operation for processing image signals, the update step of $\mathbf{u}^n$ in CS is different from Eq.~\eqref{un_sc} in sparse recovery. %Specifically, we have
\begin{equation}
\mathbf{u}^{n}=N_{\mathcal{W}^{n}}(\mathbf{v}^{n})=\mathbf{v}^{n}+\mathbf{D}_{inv}R'({\rm CvRL}_3(R(\mathbf{D}\mathbf{v}^{n}))),
\end{equation}
where $R(\cdot)$ (resp. $R'(\cdot)$) denotes the reshape operation that transforms the vectorized $\mathbf{D}\mathbf{v}^{n}$ (resp. square matrix) into the square matrix (resp. vectorized $\mathbf{D}\mathbf{v}^{n}$), and $\mathbf{D}_{inv}$ is a learned matrix initialized by the Moore-Penrose pseudo-inverse of $\mathbf{D}$. Note that $\mathbf{D}_{inv}$ is reused throughout all the iterations.

Since we have no ground-truth sparse signals as the target for the loss function in Eq.~\eqref{loss_function}, we utilize the same training strategy as LISTA-CP and LISTA-CPSS~\cite{NIPS2018_8120}. We first perform layer-wise pre-training. For the $n$th iteration, the loss function $L^{n}(\Theta^n, \mathbf{D}_{inv})$ is 
\begin{equation}\label{cs_loss1}
L^{n}(\Theta^n, \mathbf{D}_{inv})=\sum_{i=1}^{N}\left\| f_i-\mathbf{D}x_i^n(\Theta^n)\right\|_2^2, 
\end{equation}
%where $n$ denotes the iteration index and $\Theta^n$ represents all parameters in the $n$th and previous layers. 
where $\Theta^n$ represents all parameters in the $n$th and previous layers.
We introduce another fully-connected layer $\mathbf{W}_{fc}$ that is initialized by $\mathbf{D}$ in the last iteration and perform an end-to-end training with the loss function $L(\Theta, \mathbf{D}_{inv}, \mathbf{W}_{fc})$.
\begin{equation}\label{cs_loss2}
L(\Theta, \mathbf{D}_{inv}, \mathbf{W}_{fc})=\sum_{i=1}^{N}\left\| f_i-\mathbf{W}_{fc}x_i^K(\Theta)\right\|_2^2 
\end{equation}
%\begin{align}
%& L^{n}(\Theta^n, \mathbf{D}_{inv})=\sum_{i=1}^{N}\left\| f_i-\mathbf{D}x_i^n(\Theta^n)\right\|_2^2 \label{cs_loss1} \\
%& L(\Theta, \mathbf{D}_{inv}, W_{fc})=\sum_{i=1}^{N}\left\| f_i-\mathbf{W}_{fc}x_i^K(\Theta)\right\|_2^2 \label{cs_loss2}
%\end{align}
Similar to sparse recovery, we also evaluate complicated DNNs in compressive sensing. We utilize the large-scale Flickr~30k dataset  ~\cite{young2014image} containing 31783 images from \textit{Flickr}\footnote{The website is \textit{https://www.flickr.com}.} for network training. 
%Due to page limits, we only show results with simple DNNs here. 
Refer to Appendix~D for more details and results on evaluations with complicated DNNs.

We first compare classical ISTA, FISTA and ADMM with HCISTA and HCISTA-F. For ISTA, FISTA, and ADMM, $\lambda=0.0025$, as natural images are not exactly sparse and a large $\lambda$ leads to poor reconstruction performance. We set $C_{\lambda}$ to 1 for HCISTA and $\lambda^0$ to 0.05 for HCISTA and HCISTA-F. Table~\ref{T4} reports the reconstruction performance on \emph{Set11} and \emph{BSD500}, respectively. Here, ISTA, FISTA, and ADMM are iterated until convergence and HCISTA unfolds 16 iterations. HCISTA and HCISTA-F are shown to outperform classical ISTA, especially at the low MR of 0.04. This fact implies that the free-form DNNs incorporated in HCISTA benefit the recovery most at extremely low MRs and complements the evaluations in sparse recovery at the fixed MR of 0.50. %\request{which is not revealed in sparse recovery experiments}. 
HCISTA achieves an enhanced convergence speed than ISTA, which validates Theorem~\ref{theorem3}. HCISTA-F can achieve a linear convergence rate, as it outperforms HCISTA and ISTA and is comparable with LISTA and HLISTA. 
%The results of Set11 and BSD500 are listed in Tables~\ref{T4} and \ref{T5}, where we implement ISTA until convergence and HCISTA with 16 iterations. HCISTA and HCISTA-F outperform classical ISTA, especially at the low MR of 0.04. This fact implies that the free-form DNNs of HCISTA benefit the recovery most at extremely low MRs, which is not revealed in sparse recovery experiments. HCISTA achieves an enhanced convergence speed than ISTA, which validates Theorem~\ref{theorem3}. HCISTA-F approximates or attains a linear convergence rate, as the recovery performance is superior to HCISTA and ISTA, and comparable with LISTA and HLISTA, as illustrated in Table~\ref{T4} and \ref{T5}. 

We further compare HLISTA-CP, HLISTA-CPSS, HALISTA, HGLISTA and HELISTA with corresponding baselines with untied weights. All the models are built with 16 iterations. For LISTA-CPSS-U and HLISTA-CPSS, we choose $p=0.2$ and $p_{max}=4$. Table~\ref{T4} shows that the proposed HLISTA models are superior at all the four MRs and the performance gains grow when MR increases.  HLISTA-CP/CPSS and HGLISTA have much fewer learnable parameters than LISTA-CP-U/CPSS-U and Gated LISTA-U, which corroborates the efficiency of the proposed hybrid ISTA. 

% \begin{figure*}[!t]
% \renewcommand{\baselinestretch}{1.0}
% \centering
% \includegraphics[width=7.1in]{figures/imgs.pdf}
% \caption{Reconstruction of test images \emph{Foreman}, \emph{Lena}, \emph{Monarch}, and \emph{Cameraman} from \emph{Set11} at the MRs of 0.5, 0.25, 0.10, and 0.04, respectively.
% }\label{imgs}
% \end{figure*}

\begin{table}[!t]
  \renewcommand{\baselinestretch}{1.2}
  \renewcommand{\arraystretch}{1.2}
  \setlength{\tabcolsep}{2pt}
  \centering
  \caption{Comparison with ISTA-Net$^+$ of average PSNR (dB)|SSIM on \emph{BSD500} (50 images for test) obtained at the measurement rates (MRs) of 0.04, 0.10, 0.25 and 0.50. The best performance is labeled in bold and the second best is underlined.}
  \label{T6}
  \begin{tabular}{lccccc}
  \toprule
  Methods & MR=0.50 & MR=0.25 & MR=0.10 & MR=0.04 \\
  \midrule
  ISTA-Net$^+$ & 31.26|0.9046 & 27.58|0.8005  & 24.33|0.6599 & 22.26|\underline{0.5643} \\
  HCISTA & 30.86|0.8998 & 27.29|0.7901  & 24.13|0.6532 & 22.19|0.5583 \\
  HLISTA-CP  & \textbf{32.78}|\textbf{0.9264} & \textbf{28.25}|\textbf{0.8202}  & \textbf{24.60}|\textbf{0.6704} & \textbf{22.39}|\textbf{0.5672} \\
   \hline
  ISTA-Net$^+$-T & 30.13|0.8823 & 26.97|0.7824 & 24.00|0.6449 & 22.05|0.5557 \\
  HCISTA-T & 30.46|0.8904 & 27.21|0.7876  & 23.82|0.6412 & 22.15|0.5545 \\
  HLISTA-CP-T  & \textbf{32.78}|\underline{0.9263} & \underline{28.08}|\underline{0.8150}  & \underline{24.46}|\underline{0.6655} & \underline{22.28}|\underline{0.5643} \\
  \bottomrule
  \end{tabular}
\end{table}

Interestingly, HCISTA is competitive with LISTA and HLISTA models in natural image CS but is inferior in sparse recovery.
%Interestingly, the recovery performance of HCISTA lags far behind LISTA and HLISTA models in sparse recovery experiments, whereas is comparable to those models in CS experiments. 
The difference probably comes from whether the incorporated DNNs are suitable for the tasks. CNNs tend to be particularly suitable for processing natural images. Despite no constraint on the architectures of DNNs in hybrid ISTA models, properly selected architectures of DNNs can effectively improve the performance for specific tasks.

Due to page limits, visualization of reconstructed images is shown in Appendix~D.
As hybrid ISTA models are built based on the corresponding baselines, we do not compare them with other CS methods in this section. Nevertheless, the proposed hybrid ISTA still achieves the state-of-the-art reconstruction performance as one can see more results of other CS methods in~\cite{NIPS2018_8120} in the same training setting.

\subsubsection{Comparison with ISTA-Net$^+$}
\label{sec:6.2.2}
As discussed in Section~\ref{sec5.3}, hybrid ISTA can be viewed as a general framework that benefits almost all DNNs for solving linear inverse problems with convergence guarantee. In this section, we evaluate the proposed framework incorporating with ISTA-Net$^+$~\cite{zhang2018ista}. 
ISTA-Net$^+$ is constructed using successive blocks with the same architecture. Specifically, the forward propagation in the $n$th block can be formulated by
\begin{align}\label{C1}
& \mathbf{r}^{n} =  \mathbf{x}^{n} - \rho^n \Phi^T (\Phi \mathbf{x}^{n} - \mathbf{b}), \nonumber \\
& \mathbf{x}^{n+1} = \mathbf{r}^{n} + \mathcal{P}^n(\mathbf{r}^{n}),
\end{align}
%where $\Phi$ is the measurement matrix, $\mathbf{b}$ is the measurements, $\rho^n$ is the learned step size, and $\mathcal{P}^n$ denotes the network with architecture of \texttt{Conv1-Conv2-ReLU-Conv3-Soft}
where $\rho^n$ is the learned step size, and $\mathcal{P}^n$ denotes the network with the architecture of \texttt{Conv1-}
\texttt{Conv2-ReLU-Conv3-Soft-Conv4-ReLU-Conv5-Conv6}. Here, \texttt{Soft} is $\mathcal{S}_{\theta^{n}}$ with learnable threshold $\theta^{n}$. The size of convolutional layers \texttt{Conv2} and \texttt{Conv5} is set to 3$\times$3$\times$32$\times$32, and \texttt{Conv1} and \texttt{Conv6} are 3$\times$3$\times$1$\times$32 and 3$\times$3$\times$32$\times$1, respectively. Let us define $\mathcal{P}_{f}$ as the operation \texttt{Conv2-ReLU-Conv3} and $\mathcal{P}_{b}$ as \texttt{Conv4-ReLU-Conv5}. The symmetry constraint $\mathcal{P}_{f}\cdot\mathcal{P}_{b}=\mathbf{I}$ is incorporated into the loss function during network training. We utilize each block formulated in Eq.~\eqref{C1} as the DNN in each iteration of hybrid ISTA models and compare the performance with ISTA-Net$^+$. For fair comparison, we adopt the same training strategy and dataset as Section~\ref{sec:6.2.1}.

Table~\ref{T6} reports the reconstruction performance on \emph{BSD500}. For simplicity, the models are constructed with $K=6$ iterations. `ISTA-Net$^+$-T' represents that the parameters in each network are shared across all the iterations, as discussed in~\cite{zhang2018ista}. The same setting is adopted in corresponding hybrid ISTA models with postfix `T'. Table~\ref{T6} shows that the HLISTA-CP and HLISTA-CP-T outperform HCISTA and HCISTA-T and ISTA-Net$^+$ and ISTA-Net$^+$-T. It is worth mentioning that HLISTA-CP-T with tied weights is superior to ISTA-Net$^+$ with untied weights under all the MRs. This fact corroborates the effectiveness of the HLISTA. HCISTA is inferior to ISTA-Net$^{+}$, as HCISTA leverages the pre-computed $\mathbf{A}^T$ and contrains $t$ and $\lambda$ in Eq.~\eqref{e4}. Thus, the efficiency of inserted DNNs might be obstructed in HCISTA. Nevertheless, it is more reasonable to view HCISTA as a conventional iterative algorithms, since it improves classical ISTA with a provable convergence that is independent of network training. Furthermore, hybrid ISTA models are less affected by weight sharing. For example, at the MR of 0.50, ISTA-Net$^{+}$-T suffers a PSNR loss of 1.13 dB in comparison to ISTA-Net$^{+}$, while the PSNR loss is only 0.40 dB for HCISTA-T and HLISTA-CP and HLISTA-CP-T are equivalent.

\section{Conclusion}\label{sec:7}
Inspired by ISTA with pre-computed and learned parameters, we develop hybrid ISTA to incorporate them with free-form DNNs. We first develop HCISTA to unfold classical ISTA based on DNNs with arbitrary architectures, which improves the efficiency and flexibility of ISTA. We prove that HCISTA retains and sometimes enhances the convergence rate of ISTA in theory. Moreover, we generalize the hybrid framework to LISTA, dubbed HLISTA, to introduce learned parameters with free architectures while still preserving the linear convergence rate. The proposed hybrid ISTA can be viewed as a general framework applied to almost all DNNs for solving inverse problems and endow the empirically constructed DNNs with theoretical convergence. Extensive experiments endorse the theories and demonstrate an improved performance and convergence rate. We believe that the methodology in this paper can be an interesting direction for designing interpretable DNNs. 

\newpage

\appendices
\section{Proofs for HCISTA}
In this section, we prove Lemmas~1 and 2 and Theorems~1-3 to guarantee the convergence of HCISTA and develop the convergence rate. 

\subsection{Detailed Description of Proposition~1}
In this subsection, we elaborate the properties of $F$, $f$, and $g$ shown in Proposition~1. According to Definition~1, we have $f(\mathbf{x})=\frac{1}{2}\|\mathbf{Ax}-\mathbf{b}\|_2^2$, $g(\mathbf{x})=\lambda\|\mathbf{x}\|_1$, and $F(\mathbf{x})=f(\mathbf{x})+g(\mathbf{x})$.

1) $f$ is a smooth convex function with $L$-Lipschitz continuous gradient. Since $\nabla f(\mathbf{x})=\mathbf{A}^T(\mathbf{Ax}-\mathbf{b})$, $\|\nabla f(\mathbf{x})-\nabla f(\mathbf{y})\|_2\leq\|\mathbf{A}\|_2^2\|\mathbf{x}-\mathbf{y}\|_2$. For arbitrary $\mathbf{x},\mathbf{y}\in\mathbb{R}^N$, there exists $L\geq\|\mathbf{A}\|_2^2>0$ such that
\begin{equation}\label{s24}
\|\nabla f(\mathbf{x})-\nabla f(\mathbf{y})\|_2\leq L\|\mathbf{x}-\mathbf{y}\|_2
\end{equation}
For $f$ with $L$-Lipschitz continuous gradient, we have
\begin{equation}\label{s23}
f(\mathbf{x})\le f(\mathbf{y})+\langle\nabla f(\mathbf{y}),\mathbf{y}-\mathbf{x}\rangle+\frac{L}{2}\|\mathbf{x}-\mathbf{y}\|_2^2.
\end{equation}

2) $g(\mathbf{x})$ is obviously convex and continuous. However, $g(\mathbf{x})$ is not differentiable for $\mathbf{x}\in\mathbb{R}^N$ when its $i$th element $x_i=0$ for arbitrary $i=1,\cdots,N$.

3) For $f(\mathbf{x})=\frac{1}{2}\|\mathbf{Ax}-\mathbf{b}\|_2^2$, the domain $\mathrm{dom}f=\{\mathbf{x}\in\mathbb{R}^N|f(\mathbf{x})<+\infty\}$ is nonempty and $f(\mathbf{x})\geq 0>-\infty$. Same results can be obtained for $g(\mathbf{x})=\lambda\|\mathbf{x}\|_1$.

4) For $\mathbf{x}\in\mathbb{R}^N$, $\|\mathbf{x}\|_1\geq\|\mathbf{x}\|_2$ and we can achieve the equality if and only $\mathbf{x}=\mathbf{0}$. Since $f(\mathbf{x})\geq 0$, $F(\mathbf{x})\geq g(\mathbf{x})\geq \lambda\|\mathbf{x}\|_2$. Therefore, $F(\mathbf{x})\rightarrow\infty$ when $\|\mathbf{x}\|_2\rightarrow\infty$.

5) The semi-algebraic function is defined in Definition~\ref{app_defn2}.
\begin{defn}[Semi-algebraic Function]\label{app_defn2} A subset $\mathcal{G}\in\mathbb{R}^N$ is a real semi-algebraic set if there exists a finite number of real polynomial functions $\psi_{ij},\psi_{ij}':\mathbb{R}^N\rightarrow\mathbb{R}$ such that
\begin{equation}
\mathcal{G}=\bigcup_{j=1}^p\bigcap_{i=1}^p\{\mathbf{x}\in\mathbb{R}^N:\psi_{ij}(\mathbf{x})=0, \psi_{ij}'(\mathbf{x})<0\}.    
\end{equation}
A function $\Phi:\mathbb{R}^N\rightarrow\mathbb{R}$ is a semi-algebraic function if its graph $\{(\mathbf{x},\Phi(\mathbf{x}))\in\mathbb{R}^{N+1}\}$ is a semi-algebraic set.
\end{defn}

For the Lasso problem defined in Eq.~(1), $F(\mathbf{x})$ is in the form of $\|\mathbf{x}\|_p$, $p\geq 0$ and is a semi-algebraic function~[48]. According to~[47], the semi-algebraic function $F(\mathbf{x})$ satisfies the Kurdyka-Łojasiewicz (KŁ) property. The definition of KŁ property is provided in Definition~\ref{defn1}. %Specifically, we give the definition of KŁ property as follows.
\begin{defn}[{Kurdyka-Łojasiewicz (KŁ) Property~[47]}]\label{defn1}
A function $\Phi$: $\mathbb{R}^{N}\rightarrow (-\infty, +\infty]$ is said to have the KŁ property at $\bar{\mathbf{u}}\in {\rm dom} \partial\Phi := \{\mathbf{x}\in \mathbb{R}^{N}: \partial\Phi\neq \emptyset\}$ if there exists $\eta\in(0, +\infty]$, a neighborhood $\mathcal{U}$ of $\bar{\mathbf{u}}$ and a function $\psi\in\Psi_{\eta}$, such that for all $\mathbf{u}\in\mathcal{U}\bigcap\{\mathbf{u}\in\mathbb{R}^{N}:\Phi(\bar{\mathbf{u}})<\Phi(\mathbf{u})<\Phi(\bar{\mathbf{u}})+\eta\}$, the following inequality holds
\begin{equation}
\psi'(\Phi(\mathbf{u})-\Phi(\bar{\mathbf{u}})){\rm dist}(0, \partial f(\mathbf{u}))\geq 1, 
\end{equation}
where ${\rm dist}(0, \partial\Phi(\mathbf{u}))=\inf\{\|\mathbf{x}_{*}\|: \mathbf{x}_*\in \partial\Phi(\mathbf{u})\}$, and $\Psi_{\eta}$ stands for a class of function $\psi: [0, \eta)\rightarrow \mathbb{R}^{+}$ satisfying: (1) $\psi$ is concave and $C^1$ on $(0, \eta)$; (2) $\psi$ is continuous at $0$ and $\psi(0)=0;$ and (3) $\psi'(x)>0$ for arbitrary $x\in (0, \eta)$.
\end{defn}

\subsection{Proof of Lemma~1}
For arbitrary $n\in\mathbb{N}$, we have $\frac{1}{4\delta^n\|\mathbf{A}\|_2^2}\leq t^{n}\leq \frac{1}{\|\mathbf{A}\|_2^2}$ and $1/4<\delta^n<1/2$ at the $n$th iteration. For the under-complete basis matrix $\mathbf{A}\in\mathbb{R}^{M\times N}$, $\|\mathbf{A}\|_{2}^{2}$ is the maximum eigenvalue of $\mathbf{A}^{T}\mathbf{A}$, and is equivalent to the smallest Lipschitz constant $L$ of the gradient of $f(\mathbf{x})=\frac{1}{2}\|\mathbf{Ax}-\mathbf{b}\|_{2}^{2}$.

%According to Proposition~1, $f$ has $L$-Lipschitz continuous gradient, which means that 
%\begin{equation}
%f(\mathbf{x})\leq f(\mathbf{y})+\left\langle \nabla f(\mathbf{y}), \mathbf{x}-\mathbf{y} \right\rangle + \frac{L}{2}\|\mathbf{x}-\mathbf{y}\|_{2}^{2}.
%\end{equation}
Let us define 
%Thus, by letting $0< t\leq \frac{1}{L}$ and 
\begin{equation}\label{sss1}
Q(\mathbf{x},\mathbf{y})=f(\mathbf{y})+\left\langle \nabla f(\mathbf{y}), \mathbf{x}-\mathbf{y} \right\rangle + \frac{1}{2t}\|\mathbf{x}-\mathbf{y}\|_{2}^{2}+g(\mathbf{x}).
\end{equation}
According to Eq.~\eqref{s23}, we have $F(\mathbf{x})\leq Q(\mathbf{x},\mathbf{y})$ when $0< t\leq 1/L$. Since the regularization parameters $\{\lambda^n\}_{n\in\mathbb{N}}$ are adaptive with $n$, $F(\mathbf{x})\leq Q(\mathbf{x},\mathbf{y})$ holds only for the same $\lambda^n$ in one iteration. Without loss of generality, we have that $F(\mathbf{w}^n)\leq Q(\mathbf{w}^n,\mathbf{u}^n)$ and $F(\mathbf{v}^n)\leq Q(\mathbf{v}^n,\mathbf{x}^n)$ for arbitrary $n\in\mathbb{N}$. 
%However, as we introduce adaptive constrained regularization parameters $\{\lambda^n\}_{n\in\mathbb{N}}$ and reformulate Eq.~(1), $F(\mathbf{x})\leq Q(\mathbf{x},\mathbf{y})$ holds only for the same $\lambda^n$ in one iteration. Hence, we have that $F(\mathbf{w}^n)\leq Q(\mathbf{w}^n,\mathbf{u}^n)$ and $F(\mathbf{v}^n)\leq Q(\mathbf{v}^n,\mathbf{x}^n)$.
According to Eq.~\eqref{sss1}, $Q(\mathbf{x},\mathbf{u}^n)$ admits a unique minimizer with $t^n$ and $\lambda^n$.
\begin{align}\label{sss6}
\mathbf{w}^n :=&\mathop{\arg\min}_{\mathbf{x}}Q(\mathbf{x},\mathbf{u}^n)\nonumber\\
%=&\mathop{\arg\min}_{\mathbf{x}}\left\{ \left\langle \nabla f(\mathbf{u}^n), \mathbf{x}-\mathbf{u}^n \right\rangle + \frac{1}{2t^n}\|\mathbf{x}-\mathbf{u}^n\|_{2}^{2}+g(\mathbf{x})\right\} \nonumber \\
=&\mathop{\arg\min}_{\mathbf{x}}\left\{\frac{1}{2t^n}\|\mathbf{x}-(\mathbf{u}^n-t^n\nabla f(\mathbf{u}^n))\|_2^2+g(\mathbf{x})\right\}\nonumber \\
=&\mathcal{S}_{\lambda^n t^{n}}(\mathbf{u}^{n}-t^{n}\nabla f(\mathbf{u}^{n}))
\end{align}
%Eq.~\eqref{sss8} shows that $Q(\mathbf{x},\mathbf{u}^n)$ admits a unique minimizer with $t^n$ and $\lambda^n$. 
%\begin{equation}\label{sss6}
%\mathbf{w}^n := \mathop{\arg\min}_{\mathbf{x}\in\mathbb{N}}Q(\mathbf{x},\mathbf{u}^n) ,
%\end{equation}
%which can be proved as follows:
%\begin{align}
%&\mathop{\arg\min}_{\mathbf{x}}Q(\mathbf{x},\mathbf{u}^n)\nonumber\\
%=&\mathop{\arg\min}_{\mathbf{x}}\left\{ \left\langle \nabla f(\mathbf{u}^n), \mathbf{x}-\mathbf{u}^n \right\rangle + \frac{1}{2t^n}\|\mathbf{x}-\mathbf{u}^n\|_{2}^{2}+g(\mathbf{x})\right\} \nonumber \\
%=&\mathop{\arg\min}_{\mathbf{x}}\left\{\frac{1}{2t^n}\|\mathbf{x}-(\mathbf{u}^n-t^n\nabla f(\mathbf{u}^n))\|_2^2+g(\mathbf{x})\right\}\nonumber \\
%=&\mathcal{S}_{\lambda^n t^{n}}(\mathbf{u}^{n}-t^{n}\nabla f(\mathbf{u}^{n}))\nonumber \\
%=&\mathbf{w}^n
%\end{align}
%Thus, we obtain Eq.~\eqref{sss4} by using the optimality conditions for $Q(\mathbf{x},\mathbf{u}^n)$.
In Eq.~\eqref{sss6}, we obtain from the optimality condition for $Q(\mathbf{x},\mathbf{u}^n)$ that
\begin{equation}\label{sss4}
    \nabla f(\mathbf{u}^n)+\frac{1}{t^n}(\mathbf{w}^n-\mathbf{u}^n)+\partial g(\mathbf{w}^n)=0.
\end{equation}

%Since $F(\mathbf{w}^n)\leq Q(\mathbf{w}^n,\mathbf{u}^n)$, we have that 
%\begin{equation}\label{sss2}
%F(\mathbf{x}^n)-F(\mathbf{w}^n)\geq F(\mathbf{x}^n)-Q(\mathbf{w}^n,\mathbf{u}^n)
%\end{equation}
For convex $f$ and $g$ defined in Eq.~(1), we have
\begin{equation}\label{sss8}
f(\mathbf{x}^n)\geq f(\mathbf{u}^n)+\left\langle \nabla f(\mathbf{u}^n), \mathbf{\mathbf{x}^n}-\mathbf{\mathbf{u}^n} \right\rangle,
\end{equation}
and
\begin{equation}\label{sss9}
g(\mathbf{x}^n)\geq g(\mathbf{w}^n)+\left\langle \partial g(\mathbf{w}^n) , \mathbf{x}^n-\mathbf{w}^n \right\rangle.
\end{equation}
From Eq.~\eqref{sss8} and Eq.~\eqref{sss9}, 
\begin{align}\label{sss3}
F(\mathbf{x}^n) \geq f(\mathbf{u}^n)&+\left\langle \nabla f(\mathbf{u}^n), \mathbf{\mathbf{x}^n}-\mathbf{\mathbf{u}^n}  \right\rangle \nonumber\\
&+g(\mathbf{w}^n)+\left\langle \partial g(\mathbf{w}^n) , \mathbf{x}^n-\mathbf{w}^n \right\rangle.
\end{align}
Since $F(\mathbf{w}^n)\leq Q(\mathbf{w}^n,\mathbf{u}^n)$, we have that 
\begin{equation}\label{sss2}
F(\mathbf{x}^n)-F(\mathbf{w}^n)\geq F(\mathbf{x}^n)-Q(\mathbf{w}^n,\mathbf{u}^n)
\end{equation}
Considering Eq.~\eqref{sss1} and Eq.~\eqref{sss3}, we obtain from Eq.~\eqref{sss2} that 
\begin{align}\label{sss7}
F(\mathbf{x}^n)-F(\mathbf{w}^n)\geq &\left\langle \nabla f(\mathbf{u}^n)+\partial g(\mathbf{w}^n) , \mathbf{x}^n-\mathbf{w}^n \right\rangle\nonumber\\
&- \frac{1}{2t^n}\|\mathbf{w}^n-\mathbf{u}^n\|_2^2.
\end{align}
Considering the optimality condition given in Eq.~\eqref{sss4}, we find from Eq.~\eqref{sss7} that 
\begin{align}\label{add1}%\label{sss7}
& F(\mathbf{x}^n)-F(\mathbf{w}^n)\nonumber\\
&\quad\geq\frac{1}{t^n}\left\langle \mathbf{u}^n-\mathbf{w}^n, \mathbf{x}^n-\mathbf{w}^n \right\rangle- \frac{1}{2t^n}\|\mathbf{w}^n-\mathbf{u}^n\|_2^2 \nonumber\\
%&\quad=\frac{1}{t^{n}}\langle\mathbf{u}^{n}-\mathbf{x}^{n},\mathbf{w}^{n}-\mathbf{u}^{n}\rangle+\frac{1}{2t^{n}}\|\mathbf{w}^{n}-\mathbf{u}^{n}\|_{2}^{2} 
&\quad=\frac{1}{2t^n}\langle\mathbf{u}^{n}-\mathbf{w}^{n},(\mathbf{x}^{n}-\mathbf{u}^{n})+(\mathbf{x}^{n}-\mathbf{w}^{n})\rangle\nonumber\\
&\quad=\frac{1}{2t^{n}}(\|\mathbf{w}^{n}-\mathbf{x}^{n}\|_{2}^{2} -\|\mathbf{u}^{n}-\mathbf{x}^{n}\|_{2}^{2}).
\end{align}

%Similarly, we obtain Eq.~\eqref{sss5} from $F(\mathbf{v}^n)\leq Q(\mathbf{v}^n,\mathbf{x}^n)$ according to Eq.~\eqref{sss6}$-$\eqref{sss7}.
Similarly, for $\mathbf{v}^n$, we obtain $F(\mathbf{v}^n)\leq Q(\mathbf{v}^n,\mathbf{x}^n)$ and
\begin{equation}
\mathbf{v}^n:=\mathop{\arg\min}_{\mathbf{v}}Q(\mathbf{v},\mathbf{x}^n)=\mathcal{S}_{\lambda^n t^{n}}(\mathbf{x}^{n}-t^{n}\nabla f(\mathbf{x}^{n})).
\end{equation}
The optimality condition for $\mathbf{v}^n$ is
\begin{equation}\label{sss5}
0\in\nabla f(\mathbf{x}^n)+\frac{1}{t^n}(\mathbf{v}^n-\mathbf{x}^n)+\partial g(\mathbf{v}^n)
\end{equation}
Since $g$ is convex, we have
\begin{equation}
g(\mathbf{x}^n)\ge g(\mathbf{v}^n)+\langle\partial g(\mathbf{v}^n),\mathbf{x}^n-\mathbf{v}^n\rangle.
\end{equation}
%Similarly, we can obtain $F(\mathbf{v}^n)\leq Q(\mathbf{v}^n,\mathbf{x}^n)$ that
Therefore, 
\begin{align}\label{e12}%\label{sss5}
F(\mathbf{x}^n)-F(\mathbf{v}^n)\geq F(\mathbf{x}^n)-Q(\mathbf{v}^n,\mathbf{x}^n)
\geq\frac{1}{2t^{n}}\|\mathbf{v}^{n}-\mathbf{x}^{n}\|_{2}^{2}.
\end{align}
As a result, we draw Lemma~1.

%Hereby, we prove Lemma~1, Theorem~1 and 2. Note that we focus on Lasso problem in the proof, but actually Lemma~1, Theorem~1 and 2 can accommodate arbitrary $F(x)$ that satisfies Assumption~\ref{assum1} $\sim$ \ref{assum5}.

\subsection{Proof of Theorem~1}
According to Eq.~(5), $\mathbf{x}^{n+1} = \alpha^{n}\mathbf{v}^{n}+(1-\alpha^{n})\mathbf{w}^{n}$ for arbitrary $n\in\mathbb{N}$. Since $F(\mathbf{x})$ is a convex function, we have for $\mathbf{x}^{n+1}$ and $\mathbf{x}^{n}$,  \begin{align}\label{e13}
&F(\mathbf{x}^{n})-F(\mathbf{x}^{n+1}) \nonumber \\
&\quad=F(\mathbf{x}^{n})-F(\alpha^{n}\mathbf{v}^{n}+(1-\alpha^{n})\mathbf{w}^{n})\nonumber\\
&\quad\geq\alpha^{n}[F(\mathbf{x}^{n})-F(\mathbf{v}^{n})]+(1-\alpha^{n})[F(\mathbf{x}^{n})-F(\mathbf{w}^{n})]
\end{align}
According to Lemma~1, we obtain from Eq.~\eqref{add1} and Eq.~\eqref{e12} that 
\begin{align}
&F(\mathbf{x}^{n})-F(\mathbf{x}^{n+1}) \nonumber\\
&\ \geq \frac{\alpha^{n}}{2t^{n}}\|\mathbf{v}^{n}-\mathbf{x}^{n}\|_{2}^{2}+ \frac{1-\alpha^{n}}{2t^{n}}\left( \|\mathbf{w}^{n}-\mathbf{x}^{n}\|_{2}^{2}-\|\mathbf{u}^{n}-\mathbf{x}^{n}\|_{2}^{2} \right).
\end{align}
%According to Eq.~(5), we decompose $\|\mathbf{x}^{n+1}-\mathbf{x}^{n}\|_{2}^{2}$ using $\mathbf{x}^{n+1} = \alpha^{n}\mathbf{v}^{n}+(1-\alpha^{n})\mathbf{w}^{n}$.
$\|\mathbf{x}^{n+1}-\mathbf{x}^{n}\|_{2}^{2}$ can be decomposed by
\begin{align}\label{add2}
\|\mathbf{x}^{n+1}-\mathbf{x}^{n}\|_{2}^{2}&=\|\alpha^{n}(\mathbf{v}^{n}-\mathbf{x}^{n})+(1-\alpha^{n})(\mathbf{w}^{n}-\mathbf{x}^{n})\|_{2}^{2} \nonumber\\
&=(\alpha^{n})^{2}\|\mathbf{v}^{n}-\mathbf{x}^{n}\|_{2}^{2}+(1-\alpha^{n})^{2}\|\mathbf{w}^{n}-\mathbf{x}^{n}\|_{2}^{2} \nonumber \\
&\quad+2\alpha^{n}(1-\alpha^{n})\langle\mathbf{v}^{n}-\mathbf{x}^{n}, \mathbf{w}^{n}-\mathbf{x}^{n}\rangle.
\end{align}
We can further rewrite $\langle\mathbf{v}^{n}-\mathbf{x}^{n},\mathbf{w}^{n}-\mathbf{x}^{n}\rangle$ as
\begin{align}\label{add3}
&2\langle\mathbf{v}^{n}-\mathbf{x}^{n},\mathbf{w}^{n}-\mathbf{x}^{n}\rangle \nonumber\\ &\quad=\|\mathbf{w}^{n}-\mathbf{x}^{n}\|_{2}^{2}+\|\mathbf{v}^{n}-\mathbf{x}^{n}\|_{2}^{2}-\|\mathbf{w}^{n}-\mathbf{v}^{n}\|_{2}^{2}.
\end{align}
Combining Eq.~\eqref{add2} and Eq.~\eqref{add3}, we have
\begin{align}\label{add4}
\|\mathbf{x}^{n+1}-\mathbf{x}^{n}\|_{2}^{2}
&=\alpha^{n}\|\mathbf{v}^{n}-\mathbf{x}^{n}\|_{2}^{2}+(1-\alpha^{n})\|\mathbf{w}^{n}-\mathbf{x}^{n}\|_{2}^{2} \nonumber \\
&\quad-\alpha^{n}(1-\alpha^{n})\|\mathbf{w}^{n}-\mathbf{v}^{n}\|_{2}^{2}.
\end{align}
Now we consider the cases of $\|\mathbf{v}^n-\mathbf{x}^n\|_2^2=0$ and $\|\mathbf{v}^n-\mathbf{x}^n\|_2^2>0$, respectively.

i) $\|\mathbf{v}^n-\mathbf{x}^n\|_2^2=0$ for $n\in\mathcal{T}$. We obtain from Eq.~(7) that $\alpha^n=1$. According to Eq.~(5), $\mathbf{x}^{n+1}=\mathbf{x}^n$ for $\mathbf{v}^n=\mathbf{x}^n$ and $\alpha^n=1$. When $\mathbf{x}^{n+1}=\mathbf{x}^n$, we have $F(\mathbf{x}^n)=F(\mathbf{x}^{n+1})$ for the Lasso problem defined in Eq.~(1). Therefore, for arbitrary $n\in\mathcal{T}$, 
\begin{equation}\label{ls1}
F(\mathbf{x}^{n})-F(\mathbf{x}^{n+1})= \frac{1}{4}\|\mathbf{A}\|_2^2\|\mathbf{x}^{n+1}-\mathbf{x}^{n}\|_{2}^{2}= 0.
\end{equation}

ii) $\|\mathbf{v}^n-\mathbf{x}^n\|_2^2\neq 0$ for $n\notin\mathcal{T}$. In this case, it follows from Eq.~(7) that
\begin{equation}\label{es29}
\alpha^{n}\left(\frac{1}{2t^{n}}-\delta^n\|\mathbf{A}\|_2^2\right)\|\mathbf{v}^{n}-\mathbf{x}^{n}\|_{2}^{2} -\frac{1-\alpha^{n}}{2t^{n}}\|\mathbf{u}^{n}-\mathbf{x}^{n}\|_{2}^{2}\geq 0.
\end{equation}
Thus, we can obtain from Eq.~\eqref{e13} and Eq.~\eqref{add4} that 
\begin{equation}\label{add6-1}
F(\mathbf{x}^{n})-F(\mathbf{x}^{n+1})\geq \delta^n\|\mathbf{A}\|_2^2\|\mathbf{x}^{n+1}-\mathbf{x}^{n}\|_{2}^{2}\geq 0.
\end{equation}
Since $1/4<\delta^n<1/2$, we obtain from Eq.~\eqref{add6-1} that
\begin{equation}\label{add6}
F(\mathbf{x}^{n})-F(\mathbf{x}^{n+1})>\frac{1}{4}\|\mathbf{A}\|_2^2\|\mathbf{x}^{n+1}-\mathbf{x}^{n}\|_{2}^{2}\geq 0.
\end{equation}
Eq.~\eqref{ls1} and Eq.~\eqref{add6} show that the sequence $\{F(\mathbf{x}^{n})\}_{n\in \mathbb{N}}$ is non-increasing. Since $f(\mathbf{x})$ and $g(\mathbf{x})$ are proper, $F(\mathbf{x})$ is also bounded. Therefore, the sequence $\{F(\mathbf{x}^{n})\}_{n\in \mathbb{N}}$ converges to $F_{*}$, \emph{i.e.}, 
\begin{equation}\label{add17}
\lim_{n\rightarrow\infty}F(\mathbf{x}^{n})=F_{*}.
\end{equation}
Under the Assumption 1 that $F$ is coercive, we have that $\{\mathbf{x}^{n}\}_{n\in\mathbb{N}}$ is bounded and thus have accumulation points. Because $\{F(\mathbf{x}^n)\}_{n\in\mathbb{N}}$ is non-increasing, $F$ achieves the same value $F_*$ at all the accumulation points. We sum up $\|\mathbf{x}^{n+1}-\mathbf{x}^{n}\|_{2}^{2}$ over $n\in\mathbb{N}$ and further obtain from Eq.~\eqref{ls1} and Eq.~\eqref{add6} that
\begin{equation}\label{add7}
\sum_{n=0}^{\infty}\|\mathbf{x}^{n+1}-\mathbf{x}^{n}\|_{2}^{2}\leq\frac{4}{\|\mathbf{A}\|_2^2}\left[F(\mathbf{x}^{0})-F_{*}\right]< + \infty,
\end{equation}
%\begin{align}
%&\sum_{n=0}^{\infty}\|\mathbf{x}^{n+1}-\mathbf{x}^{n}\|_{2}^{2}\nonumber\\
%&\quad=\sum_{n\not\in T}\|\mathbf{x}^{n+1}-\mathbf{x}^{n}\|_{2}^{2} + \sum_{n\in T}\|\mathbf{x}^{n+1}-\mathbf{x}^{n}\|_{2}^{2} \nonumber \\
%&\quad\leq \frac{4}{\|\mathbf{A}\|_2^2}\left[F(\mathbf{x}^{0})-F_{*}\right]< + \infty,
%\end{align}
Eq.~\eqref{add7} implies that, when $n\rightarrow\infty$, 
\begin{equation}\label{s1}
\|\mathbf{x}^{n+1}-\mathbf{x}^{n}\|_{2}^{2} \rightarrow 0.
\end{equation}
As a result, we draw Theorem~1.

\subsection{Proof of Lemma~2}
Since $\mathbf{x}^{n+1}=\alpha^{n}\mathbf{v}^{n}+(1-\alpha^{n})\mathbf{w}^{n}$, we have that
\begin{align}\label{add8}
\|{\mathbf{v}^n-\mathbf{x}^n}\|_{2}
&=\|\mathbf{x}^{n+1}+\frac{1-\alpha^n}{\alpha^n}\mathbf{x}^{n+1}-\frac{1-\alpha^n}{\alpha^n}\mathbf{w}^n-\mathbf{x}^n\|_{2} \nonumber \\
&\leq\|\mathbf{x}^{n+1}-\mathbf{x}^{n}\|_2+\frac{1-\alpha^n}{\alpha^n}\|\mathbf{x}^{n+1}-\mathbf{w}^n\|_{2} \nonumber \\
&\leq\|\mathbf{x}^{n+1}-\mathbf{x}^{n}\|_2 + (1-\alpha^n)\|{\mathbf{v}^{n}-\mathbf{w}^n}\|_{2}, 
\end{align}
and
\begin{align}\label{add9}
\|\mathbf{w}^n-\mathbf{x}^n\|_{2}
&=\|\mathbf{x}^{n+1}+\frac{\alpha^n}{1-\alpha^n}\mathbf{x}^{n+1}-\frac{\alpha^n}{1-\alpha^n}\mathbf{v}^n-\mathbf{x}^n\|_{2} \nonumber\\
&\leq\|\mathbf{x}^{n+1}-\mathbf{x}^{n}\|_2+\frac{\alpha^n}{1-\alpha^n}\|\mathbf{x}^{n+1}-\mathbf{v}^n\|_{2} \nonumber\\
&\leq\|\mathbf{x}^{n+1}-\mathbf{x}^{n}\|_2+\alpha^n\|\mathbf{v}^{n}-\mathbf{w}^n\|_{2}.
\end{align}
From Eq.~\eqref{add8} and Eq.~\eqref{add9}, we obtain for finite positive numbers $a$ and $b$ that
\begin{align}\label{add10}
&a\|\mathbf{v}^n-\mathbf{x}^n\|_{2} +b\|\mathbf{w}^n-\mathbf{x}^n\|_{2} \nonumber\\
&\quad\leq (a+b)\|\mathbf{x}^{n+1}-\mathbf{x}^{n}\|_2 + (a-a\alpha^n+b \alpha^n)\|\mathbf{v}^{n}-\mathbf{w}^n\|_{2}.
\end{align}
%where $a>0$, $b>0$, and $a, b$ are upper bounded. 

%According to the nonexpansivity of proximal operator, we have that
Since the proximal operator $\mathcal{S}_{\lambda t}$ is nonexpansive, for arbitrary $\mathbf{x}\in\mathbb{R}^N$ and $\mathbf{y}\in\mathbb{R}^N$, we have
\begin{equation}
\|\mathcal{S}_{\lambda t}(\mathbf{x})-\mathcal{S}_{\lambda t}(\mathbf{y})\|_2\leq\|\mathbf{x}-\mathbf{y}\|_2.
\end{equation}
Thus, we can obtain from Eq.~(5) that
\begin{align}
&\|\mathbf{v}^n-\mathbf{w}^n\|_{2} \nonumber\\
&\quad=\|\mathcal{S}_{\lambda^n t^n}(\mathbf{x}^n-t^n\nabla f(\mathbf{x}^n))-\mathcal{S}_{\lambda^n t^n}(\mathbf{u}^n-t^n\nabla f(\mathbf{u}^n))\|_{2} \nonumber\\
&\quad\leq\|\mathbf{x}^n-\mathbf{u}^n\|_{2}+\|t^n\nabla f(\mathbf{x}^n)-t^n\nabla f(\mathbf{u}^n)\|_{2} 
\end{align}
Since $0<t^n\leq 1/L$, 
\begin{align}\label{add11}
\|\mathbf{v}^n-\mathbf{w}^n\|_{2}\leq\|\mathbf{x}^n-\mathbf{u}^n\|_{2}+\frac{1}{L}\|\nabla f(\mathbf{x}^n)-\nabla f(\mathbf{u}^n)\|_{2}
\end{align}
According to Proposition~1, $f$ has $L$-Lipschitz continuous gradient. For arbitrary $\mathbf{x}\in\mathbb{R}^N$ and $\mathbf{y}\in\mathbb{R}^N$ 
\begin{equation}\label{s2}
\|\nabla f(\mathbf{x})-\nabla f(\mathbf{y})\|_2\leq L\|\mathbf{x}-\mathbf{y}\|_2
\end{equation}
From Eq.~\eqref{add11} and Eq.~\eqref{s2}, we obtain that
\begin{equation}\label{add12}
\|\mathbf{v}^n-\mathbf{w}^n\|_{2}\leq 2\|\mathbf{x}^n-\mathbf{u}^n\|_{2}.
\end{equation}

Since $\|\mathbf{v}^{n}-\mathbf{x}^{n}\|\neq 0$ and $\|\mathbf{u}^{n}-\mathbf{x}^{n}\|_{2}=\eta^{n}\|\mathbf{v}^{n}-\mathbf{x}^{n}\|_{2}$, we obtain from Eq.~\eqref{add10} and Eq.~\eqref{add12},
\begin{align}\label{add13}
&a\|\mathbf{v}^n-\mathbf{x}^n\|_{2} +b\|\mathbf{w}^n-\mathbf{x}^n\|_{2} \nonumber\\
&\quad\leq (a+b)\|\mathbf{x}^{n+1}-\mathbf{x}^{n}\|_2 \nonumber\\
&\quad\quad\ + 2\eta^n(a-a\alpha^n+b \alpha^n)\|\mathbf{v}^{n}-\mathbf{x}^n\|_{2},
\end{align}
We have Eq.~(14) by letting $c^n=2a(1-\alpha^n)+2b\alpha^n$ in Eq.~\eqref{add13}. 
%\begin{align}
%(a-\eta^nc^n)\|\mathbf{v}^{n}-\mathbf{x}^{n}\|_{2}+b\|\mathbf{w}^{n}-\mathbf{x}^{n}\|_{2} \nonumber\\
%\leq (a+b)\|\mathbf{x}^{n+1}-\mathbf{x}^n\|_{2}.
%\end{align}
As a result, we draw Lemma~2.

\subsection{Proof of Theorem~2}
%From the optimality condition of $\mathbf{v}^{n}$ in Eq.~(5), we have
From Eq.~\eqref{sss5}, we have for $\mathbf{v}^n$ that
\begin{align}\label{add14}
0%&\in\nabla f(\mathbf{x}^n)+\frac{1}{t^{n}}(\mathbf{v}^n-\mathbf{x}^n)+\partial g(\mathbf{v}^n)\nonumber \\
&= \nabla f(\mathbf{v}^n)+\nabla f(\mathbf{x}^n)-\nabla f(\mathbf{v}^n)+\frac{1}{t^{n}}(\mathbf{v}^n-\mathbf{x}^n)+\partial g(\mathbf{v}^n)\nonumber\\
&=\partial F(\mathbf{v}^n)+\nabla f(\mathbf{x}^n)-\nabla f(\mathbf{v}^n)+\frac{1}{t^{n}}(\mathbf{v}^n-\mathbf{x}^n).
\end{align}
Here, $f$, $g$, and $F$ are defined in Eq.~(1). Eq.~\eqref{add14} implies that
\begin{equation}\label{s3}
\nabla f(\mathbf{v}^n)-\nabla f(\mathbf{x}^n)-\frac{1}{t^{n}}(\mathbf{v}^n-\mathbf{x}^n)\in\partial F(\mathbf{v}^n).
\end{equation}
According to Eq.~\eqref{s2}, we have that
\begin{align}\label{s4}
\|\nabla f(\mathbf{x}^n)-\nabla f(\mathbf{v}^n)+\frac{1}{t^{n}}(\mathbf{v}^n-\mathbf{x}^n)\|_{2} \nonumber\\
\leq (\|\mathbf{A}\|_2^2+\frac{1}{t^n}) \|\mathbf{v}^{n}-\mathbf{x}^{n}\|_{2}.
\end{align}
Note that we take $L=\|\mathbf{A}\|_2^2$ in Eq.~\eqref{s2}.
Similarly, we have for $\mathbf{u}^n$ and $\mathbf{w}^n$ that
\begin{equation}\label{s5}
\nabla f(\mathbf{w}^n)-\nabla f(\mathbf{u}^n)-\frac{1}{t^{n}}(\mathbf{w}^n-\mathbf{u}^n)\in\partial F(\mathbf{w}^n)
\end{equation}
and 
\begin{align}\label{s6}
\|\nabla f(\mathbf{u}^n)-\nabla f(\mathbf{w}^n)+\frac{1}{t^{n}}(\mathbf{w}^n-\mathbf{u}^n)\|_{2}  \nonumber\\
\quad\leq (\|\mathbf{A}\|_2^2+\frac{1}{t^n}) \|\mathbf{w}^{n}-\mathbf{u}^{n}\|_{2}.
\end{align}

Theorem~1 shows that the sequence $\{\mathbf{x}^n\}_{n\in\mathbb{N}}$ has accumulation points. Let $\mathbf{x}^*$ denote an arbitrary accumulation point of $\{\mathbf{x}^n\}_{n\in\mathbb{N}}$. There exists a subsequence $\{n_j\}_{j\in\mathbb{N}}$ that makes $\{\mathbf{x}^{n_j}\}\rightarrow\mathbf{x}^*$ as $j\rightarrow \infty$. According to Eq.~(1), $\partial F(\mathbf{x})=\mathbf{A}^T\mathbf{Ax}-\mathbf{A}^T\mathbf{b}+\lambda\mathrm{sgn}(\mathbf{x})$.
Therefore,
\begin{align}\label{new_eq1}
&\|\partial F(\mathbf{x}^{\nj+1})\|_{2} \nonumber \\
&\ \leq\alpha^{\nj}\|\partial F(\mathbf{v}^{\nj})\|_{2}+(1-\alpha^{\nj})\|\partial F(\mathbf{w}^{\nj})\|_{2} \nonumber \\
&\quad + \lambda^{\nj}\|\mathrm{sgn}(\mathbf{x}^{n_j+1})-\alpha^{\nj}\mathrm{sgn}(\mathbf{v}^{n_j})-(1-\alpha^{\nj})\mathrm{sgn}(\mathbf{w}^{n_j})\|_2 \nonumber\\
&\ \leq\alpha^{\nj}\|\partial F(\mathbf{v}^{\nj})\|_{2}+(1-\alpha^{\nj})\|\partial F(\mathbf{w}^{\nj})\|_{2}+2\lambda^{\nj}\sqrt{N}
\end{align}
From Eq.~\eqref{s3}$\sim$\eqref{s6}, we have
\begin{align}\label{s15}
&\alpha^{\nj}\|\partial F(\mathbf{v}^{\nj})\|_{2}+(1-\alpha^{\nj})\|\partial F(\mathbf{w}^{\nj})\|_{2} \nonumber\\
&\ \leq \alpha^{\nj}(\|\mathbf{A}\|_2^2+\frac{1}{t^{\nj}})\|\mathbf{v}^{\nj}-\mathbf{x}^{\nj}\|_{2} \nonumber \\
&\quad\ +(1-\alpha^{\nj})(\|\mathbf{A}\|_2^2+\frac{1}{t^{\nj}})\|\mathbf{w}^{\nj}-\mathbf{u}^{\nj}\|_{2} \nonumber \\
&\ \leq(\|\mathbf{A}\|_2^2+\frac{1}{t^{\nj}})[\alpha^{\nj}\|\mathbf{v}^{\nj}-\mathbf{x}^{\nj}\|_{2} +(1-\alpha^{\nj})\|\mathbf{w}^{\nj}-\mathbf{x}^{\nj}\|_{2}\nonumber\\
&\quad\ +(1-\alpha^{\nj})\|\mathbf{u}^{\nj}-\mathbf{x}^{\nj}\|_{2}].
\end{align}
%\begin{align}\label{s15}
%&\|\partial F(x^{\nj+1})\|_{2} \nonumber \\
%&\quad\leq \alpha^{\nj}\|\partial F(v^{\nj})\|_{2}+(1-\alpha^{\nj})\|\partial F(w^{\nj})\|_{2} \nonumber \\
%&\quad\leq \alpha^{\nj}(\|A\|_2^2+\frac{1}{t^{\nj}})\|v^{\nj}-x^{\nj}\|_{2} \nonumber \\ &\qquad\quad+(1-\alpha^{\nj})(\|A\|_2^2+\frac{1}{t^{\nj}})\|w^{\nj}-u^{\nj}\|_{2} \nonumber \\
%&\quad\leq (\|A\|_2^2+\frac{1}{t^{\nj}})\left[\alpha^{\nj}\|v^{\nj}-x^{\nj}\|_{2} \right.\nonumber \\
%&\qquad\quad\left.+(1-\alpha^{\nj})(\|w^{\nj}-x^{\nj}\|_{2}+\|u^{\nj}-x^{\nj}\|_{2})\right].
%\end{align}

%Define a set $T:=\{i|i\in\mathbb{N}, \|v^i-x^i\|=0\}$. \textbf{We first prove the conclusion of Theorem~1 when \bm{$T=\emptyset$}.} .

We consider the set $\mathcal{T}=\{n|n\in\mathbb{N}, \|\mathbf{v}^n-\mathbf{x}^n\|_2=0\}$.

\textbf{i) $\mathcal{T}=\emptyset$. }
According to Assumption~1, for arbitrary $n\in\mathbb{N}$, $\|\mathbf{u}^{n}-\mathbf{x}^{n}\|=\eta^{n}\|\mathbf{v}^{n}-\mathbf{x}^{n}\|$ with $\eta^{n}\leq\eta_c$. Then we obtain from Eq.~\eqref{s15} that
\begin{align}\label{s7}
&\alpha^{\nj}\|\partial F(\mathbf{v}^{\nj})\|_{2}+(1-\alpha^{\nj})\|\partial F(\mathbf{w}^{\nj})\|_{2} \nonumber\\
&\quad\leq (\|\mathbf{A}\|_2^2+\frac{1}{t^{\nj}})\{[\alpha^{\nj}+(1-\alpha^{\nj})\eta^{\nj}]\|\mathbf{v}^{\nj}-\mathbf{x}^{\nj}\|_{2} \nonumber\\
&\quad\quad\ +(1-\alpha^{\nj})\|\mathbf{w}^{\nj}-\mathbf{x}^{\nj}\|_{2}\}.
\end{align}
According to Lemma~2, we obtain from Eq.~\eqref{s7} that
%Applying Lemma~2 to the right-hand side of the Eq.~\eqref{s7} with $a-2a\eta^{\nj}+2a\eta^{\nj}\alpha^{\nj}-2b\eta^{\nj}\alpha^{\nj}=\alpha^{\nj}+(1-\alpha^{\nj})\eta^{\nj}$ and $b=1-\alpha^{\nj}$, we thus get
\begin{equation}\label{s8}
\alpha^{\nj}\|\partial F(\mathbf{v}^{\nj})\|_{2}+(1-\alpha^{\nj})\|\partial F(\mathbf{w}^{\nj})\|_{2}\leq(a+b)(\|\mathbf{A}\|_2^2+\frac{1}{t^{\nj}})\|\mathbf{x}^{\nj+1}-\mathbf{x}^{\nj}\|_{2},
\end{equation}
where 
\begin{equation}\label{ls3}
a=\frac{\alpha^{\nj}+(1-\alpha^{\nj})\eta^{\nj}+2(1-\alpha^{\nj})\eta^{\nj}\alpha^{\nj}}{1-2\eta^{\nj}+2\eta^{\nj}\alpha^{\nj}},
\end{equation}
and
\begin{equation}\label{lsb}
b=1-\alpha^{\nj}.
\end{equation}
Subsequently, we show $a$ and $b$ are positive and bounded to guarantee Eq.~\eqref{s8}. According to Eq.~\eqref{lsb}, it is obvious that $0<b<1$. Therefore, we focus on $a$.
%However, Eq.~\eqref{s8} holds when $a>0$, $b>0$, and $a, b$ are upper bounded. It is obvious that $0<b<1$, thus we have to prove that $a>0$ and $a$ is upper bounded. 

When $\eta^{\nj}=0$, we have that $a=\alpha^{\nj}$. Thus, $0<a<1$ according to Eq.~(7). When $0<\eta^{\nj}\leq \eta_c$, we consider the denominator $1-2\eta^{\nj}+2\eta^{\nj}\alpha^{\nj}$ in Eq.~\eqref{ls3}. From Eq.~(7), we obtain that
%When $0<\eta^{\nj}\leq \eta_c$, as the numerator of $a$ is greater than 0, we apply the minimum of $\alpha^n$ in Eq.~\eqref{alpha_bound} to the denominator of $a$. Under this condition, if the denominator of $a$ is greater than 0, we have $a>0$.
\begin{align}\label{ls2}
&1-2\eta^{\nj}+2\eta^{\nj}\alpha^{\nj}\nonumber\\
&\quad\geq 1-2\eta^{\nj}+\frac{2\eta^{\nj}\|\mathbf{u}^{\nj}-\mathbf{x}^{\nj}\|_{2}^{2}}{\|\mathbf{u}^{\nj}-\mathbf{x}^{\nj}\|_{2}^{2}+(1-2t^{\nj}\delta^{\nj}\|\mathbf{A}\|_2^2)\|\mathbf{v}^{\nj}-\mathbf{x}^{\nj}\|_{2}^{2}} \nonumber \\
&\quad=1-\frac{2\eta^{\nj}(1-2t^{\nj}\delta^{\nj}\|\mathbf{A}\|_2^2)}{(\eta^{\nj})^2+(1-2t^{\nj}\delta^{\nj}\|\mathbf{A}\|_2^2)} %\nonumber \\
%&\quad=1-\frac{2}{1/\eta^{\nj}+1/(1-2t^{\nj}\delta^n\|\mathbf{A}\|_2^2)}
\end{align}
Since $\eta^{\nj}> 0$, $1/(4\delta^n\|\mathbf{A}\|_2^2)\leq t^{\nj}\leq 1/\|\mathbf{A}\|_2^2$ and $1/4<\delta^n<1/2$, we obtain that
\begin{align}\label{ls2-2}
1-2\eta^{\nj}+2\eta^{\nj}\alpha^{\nj}&\ge 1-\sqrt{1-2t^{\nj}\delta^{\nj}\|\mathbf{A}\|_2^2}\ge 1-\frac{\sqrt{2}}{2}
%1-2\eta^{\nj}+2\eta^{\nj}\alpha^{\nj}&\geq 1-\frac{2}{1/\eta_c+1/(1-2\frac{1}{4\delta^n\|A\|_2^2}\delta^n\|\mathbf{A}\|_2^2)}\nonumber \\
%&=\frac{1}{1+2\eta_c}>0.
\end{align}
Therefore, $a>0$. From Eq.~\eqref{ls3} and Eq.~\eqref{ls2-2}, we obtain for $0<\eta^{\nj}\leq \eta_c$ that
\begin{equation}\label{ls2-3}
a<\frac{1+\eta_c+2\eta_c}{1-\frac{\sqrt{2}}{2}}=(1+3\eta_c)(2+\sqrt{2})
\end{equation}
Let us define $a_0=(1+3\eta_c)(2+\sqrt{2})$.
When $\eta^{\nj}\geq 0$, we obtain from Eq.~\eqref{s8}, Eq.~\eqref{lsb}, and Eq.~\eqref{ls2-3} that
\begin{align}\label{s12}
&\alpha^{\nj}\|\partial F(\mathbf{v}^{\nj})\|_{2}+(1-\alpha^{\nj})\|\partial F(\mathbf{w}^{\nj})\|_{2}\nonumber\\
&\quad\leq\frac{(1+a_0)(1+\|\mathbf{A}\|_2^2t^{\nj})}{t^{\nj}}\|\mathbf{x}^{\nj+1}-\mathbf{x}^{\nj}\|_{2}\nonumber\\
&\quad\leq 3(1+a_0)\|\mathbf{A}\|_2^2\|\mathbf{x}^{\nj+1}-\mathbf{x}^{\nj}\|_{2}.
\end{align}

Recall that $0<\lambda^n\leq \min\{\lambda^{n-1}, C_{\lambda}\|\mathbf{x}^n-\mathbf{x}^{n-1}\|_2\}, ~\forall n \in\mathbb{N}_+.$
Combining Eq.~\eqref{new_eq1} and Eq.~\eqref{s12}, we have that 
\begin{align}\label{new_eq2}
&\|\partial F(\mathbf{x}^{\nj+1})\|_{2} \nonumber \\
&\quad\leq 3(1+a_0)\|\mathbf{A}\|_2^2\|\mathbf{x}^{\nj+1}-\mathbf{x}^{\nj}\|_{2} +2\sqrt{N} C_{\lambda}\|\mathbf{x}^{\nj}-\mathbf{x}^{\nj-1}\|_2\nonumber \\
&\quad\leq C_{max}(\|\mathbf{x}^{\nj+1}-\mathbf{x}^{\nj}\|_{2}+\frac{1}{2}\|\mathbf{x}^{\nj}-\mathbf{x}^{\nj-1}\|_2), %\nj\geq 1.
\end{align}
where $C_{max}=\max\{3(1+a_0)\|\mathbf{A}\|_2^2, 4\sqrt{N} C_{\lambda}\}$ is a constant.

%Hence, we have from Eq.~\eqref{s8} when $\eta^{\nj}\geq 0$
%\begin{align}\label{s12}
%&\|\partial F(x^{\nj+1})\|_{2}\nonumber\\
%\leq &\frac{(6\eta_c^2+5\eta_c+2)(\|A\|_2^2t^{\nj}+1)}{t^{\nj}}\|x^{\nj+1}-x^{\nj}\|_{2}\nonumber\\
%\leq &3\|A\|_2^2(6\eta_c^2+5\eta_c+2)\|x^{\nj+1}-x^{\nj}\|_{2}.
%\end{align}

As $F$ is continuous and $\{\mathbf{x}^{\nj}\}\rightarrow\mathbf{x}^*$, we have that 
\begin{equation}\label{s13}
\lim\limits_{j\rightarrow\infty}F(\mathbf{x}^{\nj+1})=F(\mathbf{x}^*)=F_*.
\end{equation}
From Eq.~\eqref{s1}, Eq.~\eqref{new_eq2}, and Eq.~\eqref{s13}, we have $0\in\partial F(\mathbf{x}^{*})$. Therefore, $\mathbf{x}^*$ is a stationary point. 

Furthermore, we prove that $\{\mathbf{x}^n\}_{n\in\mathbb{N}}$ is a Cauchy sequence. Let $\Omega$ denote the set that contains all the accumulation points of $\{\mathbf{x}^n\}_{n\in\mathbb{N}}$. As shown in Theorem~1, $\{F(\mathbf{x}^n)\}_{n\in\mathbb{N}}$ is non-increasing and converges to $F_*$ as $n\rightarrow\infty$. This fact implies that $F(\mathbf{x}^n)\ge F_*$ for arbitrary $n\in\mathbb{N}$. 
%From Eq.s~\eqref{add6} and \eqref{add17}, we have that $F(x^n)\geq F_*, F(x^n)\rightarrow F_*$. 

If there exists a positive integer $k'$ such that $F(\mathbf{x}^{k'})=F_*$, $F(\mathbf{x}^{n})=F_*$ for arbitrary $n\geq k'$. Therefore, the algorithm terminates in finite steps. 

If $F(\mathbf{x}^{n})>F_*$ for arbitrary $n\in\mathbb{N}$, given arbitrary $\kappa\in(0,+\infty]$, there exists $\bar{k}_1\in\mathbb{N}$ such that $F(\mathbf{x}^{n})<F_*+\eta$ whenever $n>\bar{k}_1$. 
%then from $F(x^n)\rightarrow F_*$ we obtain that there exists $\bar{k}_1$ such that $F(x^{n})<F_*+\eta$ whenever $n>\bar{k}_1$ ($\eta$ is defined in Definition~\ref{defn1}). 
Moreover, since $\rm{dist}(\mathbf{x}^n, \Omega)\rightarrow 0$ as $n\rightarrow\infty$, there exists $\bar{k}_2$ such that $\rm{dist}(\mathbf{x}^n, \Omega)<\epsilon'$ for arbitrary $\epsilon'>0$ whenever $n>\bar{k}_2$. Let $k_0=\max\{\bar{k}_1, \bar{k}_2\}$. For arbitrary $n>k_0$, we have
\begin{equation}\label{s18}
\mathbf{x}^n\in\{\mathbf{x}|\mathrm{dist}(\mathbf{x},\Omega)\leq \epsilon'\}\bigcap\{\mathbf{x}|F_*<F(\mathbf{x})<F_*+\eta\}.
\end{equation}
Since $F$ is a KŁ function, we obtain from Definition~\ref{defn1} that
\begin{equation}\label{s17}
\varphi'(F(\mathbf{x}^{n+1})-F(\mathbf{x}^*)){\rm dist}(0, \partial F(\mathbf{x}^{n+1}))\geq 1, 
\end{equation}
From Eq.~\eqref{new_eq2} and Eq.~\eqref{s17}, we have
\begin{equation}\label{s14}
\varphi'(F(\mathbf{x}^{n+1})-F(\mathbf{x}^*)) \geq  \frac{1}{C_{max}(\|\mathbf{x}^{n+1}-\mathbf{x}^{n}\|_{2}+\frac{1}{2}\|\mathbf{x}^{n}-\mathbf{x}^{n-1}\|_2)}.
\end{equation}
Since $\varphi$ is concave and $\{F(\mathbf{x}^n)\}_{n\in\mathbb{N}}$ is non-increasing, we obtain that 
\begin{align}
&\varphi(F(\mathbf{x}^{n+1})-F(\mathbf{x}^*))-\varphi(F(\mathbf{x}^{n+2})-F(\mathbf{x}^*)) \nonumber \\
&\quad\geq\varphi'(F(\mathbf{x}^{n+1})-F(\mathbf{x}^*))(F(\mathbf{x}^{n+1})-F(\mathbf{x}^{n+2}))
\end{align}
From Eq.~\eqref{add6-1} and  Eq.~\eqref{s14},  
\begin{align}\label{app_ls5}
&\varphi(F(\mathbf{x}^{n+1})-F(\mathbf{x}^*))-\varphi(F(\mathbf{x}^{n+2})-F(\mathbf{x}^*)) \nonumber \\
&\quad\geq\frac{\delta^n\|\mathbf{A}\|_2^2\|\mathbf{x}^{n+2}-\mathbf{x}^{n+1}\|_{2}^{2}}{C_{max}(\|\mathbf{x}^{n+1}-\mathbf{x}^{n}\|_{2}+\frac{1}{2}\|\mathbf{x}^{n}-\mathbf{x}^{n-1}\|_2)}\nonumber \\
&\quad\geq\frac{\|\mathbf{A}\|_2^2}{4C_{max}}\frac{\|\mathbf{x}^{n+2}-\mathbf{x}^{n+1}\|_{2}^{2}}{\|\mathbf{x}^{n+1}-\mathbf{x}^{n}\|_{2}+\frac{1}{2}\|\mathbf{x}^{n}-\mathbf{x}^{n-1}\|_2}.
\end{align}
%For convenience, we define for all $p,q\in\mathbb{N}$ and $x^*$ the following quantities
For simplicity, given $p,q\in\mathbb{N}$ and $\mathbf{x}^*\in\mathbb{R}^N$, we define 
\begin{equation}\label{Delta}
\Delta_{p,q}=\varphi(F(\mathbf{x}^{p})-F(\mathbf{x}^*))-\varphi(F(\mathbf{x}^{q})-F(\mathbf{x}^*)),    
\end{equation} 
and 
\begin{equation}\label{E}
E=\frac{4C_{max}}{\|\mathbf{A}\|_2^2}.
\end{equation}
Eq.~\eqref{app_ls5} can be rewritten as
\begin{equation}\label{ls6}
\Delta_{n+1,n+2} \geq \frac{\|\mathbf{x}^{n+2}-\mathbf{x}^{n+1}\|_{2}^{2}}{E(\|\mathbf{x}^{n+1}-\mathbf{x}^{n}\|_{2}+\frac{1}{2}\|\mathbf{x}^{n}-\mathbf{x}^{n-1}\|_2)}.
\end{equation}
Eq.~\eqref{ls6} implies that
\begin{align}\label{ls7}
&\|\mathbf{x}^{n+2}-\mathbf{x}^{n+1}\|_{2}\nonumber\\
&\quad\leq
\frac{1}{2}\left( E\Delta_{n+1,n+2} +  \|\mathbf{x}^{n+1}-\mathbf{x}^{n}\|_{2}+\frac{1}{2}\|\mathbf{x}^{n}-\mathbf{x}^{n-1}\|_2\right).
\end{align}
Given arbitrary $k,l\in\mathbb{N}_{+}$ with $k>l$, we sum up Eq.~\eqref{ls7} for $n=l,\cdots,k$ and yield that
\begin{align}
&\sum_{n=l}^{k}\|\mathbf{x}^{n+2}-\mathbf{x}^{n+1}\|_{2}\nonumber\\
&\quad\leq\frac{1}{2}\sum_{n=l}^{k}\|\mathbf{x}^{n+1}-\mathbf{x}^{n}\|_{2}+\frac{1}{4}\sum_{n=l}^{k}\|\mathbf{x}^{n}-\mathbf{x}^{n-1}\|_{2}\nonumber\\
&\qquad+\frac{E}{2}\sum_{n=l}^{k}\Delta_{n+1,n+2} \nonumber\\
&\quad\leq\frac{1}{2}\sum_{n=l-1}^{k}\|\mathbf{x}^{n+2}-\mathbf{x}^{n+1}\|_{2}+\frac{1}{4}\sum_{n=l-2}^{k}\|\mathbf{x}^{n+2}-\mathbf{x}^{n+1}\|_{2}\nonumber\\
&\qquad+\frac{E}{2}\Delta_{l+1,k+2}.
%&\quad\leq\sum_{i=l}^{k}\|x^{i+2}-x^{i+1}\|_{2}+\|x^{l+1}-x^{l}\|_{2}+E\Delta_{l+1,k+2},
\end{align}
%where $l\in\mathbb{N}$, $k\in\mathbb{N}$ and $k>l$. Since $\varphi>0$, we thus have that
Consequently, we obtain that
\begin{align}\label{ls8}
&\sum_{n=l}^{k}\|\mathbf{x}^{n+2}-\mathbf{x}^{n+1}\|_{2}\nonumber\\
&\quad\leq 2E\left[\varphi(F(\mathbf{x}^{l+1})-F(\mathbf{x}^*))-\varphi(F(\mathbf{x}^{k+2})-F(\mathbf{x}^*))\right]\nonumber\\
&\qquad +3\|\mathbf{x}^{l+1}-\mathbf{x}^{l}\|_{2}+\|\mathbf{x}^{l}-\mathbf{x}^{l-1}\|_{2},
\end{align}
According to Definition~\ref{defn1}, $\varphi$ is continuous at 0 and $\varphi(0)=0$. In Eq.~\eqref{ls8}, making $l=1$ and $k\rightarrow\infty$, we have
%\begin{equation}
%\sum_{i=l}^{k}\|x^{i+2}-x^{i+1}\|_{2}\leq \|x^{l+1}-x^{l}\|_{2}+E\varphi(F(x^{l+1})-F(x^*)),
%\end{equation}
%which means that
\begin{align}\label{ls9}
\sum_{n=1}^{\infty}\|\mathbf{x}^{n+1}-\mathbf{x}^{n}\|_{2}&\leq 4\|\mathbf{x}^{2}-\mathbf{x}^{1}\|_{2}+\|\mathbf{x}^{1}-\mathbf{x}^{0}\|_{2}\nonumber\\
&\quad+2E\varphi(F(\mathbf{x}^{2})-F(\mathbf{x}^*)).
\end{align}
Eq.~\eqref{ls9} suggests that $\sum_{n=1}^{\infty}\|\mathbf{x}^{n+1}-\mathbf{x}^{n}\|_{2}<+\infty$. Therefore, the sequence $\{\mathbf{x}^n\}_{n\in\mathbb{N}}$ is a Cauchy sequence and converges to the stationary point $\mathbf{x}^*$ of $F$ as $n\rightarrow\infty$. %It converges to the stationary point $x^*$ of $F$. 

%\textbf{Next, we prove the conclusion of Theorem~1 when \bm{$T\neq \emptyset$}.} If $T$ is a finite set, we easily deduce that $x^*$ is still a stationary point of $F$, and there exists a positive integer $N_f$ such that $\sum_{n=n'}^{\infty}\|x^{n+1}-x^{n}\|_{2}<+\infty$ for any $n'>N_f$. Hence, $\{x^n\}_{n\in\mathbb{N}}$ is still a Cauchy sequence. 
\textbf{ii) $\mathcal{T}\neq\emptyset$.} If $\mathcal{T}$ is a finite set, we can also obtain that $\mathbf{x}^*$ is a stationary point of $F$. We can easily extend Eq.~\eqref{ls9}. There exists a positive integer $N_f$ such that, for arbitrary $n'>N_f$, $\sum_{n=n'}^{\infty}\|\mathbf{x}^{n+1}-\mathbf{x}^{n}\|_{2}<+\infty$. Therefore, $\{\mathbf{x}^n\}_{n\in\mathbb{N}}$ is a Cauchy sequence. 

If $\mathcal{T}$ is an infinite set, there exists a subsequence $\{n_j\}_{j\in\mathbb{N}}\subset\mathcal{T}$ such that 
\begin{equation}
\lim_{j\rightarrow\infty}\mathbf{v}^{\nj} = \lim_{j\rightarrow\infty}\mathbf{x}^{\nj}=\mathbf{x}^*.
\end{equation}
Note that $\alpha^{\nj}=1$ for arbitrary $\nj\in\mathcal{T}$.
From Eq.~\eqref{new_eq1} and \eqref{s15}, as $j\rightarrow \infty$, we have that 
\begin{align}\label{s16}
\|\partial F(\mathbf{x}^{\nj+1})\|_{2}\leq&\alpha^{\nj}(\|\mathbf{A}\|_2^2+\frac{1}{t_1^{\nj}})\|\mathbf{v}^{\nj}-\mathbf{x}^{\nj}\|_{2}
\nonumber\\
&+2\sqrt{N} C_{\lambda}\|\mathbf{x}^{\nj}-\mathbf{x}^{\nj-1}\|_2
\rightarrow 0.
\end{align}
From Eq.~\eqref{s13} and Eq.~\eqref{s16}, $\mathbf{x}^*$ is a stationary point. For any integer $k>n_j$, $k\in\mathcal{T}$ and the sequence $\{\mathbf{x}^n\}_{n\in\mathbb{N}}$ converges to the stationary point $\mathbf{x}^*$. 

Since $F(\mathbf{x})$ is convex for the Lasso problem, each stationary point is an optimum. Therefore, we draw Theorem~2.

\subsection{Proof of Theorem~3}
%Recall Eq.~\eqref{s18}, \eqref{s17} and the corresponding discussion. Define $r^{n}=F(x^n)-F_*$. We suppose $r^{n}>0$ for all $n$. Otherwise $F(x^n)=F(x^{n+1})=\cdots=F_*$ and the algorithm terminates in finite steps. By supposing this we have from Eq.~\eqref{s17}, \eqref{s12} and \eqref{add6}
Let us define $r^{n}=F(\mathbf{x}^n)-F_*$. Without loss of generality, we suppose $r^{n}>0$ for $n\in\mathbb{N}$. In fact, since $\{F(\mathbf{x}^n)\}_{n\in\mathbb{N}}$ is non-increasing, when there exists $N_0\in\mathbb{N}$ such that $r^{N_0}=0$, we can obtain $F(\mathbf{x}^n)=F_*$ for $n\ge N_0$. 

From Eq.~\eqref{add6-1} and Eq.~\eqref{new_eq2}, for arbitrary $n > k_0+1$, we have 
\begin{align}\label{new_eq3}
&{\rm dist}(0, \partial F(\mathbf{x}^{n+1}))^2 \nonumber\\
&\quad\leq C_{max}^2(\|\mathbf{x}^{n+1}-\mathbf{x}^{n}\|_2^2+\frac{1}{4}\|\mathbf{x}^{n}-\mathbf{x}^{n-1}\|_2^2 \nonumber \\
&\qquad+\|\mathbf{x}^{n+1}-\mathbf{x}^{n}\|_2\|\mathbf{x}^{n}-\mathbf{x}^{n-1}\|_2)\nonumber \\
&\quad\leq\frac{C_{max}^2}{\delta^n \|\mathbf{A}\|_2^2}[r^n-r^{n+1}+\frac{1}{4}(r^{n-1}-r^n)\nonumber\\
&\qquad+(r^n-r^{n+1})^{\frac{1}{2}}(r^{n-1}-r^{n})^{\frac{1}{2}}]
\end{align}
As $r^{n+1}\leq r^{n}\leq r^{n-1}$ and $\delta\in(0.25, 0.5)$, we obtain from Eq.~\eqref{new_eq3} that, for arbitrary $n>k_0+1$,
\begin{equation}\label{new_eq4}
{\rm dist}(0, \partial F(\mathbf{x}^{n+1}))^2\leq \frac{9C_{max}^2}{ \|\mathbf{A}\|_2^2}(r^{n-1}-r^{n+1}).
\end{equation}
From Eq.~\eqref{s17} and Eq.~\eqref{new_eq4}, for arbitrary $n > k_0+1$,
\begin{align}\label{s19}
1 &\leq [\varphi'(F(\mathbf{x}^{n+1})-F_*){\rm dist}(0, \partial F(\mathbf{x}^{n+1}))]^2 \nonumber\\
&\leq\frac{9C_{max}^2}{ \|\mathbf{A}\|_2^2}[\varphi'(r^{n+1})]^2 (r^{n-1}-r^{n+1}).
\end{align}
%\textcolor{red}{Here, $E$ is defined in Eq.~\eqref{E}.}
From Eq.~(10), we have that %$\varphi(t)=\frac{C}{\theta}t^{\theta}$ with $\theta\in(0,1]$, we have 
$\varphi'(t)=Ct^{\theta-1}$ for $\theta\in(0,1]$. Thus, we obtain from Eq.~\eqref{s19} that
\begin{equation}\label{s20}
1\leq P(r^{n+1})^{2\theta-2}(r^{n-1}-r^{n+1}),
\end{equation}
where $P=9C^2C_{max}^2/\|\mathbf{A}\|_2^2$ is a constant. Consequently, we consider the cases that $\mathcal{T}$ is an infinite set, a finite and non-empty set, and an empty set, respectively.

\textbf{i) $\theta=1$ or $\mathcal{T}$ is an infinite set.}
If $\theta=1$, Eq.~\eqref{s20} can be rewritten as 
\begin{equation}
1\leq P(r^{n-1}-r^{n+1}).
\end{equation}
Since $r^n\rightarrow 0$ as $n\rightarrow\infty$ and $P>0$, there exists $N_0\in\mathbb{N}$ such that $r^{n-1}-r^{n+1}<1/P$ whenever $n>N_0$. Thus, the algorithm has to terminate in finite number of steps. This fact implies that there exists $\hat{k}_1\in\mathbb{N}$ such that $r^n=0$ for arbitrary $n>\hat{k}_1$. 
%The algorithm terminates in finite steps.
If $\mathcal{T}$ is an infinite set, there exists a positive integer $\hat{k}_2\in\mathcal{T}$ such that $F(\mathbf{x}^{\hat{k}_2})=F_*$. Thus, $F(\mathbf{x}^{n})=F_*$ for arbitrary $n\geq \hat{k}_2$. Therefore, when $\theta=1$ or $\mathcal{T}$ is an infinite set, there exists $k_1=\max\{\hat{k}_1, \hat{k}_2\}$ such that the algorithm terminates in finite steps.
%Thus for all $k_1>\max\{\hat{k}_1, \hat{k}_2\}$, the algorithm terminates in finite steps. The following proof is based on that $T=\emptyset$ or $T$ is a finite set when $T\neq\emptyset$.

\textbf{ii) $\theta\in[\frac{1}{2},1)$}.
In this case, $0<2-2\theta\leq1$. Since $r^n\rightarrow 0$ as $n\rightarrow\infty$, there exists $\hat{k}_3\in\mathbb{N}$ such that, for arbitrary $n>\hat{k}_3$, $r^{n}\leq 1$ and $(r^{n})^{2-2\theta}\geq r^{n}$. Thus, we obtain from Eq.~\eqref{s20} that 
\begin{equation}
r^{n+1}\leq P(r^{n-1}-r^{n+1}).
\end{equation}
Therefore,
\begin{equation}
r^{n+1}\leq\frac{P}{1+P}r^{n-1}.
\end{equation}
There exists $k_2>\max\{k_0, \hat{k}_3\}$ such that, for arbitrary $n=k_2+2l$ and $l \in \mathbb{N}_{+}$, 
\begin{align}
r^{n}\leq\left( \frac{P}{1+P}\right)^{l} r^{n-2l}\leq \left( \frac{P}{1+P}\right)^{l} r^{k_2},
\end{align}
where $k_0$ is defined for Eq.~\eqref{s18}. 
Since $r^n=F(\mathbf{x}^n)-F_*$, we have for arbitrary $n=k_2+2l$ that
\begin{equation}
F(\mathbf{x}^n)-F_*\leq \left( \frac{P}{1+P}\right)^{l} r^{k_2}.
\end{equation}

\textbf{iii) $\theta\in(0,\frac{1}{2})$}.
In this case, $-2<2\theta-2<-1$ and $-1<2\theta-1<0$. As $r^{n-1}\geq r^n\geq r^{n+1}$, we have 
\begin{equation}
(r^{n-1})^{2\theta-2}\leq (r^{n+1})^{2\theta-2},
\end{equation}
and
\begin{equation}
(r^0)^{2\theta-1}\leq\cdots\leq(r^n)^{2\theta-1}\leq(r^{n+1})^{2\theta-1}.
\end{equation}
Let us define $\phi(t)=Ct^{2\theta-1}/(1-2\theta)$. Thus, $\phi'(t)=-Ct^{2\theta-2}$. When $(r^{n+1})^{2\theta-2}\leq 2(r^{n-1})^{2\theta-2}$, for arbitrary $n>k_0+1$,
\begin{align}
\phi(r^{n+1})-\phi(r^{n-1})&=\int_{r^{n-1}}^{r^{n+1}}\phi'(t)\mathrm{d}t=C\int_{r^{n+1}}^{r^{n-1}}t^{2\theta-2}\mathrm{d}t \nonumber \\
&\geq C(r^{n-1}-r^{n+1})(r^{n-1})^{2\theta-2} \nonumber \\
&\geq \frac{C}{2}(r^{n-1}-r^{n+1})(r^{n+1})^{2\theta-2} \nonumber \\
&\geq \frac{C}{2P}
\end{align}
%for all $n>k_0$.

When $(r^{n+1})^{2\theta-2}> 2(r^{n-1})^{2\theta-2}$, we obtain that $(r^{n+1})^{2\theta-1}> 2^{\frac{2\theta-1}{2\theta-2}}(r^{n-1})^{2\theta-1}$. Thus, 
\begin{align}
\phi(r^{n+1})-\phi(r^{n-1})
&=\frac{C}{1-2\theta}[(r^{n+1})^{2\theta-1}-(r^{n-1})^{2\theta-1}] \nonumber\\
&>\frac{C}{1-2\theta}(2^{\frac{2\theta-1}{2\theta-2}}-1)(r^{n-1})^{2\theta-1} \nonumber \\
&\geq \frac{C}{1-2\theta}(2^{\frac{2\theta-1}{2\theta-2}}-1)(r^{0})^{2\theta-1}.
\end{align}
Let 
\begin{equation}
C_r = \min\left\{ \frac{C}{2P}, \frac{C}{1-2\theta}(2^{\frac{2\theta-1}{2\theta-2}}-1)(r^{0})^{2\theta-1} \right\}. 
\end{equation}
For arbitrary integer $m>k_0+1$, 
\begin{equation}\label{ls11}
\phi(r^{m+1})-\phi(r^{m-1})>C_r.
\end{equation}
Since $\phi(t)$ is monotonically decreasing for $\theta\in(0,\frac{1}{2})$ and $\{r^n\}_{n\in\mathbb{N}}$ is non-increasing, for arbitrary $n> k_0+2$, we have 
\begin{align}
2\phi(r^{n})&\geq \phi(r^{n})+\phi(r^{n-1})-\phi(r^{k_0})-\phi(r^{k_0+1}) \nonumber\\
&\geq \sum_{i=k_0+1}^{n-1}\phi(r^{i+1})-\phi(r^{i-1}).
\end{align}
From Eq.~\eqref{ls11}, 
\begin{equation}\label{ls10}
    2\phi(r^{n})>(n-k_0-2)C_r.
\end{equation}
Eq.~\eqref{ls10} suggests that \begin{equation}
(r^n)^{2\theta-1}\geq \frac{(n-k_0-2)C_r(1-2\theta)}{2C},
\end{equation}
and 
\begin{equation}
r^n\leq\left[\frac{2C}{(n-k_0-2)C_r(1-2\theta)}\right] ^{\frac{1}{1-2\theta}}.
\end{equation}
Let $k_3=k_0+2$,
%We obtain by taking $k_3=k_0+2$: 
\begin{equation}
F(\mathbf{x}^n)-F_*\leq\left[\frac{2C}{(n-k_3)C_r(1-2\theta)}\right]^{\frac{1}{1-2\theta}}.
\end{equation}

As a result, we draw Theorem~3.

\section{Proofs for HLISTA}
%Before proving Theorem~3, we would first introduce same notations and a definition. 
%We first introduce some notations. For vectors/matrices originally introduced in the manuscript, we add a subscript (e.g., $i$) to indicate its element/column at the corresponding position. For example, $x_{i}$ represents the $i$-th element of the vector  $\mathbf{x}\in\mathbb{R}^{N}$, and $\mathbf{W}_{i}$ is the $i$-th column of a matrix $\mathbf{W}$. 

%\textbf{Notations:} For vectors/matrices originally introduced without any subscript, adding a subscript (e.g., $i$) indicates its element/column at the corresponding position. For instance,
%for $x\in \mathbb{R}^{N}$, $x_{i}$ represents the $i$-th element of the vector, and $W_{i}$ denote the $i$-th column of a matrix $W$.

%Recall the properties of $\mathbf{x}^*$ in Assumption~2. Recall $\mu(\mathbf{A})$ and $\mathcal{W}_s(\mathbf{A})$ defined in Definition~2. Note that there exists matrices $\mathbf{W}$ that $\mathbf{W}\in \mathcal{W}_s(\mathbf{A})$, \emph{i.e.}, $\mathcal{W}_s(\mathbf{A})\neq \emptyset$, which was proven in Lemma~1 in~[30]. For simplicity, we use $\mu$ to replace $\mu(\mathbf{A})$ in the following proofs.

%Subsequently, we prove Theorem~3.
\subsection{Proof of Theorem~4}
%We first propose a choice of parameters $\{\overline{\mathbf{W}}^n,\hat{\mathbf{W}}^n,\theta_1^n,\theta_2^n\}_{n\in\mathbb{N}}$ that are uniform for $\mathbf{x}^{*}\in \mathcal{X}(B_{\mathbf{x}}, \mathbb{S})$. For arbitray $n\in\mathbb{N}$,
%\begin{equation}\label{new1}
%\overline{\mathbf{W}}^{n} \in W_s(A),\ 
%\theta_{1}^{n}=\sup_{x^{*}\in\mathcal{X}(B_{x}, \mathbb{S})}\{\mu\|x^{n}-x^{*}\|_{1}\},
%\end{equation}
%and
%\begin{align}\label{new4}
%&\widehat{W}^{n} \in W_s(A), \nonumber\\ &\theta_{2}^{n} = \sup_{x^{*}\in\mathcal{X}(B_{x}, \mathbb{S})}\{\mu\|u^{n}-x^{*}\|_{1}\}+\frac{\mu(N-|\mathbb{S}|)}{|\mathbb{S}|-1}\|u^n\|_{1}.
%\end{align}
%where $N$ is the dimension of signal $x^*$, $\mathbb{S}$ is the support of $x^*$ and $|\mathbb{S}|\geq 2$. The criterion for selecting $\theta_{1}$, $\theta_{2}$, $\overline{W}$, and $\widehat{W}$ is similar to~[30]. We prove that Theorem~3 holds if Eq.~\eqref{new1} and \eqref{new4} are satisfied.
%which are uniform for all $x^{*}\in \mathcal{X}(B_{x}, s)$. The criterion of the choice of $\theta_{1}$ and $\theta_{2}$ is similar to the counterpart in LISTA-CP(SS) \cite{NIPS2018_8120}. We shall prove that Theorem~3 holds when Eq.~\eqref{new1} and Eq.~\eqref{new4} are satisfied.

%\textbf{(1) We firstly prove that there is "no false positive".}
%\textbf{i) ``No false positive''.}
\subsubsection{``No False Positive''}
%For arbitrary $n\in\mathbb{N}$, we assume that $x_{i}^{n}=0$ for arbitrary $i \not \in \mathbb{S}$. Here, $x_i^n$ is the $i$-th element of $\mathbf{x}^n$. We can obtain the $i$-th element $v_i^n$ of $\mathbf{v}^n$.
From Eq.~(20), for arbitrary $n\in\mathbb{N}$, the $i$th element $v_i^n$ of $\mathbf{v}^n$ is
\begin{align}\label{t8}
v_{i}^{n}
%=&\mathcal{S}_{\theta_{1}^{n}}\left(x_{i}^{n}+\sum_{j}(\overline{\mathbf{W}}_{i}^{n})^{T}(\mathbf{b}-\mathbf{Ax}^{n})\right)
=\mathcal{S}_{\theta_{1}^{n}}\Bigg(x_{i}^{n}-&\sum_{j\not\in \mathbb{S}}(\overline{\mathbf{W}}_{i}^{n})^{T}\mathbf{A}_{j}(x_{j}^{n}-x_{j}^{*}) \nonumber\\
&\qquad -  \sum_{j\in \mathbb{S}}(\overline{\mathbf{W}}_{i}^{n})^{T}\mathbf{A}_{j}(x_{j}^{n}-x_{j}^{*})\Bigg),
\end{align}
where $x_i^n$ is the $i$th element of $\mathbf{x}^n$, $\overline{\mathbf{W}}_{i}^{n}$ is the $i$th column of $\overline{\mathbf{W}}$, and $\mathbf{A}_j$ is the $j$th column of $\mathbf{A}$. Let us assume that $x_{i}^{n}=0$ for arbitrary $i \not \in \mathbb{S}$. 
\begin{align}
v_i^n=\mathcal{S}_{\theta_{1}^{n}}\left( - \sum_{j\in \mathbb{S}}(\overline{\mathbf{W}}_{i}^{n})^{T}\mathbf{A}_{j}(x_{j}^{n}-x_{j}^{*})\right)
\end{align}
%Recall Eq.~\eqref{new1}, it follows that
%According to Eq.~\eqref{new1}, for arbitrary $i \not\in \mathbb{S}$,
According to Eq.~(23) and Eq.~(25), for arbitrary $i \not\in \mathbb{S}$,
\begin{align}\label{t9}
\theta_{1}^{n}&\geq\mu\|\mathbf{x}^{n}-\mathbf{x}^{*}\|_{1}\geq\sum_{j}\left|(\overline{\mathbf{W}}_{i}^{n})^{T}\mathbf{A}_{j}\right|\left|x_{j}^{n}-x_{j}^{*}\right|\nonumber\\
&\geq\left|-\sum_{j\in \mathbb{S}}(\overline{\mathbf{W}}_{i}^{n})^{T}\mathbf{A}_{j}(x_{j}^{n}-x_{j}^{*})\right|.
\end{align}
According to the definition of $\mathcal{S}_{\theta_{1}^{n}}$, $v_{i}^{n}=0$ for arbitrary $i \not \in \mathbb{S}$. Therefore, when $\theta_1^n$ is determined by Eq.~(25),  $v^{n}_{i}=0$ for $x^{n}_{i}=0$, $\forall i \not\in \mathbb{S}, \forall n\in\mathbb{N}$.
%By the definition of $\mathcal{S}_{\theta_{1}^{n}}$, we easily deduce that $v_{i}^{n}=0, \forall i \not \in S$. Thus we conclude that as long as Eq.~\eqref{new1} holds, if $x^{n}_{i}=0$, it follows that $v^{n}_{i}=0$, $\forall i \not\in S, \forall n$. 

Subsequently, we similarly consider $w_{i}^{n}$ for $i\not \in \mathbb{S}$. 
\begin{align}\label{es76}
w_{i}^{n}=&\mathcal{S}_{\theta_{2}^{n}}\left(u_{i}^{n}+\sum_{j}(\widehat{\mathbf{W}}_{i}^{n})^{T}(\mathbf{b}-\mathbf{Au}^{n})\right) \nonumber\\
=&\mathcal{S}_{\theta_{2}^{n}}\left(u_{i}^{n}-\sum_{j\not\in \mathbb{S}}(\widehat{\mathbf{W}}_{i}^{n})^{T}\mathbf{A}_{j}u_{j}^{n} \right. \nonumber \\
&\qquad\qquad\qquad\left.- \sum_{j\in \mathbb{S}}(\widehat{\mathbf{W}}_{i}^{n})^{T}\mathbf{A}_{j}(u_{j}^{n}-x_{j}^{*})\right)\nonumber\\
=&\mathcal{S}_{\theta_{2}^{n}}\left( -\sum_{j\not\in \mathbb{S}, j\neq i}(\widehat{\mathbf{W}}_{i}^{n})^{T}\mathbf{A}_{j}u_{j}^{n} \right.\nonumber\\
&\left.\qquad\qquad\qquad-\sum_{j\in \mathbb{S}}(\widehat{\mathbf{W}}_{i}^{n})^{T}\mathbf{A}_{j}(u_{j}^{n}-x_{j}^{*})\right).
\end{align}
From Eq.~(23) and Eq.~(25), we have
\begin{align}\label{es77}
\theta_{2}^{n} &\geq\mu\|\mathbf{u}^{n}-\mathbf{x}^{*}\|_{1} \nonumber\\
&\geq\sum_{j\not\in \mathbb{S}, j\neq i}\left|(\widehat{\mathbf{W}}_{i}^{n})^{T}\mathbf{A}_{j}\right|\left|u_{j}^{n}\right|+\sum_{j\in \mathbb{S}}\left|(\widehat{\mathbf{W}}_{i}^{n})^{T}\mathbf{A}_{j}\right|\left|u_{j}^{n}-x_{j}^{*}\right| \nonumber\\
&\geq\left|-\sum_{j\not\in \mathbb{S}, j\neq i}((\widehat{\mathbf{W}}_{i}^{n})^{T}\mathbf{A}_{j}u_{j}^{n} - \sum_{j\in \mathbb{S}}((\widehat{\mathbf{W}}_{i}^{n})^{T}\mathbf{A}_{j}(u_{j}^{n}-x_{j}^{*})\right|.
\end{align}
From Eq.~\eqref{es76} and Eq.~\eqref{es77}, we obtain that $w_{i}^{n}=0$ for $i \not \in \mathbb{S}$. Therefore, for arbitrary $n\in\mathbb{N}$ and $i\notin\mathbb{S}$, we have $x_{i}^{n+1}=\alpha^{n}v^{n}_{i}+(1-\alpha^{n})w^{n}_{i}=0$, when $x^{n}_{i}=0$. Introducing $\mathbf{x}^{0}=0$, we obtain $x_{i}^{n}=0$ for arbitrary $n\in\mathbb{N}$ and $i \not\in \mathbb{S}$.
%In a conclusion, if $x^{n}_{i}=0$, then we have $x_{i}^{n+1}=\alpha^{n}v^{n}_{i}+(1-\alpha^{n})w^{n}_{i}=0$, $\forall i \not\in S, \forall n$.  Since $x^{0}=0$, by induction we obtain
%\begin{equation}
%x_{i}^{n}=0, \forall i \not\in \mathbb{S}, \forall n.
%\end{equation}
Therefore, the ``no false positive'' has been proved, \emph{i.e.}, 
\begin{equation}
{\rm support}(\mathbf{x}^n) \subset \mathbb{S}.
\end{equation}

Note that, when $\theta_{1}^{n}$ and $\theta_{2}^{n}$ are adaptively trained, ``no false positives'' might not be guaranteed for all layers. As demonstrated in~[30], however, the magnitudes on the "false positives" are tiny compared to those on "true positives". As a result, the proof can describe the HLISTA-CP with learned $\theta_{1}^{n}$ and $\theta_{2}^{n}$ qualitatively.
%The "no false positive" has been proved. Note that if $\theta_{1}^{n}$ and $\theta_{2}^{n}$ are obtained by training, they may not guarantee "no false positives" for all layers. However, as demonstrated in \cite{NIPS2018_8120}, the magnitudes on the "false positives" are tiny compared to those on true positives. Thus the proof could describe the HLISTA with learned $\theta_{1}^{n}, \theta_{2}^{n}$ qualitatively.

%\textbf{(2) Second, we develop the error bound and convergence rate.}
%\textbf{ii) Upper bound of recovery error.}
\subsubsection{Upper Bound of Recovery Error}
%Let's consider the components on $S$. For all $i\in S$, we have
For arbitrary $i\in \mathbb{S}$, we have
\begin{align}\label{s21}
v_{i}^{n}=&\mathcal{S}_{\theta_{1}^{n}}\left(x_{i}^{n}-\sum_{j\in \mathbb{S}}(\overline{\mathbf{W}}_{i}^{n})^{T}\mathbf{A}_{j}(x_{j}^{n}-x_{j}^{*})\right)\nonumber\\
=&\mathcal{S}_{\theta_{1}^{n}}\left(x_{i}^{n}-\sum_{j\in \mathbb{S}, j\neq i}(\overline{\mathbf{W}}_{i}^{n})^{T}\mathbf{A}_{j}(x_{j}^{n}-x_{j}^{*})-(x_{i}^{n}-x_{i}^{*})\right)\nonumber\\
=&\mathcal{S}_{\theta_{1}^{n}}\left(x_{i}^{*}-\sum_{j\in \mathbb{S}, j\neq i}(\overline{\mathbf{W}}_{i}^{n})^{T}\mathbf{A}_{j}(x_{j}^{n}-x_{j}^{*})\right)\nonumber\\
\in&x_{i}^{*}-\sum_{j\in \mathbb{S}, j\neq i}(\overline{\mathbf{W}}_{i}^{n})^{T}\mathbf{A}_{j}(x_{j}^{n}-x_{j}^{*})-\theta_{1}^{n}\partial \mathit{l}_{1}(v_{i}^{n}), 
\end{align}
where $\partial \mathit{l}_{1}(v_{i}^{n})$ denotes the sub-gradient of $\|v_{i}^{n}\|_{1}$ that is defined by 
%It is a set which can be defined in a component-wisely fashion.
\begin{equation}\label{new2}
\partial \mathit{l}_{1}(v_{i}^{n})=
\begin{cases}
\mathrm{sign}(v_{i}^{n}), & ~~~{\rm if}~v_{i}^{n}\neq 0, \\
[-1,1], & ~~~{\rm if}~v_{i}^{n}= 0.
\end{cases}
\end{equation}
Eq.~\eqref{new2} suggests that $\partial \mathit{l}_{1}(v_{i}^{n})$ has a magnitude not greater than 1. Thus, we obtain for $i\in \mathbb{S}$,
%From the above equation it's clear that $\partial \mathit{l}_{1}(x)$ has a magnitude less than or equal to 1. Hence, we obtain
\begin{align}
\left|v_{i}^{n}-x_{i}^{*}\right| \leq & \sum_{j\in \mathbb{S}, j\neq i}\left|(\overline{\mathbf{W}}_{i}^{n})^{T}\mathbf{A}_{j}\right|\left|x_{j}^{n}-x_{j}^{*}\right|+\theta_{1}^{n} \nonumber\\
\leq&\mu\sum_{j\in \mathbb{S}, j\neq i}\left|x_{j}^{n}-x_{j}^{*}\right|+\theta_{1}^{n}.
\end{align}
Consequently,
\begin{align}\label{new5}
\|\mathbf{v}^{n}-\mathbf{x}^{*}\|_{1}&=\sum_{i\in \mathbb{S}}|v^{n}_{i}-x^{*}_{i}| \leq\sum_{i\in \mathbb{S}}\left(\mu\sum_{j\in \mathbb{S}, j\neq i}\left|x_{j}^{n}-x_{j}^{*}\right|+\theta_{1}^{n}\right) \nonumber\\
&=\mu(|\mathbb{S}|-1)\sum_{i\in \mathbb{S}}\left|x_{i}^{n}-x_{i}^{*}\right|+|\mathbb{S}|\theta_{1}^{n}  \nonumber \\
&\leq \mu(|\mathbb{S}|-1)\|\mathbf{x}^{n}-\mathbf{x}^{*}\|_{1}+|\mathbb{S}|\theta_{1}^{n} \nonumber\\
&\leq\mu\left(2|\mathbb{S}|-1\right)\sup_{\mathbf{x}^{*}\in\mathcal{X}(B_{\mathbf{x}},\mathbb{S})}\|\mathbf{x}^{n}-\mathbf{x}^{*}\|_{1}.
\end{align}
Similarly, for arbitrary $i\in \mathbb{S}$, we have
\begin{align}\label{es83}
w_{i}^{n}
=&\mathcal{S}_{\theta_{2}^{n}}\left(u_{i}^{n}-\sum_{j\not\in \mathbb{S}}(\widehat{\mathbf{W}}_{i}^{n})^{T}\mathbf{A}_{j}u_{j}^{n} - \sum_{j\in \mathbb{S}}(\widehat{\mathbf{W}}_{i}^{n})^{T}\mathbf{A}_{j}(u_{j}^{n}-x_{j}^{*})\right)\nonumber\\
=&\mathcal{S}_{\theta_{2}^{n}}\left( x_{i}^{*}-\sum_{j\not\in \mathbb{S}}(\widehat{\mathbf{W}}_{i}^{n})^{T}\mathbf{A}_{j}u_{j}^{n} - \sum_{\substack{j\in \mathbb{S},\\ j\neq i}}(\widehat{\mathbf{W}}_{i}^{n})^{T}\mathbf{A}_{j}(u_{j}^{n}-x_{j}^{*})\right)\nonumber\\
\in & x_{i}^{*}-\sum_{j\not\in \mathbb{S}}(\widehat{\mathbf{W}}_{i}^{n})^{T}\mathbf{A}_{j}u_{j}^{n} \nonumber \\
& \quad- \sum_{j\in \mathbb{S}, j\neq i}(\widehat{\mathbf{W}}_{i}^{n})^{T}\mathbf{A}_{j}(u_{j}^{n}-x_{j}^{*}) -
\theta_{2}^{n}\partial \mathit{l}_{1}(w_{i}^{n}). 
\end{align}
From Eq.~\eqref{new2} and Eq.~\eqref{es83}, we obtain that
\begin{align}\label{es84}
\left|w_{i}^{n}-x_{i}^{*}\right| &\leq \sum_{j\in \mathbb{S}, j\neq i}\left|(\widehat{\mathbf{W}}_{i}^{n})^{T}\mathbf{A}_{j}\right|\left|u_{j}^{n}-x_{j}^{*}\right|+\theta_{2}^{n} \nonumber\\
&\qquad+\sum_{j\not\in \mathbb{S}}\left|(\widehat{\mathbf{W}}_{i}^{n})^{T}\mathbf{A}_{j}\right|\left|u_{j}^{n}\right| \nonumber\\
&\leq\mu\sum_{j\in \mathbb{S}, j\neq i}\left|u_{j}^{n}-x_{j}^{*}\right|+\theta_{2}^{n}+ \mu\sum_{j\not\in \mathbb{S}}\left|u_{j}^{n}\right|.
\end{align}
From Eq.~\eqref{es84}, we have
\begin{align}\label{new6}
&\|\mathbf{w}^{n}-\mathbf{x}^{*}\|_{1} \nonumber\\ &\quad=\sum_{i\in \mathbb{S}}|w^{n}_{i}-x^{*}_{i}| \nonumber \\
&\quad\leq\sum_{i\in \mathbb{S}}\left(
\mu\sum_{j\in \mathbb{S}, j\neq i}\left|u_{j}^{n}-x_{j}^{*}\right|+\theta_{2}^{n}+ \mu\sum_{j\not\in \mathbb{S}}\left|u_{j}^{n}\right|
\right)\nonumber\\
&\quad =\mu(|\mathbb{S}|-1)\sum_{i\in \mathbb{S}}\left|u_{i}^{n}-x_{i}^{*}\right|+|\mathbb{S}|\theta_{2}^{n} +\mu(N-|\mathbb{S}|)\sum_{i\in \mathbb{S}}\left|u_{i}^{n}\right|\nonumber\\
&\quad\leq\mu(|\mathbb{S}|-1)\|\mathbf{u}^{n}-\mathbf{x}^{*}\|_{1}+|\mathbb{S}|\theta_{2}^{n}+\mu(N-|\mathbb{S}|)\|\mathbf{u}^{n}\|_{1} \nonumber\\
\end{align}
Considering Eq.~(25) in Eq.~\eqref{new6}, we have
\begin{align}\label{new9}
&\|\mathbf{w}^{n}-\mathbf{x}^{*}\|_{1}\nonumber\\
&\quad\leq\mu \left(2|\mathbb{S}|-1\right)\left(\sup_{\mathbf{x}^{*}\in\mathcal{X}(B_{\mathbf{x}},\mathbb{S})}\|\mathbf{u}^{n}-\mathbf{x}^{*}\|_{1}+\frac{N-|\mathbb{S}|}{|\mathbb{S}|-1}\|\mathbf{u}^{n}\|_{1} \right).
\end{align}
%Recalling the bound of $\alpha^{n}$ in Eq.~(17), 
For arbitrary $n\in\mathbb{N}$, we obtain from Eq.~\eqref{new5} and Eq.~\eqref{new9} that
%\begin{equation}\label{alpha}
%\frac{\theta_{2}^{n}}{ \theta_{1}^{n}+ \theta_{2}^{n}}\leq\alpha^{n} <1, \forall n \in \mathbb{N}.
%\end{equation}
%Combining Eq.~\eqref{new5} and \eqref{new6}, we obtain that
\begin{align}\label{es87}
&\sup_{\mathbf{x}^{*}\in\mathcal{X}(B_{\mathbf{x}},\mathbb{S})}\|\mathbf{x}^{n+1}-\mathbf{x}^{*}\|_{1}\nonumber\\
&\quad\leq\sup_{\mathbf{x}^{*}\in\mathcal{X}(B_{\mathbf{x}},\mathbb{S})}\left(\alpha^{n}\|\mathbf{v}^{n}-\mathbf{x}^{*}\|_{1}+(1-\alpha^{n})\|\mathbf{w}^{n}-\mathbf{x}^{*}\|_{1}\right) \nonumber\\
&\quad\leq\alpha^{n}\mu\left(2|\mathbb{S}|-1\right)\sup_{\mathbf{x}^{*}\in\mathcal{X}(B_{\mathbf{x}},\mathbb{S})}\|\mathbf{x}^{n}-\mathbf{x}^{*}\|_{1} \nonumber \\
&\qquad+(1-\alpha^{n})\mu\left(2|\mathbb{S}|-1\right)\nonumber \\
&\qquad\qquad\qquad\cdot\left(\sup_{\mathbf{x}^{*}\in\mathcal{X}(B_{\mathbf{x}},\mathbb{S})}\|\mathbf{u}^{n}-\mathbf{x}^{*}\|_{1}+\frac{N-|\mathbb{S}|}{|\mathbb{S}|-1}\|\mathbf{u}^{n}\|_{1}\right).
\end{align}
%Introducing Eq.~\eqref{alpha} in Eq.~\eqref{es87}, 
Recalling Eq.~(25) and the choice of $\alpha^{n}$ in Eq.~(21), we obtain from Eq.~\eqref{es87} that
\begin{align}
&\sup_{\mathbf{x}^{*}\in\mathcal{X}(B_{\mathbf{x}})}\|\mathbf{x}^{n+1}-\mathbf{x}^{*}\|_{1}\nonumber\\ 
&\qquad\leq (\alpha^{n}+1)\mu\left(2|\mathbb{S}|-1\right)\sup_{\mathbf{x}^{*}\in\mathcal{X}(B_{\mathbf{x}})}\|\mathbf{x}^{n}-\mathbf{x}^{*}\|_{1} \nonumber \\
&\qquad\leq \left(4\mu|\mathbb{S}|-2\mu\right)^{n+1}\sup_{\mathbf{x}^{*}\in\mathcal{X}(B_{\mathbf{x}})}\|\mathbf{x}^{0}-\mathbf{x}^{*}\|_{1}. 
\end{align}
Note that $\mathbf{x}^0=0$ here. According to Assumption~3, we have
\begin{equation}\label{es88}
\sup_{\mathbf{x}^{*}\in\mathcal{X}(B_{\mathbf{x}})}\|\mathbf{x}^{n+1}-\mathbf{x}^{*}\|_{1}\leq \left(4\mu|\mathbb{S}|-2\mu\right)^{n+1}|\mathbb{S}|B_{\mathbf{x}}.
\end{equation}

%Let us define
%\begin{equation}\label{c}
%c=-\log\left(4\mu|\mathbb{S}|-2\mu\right).
%\end{equation}
%Then, Eq.~\eqref{es88} is simplified as
Eq.~\eqref{es88} can be rewritten as
\begin{align}
\|\mathbf{x}^{n}-\mathbf{x}^{*}\|_{1}&\leq\sup_{\mathbf{x}^{*}\in\mathcal{X}(B_{\mathbf{x},\mathbb{S}})}\|\mathbf{x}^{n}-\mathbf{x}^{*}\|_{1}\nonumber\\
&\leq|\mathbb{S}|B_{\mathbf{x}}\exp\left(n\log\left(4\mu|\mathbb{S}|-2\mu\right)\right).
\end{align}
Since $\|\mathbf{x}\|_{2}\leq\|\mathbf{x}\|_{1}$, for arbitrary $n\in \mathbb{N}$, 
\begin{equation}\label{new7}
\|\mathbf{x}^{n}-\mathbf{x}^{*}\|_{2}\leq|\mathbb{S}|B_{\mathbf{x}}\exp\left(n\log\left(4\mu|\mathbb{S}|-2\mu\right)\right).
\end{equation}
Therefore, we develop the upper bound of $\|\mathbf{x}_n-\mathbf{x}^*\|_{2}$. Eq.~\eqref{new7} holds uniformly for arbitrary $\mathbf{x}^{*}\in \mathcal{X}(B_{x},s)$ and $n\in\mathbb{N}$, when
\begin{equation}
|\mathbb{S}|<\frac{1}{2}+\frac{1}{4\mu}.
\end{equation}
%then Eq.~\eqref{new7} holds uniformly for all $x^{*}\in \mathcal{X}(B_{x},s)$. 
As a result, we draw Theorem~4.
%\end{proof}

\subsection{Proof of Theorem~5}
We can easily extend Theorem~4 to prove that HLISTA-CPSS with $\mathcal{S}^{p^n}_{ss, \theta^{n}}$ specified in Eq.~(29) satisfies ``no false positive'', when the learnable parameters $\{\overline{\mathbf{W}}^n,\widehat{\mathbf{W}}^n,\theta_1^n,\theta_2^n\}_{n\in\mathbb{N}}$ are determined according to Eq.~(25). Therefore, we focus on the upper bound of recovery error.
%By the definition of $\mathcal{S}^{p^n}_{ss, \theta^{n}}$, using the same argument with the proof of Theorem~3, we easily obtain that HLISTA-CPSS also satisfies "no false positive" if Eq.~\eqref{new1} and \eqref{new4} are satisfied. Furthermore, we prove the upper bound of the recovery error.

%For arbitrary $i \in \mathbb{S}$, by the definition of $\mathcal{S}^{p^n}_{ss, \theta^{n}}$, we have following the same argument of Eq.~\eqref{s21} and \eqref{es83}
For arbitrary $i \in \mathbb{S}$, we can reformulate Eq.~\eqref{s21} and Eq.~\eqref{es83} for $v_i^n$ and $w_i^n$, respectively.
\begin{equation}
v_{i}^{n} \in x_{i}^{*}-\sum_{j\in \mathbb{S}, j\neq i}(\overline{\mathbf{W}}_{i}^{n})^{T}\mathbf{A}_{j}(x_{j}^{n}-x_{j}^{*})-\theta_{1}^{n}\xi(v_{i}^n), 
\end{equation}
and
\begin{align}
w_{i}^{n}
\in& x_{i}^{*}-\sum_{j\not\in \mathbb{S}}(\widehat{\mathbf{W}}_{i}^{n})^{T}\mathbf{A}_{j}u_{j}^{n} \nonumber\\
&-\sum_{j\in \mathbb{S}, j\neq i}(\widehat{\mathbf{W}}_{i}^{n})^{T}\mathbf{A}_{j}(u_{j}^{n}-x_{j}^{*}) -
\theta_{2}^{n}\xi(w_{i}^n),
\end{align}
where 
\begin{equation}
\xi(x_{i}^n)=
\begin{cases}
0, & ~{\rm if}~i \not \in \mathbb{S},\\
[-1,1], & ~{\rm if}~i \in \mathbb{S}, x^n_{i}= 0, \\
\mathrm{sign}(x_{i}), & ~{\rm if}~i \in \mathbb{S}, x^n_{i}\neq 0, i\notin S^{p^n}(\mathbf{x}^n),\\
0, & ~{\rm if}~i \in \mathbb{S}, x^n_{i}\neq 0, i\in S^{p^n}(\mathbf{x}^n).
\end{cases}
\end{equation}
Here, $\mathcal{I}^{p^n}(\mathbf{x}^n)$ includes the indices of the largest $p^n \%$ magnitudes in vector $\mathbf{x}^n$.
Let 
\begin{equation}\label{Psi}
\Psi_{\mathbf{x}^n}=\left\{i|i \in \mathbb{S}, x^n_{i}\neq 0, i\in \mathcal{I}^{p^n}(\mathbf{x}^n)\right\}.
\end{equation}

%Then, using the same argument of Eq.~\eqref{new5} and \eqref{new6}, we obtain
Similar to Eq.~\eqref{new5} and Eq.~\eqref{new6}, we can obtain that
\begin{align}\label{new10}
\|\mathbf{v}^{n}-\mathbf{x}^{*}\|_{1}&\leq\mu(|\mathbb{S}|-1)\|\mathbf{x}^{n}-\mathbf{x}^{*}\|_{1}+(|\mathbb{S}|-|\Psi_{\mathbf{v}^n}|)\theta_{1}^{n} \nonumber\\
&\leq\mu{}{}\left(2|\mathbb{S}|-1-|\Psi_{\mathbf{v}^n}|\right)\sup_{\mathbf{x}^{*}\in\mathcal{X}(B_{\mathbf{x}},\mathbb{S})}\|\mathbf{x}^{n}-\mathbf{x}^{*}\|_{1},
\end{align}
and
\begin{align}\label{new11}
\|\mathbf{w}^{n}-\mathbf{x}^{*}\|_{1}&\leq\mu(|\mathbb{S}|-1)\|\mathbf{u}^{n}-\mathbf{x}^{*}\|_{1}+(|\mathbb{S}|-|\Psi_{\mathbf{w}^n}|)\theta_{2}^{n}\nonumber\\
&\quad+\mu(N-|\mathbb{S}|)\|\mathbf{u}^{n}\|_{1} \nonumber\\
&\leq\mu\left(2|\mathbb{S}|-1- |\Psi_{\mathbf{w}^n}|\right)\nonumber\\
&\quad\cdot\left(\sup_{\mathbf{x}^{*}\in\mathcal{X}(B_{\mathbf{x}},\mathbb{S})}\|\mathbf{u}^{n}-\mathbf{x}^{*}\|_{1}+\frac{N-|\mathbb{S}|}{|\mathbb{S}|-1}\|\mathbf{u}^{n}\|_{1}\right).
\end{align}
From Eq.~\eqref{new10} and Eq.~\eqref{new11}, we obtain that 
%Combining the above two equations, we have following Eq.~\eqref{es87}
\begin{align}\label{s22}
&\sup_{\mathbf{x}^{*}\in\mathcal{X}(B_{\mathbf{x}},\mathbb{S})}\|\mathbf{x}^{n+1}-\mathbf{x}^{*}\|_{1}\nonumber\\
&\qquad\leq\alpha^{n}\mu\left(2|\mathbb{S}|-1-|\Psi_{v^n}|\right)\sup_{\mathbf{x}^{*}\in\mathcal{X}(B_{\mathbf{x}},\mathbb{S})}\|\mathbf{x}^{n}-\mathbf{x}^{*}\|_{1} \nonumber \\
&\qquad\quad+(1-\alpha^{n})\mu\left(2|\mathbb{S}|-1-|\Psi_{\mathbf{w}^n}|\right) \nonumber \\
&\qquad\quad\cdot \left(\sup_{\mathbf{x}^{*}\in\mathcal{X}(B_{\mathbf{x}},\mathbb{S})}\|\mathbf{u}^{n}-\mathbf{x}^{*}\|_{1}+\frac{N-|\mathbb{S}|}{|\mathbb{S}|-1}\|\mathbf{u}^{n}\|_{1}\right).
\end{align}
Let $|\Psi_{*^n}|=\min \{|\Psi_{\mathbf{v}^n}|, |\Psi_{\mathbf{w}^n}|\}$.
%and
%\begin{equation}
%c_{ss}^{k}=-\log\left(4\mu|\mathbb{S}|-2\mu-2\mu|\Psi_{*^k}|\right).
%\end{equation}
Then, from Eq.~\eqref{s22}, Eq.~(21), and Eq.~(25), we have
\begin{align}
&\sup_{\mathbf{x}^{*}\in\mathcal{X}(B_{\mathbf{x}},\mathbb{S})}\|\mathbf{x}^{n+1}-\mathbf{x}^{*}\|_{1}\nonumber\\
&\qquad\leq(\alpha^{n}+1)\mu\left(2|\mathbb{S}|-1-|\Psi_{*^n}|\right)\sup_{\mathbf{x}^{*}\in\mathcal{X}(B_{\mathbf{x}},\mathbb{S})}\|\mathbf{x}^{n}-\mathbf{x}^{*}\|_{1} \nonumber \\
&\qquad\leq\prod_{k=0}^{n}2\mu \left(2|\mathbb{S}|-1-|\Psi_{*^k}|\right)\sup_{\mathbf{x}^{*}\in\mathcal{X}(B_{\mathbf{x}},\mathbb{S})}\|\mathbf{x}^{0}-\mathbf{x}^{*}\|_{1} \nonumber \\
&\qquad\leq|\mathbb{S}|B_{\mathbf{x}} \exp\left(\sum_{k=0}^{n-1}\log\left(4\mu|\mathbb{S}|-2\mu-2\mu|\Psi_{*^k}|\right)\right).
\end{align}
%Since $\|\mathbf{x}\|_{2}\leq\|\mathbf{x}\|_{1}$, we develop the upper bound for $\|x_n-x^*\|_{2}$ for arbitrary $n\in \mathbb{N}$.
Since $\|\mathbf{x}\|_{2}\leq\|\mathbf{x}\|_{1}$, for arbitrary $n\in \mathbb{N}$,
\begin{equation}\label{new12}
\|\mathbf{x}^{n}-\mathbf{x}^{*}\|_{2}\leq|\mathbb{S}|B_{\mathbf{x}}\exp\left(\sum_{k=0}^{n-1}\log\left(4\mu|\mathbb{S}|-2\mu-2\mu|\Psi_{*^k}|\right)\right).
\end{equation}
Therefore, we develop the upper bound of $\|\mathbf{x}_n-\mathbf{x}^*\|_{2}$. Note that Eq.~\eqref{Psi} implies that $|\mathbb{S}|>|\Psi_{*^k}|$. Thus, $4\mu|\mathbb{S}|-2\mu-2\mu|\Psi_{*^k}|>0$ if $|\mathbb{S}|\geq 2$.  
Comparing Eq.~\eqref{new7} and Eq.~\eqref{new12}, HLISTA-CPSS achieves a tighter upper bound than HLISTA-CP (\emph{i.e.}, $n\log(4\mu|\mathbb{S}|-2\mu)<\sum_{k=0}^{n-1}\log(4\mu|\mathbb{S}|-2\mu-2\mu|\Psi_{*^k}|)$, when
\begin{equation}
|\mathbb{S}|<\frac{1}{2}+\frac{1}{4\mu}+\frac{\min_{n}\{|\Psi_{*^n}|\}}{2}.
\end{equation}
%we have that $c_{ss}^k\geq c$.
As a result, we draw Theorem~5.

\subsection{Proof of Theorem~6}
%We propose a choice of parameters that are uniform for $x^{*}\in \mathcal{X}(B_{x}, \mathbb{S})$ with $W\in W_s(A)$ and $\gamma_1^n, \gamma_2^n>0$, as
%\begin{align}\label{t5}
%&\overline{W}^{n} = \gamma_1^n W, ~\widehat{W}^{n} = \gamma_2^n W, \nonumber \\
%&\theta_{1}^{n}=\gamma_1^n\sup_{x^{*}\in\mathcal{X}(B_{x}, \mathbb{S})}\{\mu\|x^{n}-x^{*}\|_{1}\}, \nonumber \\
%&\theta_{2}^{n} =\gamma_2^n \sup_{x^{*}\in\mathcal{X}(B_{x}, \mathbb{S})}\{\mu\|u^{n}-x^{*}\|_{1}\}+\frac{\gamma_2^n\mu(N-|\mathbb{S}|)}{|\mathbb{S}|-1}\|u^n\|_{1}.
%\end{align}

Similar to the proof of Theorem~4, we first prove that HALISTA satisfies ``no false positive''.

%\textbf{i) ``No false positive''.}
\subsubsection{``No False Positive''}
%Recall the steps in HLISTA. Let $S$ = support($x^{*}$).  Fixing $n$, and assuming $x_{i}^{n}=0, \forall i \not \in S$, we obtain
For arbitrary $n\in\mathbb{N}$, we assume that $x_{i}^{n}=0$ for arbitrary $i \not \in \mathbb{S}$. Thus, we have
\begin{align}\label{t6}
v_{i}^{n}=&\mathcal{S}_{\theta_{1}^{n}}\left(x_{i}^{n}+\gamma_1^n\sum_{j}(\mathbf{W}_{i})^{T}(\mathbf{b}-\mathbf{Ax}^{n})_j\right)\nonumber\\
=&\mathcal{S}_{\theta_{1}^{n}}\left(x_{i}^{n}-\gamma_1^n\sum_{j\not\in \mathbb{S}}(\mathbf{W}_{i})^{T}\mathbf{A}_{j}(x_{j}^{n}-x_{j}^{*}) \right.  \nonumber \\
& \qquad\qquad \left. -  \gamma_1^n\sum_{j\in \mathbb{S}}(\mathbf{W}_{i})^{T}\mathbf{A}_{j}(x_{j}^{n}-x_{j}^{*})\right)\nonumber\\
=&\mathcal{S}_{\theta_{1}^{n}}\left( -\gamma_1^n \sum_{j\in \mathbb{S}}(\mathbf{W}_{i})^{T}\mathbf{A}_{j}(x_{j}^{n}-x_{j}^{*})\right).
\end{align}
%Recall Eq.~\eqref{new1}, it follows that
According to Eq.~(36), for arbitrary $i \not\in \mathbb{S}$,
\begin{align}\label{t7}
\theta_{1}^{n}&=\gamma_1^n \mu\sup_{\mathbf{x}^{*}\in\mathcal{X}(B_{\mathbf{x}},\mathbb{S})}\{\|\mathbf{x}^{n}-\mathbf{x}^{*}\|_{1}\}\geq\gamma_1^n \mu\|\mathbf{x}^{n}-\mathbf{x}^{*}\|_{1} \nonumber \\
&\geq\gamma_1^n \sum_{j=1}^N\left|(\mathbf{W}_{i})^{T}\mathbf{A}_{j}\right|\left|x_{j}^{n}-x_{j}^{*}\right| \nonumber\\
&\geq\left|-\gamma_1^n \sum_{j\in \mathbb{S}}(\mathbf{W}_{i})^{T}\mathbf{A}_{j}(x_{j}^{n}-x_{j}^{*})\right|.
\end{align}
According to the definition of $\mathcal{S}_{\theta_{1}^{n}}$, we obtain that $v_{i}^{n}=0$ for arbitrary $i \not \in \mathbb{S}$. Therefore, when $\theta_1^n$ is determined by Eq.~(36),  $v^{n}_{i}=0$ for $x^{n}_{i}=0$, $\forall i \not\in \mathbb{S}, \forall n\in\mathbb{N}$.

Subsequently, we discuss $w_{i}^{n}$ for $i\not \in \mathbb{S}$. Here, we fix $\gamma_2^n$ to $1$, and consequently, obtain $w_i^n$ for $i\not \in \mathbb{S}$ similar to Eq.~\eqref{es76} and Eq.~\eqref{es77} as in Theorem~3. 
%Though the bound of $\gamma_2^n$ is proved in part \textbf{ii)}, it actually affects the discussion here. Thus, we directly apply the result $\gamma_2^n\equiv1$ and consequently $w_{i}^{n}$ for $i\not \in \mathbb{S}$  is similar to Theorem~3 as Eq.s~\eqref{es76} and~\eqref{es77}. More details about the bound of $\gamma_2^n$ is given in part \textbf{ii)}.
Then, we obtain that $w_{i}^{n}=0$ for $i \not \in \mathbb{S}$ and $x_{i}^{n+1}=\alpha^{n}v^{n}_{i}+(1-\alpha^{n})w^{n}_{i}=0$ for $x^{n}_{i}=0$, $\forall i \not\in \mathbb{S}, \forall n\in\mathbb{N}$. Since $\mathbf{x}^{0}=0$, we obtain $x_{i}^{n}=0$ for arbitrary $i \not\in \mathbb{S}$ and $n\in\mathbb{N}$.
%In a conclusion, if $x^{n}_{i}=0$, then we have $x_{i}^{n+1}=\alpha^{n}v^{n}_{i}+(1-\alpha^{n})w^{n}_{i}=0$, $\forall i \not\in S, \forall n$.  Since $x^{0}=0$, by induction we obtain
%\begin{equation}
%	x_{i}^{n}=0, \forall i \not\in \mathbb{S}, \forall n.
%\end{equation}
Therefore, the ``no false positive'' has been proved, \emph{i.e.}, 
\begin{equation}
{\rm support}(\mathbf{x}^n) \subset \mathbb{S}.
\end{equation}
Note that, however, the bound of $\gamma_2^n$ actually has an impact and we further develop the bound of $\gamma_2^n$ in Appendix~\ref{B32}. 
%\textbf{ii) Upper bound of recovery error.}
\subsubsection{Upper Bound of Recovery Error}\label{B32}
%Let's consider the components on $S$. For all $i\in S$, we have
For arbitrary $i\in \mathbb{S}$, we have
\begin{align}\label{t10}
v_{i}^{n} &=\mathcal{S}_{\theta_{1}^{n}}\left(x_{i}^{n}-\gamma_1^n \sum_{j\in \mathbb{S}}(\mathbf{W}_i)^{T}\mathbf{A}_{j}(x_{j}^{n}-x_{j}^{*})\right)\nonumber\\
&=\mathcal{S}_{\theta_{1}^{n}}\left(x_{i}^{n}-\gamma_1^n \sum_{\substack{j\in \mathbb{S},\\ j\neq i}}(\mathbf{W}_i)^{T}\mathbf{A}_{j}(x_{j}^{n}-x_{j}^{*})-\gamma_1^n(x_{i}^{n}-x_{i}^{*})\right)
\end{align}
%Note that 
%\begin{equation}
%x_{i}^{n}-\gamma_1^n(x_{i}^{n}-x_{i}^{*}) = x_{i}^{*}+(1-\gamma_1^n)(x_{i}^{n}-x_{i}^{*}).
%\end{equation}	
%Thus, we obtain from Eq.~\eqref{t10} 
For arbitrary $i \in \mathbb{S}$, since $x_{i}^{n}-\gamma_1^n(x_{i}^{n}-x_{i}^{*}) = x_{i}^{*}+(1-\gamma_1^n)(x_{i}^{n}-x_{i}^{*})$, we obtain from Eq.~\eqref{t10} that
\begin{align}\label{new13}
v_{i}^{n}&\in x_{i}^{*}-\gamma_1^n\sum_{j\in \mathbb{S}, j\neq i}(\mathbf{W}_i)^{T}\mathbf{A}_{j}(x_{j}^{n}-x_{j}^{*}) \nonumber\\
&\quad+(1-\gamma_1^n)(x_{i}^{n}-x_{i}^{*})-\theta_{1}^{n}\partial \mathit{l}_{1}(v_{i}^{n}).
\end{align}
As defined in Eq.~\eqref{new2}, $\partial \mathit{l}_{1}(x)$ has a magnitude not greater than $1$. Thus, we obtain from Eq.~\eqref{new13} that, for arbitrary $i\in \mathbb{S}$, 
%where $\partial \mathit{l}_{1}(x)$ is defined in Eq.~\eqref{new2} that shows $\partial \mathit{l}_{1}(x)$ has a magnitude not greater than 1. From the above equation, we obtain for $i\in \mathbb{S}$,
%From the above equation it's clear that $\partial \mathit{l}_{1}(x)$ has a magnitude less than or equal to 1. Hence, we obtain
\begin{align}
&\left|v_{i}^{n}-x_{i}^{*}\right| \nonumber \\
&\quad\leq\gamma_1^n\sum_{\substack{j\in \mathbb{S},\\ j\neq i}}\left|(\mathbf{W}_i)^{T}\mathbf{A}_{j}\right|\left|x_{j}^{n}-x_{j}^{*}\right|+\theta_{1}^{n} +\left|1-\gamma_1^n\right|\left|x_{i}^{n}-x_{i}^{*}\right|\nonumber\\
&\quad\leq\mu\gamma_1^n\sum_{\substack{j\in \mathbb{S},\\ j\neq i}}\left|x_{j}^{n}-x_{j}^{*}\right|+\theta_{1}^{n}+\left|1-\gamma_1^n\right|\left|x_{i}^{n}-x_{i}^{*}\right|.
\end{align}
Consequently,
\begin{align}
&\|\mathbf{v}^{n}-\mathbf{x}^{*}\|_{1} \nonumber \\
&=\sum_{i\in \mathbb{S}}|v^{n}_{i}-x^{*}_{i}| \nonumber \\
&\leq\sum_{i\in \mathbb{S}}\left(\mu\gamma_1^n\sum_{j\in \mathbb{S}, j\neq i}\left|x_{j}^{n}-x_{j}^{*}\right|+\theta_{1}^{n}+\left|1-\gamma_1^n\right|\left|x_{i}^{n}-x_{i}^{*}\right|\right) \nonumber\\
&=\left[\mu\gamma_1^n(|\mathbb{S}|-1)+\left|1-\gamma_1^n\right|\right]\sum_{i\in \mathbb{S}}\left|x_{i}^{n}-x_{i}^{*}\right|+|\mathbb{S}|\theta_{1}^{n}  \nonumber \\
&\leq\left[\mu\gamma_1^n(|\mathbb{S}|-1)+\left|1-\gamma_1^n\right|\right]\|\mathbf{x}^{n}-\mathbf{x}^{*}\|_{1}+|\mathbb{S}|\theta_{1}^{n}
\end{align}
From Eq.~(36), we have
\begin{align}\label{t11}
\|\mathbf{v}^n-\mathbf{x}^*\|_1\leq\left[\mu\gamma_1^n(2|\mathbb{S}|-1)+\left|1-\gamma_1^n\right|\right]\sup_{\mathbf{x}^{*}\in\mathcal{X}(B_{\mathbf{x}},\mathbb{S})}\|\mathbf{x}^{n}-\mathbf{x}^{*}\|_{1}.
\end{align}
Similarly, for all $i\in \mathbb{S}$, we have
\begin{align}\label{new14}
w_{i}^{n}=&\mathcal{S}_{\theta_{2}^{n}}\left(u_{i}^{n} -\gamma_2^n \sum_{j\not\in \mathbb{S}}(\mathbf{W}_i)^{T}\mathbf{A}_{j}u_{j}^{n} \nonumber\right.\\
&\left.\qquad\qquad\qquad- \gamma_2^n \sum_{j\in \mathbb{S}}(\mathbf{W}_i)^{T}\mathbf{A}_{j}(u_{j}^{n}-x_{j}^{*})\right)\nonumber\\
=&\mathcal{S}_{\theta_{2}^{n}}\left( u_{i}^{n}-\gamma_2^n\sum_{j\not\in \mathbb{S}}(\mathbf{W}_i)^{T}\mathbf{A}_{j}u_{j}^{n}-\gamma_2^n(u_{i}^{n}-x_{i}^{*}) \right. \nonumber \\
&\left.\qquad\qquad\qquad - \gamma_2^n\sum_{j\in \mathbb{S}, j\neq i}(\mathbf{W}_i)^{T}\mathbf{A}_{j}(u_{j}^{n}-x_{j}^{*})\right)\nonumber\\
\in & x_{i}^{*}-\gamma_2^n\sum_{j\not\in \mathbb{S}}(\mathbf{W}_i)^{T}\mathbf{A}_{j}u_{j}^{n}+(1-\gamma_2^n)(u_{i}^{n}-x_{i}^{*}) \nonumber \\
&-\gamma_2^n \sum_{\substack{j\in \mathbb{S},\\ j\neq i}}(\mathbf{W}_i)^{T}\mathbf{A}_{j}(u_{j}^{n}-x_{j}^{*}) - \theta_{2}^{n}\partial \mathit{l}_{1}(w_{i}^{n}). 
\end{align}
From Eq.~\eqref{new14}, we obtain that
\begin{align}
&\left|w_{i}^{n}-x_{i}^{*}\right| \nonumber \\
&\quad\leq \gamma_2^n\sum_{j\in \mathbb{S}, j\neq i}\left|(\mathbf{W}_i)^{T}\mathbf{A}_{j}\right|\left|u_{j}^{n}-x_{j}^{*}\right|+\theta_{2}^{n} \nonumber\\
&\qquad\qquad\quad+\left|1-\gamma_2^n\right|\left|u_{i}^{n}-x_{i}^{*}\right| +\gamma_2^n\sum_{j\not\in \mathbb{S}}\left|(\mathbf{W}_i)^{T}\mathbf{A}_{j}\right|\left|u_{j}^{n}\right| \nonumber\\
&\quad\leq\mu\gamma_2^n\left(\sum_{j\in \mathbb{S}, j\neq i}\left|u_{j}^{n}-x_{j}^{*}\right|+\sum_{j\not\in \mathbb{S}}\left|u_{j}^{n}\right|\right)\nonumber \\
&\qquad\qquad\qquad\qquad\quad+\left|1-\gamma_2^n\right|\left|u_{i}^{n}-x_{i}^{*}\right|+\theta_{2}^{n}.
\end{align}
Then we have
\begin{align}\label{t12}
&\|\mathbf{w}^{n}-\mathbf{x}^{*}\|_{1}  \nonumber\\
&\quad=\sum_{i\in \mathbb{S}}|w^{n}_{i}-x^{*}_{i}| \nonumber \\
&\quad\leq\sum_{i\in \mathbb{S}}\left(\mu\gamma_2^n\sum_{j\in \mathbb{S}, j\neq i}\left|u_{j}^{n}-x_{j}^{*}\right|+\theta_{2}^{n} \right.\nonumber \\
&\qquad\qquad\quad\left.+ \mu\gamma_2^n\sum_{j\not\in \mathbb{S}}\left|u_{j}^{n}\right|+\left|1-\gamma_2^n\right|\left|u_{i}^{n}-x_{i}^{*}\right|\right) \nonumber\\
&\quad=\left[\mu\gamma_2^n(|\mathbb{S}|-1)+\left|1-\gamma_2^n\right|\right]\sum_{i\in \mathbb{S}}\left|u_{i}^{n}-x_{i}^{*}\right|+|\mathbb{S}|\theta_{2}^{n} \nonumber \\
&\quad\quad\ +\mu\gamma_2^n(N-|\mathbb{S}|)\sum_{i\in \mathbb{S}}\left|u_{i}^{n}\right|\nonumber\\
&\quad\leq\left[\mu\gamma_2^n\left(2|\mathbb{S}|-1\right)+\left|1-\gamma_2^n\right|\right]\sup_{\mathbf{x}^{*}\in\mathcal{X}(B_{\mathbf{x}},\mathbb{S})}\|\mathbf{u}^{n}-\mathbf{x}^{*}\|_{1}\nonumber\\
&\quad\quad\ +\frac{\mu\gamma_2^n(2|\mathbb{S}|-1)(N-|\mathbb{S}|)}{|\mathbb{S}|-1}\|\mathbf{u}^{n}\|_{1} .
\end{align}
Combining Eq.~\eqref{t11} and Eq.~\eqref{t12}, we obtain that
\begin{align}\label{t15}
&\sup_{\mathbf{x}^{*}\in\mathcal{X}(B_{\mathbf{x}},\mathbb{S})}\|\mathbf{x}^{n+1}-\mathbf{x}^{*}\|_{1}\nonumber\\
&\quad\leq\sup_{\mathbf{x}^{*}\in\mathcal{X}(B_{\mathbf{x}},\mathbb{S})}\left(\alpha^{n}\|\mathbf{v}^{n}-\mathbf{x}^{*}\|_{1}+(1-\alpha^{n})\|\mathbf{w}^{n}-\mathbf{x}^{*}\|_{1}\right) \nonumber\\
&\quad\leq\alpha^{n}\left[\mu\gamma_1^n(2|\mathbb{S}|-1)+|1-\gamma_1^n|\right]\sup_{\mathbf{x}^{*}\in\mathcal{X}(B_{\mathbf{x}},\mathbb{S})}\|\mathbf{x}^{n}-\mathbf{x}^{*}\|_{1}\nonumber \\
&\qquad+(1-\alpha^{n})\frac{\mu\gamma_2^n(2|\mathbb{S}|-1)(N-|\mathbb{S}|)}{|\mathbb{S}|-1}\|\mathbf{u}^{n}\|_{1}\nonumber\\
&\qquad+(1-\alpha^{n})\left[\mu\gamma_2^n(2|\mathbb{S}|-1)+|1-\gamma_2^n|\right] \nonumber\\
&\qquad\qquad\qquad\qquad\qquad\quad\cdot\sup_{\mathbf{x}^{*}\in\mathcal{X}(B_{\mathbf{x}},\mathbb{S})}\|\mathbf{u}^{n}-\mathbf{x}^{*}\|_{1}.
\end{align}
Recalling $\alpha^{n}$ specified in Eq.~(21), we have
\begin{align}\label{t13}
&(1-\alpha^{n})\mu\gamma_2^n\left(\sup_{\mathbf{x}^{*}\in\mathcal{X}(B_{\mathbf{x}},\mathbb{S})}\|\mathbf{u}^{n}-\mathbf{x}^{*}\|_{1}+\frac{N-|\mathbb{S}|}{|\mathbb{S}|-1}\|\mathbf{u}^{n}\|_{1} \right)\nonumber\\
&\qquad\leq\frac{\theta_{1}^n\theta_{2}^n}{\theta_{1}^n+\theta_{2}^n}\leq \mu\gamma_1^n\sup_{\mathbf{x}^{*}\in\mathcal{X}(B_{\mathbf{x}},\mathbb{S})}\|\mathbf{x}^{n}-\mathbf{x}^{*}\|_{1},
\end{align}
and 
\begin{align}\label{t14}
&(1-\alpha^{n})\left|1-\gamma_2^n\right|\sup_{\mathbf{x}^{*}\in\mathcal{X}(B_{\mathbf{x}},\mathbb{S})}\|\mathbf{u}^{n}-\mathbf{x}^{*}\|_{1}\nonumber \\
&\qquad\qquad\leq \frac{\theta_{1}^n\left|1-\gamma_2^n\right|}{\theta_{1}^n+\theta_{2}^n}\frac{\theta_{2}^n}{\mu\gamma_2^n} \nonumber \\
&\qquad\qquad\leq \frac{\gamma_1^n}{\gamma_2^n}\left|1-\gamma_2^n\right|\sup_{\mathbf{x}^{*}\in\mathcal{X}(B_{\mathbf{x}},\mathbb{S})}\|\mathbf{x}^{n}-\mathbf{x}^{*}\|_{1}.
\end{align}
From Eq.~\eqref{t15}, Eq.~\eqref{t13} and Eq.~\eqref{t14}, we have 
\begin{align}\label{t16}
&\sup_{\mathbf{x}^{*}\in\mathcal{X}(B_{\mathbf{x}},\mathbb{S})}\|\mathbf{x}^{n+1}-\mathbf{x}^{*}\|_{1} \nonumber\\
&\quad\leq(\alpha^{n}+1)\mu\gamma_1^n(2|\mathbb{S}|-1)\sup_{\mathbf{x}^{*}\in\mathcal{X}(B_{\mathbf{x}},\mathbb{S})}\|\mathbf{x}^{n}-\mathbf{x}^{*}\|_{1} \nonumber \\
&\quad\quad+\left(\alpha^n\left|1-\gamma_1^n\right|+\frac{\gamma_1^n}{\gamma_2^n}\left|1-\gamma_2^n\right|\right)\sup_{\mathbf{x}^{*}\in\mathcal{X}(B_{\mathbf{x}},\mathbb{S})}\|\mathbf{x}^{n}-\mathbf{x}^{*}\|_{1} \nonumber \\
&\quad\leq\left[ 2\mu\gamma_1^n(2|\mathbb{S}|-1)+\left(\left|1-\gamma_1^n\right|+\frac{\gamma_1^n}{\gamma_2^n}\left|1-\gamma_2^n\right|\right)\right] \nonumber\\
&\quad\quad\cdot\sup_{\mathbf{x}^{*}\in\mathcal{X}(B_{\mathbf{x}},\mathbb{S})}\|\mathbf{x}^{n}-\mathbf{x}^{*}\|_{1} .
\end{align}
Let us define
\begin{equation}
c^k_a=-\log\left[ 2\mu\gamma_1^k(2|\mathbb{S}|-1)+\left(\left|1-\gamma_1^k\right|+\frac{\gamma_1^k}{\gamma_2^k}\left|1-\gamma_2^k\right|\right)\right].
\end{equation}
From Eq.~\eqref{t16}, we have
\begin{align}
&\sup_{\mathbf{x}^{*}\in\mathcal{X}(B_{\mathbf{x}},\mathbb{S})}\|\mathbf{x}^{n+1}-\mathbf{x}^{*}\|_{1}\nonumber\\
&\qquad\qquad\leq\exp\left(-c_a^n\right)\sup_{\mathbf{x}^{*}\in\mathcal{X}(B_{\mathbf{x}},\mathbb{S})}\|\mathbf{x}^{n}-\mathbf{x}^{*}\|_{1} \nonumber \\
&\qquad\qquad\leq\exp\left(-\sum_{k=0}^{n}c_a^k\right)\sup_{\mathbf{x}^{*}\in\mathcal{X}(B_{\mathbf{x}},\mathbb{S})}\|\mathbf{x}^{0}-\mathbf{x}^{*}\|_{1} \nonumber \\
&\qquad\qquad\leq |\mathbb{S}|B_{\mathbf{x}} \exp\left(-\sum_{k=0}^{n}c_{a}^{k}\right).
\end{align}
Since $\|\mathbf{x}\|_{2}\leq\|\mathbf{x}\|_{1}$, for arbitrary $n\in \mathbb{N}$,
\begin{equation}
\|\mathbf{x}^n-\mathbf{x}^*\|_{2}\leq |\mathbb{S}|B_{\mathbf{x}} \exp\left(-\sum_{k=0}^{n-1}c_{a}^{k}\right).
\end{equation}
Therefore, we develop the upper bound of $\|\mathbf{x}^n-\mathbf{x}^*\|_{2}$ 
However, to guarantee $c_a^k>0$, the following criterion needs to be satisfied.
\begin{equation}\label{gamma}
0<2\mu\gamma_1^k(2|\mathbb{S}|-1)+\left|1-\gamma_1^k\right|+\frac{\gamma_1^k}{\gamma_2^k}\left|1-\gamma_2^k\right|<1
\end{equation}  
Based on $\gamma_1^k, \gamma_2^k>0$, we further prove the sharp bound of $\gamma_1^k$ and $\gamma_2^k$.
 The assumption $|\mathbb{S}|<(2+1/\mu)/4$ gives $2\mu(2|\mathbb{S}|-1)<1$. Thus, we have 
\begin{equation}\label{t17}
2\mu\gamma_1^k(2|\mathbb{S}|-1)+\left|1-\gamma_1^k\right|+\frac{\gamma_1^k}{\gamma_2^k}\left|1-\gamma_2^k\right| \\
< \gamma_1^k+\left|1-\gamma_1^k\right|+\frac{\gamma_1^k}{\gamma_2^k}\left|1-\gamma_2^k\right|
\end{equation}
Thus, we consider the cases that $0<\gamma_2^k\le 1$ and $\gamma_2^k>1$.

i) $0<\gamma_2^k\leq 1$. We have 
\begin{align}\label{t18}
&\gamma_1^k+\left|1-\gamma_1^k\right|+\frac{\gamma_1^k}{\gamma_2^k}\left|1-\gamma_2^k\right|\nonumber\\
&\qquad=\left|1-\gamma_1^k\right|+\frac{\gamma_1^k}{\gamma_2^k} \geq \left|1-\gamma_1^k\right|+\gamma_1^k \geq 1.
\end{align}
Eq.~\eqref{t18} holds if and only if $\gamma_2^k=1$. 

ii) $\gamma_2^k>1$. We have 
\begin{align}\label{t19}
&\gamma_1^k+\left|1-\gamma_1^k\right|+\frac{\gamma_1^k}{\gamma_2^k}\left|1-\gamma_2^k\right|\nonumber\\
&\qquad=2\gamma_1^k+\left|1-\gamma_1^k\right|-\frac{\gamma_1^k}{\gamma_2^k} > \left|1-\gamma_1^k\right|+\gamma_1^k \geq 1.
\end{align}

From Eq.~\eqref{t17}, Eq.~\eqref{t18}, and Eq.~\eqref{t19}, we obtain that Eq.~\eqref{gamma} holds only when $\gamma_2=1$. Note that the constraint on $|\mathbb{S}|$ is not required any longer. When $\gamma_2=1$,  Eq.~\eqref{gamma} holds if 
\begin{equation}
0<\gamma_1^k<\frac{2}{1+4\mu|\mathbb{S}|-2\mu}.
\end{equation}

As a result, we draw Theorem~6.

\subsection{Proof of Theorem~7}
Recall that the $n$th iteration of HGLISTA with gain gates can be written as follows for $n\in\mathbb{N}_{+}$.
\begin{equation}\label{hg_eq1}
\begin{aligned}
&\mathbf{v}^{n}=\mathcal{S}_{\theta_{1}^{n}}\left(\Delta_{g^n}\mathbf{x}^{n}+(\overline{\mathbf{W}}^{n})^{T}(\mathbf{b}-\mathbf{A}\Delta_{g^n}\mathbf{x}^{n})\right), \\
&\mathbf{u}^{n}=N_{\mathcal{W}^{n}}(\mathbf{v}^{n}), \\
&\mathbf{w}^{n}=\mathcal{S}_{\theta_{2}^{n}}\left(\Delta_{g^n}\mathbf{u}^{n}+(\widehat{\mathbf{W}}^{n})^{T}(\mathbf{b}-\mathbf{A}\Delta_{g^n}\mathbf{u}^{n})\right),\\
&\mathbf{x}^{n+1}=\alpha^{n}\mathbf{v}^{n}+(1-\alpha^{n})\mathbf{w}^{n},
\end{aligned}
\end{equation}
where $\Delta_{g^n}\mathbf{x}^{n}= g_t(\mathbf{v}^{n-1}, \mathbf{w}^{n-1}, \mathbf{b}|\Lambda_{g}^n)\odot\mathbf{x}^{n}$, and $\Delta_{g^n}\mathbf{u}^{n}= g_t(\mathbf{v}^{n-1}, \mathbf{w}^{n-1}, \mathbf{b}|\Lambda_{g}^n)\odot\mathbf{u}^{n}$.

\subsubsection{``No False Positive''}
%For arbitrary $n\in\mathbb{N}$, we assume that $x_{i}^{n}=0$ for arbitrary $i \not \in \mathbb{S}$. Here, $x_i^n$ is the $i$-th element of $\mathbf{x}^n$. We can obtain the $i$-th element $v_i^n$ of $\mathbf{v}^n$.
From Eq.~\eqref{hg_eq1}, for arbitrary $n\in\mathbb{N}_{+}$, the $i$th element $v_i^n$ of $\mathbf{v}^n$ is
\begin{align}
v_{i}^{n}
=\mathcal{S}_{\theta_{1}^{n}}\Bigg(\Delta_{g^n}^{i}x_{i}^{n}-&\sum_{j\not\in \mathbb{S}}(\overline{\mathbf{W}}_{i}^{n})^{T}\mathbf{A}_{j}(\Delta_{g^n}^{j}x_{j}^{n}-x_{j}^{*}) \nonumber\\
&- \sum_{j\in \mathbb{S}}(\overline{\mathbf{W}}_{i}^{n})^{T}\mathbf{A}_{j}(\Delta_{g^n}^{j}x_{j}^{n}-x_{j}^{*})\Bigg),
\end{align}
where $\Delta_{g^n}^{i}x_{i}^{n}$ represents the $i$th element of $\Delta_{g^n}\mathbf{x}^{n}$.
Let us assume that $x_{i}^{n}=0$ for arbitrary $i \not \in \mathbb{S}$. Then we have
\begin{align}
v_i^n=\mathcal{S}_{\theta_{1}^{n}}\left( - \sum_{j\in \mathbb{S}}(\overline{\mathbf{W}}_{i}^{n})^{T}\mathbf{A}_{j}(\Delta_{g^n}^{j}x_{j}^{n}-x_{j}^{*})\right)
\end{align}
as $\Delta_{g^n}^{i}x_{i}^{n}=0$ for arbitrary $i \not \in \mathbb{S}$.
According to Eq.~(47), for arbitrary $i \not\in \mathbb{S}$,
\begin{align}
\theta_{1}^{n}&\geq\mu\|\Delta_{g^n}\mathbf{x}^{n}-\mathbf{x}^{*}\|_{1}\geq\sum_{j}\left|(\overline{\mathbf{W}}_{i}^{n})^{T}\mathbf{A}_{j}\right|\left|\Delta_{g^n}^{j}x_{j}^{n}-x_{j}^{*}\right|\nonumber\\
&\geq\left|-\sum_{j\in \mathbb{S}}(\overline{\mathbf{W}}_{i}^{n})^{T}\mathbf{A}_{j}(\Delta_{g^n}^{j}x_{j}^{n}-x_{j}^{*})\right|.
\end{align}
According to the definition of $\mathcal{S}_{\theta_{1}^{n}}$, $v_{i}^{n}=0$ for arbitrary $i \not \in \mathbb{S}$. Therefore, when $\theta_1^n$ is determined by Eq.~(47),  $v^{n}_{i}=0$ for $x^{n}_{i}=0$, $\forall i \not\in \mathbb{S}, \forall n\in\mathbb{N}_{+}$.
%By the definition of $\mathcal{S}_{\theta_{1}^{n}}$, we easily deduce that $v_{i}^{n}=0, \forall i \not \in S$. Thus we conclude that as long as Eq.~\eqref{new1} holds, if $x^{n}_{i}=0$, it follows that $v^{n}_{i}=0$, $\forall i \not\in S, \forall n$. 

Subsequently, we similarly consider $w_{i}^{n}$ for $i\not \in \mathbb{S}$. 
\begin{align}\label{hg_eq3}
w_{i}^{n}=&\mathcal{S}_{\theta_{2}^{n}}\left(\Delta_{g^n}^{i}u_{i}^{n}-\sum_{j\not\in \mathbb{S}}(\widehat{\mathbf{W}}_{i}^{n})^{T}\mathbf{A}_{j}\Delta_{g^n}^{j}u_{j}^{n} \right. \nonumber \\
&\qquad\qquad\qquad\left.- \sum_{j\in \mathbb{S}}(\widehat{\mathbf{W}}_{i}^{n})^{T}\mathbf{A}_{j}(\Delta_{g^n}^{j}u_{j}^{n}-x_{j}^{*})\right)\nonumber\\
=&\mathcal{S}_{\theta_{2}^{n}}\left( -\sum_{j\not\in \mathbb{S}, j\neq i}(\widehat{\mathbf{W}}_{i}^{n})^{T}\mathbf{A}_{j}\Delta_{g^n}^{j}u_{j}^{n} \right.\nonumber\\
&\left.\qquad\qquad\qquad-\sum_{j\in \mathbb{S}}(\widehat{\mathbf{W}}_{i}^{n})^{T}\mathbf{A}_{j}(\Delta_{g^n}^{j}u_{j}^{n}-x_{j}^{*})\right),
\end{align}
where $\Delta_{g^n}^{i}u_{i}^{n}$ represents the $i$-th element of $\Delta_{g^n}\mathbf{u}^{n}$.
From Eq.~(47), we have
\begin{align}\label{hg_eq4}
\theta_{2}^{n} &\geq\mu\|\Delta_{g^n}\mathbf{u}^{n}-\mathbf{x}^{*}\|_{1} \nonumber\\
&\geq\sum_{j\not\in \mathbb{S}, j\neq i}\left|(\widehat{\mathbf{W}}_{i}^{n})^{T}\mathbf{A}_{j}\right|\left|\Delta_{g^n}^{j}u_{j}^{n}\right|\nonumber\\
&\qquad\qquad\qquad+\sum_{j\in \mathbb{S}}\left|(\widehat{\mathbf{W}}_{i}^{n})^{T}\mathbf{A}_{j}\right|\left|\Delta_{g^n}^{j}u_{j}^{n}-x_{j}^{*}\right| \nonumber\\
&\geq\left|-\sum_{j\not\in \mathbb{S}, j\neq i}(\widehat{\mathbf{W}}_{i}^{n})^{T}\mathbf{A}_{j}\Delta_{g^n}^{j}u_{j}^{n} \right.\nonumber\\
&\qquad\qquad\qquad\left.- \sum_{j\in \mathbb{S}}(\widehat{\mathbf{W}}_{i}^{n})^{T}\mathbf{A}_{j}(\Delta_{g^n}^{j}u_{j}^{n}-x_{j}^{*})\right|.
\end{align}
From Eq.~\eqref{hg_eq3} and Eq.~\eqref{hg_eq4}, we obtain that $w_{i}^{n}=0$ for $i \not \in \mathbb{S}$. Therefore, for arbitrary $n\in\mathbb{N}_{+}$ and $i\notin\mathbb{S}$, we have $x_{i}^{n+1}=\alpha^{n}v^{n}_{i}+(1-\alpha^{n})w^{n}_{i}=0$, when $x^{n}_{i}=0$. Note that the first iteration of HGLISTA is the same as HLISTA and we have proven that HLISTA-CP achieves ``no false positive''. Thus, introducing $\mathbf{x}^{0}=0$, we obtain $x_{i}^{n}=0$ for arbitrary $n\in\mathbb{N}$ and $i \not\in \mathbb{S}$.
%In a conclusion, if $x^{n}_{i}=0$, then we have $x_{i}^{n+1}=\alpha^{n}v^{n}_{i}+(1-\alpha^{n})w^{n}_{i}=0$, $\forall i \not\in S, \forall n$.  Since $x^{0}=0$, by induction we obtain
%\begin{equation}
%x_{i}^{n}=0, \forall i \not\in \mathbb{S}, \forall n.
%\end{equation}
Therefore, the ``no false positive'' has been proved, \emph{i.e.}, 
\begin{equation}
{\rm support}(\mathbf{x}^n) \subset \mathbb{S}.
\end{equation}

\subsubsection{Upper Bound of Recovery Error}
%Let's consider the components on $S$. For all $i\in S$, we have
Following the definition of $\Delta_{g^n}\mathbf{x}^{n}$ and $\Delta_{g^n}\mathbf{u}^{n}$ , we then define $\Delta_{g^{n+1}}\mathbf{v}^{n}$, $\Delta_{g^{n+1}}\mathbf{w}^{n}$ and $\Delta_{g^{n+1}}\mathbf{x}^{n+1}$, and the $i$th elements are denoted by $\Delta_{g^{n+1}}^{i}v_{i}^{n}$, $\Delta_{g^{n+1}}^{i}w_{i}^{n}$, and $\Delta_{g^{n+1}}^{i}x_{i}^{n+1}$, respectively.

Note that the gain gate function can be written as 
\begin{equation}\label{gain_gate}
g_t(\mathbf{v}^{n},\mathbf{w}^{n},  \mathbf{b}|\Lambda_{g}^{n+1})_i = 1 + \kappa_t(\mathbf{v}^{n},\mathbf{w}^{n},  \mathbf{b}|\Lambda_{g}^{n+1})_i,    
\end{equation}
Let 
\begin{align}
   &\theta_{max}^{n}= \max\{\theta_1^{n},\theta_2^{n}\},~ \theta_{min}^{n}= \min\{\theta_1^{n},\theta_2^{n}\},\nonumber\\
   &\Xi_{i}^{n}=\max\{|v_i^n|, |w_i^n|\},~
   \Upsilon_{i}^{n}= \min\{|v_i^n|, |w_i^n|\},
\end{align}
then the range is specified as for $i\in(\mathbb{S}\cap\mathrm{supp}(\mathbf{v}^n))\cup(\mathbb{S}\cap\mathrm{supp}(\mathbf{w}^n))$
\begin{equation}\label{app_kappa_cond}
    \frac{(1-\varrho^{n})\theta_{max}^{n}}{\Upsilon_{i}^{n}}\leq\kappa_t(\mathbf{v}^{n}, \mathbf{w}^{n}, \mathbf{b}|\Lambda_{g}^{n+1})_i \leq \frac{(1+\varrho^{n})\theta_{min}^{n}}{\Xi_{i}^{n}},
\end{equation}
where $\varrho^{n}$ is a constant and satisfies that 
\begin{equation}
\sup_{i\in\mathbb{Q}}\left\{\frac{\theta_{max}^{n}\Xi_{i}^{n}-\theta_{min}^{n}\Upsilon_{i}^{n}}{\theta_{max}^{n}\Xi_{i}^{n}+\theta_{min}^{n}\Upsilon_{i}^{n}}\right\}\leq \varrho^{n} < 1,
\end{equation}
where $\mathbb{Q}=(\mathbb{S}\cap\mathrm{supp}(\mathbf{v}^n))\cup(\mathbb{S}\cap\mathrm{supp}(\mathbf{w}^n))$. And we further define the following vectors for the subsequent proofs. 
\begin{align}
&\Delta_{\kappa^{n+1}}\mathbf{v}^{n}= \kappa_t(\mathbf{v}^{n}, \mathbf{w}^{n},  \mathbf{b}|\Lambda_{g}^{n+1})\odot\mathbf{v}^{n}, \nonumber\\ &\Delta_{\kappa^{n+1}}|\mathbf{v}^{n}|= \kappa_t(\mathbf{v}^{n}, \mathbf{w}^{n},  \mathbf{b}|\Lambda_{g}^{n+1})\odot|\mathbf{v}^{n}|, \nonumber\\ 
&\Delta_{\kappa^{n+1}}\mathbf{w}^{n}= \kappa_t(\mathbf{v}^{n}, \mathbf{w}^{n},  \mathbf{b}|\Lambda_{g}^{n+1})\odot\mathbf{w}^{n}, \nonumber\\ &\Delta_{\kappa^{n+1}}|\mathbf{w}^{n}|= \kappa_t(\mathbf{v}^{n}, \mathbf{w}^{n},  \mathbf{b}|\Lambda_{g}^{n+1})\odot|\mathbf{w}^{n}|.
\end{align}

For arbitrary $i\in \mathbb{S}$, we have
\begin{align}\label{hg_eq5}
v_{i}^{n}=&\mathcal{S}_{\theta_{1}^{n}}\left(\Delta_{g^n}^{i}x_{i}^{n}-\sum_{j\in \mathbb{S}}(\overline{\mathbf{W}}_{i}^{n})^{T}\mathbf{A}_{j}(\Delta_{g^n}^{j}x_{j}^{n}-x_{j}^{*})\right)\nonumber\\
=&\mathcal{S}_{\theta_{1}^{n}}\left(\Delta_{g^n}^{i}x_{i}^{n}-\sum_{j\in \mathbb{S}, j\neq i}(\overline{\mathbf{W}}_{i}^{n})^{T}\mathbf{A}_{j}(\Delta_{g^n}^{j}x_{j}^{n}-x_{j}^{*})\right.\nonumber\\
&\qquad\qquad\qquad\qquad\qquad\quad-(\Delta_{g^n}^{i}x_{i}^{n}-x_{i}^{*})\Bigg)\nonumber\\
=&\mathcal{S}_{\theta_{1}^{n}}\left(x_{i}^{*}-\sum_{j\in \mathbb{S}, j\neq i}(\overline{\mathbf{W}}_{i}^{n})^{T}\mathbf{A}_{j}(\Delta_{g^n}^{j}x_{j}^{n}-x_{j}^{*})\right)\nonumber\\
\in&x_{i}^{*}-\sum_{j\in \mathbb{S}, j\neq i}(\overline{\mathbf{W}}_{i}^{n})^{T}\mathbf{A}_{j}(\Delta_{g^n}^{j}x_{j}^{n}-x_{j}^{*})-\theta_{1}^{n}\partial \mathit{l}_{1}(v_{i}^{n}), 
\end{align}
where $\partial \mathit{l}_{1}(v_{i}^{n})$ is the sub-gradient of $\|v_{i}^{n}\|_{1}$ that is defined by 
%It is a set which can be defined in a component-wisely fashion.
\begin{equation}
\partial \mathit{l}_{1}(v_{i}^{n})=
\begin{cases}
\mathrm{sign}(v_{i}^{n}), & ~~~{\rm if}~v_{i}^{n}\neq 0, \\
[-1,1], & ~~~{\rm if}~v_{i}^{n}= 0.
\end{cases}
\end{equation}
According to Eq.~\eqref{gain_gate} and Eq.~\eqref{hg_eq5}, we obtain for $i\in \mathbb{S}$,
\begin{multline}\label{hg_eq7}
    \Delta_{g^{n+1}}^{i} v_i^n - x_{i}^{*}\in \\ \Delta_{\kappa^{n+1}}^{i}v_i^n-\sum_{j\in \mathbb{S}, j\neq i}(\overline{\mathbf{W}}_{i}^{n})^{T}\mathbf{A}_{j}(\Delta_{g^n}^{j}x_{j}^{n}-x_{j}^{*})-\theta_{1}^{n}\partial \mathit{l}_{1}(v_{i}^{n}),
\end{multline}
where $\Delta_{\kappa^{n+1}}^{i}v_i^n$ represents the $i$th element of $\Delta_{\kappa^{n+1}}\mathbf{v}^n$.
To calculate $|\Delta_{g^{n+1}}^{i}v_i^n - x_{i}^{*}|$, we consider two cases for the index $i$: i) $i\in \mathbb{S}$ but $i \not \in \mathrm{supp}(\mathbf{v}^n)$, and ii) $i\in \mathbb{S}$ and $i\in \mathrm{supp}(\mathbf{v}^n)$.

i) $i\in \mathbb{S}$ but $i \not \in \mathrm{supp}(\mathbf{v}^n)$. We have that $v_i^n=0$ and $-1\leq \partial \mathit{l}_{1}(v_{i}^{n})\leq 1$. Thus, we obtain 
\begin{align}
\left|\Delta_{g^{n+1}}^{i}v_i^n - x_{i}^{*}\right|&\leq\left|\sum_{j\in \mathbb{S}, j\neq i}(\overline{\mathbf{W}}_{i}^{n})^{T}\mathbf{A}_{j}(\Delta_{g^n}^{j}x_{j}^{n}-x_{j}^{*})\right|+\theta_{1}^{n}\nonumber\\
&\leq \mu\sum_{j\in \mathbb{S}, j\neq i}\left|\Delta_{g^n}^{j}x_{j}^{n}-x_{j}^{*}\right|+\theta_{1}^{n}.
\end{align}
ii) $i\in \mathbb{S}$ and $i \in \mathrm{supp}(\mathbf{v}^n)$. We have that $v_i^n\neq 0$ and $\partial \mathit{l}_{1}(v_{i}^{n})=\mathrm{sign}(v_{i}^{n})$. Thus, we obtain 
\begin{equation}\label{hg_eq6}
    \left|\Delta_{g^{n+1}}^{i}v_i^n - x_{i}^{*}\right|\leq \mu\sum_{j\in \mathbb{S}, j\neq i}\left|\Delta_{g^n}^{j}x_{j}^{n}-x_{j}^{*}\right|+\left|\Delta_{\kappa^{n+1}}^{i}|v_i^n|-\theta_{1}^{n}\right|,
\end{equation}
where $\Delta_{\kappa^{n+1}}^{i}|v_i^n|$ is the $i$th element of $\Delta_{\kappa^{n+1}}|\mathbf{v}^n|$.
According to Eq.~\eqref{app_kappa_cond}, we have
\begin{align}\label{hg_eq13n}
\Delta_{\kappa^{n+1}}^{i}|v_i^n|-\theta_1^n 
 &\leq \frac{(1+\varrho^{n})|v_i^n|\min\{\theta_1^{n},\theta_2^{n}\}}{\max\{|v_i^n|, |w_i^n|\}}-\theta_1^n \nonumber \\
 &\leq (1+\varrho^{n})\min\{\theta_1^{n},\theta_2^{n}\}-\theta_1^n  \nonumber \\
  &\leq \varrho^{n}\theta_1^n,
\end{align}
and
\begin{align}\label{hg_eq14}
\Delta_{\kappa^{n+1}}^{i}|v_i^n|-\theta_1^n 
 &\geq \frac{(1-\varrho^{n})|v_i^n|\max\{\theta_1^{n},\theta_2^{n}\}}{\min\{|v_i^n|, |w_i^n|\}}-\theta_1^n \nonumber \\
 &\geq (1-\varrho^{n})\max\{\theta_1^{n},\theta_2^{n}\}-\theta_1^n  \nonumber \\
 &\geq -\varrho^{n}\theta_1^n,
\end{align}
Thus, we have $|\Delta_{\kappa^{n+1}}^{i}|v_i^n|-\theta_1^n |\leq \varrho^{n}\theta_1^n$.
Substituting it to Eq.~\eqref{hg_eq6}, we obtain
\begin{equation}\label{hg_eq9}
   \left|\Delta_{g^{n+1}}^{i} v_i^n - x_{i}^{*}\right|\leq \mu\sum_{j\in \mathbb{S}, j\neq i}\left|\Delta_{g^n}^{j}x_{j}^{n}-x_{j}^{*}\right|+\varrho^{n}\theta_1^n.
\end{equation}
Accumulate all the $\left|\Delta_{g^{n+1}}^{i} v_i^n - x_{i}^{*}\right|$ with all $i\in\mathbb{S}$, and define $s_\mathbf{v}^n=|\mathrm{supp}(\mathbf{v}^n)|$ as the cardinality of $\mathrm{supp}(\mathbf{v}^n)$, 
\begin{align}\label{hg_eq11}
&\left\|\Delta_{g^{n+1}} \mathbf{v}^n - \mathbf{x}^{*}\right\|_1 \nonumber\\
&\quad\leq \sum_{i\in \mathbb{S}}\mu\sum_{j\in \mathbb{S}, j\neq i}\left|\Delta_{g^n}^{j}x_{j}^{n}-x_{j}^{*}\right|+(s_\mathbf{v}^n\varrho^{n}+|\mathbb{S}|-s_\mathbf{v}^n)\theta_1^n \nonumber\\
&\quad\leq (|\mathbb{S}|-1)\mu\left\|\Delta_{g^n}\mathbf{x}^{n}-\mathbf{x}^{*}\right\|_1\nonumber\\
&\qquad+(s_\mathbf{v}^n\varrho^{n}+|\mathbb{S}|-s_\mathbf{v}^n)\sup_{\mathbf{x}^{*}\in\mathcal{X}(B_{\mathbf{x}},\mathbb{S})}\{\mu\|\Delta_{g^n}\mathbf{x}^{n}-\mathbf{x}^{*}\|_{1}\} \nonumber\\
&\quad\leq \left[2|\mathbb{S}|-1-(1-\varrho^{n})s_\mathbf{v}^n\right]\mu\sup_{\mathbf{x}^{*}}\{\mu\|\Delta_{g^n}\mathbf{x}^{n}-\mathbf{x}^{*}\|_{1}\}.
\end{align}

Similarly, for arbitrary $i\in \mathbb{S}$, we have
\begin{align}\label{hg_eq10}
w_{i}^{n}
=&\mathcal{S}_{\theta_{2}^{n}}\left(\Delta_{g^n} ^{i}u_{i}^{n}-\sum_{j\not\in \mathbb{S}}(\widehat{\mathbf{W}}_{i}^{n})^{T}\mathbf{A}_{j}\Delta_{g^n}^{j}u_{j}^{n} \right.\nonumber\\
&\qquad\qquad\qquad\left.- \sum_{j\in\mathbb{S}}(\widehat{\mathbf{W}}_{i}^{n})^{T}\mathbf{A}_{j}(\Delta_{g^n}^{j}u_{j}^{n}-x_{j}^{*})\right)\nonumber\\
=&\mathcal{S}_{\theta_{2}^{n}}\left( x_{i}^{*}-\sum_{j\not\in \mathbb{S}}(\widehat{\mathbf{W}}_{i}^{n})^{T}\mathbf{A}_{j}\Delta_{g^n}^{j}u_{j}^{n} \right.\nonumber\\
&\qquad\qquad\qquad\left.- \sum_{j\in \mathbb{S}, j\neq i}(\widehat{\mathbf{W}}_{i}^{n})^{T}\mathbf{A}_{j}(\Delta_{g^n}^{j}u_{j}^{n}-x_{j}^{*})\right)\nonumber\\
\in & x_{i}^{*}-\sum_{j\not\in \mathbb{S}}(\widehat{\mathbf{W}}_{i}^{n})^{T}\mathbf{A}_{j}\Delta_{g^n}^{j}u_{j}^{n} \nonumber \\
& \quad- \sum_{j\in \mathbb{S}, j\neq i}(\widehat{\mathbf{W}}_{i}^{n})^{T}\mathbf{A}_{j}(\Delta_{g^n}^{j}u_{j}^{n}-x_{j}^{*}) -
\theta_{2}^{n}\partial \mathit{l}_{1}(w_{i}^{n}). 
\end{align}
Similar to Eq.~\eqref{hg_eq7}, we obtain 
\begin{multline}\label{hg_eq8}
    \Delta_{g^{n+1}}^{i} w_i^n - x_{i}^{*}\in \Delta_{\kappa^{n+1}}^{i}w_i^n-\sum_{j\not\in \mathbb{S}}(\widehat{\mathbf{W}}_{i}^{n})^{T}\mathbf{A}_{j}\Delta_{g^n}^{j}u_{j}^{n}\\-\sum_{j\in \mathbb{S}, j\neq i}(\widehat{\mathbf{W}}_{i}^{n})^{T}\mathbf{A}_{j}(\Delta_{g^n}^{j}u_{j}^{n}-x_{j}^{*})-\theta_{2}^{n}\partial \mathit{l}_{1}(w_{i}^{n})
\end{multline}
To calculate $|\Delta_{g^{n+1}}^{i} w_i^n - x_{i}^{*}|$, we also consider the two cases for the index $i$ as below.

i) $i\in \mathbb{S}$ but $i \not \in \mathrm{supp}(\mathbf{w}^n)$. We have that $w_i^n=0$ and $-1\leq \partial \mathit{l}_{1}(w_{i}^{n})\leq 1$. Thus, we obtain 
\begin{align}
&\left|\Delta_{g^{n+1}}^{i}w_{i}^{n}-x_{i}^{*}\right| \nonumber \\
&\ \leq \sum_{j\in \mathbb{S}, j\neq i}\left|(\widehat{\mathbf{W}}_{i}^{n})^{T}\mathbf{A}_{j}\right|\left|\Delta_{g^n}^{j}u_{j}^{n}-x_{j}^{*}\right|+\theta_{2}^{n} \nonumber\\
&\quad\qquad\qquad\qquad\qquad\qquad+\sum_{j\not\in \mathbb{S}}\left|(\widehat{\mathbf{W}}_{i}^{n})^{T}\mathbf{A}_{j}\right|\left|\Delta_{g^n}^{j}u_{j}^{n}\right| \nonumber\\
&\ \leq\mu\sum_{j\in \mathbb{S}, j\neq i}\left|\Delta_{g^n}^{j}u_{j}^{n}-x_{j}^{*}\right|+\theta_{2}^{n}+ \mu\sum_{j\not\in \mathbb{S}}\left|\Delta_{g^n}^{j}u_{j}^{n}\right|.
\end{align}
ii) $i\in \mathbb{S}$ and $i \in \mathrm{supp}(\mathbf{w}^n)$. We have that $w_i^n\neq 0$ and $\partial \mathit{l}_{1}(w_{i}^{n})=\mathrm{sign}(w_{i}^{n})$. Thus, we obtain 
\begin{equation}
\left|\Delta_{g^{n+1}}^{i}w_{i}^{n}-x_{i}^{*}\right| \leq  \mu\sum_{j\in \mathbb{S}, j\neq i}\left|\Delta_{g^n}^{j}u_{j}^{n}-x_{j}^{*}\right|+\left|\Delta_{\kappa^{n+1}}^{i}|w_i^n|-\theta_{2}^{n}\right|+ \mu\sum_{j\not\in \mathbb{S}}\left|\Delta_{g^n}^{j}u_{j}^{n}\right|,
\end{equation}
where $\Delta_{\kappa^{n+1}}^{i}|w_i^n|$ is the $i$th element of $\Delta_{\kappa^{n+1}}|\mathbf{w}^n|$.
Using the same argument as Eq.~\eqref{hg_eq13n}, \eqref{hg_eq14} and \eqref{hg_eq9}, we obtain 
\begin{equation}
\left|\Delta_{g^{n+1}}^{i}w_{i}^{n}-x_{i}^{*}\right| \leq  \mu\sum_{j\in \mathbb{S}, j\neq i}\left|\Delta_{g^n}^{j}u_{j}^{n}-x_{j}^{*}\right|+\varrho^{n}\theta_2^n+ \mu\sum_{j\not\in \mathbb{S}}\left|\Delta_{g^n}^{j}u_{j}^{n}\right|.
\end{equation}
Accumulate all the $\left|\Delta_{g^{n+1}}^{i} w_i^n - x_{i}^{*}\right|$ with all $i\in\mathbb{S}$, and define $s_\mathbf{w}^n=|\mathrm{supp}(\mathbf{w}^n)|$ as the cardinality of $\mathrm{supp}(\mathbf{w}^n)$, 
\begin{align}\label{hg_eq12}
&\left\|\Delta_{g^{n+1}} \mathbf{w}^n - \mathbf{x}^{*}\right\|_1 \nonumber\\
&\quad\leq \sum_{i\in \mathbb{S}}\mu\sum_{j\in \mathbb{S}, j\neq i}\left|\Delta_{g^n}^{j}u_{j}^{n}-x_{j}^{*}\right|+(s_\mathbf{w}^n\varrho^{n}+|\mathbb{S}|-s_\mathbf{w}^n)\theta_2^n \nonumber\\
&\quad\quad+\sum_{i\in \mathbb{S}}\mu\sum_{j\not\in \mathbb{S}}\left|\Delta_{g^n}^{j}u_{j}^{n}\right| \nonumber\\
&\quad\leq (|\mathbb{S}|-1)\mu\left\|\Delta_{g^n}\mathbf{u}^{n}-\mathbf{x}^{*}\right\|_1+\mu(N-|\mathbb{S}|)\|\Delta_{g^n}\mathbf{u}^{n}\|_1\nonumber\\
&\quad\quad+(s_\mathbf{w}^n\varrho^{n}+|\mathbb{S}|-s_\mathbf{w}^n)\left(\sup_{\mathbf{x}^{*}\in\mathcal{X}(B_{\mathbf{x}},\mathbb{S})}\{\mu\|\Delta_{g^n}\mathbf{u}^{n}-\mathbf{x}^{*}\|_{1}\}\right.\nonumber\\
&\qquad\qquad\qquad\qquad\qquad\qquad\quad\left. +\frac{\mu(N-|\mathbb{S}|)}{|\mathbb{S}|-1}\|\Delta_{g^n}\mathbf{u}^{n}\|_{1}\right)\nonumber\\
&\quad\leq \left[2|\mathbb{S}|-1-(1-\varrho^{n})s_\mathbf{w}^n\right]\mu\left(\sup_{\mathbf{x}^{*}}\{\mu\|\Delta_{g^n}\mathbf{u}^{n}-\mathbf{x}^{*}\|_{1}\}\right.\nonumber\\
&\qquad\qquad\qquad\qquad\qquad\qquad\quad\left. +\frac{\mu(N-|\mathbb{S}|)}{|\mathbb{S}|-1}\|\Delta_{g^n}\mathbf{u}^{n}\|_{1}\right).
\end{align}

Let $s_{*}^n = \min\{s_{\mathbf{v}}^n, s_{\mathbf{w}}^n\}.$
Then for arbitrary $n\in\mathbb{N}$, we obtain from Eq.~\eqref{hg_eq11} and Eq.~\eqref{hg_eq12} that
\begin{align}\label{hg_eq13}
&\sup_{\mathbf{x}^{*}\in\mathcal{X}(B_{\mathbf{x}},\mathbb{S})}\|\Delta_{g^{n+1}}\mathbf{x}^{n+1}-\mathbf{x}^{*}\|_{1}\nonumber\\
&\quad\leq\sup_{\mathbf{x}^{*}\in\mathcal{X}(B_{\mathbf{x}},\mathbb{S})}(\alpha^{n}\|\Delta_{g^{n+1}}\mathbf{v}^{n}-\mathbf{x}^{*}\|_{1} \nonumber\\
&\qquad\qquad\qquad\qquad\qquad\quad+(1-\alpha^{n})\|\Delta_{g^{n+1}}\mathbf{w}^{n}-\mathbf{x}^{*}\|_{1}) \nonumber\\
&\quad\leq\alpha^{n}\left[2|\mathbb{S}|-1-(1-\varrho^{n})s_\mathbf{*}^n\right]\mu\sup_{\mathbf{x}^{*}}\{\mu\|\Delta_{g^n}\mathbf{x}^{n}-\mathbf{x}^{*}\|_{1}\} \nonumber \\
&\qquad+(1-\alpha^{n})\left[2|\mathbb{S}|-1-(1-\varrho^{n})s_\mathbf{*}^n\right]\mu\nonumber \\
&\qquad \cdot\left(\sup_{\mathbf{x}^{*}}\{\mu\|\Delta_{g^n}\mathbf{u}^{n}-\mathbf{x}^{*}\|_{1}\} +\frac{\mu(N-|\mathbb{S}|)}{|\mathbb{S}|-1}\|\Delta_{g^n}\mathbf{u}^{n}\|_{1}\right).
\end{align}
Recalling the choice of $\alpha^{n}$ in Eq.~(21), we obtain from Eq.~\eqref{hg_eq13} that
\begin{align}\label{hg_eq15}
&\sup_{\mathbf{x}^{*}\in\mathcal{X}(B_{\mathbf{x}},\mathbb{S})}\|\Delta_{g^{n+1}}\mathbf{x}^{n+1}-\mathbf{x}^{*}\|_{1}\nonumber\\ 
&\ \leq (\alpha^{n}+1)\mu\left[2|\mathbb{S}|-1-(1-\varrho^{n})s_\mathbf{*}^n\right] \cdot\sup_{\mathbf{x}^{*}\in\mathcal{X}(B_{\mathbf{x}})}\|\Delta_{g^n}\mathbf{x}^{n}-\mathbf{x}^{*}\|_{1} \nonumber \\
&\ \leq \prod_{k=1}^{n}2\mu\left[2|\mathbb{S}|-1-(1-\varrho^{k})s_\mathbf{*}^k\right]\cdot \sup_{\mathbf{x}^{*}\in\mathcal{X}(B_{\mathbf{x}})}\|\Delta_{g^1}\mathbf{x}^{1}-\mathbf{x}^{*}\|_{1}. 
\end{align}
Note that we do not utilize the gain gates in the first iteration to generate $\mathbf{x}^{1}$, thus the first iteration is the same as HLISTA-CP. Similar to Eq.~\eqref{hg_eq7} and \eqref{hg_eq8}, we obtain
\begin{equation}
\begin{aligned}
&\Delta_{g^1}^{i} v_i^0 - x_{i}^{*}\in \Delta_{\kappa^1}^{i}v_i^0-\sum_{j\in \mathbb{S}, j\neq i}(\overline{\mathbf{W}}_{i}^{0})^{T}\mathbf{A}_{j}(x_{j}^{0}-x_{j}^{*})-\theta_{1}^{0}\partial \mathit{l}_{1}(v_{i}^{0}), \\
    &\Delta_{g^1}^{i} w_i^0 - x_{i}^{*}\in \Delta_{\kappa^1}^{i}w_i^0-\sum_{j\not\in \mathbb{S}}(\widehat{\mathbf{W}}_{i}^{0})^{T}\mathbf{A}_{j}u_{j}^{0}\\
    &\qquad-\sum_{j\in \mathbb{S}, j\neq i}(\widehat{\mathbf{W}}_{i}^{0})^{T}\mathbf{A}_{j}(u_{j}^{0}-x_{j}^{*})-\theta_{2}^{0}\partial \mathit{l}_{1}(w_{i}^{0}).
\end{aligned}
\end{equation}
Using the same process of obtaining Eq.~\eqref{hg_eq11}, \eqref{hg_eq12}, \eqref{hg_eq13} and \eqref{hg_eq15}, we can obtain that 
\begin{align}\label{hg_eq16}
   &\sup_{\mathbf{x}^{*}\in\mathcal{X}(B_{\mathbf{x}})}\|\Delta_{g^{1}}\mathbf{x}^{1}-\mathbf{x}^{*}\|_{1}\nonumber\\
   &\quad\leq 2\mu\left[2|\mathbb{S}|-1-(1-\varrho^{0})s_\mathbf{*}^0\right] \sup_{\mathbf{x}^{*}\in\mathcal{X}(B_{\mathbf{x}})}\|\mathbf{x}^{0}-\mathbf{x}^{*}\|_{1}\nonumber\\
   &\quad\leq2\mu\left[2|\mathbb{S}|-1-(1-\varrho^{0})s_\mathbf{*}^0\right]|\mathbb{S}|B_{\mathbf{x}}.
\end{align}
Substituting Eq.~\eqref{hg_eq16} to Eq.~\eqref{hg_eq15}, we obtain for arbitrary $n\in\mathbb{N}$,
\begin{equation}\label{hg_eq19}
\sup_{\mathbf{x}^{*}\in\mathcal{X}(B_{\mathbf{x}},\mathbb{S})}\|\Delta_{g^{n+1}}\mathbf{x}^{n+1}-\mathbf{x}^{*}\|_{1}\leq\prod_{k=0}^{n}2\mu\left[2|\mathbb{S}|-1-(1-\varrho^{k})s_\mathbf{*}^k\right]|\mathbb{S}|B_{\mathbf{x}}.
\end{equation}

Next, we shall establish the relationship between $\sup_{\mathbf{x}^{*}\in\mathcal{X}(B_{\mathbf{x}},\mathbb{S})}\|\Delta_{g^{n+1}}\mathbf{x}^{n+1}-\mathbf{x}^{*}\|_{1}$ and \\
$\sup_{\mathbf{x}^{*}\in\mathcal{X}(B_{\mathbf{x}},\mathbb{S})}\|\mathbf{x}^{n+1}-\mathbf{x}^{*}\|_{1}$.
According to Eq.~\eqref{hg_eq5}, we have for arbitrary $n\in\mathbb{N}_{+}$,
\begin{align}\label{hg_eq17}
    \|\mathbf{v}^n-\mathbf{x}^*\|_1&\leq |\sum_{i\in \mathbb{S}}\sum_{j\in \mathbb{S}, j\neq i}(\overline{\mathbf{W}}_{i}^{n})^{T}\mathbf{A}_{j}(\Delta_{g^n}^{j}x_{j}^{n}-x_{j}^{*})|+|\mathbb{S}|\theta_{1}^{n}\nonumber\\
    &\leq (2|\mathbb{S}|-1)\mu\sup_{\mathbf{x}^{*}\in\mathcal{X}(B_{\mathbf{x}},\mathbb{S})}\|\Delta_{g^n}\mathbf{x}^n-\mathbf{x}^*\|_1.
\end{align}
Similar to the above inequality, we also have from Eq.~\eqref{hg_eq10}
\begin{equation}\label{hg_eq18}
    \|\mathbf{w}^n-\mathbf{x}^*\|_1\leq  (2|\mathbb{S}|-1)\mu\Bigg(\sup_{\mathbf{x}^{*}\in\mathcal{X}(B_{\mathbf{x}},\mathbb{S})}\|\Delta_{g^n}\mathbf{u}^n-\mathbf{x}^*\|_1 
    +\frac{\mu(N-|\mathbb{S}|)}{|\mathbb{S}|-1}\|\Delta_{g^n}\mathbf{u}^{n}\|_{1}\Bigg).
\end{equation}
Recalling the choice of $\alpha^{n}$ in Eq.~(21), we obtain from Eq.~\eqref{hg_eq17} and \eqref{hg_eq18} for arbitrary $n\in\mathbb{N}_{+}$,
\begin{align}\label{hg_eq20}
&\sup_{\mathbf{x}^{*}\in\mathcal{X}(B_{\mathbf{x}},\mathbb{S})}\|\mathbf{x}^{n+1}-\mathbf{x}^{*}\|_{1}\nonumber\\
&\quad\leq\sup_{\mathbf{x}^{*}\in\mathcal{X}(B_{\mathbf{x}},\mathbb{S})}\left(\alpha^{n}\|\mathbf{v}^{n}-\mathbf{x}^{*}\|_{1}+(1-\alpha^{n})\|\mathbf{w}^{n}-\mathbf{x}^{*}\|_{1}\right) \nonumber\\
&\quad\leq\alpha^{n}\mu\left(2|\mathbb{S}|-1\right)\sup_{\mathbf{x}^{*}\in\mathcal{X}(B_{\mathbf{x}},\mathbb{S})}\|\Delta_{g^n}\mathbf{x}^{n}-\mathbf{x}^{*}\|_{1} \nonumber \\
&\qquad+(1-\alpha^{n})\mu\left(2|\mathbb{S}|-1\right)\Bigg(\sup_{\mathbf{x}^{*}\in\mathcal{X}(B_{\mathbf{x}},\mathbb{S})}\|\Delta_{g^n}\mathbf{u}^{n}-\mathbf{x}^{*}\|_{1}\nonumber\\
&\qquad\qquad\qquad\qquad\qquad\qquad\qquad\quad+\frac{N-|\mathbb{S}|}{|\mathbb{S}|-1}\|\Delta_{g^n}\mathbf{u}^{n}\|_{1}\Bigg)\nonumber\\
&\quad \leq 2\mu\left(2|\mathbb{S}|-1\right)\sup_{\mathbf{x}^{*}\in\mathcal{X}(B_{\mathbf{x}},\mathbb{S})}\|\Delta_{g^n}\mathbf{x}^{n}-\mathbf{x}^{*}\|_{1} .
\end{align}
Substituting Eq.~\eqref{hg_eq19} to Eq.~\eqref{hg_eq20}, we obtain for arbitrary $n\in\mathbb{N}_{+}$,
\begin{align}\label{hg_eq21}
&\sup_{\mathbf{x}^{*}\in\mathcal{X}(B_{\mathbf{x}},\mathbb{S})}\|\mathbf{x}^{n+1}-\mathbf{x}^{*}\|_{1}\nonumber\\
&\quad\leq 2\mu\left(2|\mathbb{S}|-1\right)\prod_{k=0}^{n-1}2\mu\left[2|\mathbb{S}|-1-(1-\varrho^{k})s_\mathbf{*}^k\right]|\mathbb{S}|B_{\mathbf{x}}\nonumber\\
&\quad\leq\exp{\left(-\sum_{k=0}^{n-1}c^k_{g}-c\right)}|\mathbb{S}|B_{\mathbf{x}},
\end{align}
where
\begin{align}
    c^k_{g}&=-\log\left(4\mu|\mathbb{S}|-2\mu-2(1-\varrho^{k})\mu s_\mathbf{*}^k\right), \nonumber\\
    c&=-\log\left(4\mu|\mathbb{S}|-2\mu\right).
\end{align}
Since $\|\mathbf{x}\|_{2}\leq\|\mathbf{x}\|_{1}$, for $n\geq2$, 
\begin{equation}
\begin{aligned}
\|\mathbf{x}^{n}-\mathbf{x}^{*}\|_{2}&\leq\sup_{\mathbf{x}^{*}\in\mathcal{X}(B_{\mathbf{x}},\mathbb{S})}\|\mathbf{x}^{n}-\mathbf{x}^{*}\|_{1}\nonumber\\
&\leq\exp{\left(-\sum_{k=0}^{n-2}c^k_{g}-c\right)}|\mathbb{S}|B_{\mathbf{x}}.
\end{aligned}
\end{equation}
The above equation holds uniformly for arbitrary $\mathbf{x}^{*}\in \mathcal{X}(B_{x},s)$ and $n\geq2$, when
\begin{equation}
|\mathbb{S}|<\frac{1}{2}+\frac{1}{4\mu}.
\end{equation}
%then Eq.~\eqref{new7} holds uniformly for all $x^{*}\in \mathcal{X}(B_{x},s)$. 
When $n=1$, one can refer to the conclusion of HLISTA-CP.

As a result, we draw Theorem~7.

\subsection{Proof of Theorem~8}
Recall that the $n$th iteration of HELISTA can be written as follows for $n\in\mathbb{N}$.
\begin{equation}\label{he_eq1}
\begin{aligned}
\mathbf{v}^{n}&=M_{\theta_{1}^{n}, \hat{\theta}_{1}^{n}}\left(\mathbf{x}^{n}+\gamma^n_1(\mathbf{W})^{T}(\mathbf{b}-\mathbf{Ax}^{n})\right), \\
\mathbf{v}^{n+\frac{1}{2}}&=M_{\theta_{2}^{n}, \hat{\theta}_{2}^{n}}\left(\mathbf{x}^{n}+\gamma^n_2(\mathbf{W})^{T}(\mathbf{b}-\mathbf{Av}^{n})\right), \\
\mathbf{u}^{n}&=N_{\mathcal{W}^{n}}(\mathbf{v}^{n+\frac{1}{2}}), \\
\mathbf{w}^{n}&=M_{\theta_{3}^{n}, \hat{\theta}_{3}^{n}}\left(\mathbf{u}^{n}+\gamma^n_3(\mathbf{W})^{T}(\mathbf{b}-\mathbf{A}\mathbf{u}^{n})\right),\\
\mathbf{w}^{n+\frac{1}{2}}&=M_{\theta_{4}^{n}, \hat{\theta}_{4}^{n}}\left(\mathbf{u}^{n}+\gamma^n_4(\mathbf{W})^{T}(\mathbf{b}-\mathbf{A}\mathbf{w}^{n})\right),\\
\mathbf{x}^{n+1}&=\alpha^{n}\mathbf{v}^{n+\frac{1}{2}}+(1-\alpha^{n})\mathbf{w}^{n+\frac{1}{2}},
\end{aligned}
\end{equation}
where $M_{\theta, \widehat{\theta}}$ is defined as
\begin{equation}
  M_{\theta, \widehat{\theta}}(x)=\left\{
  \begin{aligned}
  0 & , & 0\leq|x|<\theta, \\
  \frac{\widehat{\theta}}{\widehat{\theta}-\theta}{\rm sgn}(x)(|x|-\theta) & , & \theta\leq|x|<\widehat{\theta}, \\
  x & , & |x|\geq\widehat{\theta}.
  \end{aligned}
  \right.
\end{equation}

\subsubsection{``No False Positive''}
For arbitrary $n\in\mathbb{N}$, we assume that $x_{i}^{n}=0$ for arbitrary $i \not \in \mathbb{S}$. Thus, we have
\begin{align}
v_{i}^{n}=&M_{\theta_{1}^{n}, \hat{\theta}_{1}^{n}}\left(x_{i}^{n}+\gamma_1^n\sum_{j}(\mathbf{W}_{i})^{T}(\mathbf{b}-\mathbf{Ax}^{n})_j\right)\nonumber\\
=&M_{\theta_{1}^{n}, \hat{\theta}_{1}^{n}}\left(x_{i}^{n}-\gamma_1^n\sum_{j\not\in \mathbb{S}}(\mathbf{W}_{i})^{T}\mathbf{A}_{j}(x_{j}^{n}-x_{j}^{*}) \right.  \nonumber \\
& \qquad\qquad \left. -  \gamma_1^n\sum_{j\in \mathbb{S}}(\mathbf{W}_{i})^{T}\mathbf{A}_{j}(x_{j}^{n}-x_{j}^{*})\right)\nonumber\\
=&M_{\theta_{1}^{n}, \hat{\theta}_{1}^{n}}\left( -\gamma_1^n \sum_{j\in \mathbb{S}}(\mathbf{W}_{i})^{T}\mathbf{A}_{j}(x_{j}^{n}-x_{j}^{*})\right).
\end{align}
According to the choice of $\theta_1^n$ in Eq.~(54), for arbitrary $i \not\in \mathbb{S}$,
\begin{align}
\theta_{1}^{n}&=\gamma_1^n \mu\sup_{\mathbf{x}^{*}\in\mathcal{X}(B_{\mathbf{x}},\mathbb{S})}\{\|\mathbf{x}^{n}-\mathbf{x}^{*}\|_{1}\}\geq\gamma_1^n \mu\|\mathbf{x}^{n}-\mathbf{x}^{*}\|_{1} \nonumber \\
&\geq\gamma_1^n \sum_{j=1}^N\left|(\mathbf{W}_{i})^{T}\mathbf{A}_{j}\right|\left|x_{j}^{n}-x_{j}^{*}\right| \nonumber\\
&\geq\left|-\gamma_1^n \sum_{j\in \mathbb{S}}(\mathbf{W}_{i})^{T}\mathbf{A}_{j}(x_{j}^{n}-x_{j}^{*})\right|.
\end{align}
According to the definition of $M_{\theta_{1}^{n}, \hat{\theta}_{1}^{n}}$, we obtain that $v_{i}^{n}=0$ for arbitrary $i \not \in \mathbb{S}$. Therefore, when $\theta_1^n$ is determined by Eq.~(54),  $v^{n}_{i}=0$ for $x^{n}_{i}=0$, $\forall i \not\in \mathbb{S}, \forall n\in\mathbb{N}$. One can see that the proof is similar to HALISTA even with the new thresholding function. We can easily obtain that $v_i^{n+\frac{1}{2}}=0$, $w_i^{n}=0$ and $w_i^{n+\frac{1}{2}}=0$ for arbitrary $i \not\in \mathbb{S}$, $\forall n\in\mathbb{N}$. Therefore, for arbitrary $n\in\mathbb{N}$ and $i\notin\mathbb{S}$, we have $x_{i}^{n+1}=\alpha^{n}v^{n+\frac{1}{2}}_{i}+(1-\alpha^{n})w^{n+\frac{1}{2}}_{i}=0$, when $x^{n}_{i}=0$.  Thus, introducing $\mathbf{x}^{0}=0$, we obtain $x_{i}^{n}=0$ for arbitrary $n\in\mathbb{N}$ and $i \not\in \mathbb{S}$.

Therefore, the ``no false positive'' has been proved, \emph{i.e.}, 
\begin{equation}
{\rm support}(\mathbf{x}^n) \subset \mathbb{S}.
\end{equation}

\subsubsection{Upper Bound of Recovery Error}
If we define $\mathbf{z}=M_{\theta, \hat{\theta}}(\mathbf{x})$, $\Tilde{\mathbf{z}}=\mathcal{S}_{\theta}(\mathbf{x})$, then we have 
\begin{equation}\label{he_eq3}
    \mathbf{z} = \mathbf{K}_{\mathbf{z}}\odot\Tilde{\mathbf{z}},
\end{equation}
where $\mathbf{K}_{\mathbf{z}}$ is a vector and the $i$th element is defined as
\begin{equation}
(\mathbf{K}_{\mathbf{z}})_i=
\begin{cases}
\frac{\hat{\theta}}{\hat{\theta}-\theta}, & ~~~{\rm if}~ 0\leq|\Tilde{z}_i|<\hat{\theta}, \\
1, & ~~~{\rm if}~|\Tilde{z}_i|\geq\hat{\theta}.
\end{cases}
\end{equation}
One can refer to Lemma~2 in ELISTA for the detailed proofs of Eq.~\eqref{he_eq3}. Hereby, we begin to develop the upper bound of recovery error.

$\mathbf{1)~The~upper~bound~for~\|\mathbf{v}^n - \mathbf{x}^{*}\|_1 .}$

Define the $i$th element of $\Tilde{\mathbf{v}}^n$ as
\begin{equation}\label{he_eq4}
\Tilde{v}^n_i=\mathcal{S}_{\theta_{1}^{n}}\left(x_{i}^{n}+\gamma_1^n\sum_{j}(\mathbf{W}_{i})^{T}(\mathbf{b}-\mathbf{Ax}^{n})_j\right)
\end{equation}
when $0\leq|\Tilde{v}^n_i|<\hat{\theta}_{1}^{n}$, and \begin{equation}\label{he_eq16}
\Tilde{v}^n_i=x_{i}^{n}+\gamma_1^n\sum_{j}(\mathbf{W}_{i})^{T}(\mathbf{b}-\mathbf{Ax}^{n})_j
\end{equation}
when $|\Tilde{v}^n_i|\geq\hat{\theta}_{1}^{n}$.
Then according to Eq.~\eqref{he_eq1} and \eqref{he_eq3}, we obtain that 
\begin{equation}
     \mathbf{v}^n = \mathbf{K}_{\mathbf{v}^n}\odot\Tilde{\mathbf{v}}^n,
\end{equation}
where
\begin{equation}
(\mathbf{K}_{\mathbf{v}^n})_i=
\begin{cases}
\frac{\hat{\theta}^n_1}{\hat{\theta}^n_1-\theta^n_1}, & ~~~{\rm if}~ 0\leq|\Tilde{v}^n_i|<\hat{\theta}^n_1, \\
1, & ~~~{\rm if}~|\Tilde{v}^n_i|\geq\hat{\theta}^n_1.
\end{cases}
\end{equation}
%Similarly, we also obtain that $\mathbf{x}^n=\mathbf{K}_{\mathbf{x}^n}\odot\Tilde{\mathbf{x}}^n$.

We first consider the case of  $0\leq|\Tilde{v}^n_i|<\hat{\theta}^n_1$. We have from Eq.~\eqref{he_eq4} for arbitrary $i\in\mathbb{S}$,
\begin{align}
\Tilde{v}^n_i &=\mathcal{S}_{\theta_{1}^{n}}\left(x_{i}^{n}-\gamma_1^n \sum_{j\in \mathbb{S}}(\mathbf{W}_i)^{T}\mathbf{A}_{j}(x_{j}^{n}-x_{j}^{*})\right)\nonumber\\
&=\mathcal{S}_{\theta_{1}^{n}}\left(x_{i}^{n}-\gamma_1^n \sum_{\substack{j\in \mathbb{S},\\ j\neq i}}(\mathbf{W}_i)^{T}\mathbf{A}_{j}(x_{j}^{n}-x_{j}^{*})-\gamma_1^n(x_{i}^{n}-x_{i}^{*})\right)\nonumber\\
&\in x_{i}^{*}-\gamma_1^n\sum_{j\in \mathbb{S}, j\neq i}(\mathbf{W}_i)^{T}\mathbf{A}_{j}(x_{j}^{n}-x_{j}^{*})\nonumber\\
&\qquad\qquad\qquad+(1-\gamma_1^n)(x_{i}^{n}-x_{i}^{*})-\theta_{1}^{n}\partial \mathit{l}_{1}(\Tilde{v}_{i}^{n}).
\end{align}
Thus, we obtain that
\begin{equation}
\begin{aligned}\label{he_eq6}
&v_i^n-x_i^*\\
=&(\mathbf{K}_{\mathbf{v}^n})_i\Tilde{v}^n_i -x_i^* \\
=&-\gamma_1^n\sum_{j\in \mathbb{S}, j\neq i}(\mathbf{W}_i)^{T}\mathbf{A}_{j}(x_{j}^{n}-x_{j}^{*})+(1-\gamma_1^n)(x_{i}^{n}-x_{i}^{*})\\
&-\theta_{1}^{n}\partial \mathit{l}_{1}(\Tilde{v}_{i}^{n})+\frac{\theta^n_1}{\hat{\theta}^n_1-\theta^n_1}\Tilde{v}^n_i.
\end{aligned}
\end{equation}
We shall calculate $|v_i^n-x_i^*|$ with the index $i$ that can be divided into two cases. One is $i\in \mathbb{S}$ but $i \not \in \mathrm{supp}(\mathbf{v}^n)$, another one is $i\in \mathbb{S}$ and $i\in \mathrm{supp}(\mathbf{v}^n)$.
For $i\in \mathbb{S}$ but $i \not \in \mathrm{supp}(\mathbf{v}^n)$, we have $\Tilde{v}_{i}^{n}=0$, and
\begin{align}\label{he_eq5}
&\left|v_{i}^{n}-x_{i}^{*}\right| \nonumber \\
&~\leq\gamma_1^n\sum_{\substack{j\in \mathbb{S},\\ j\neq i}}\left|(\mathbf{W}_i)^{T}\mathbf{A}_{j}\right|\left|x_{j}^{n}-x_{j}^{*}\right|+\theta_{1}^{n} +\left|1-\gamma_1^n\right|\left|x_{i}^{n}-x_{i}^{*}\right|\nonumber\\
&~\leq\mu\gamma_1^n\sum_{\substack{j\in \mathbb{S},\\ j\neq i}}\left|x_{j}^{n}-x_{j}^{*}\right|+\theta_{1}^{n}+\left|1-\gamma_1^n\right|\left|x_{i}^{n}-x_{i}^{*}\right|.
\end{align}
For $i\in \mathbb{S}$ and $i \in \mathrm{supp}(\mathbf{v}^n)$, we have $\Tilde{v}_{i}^{n}\not=0$, and
\begin{equation}
\left|v_{i}^{n}-x_{i}^{*}\right| \leq\mu\gamma_1^n\sum_{\substack{j\in \mathbb{S},\\ j\neq i}}\left|x_{j}^{n}-x_{j}^{*}\right| +\left|1-\gamma_1^n\right|\left|x_{i}^{n}-x_{i}^{*}\right|+\left|\frac{\theta^n_1}{\hat{\theta}^n_1-\theta^n_1}|\Tilde{v}^n_i|-\theta_{1}^{n}\right|.
\end{equation}
Note that the discussion is based on $0\leq|\Tilde{v}^n_i|<\hat{\theta}^n_1$. When $0<|\Tilde{v}^n_i|\leq\hat{\theta}^n_1-\theta^n_1$ ($\Tilde{v}_{i}^{n}\not=0$ for $i \in \mathrm{supp}(\mathbf{v}^n)$), 
%(\request{Based on the assumption that $\|\Tilde{\mathbf{v}}^n\|_1$ is bounded for $\forall n\in \mathbb{N}$.}) We assume $\hat{\theta}_{1}^{n}\geq\theta_{1}^{n}+|\Tilde{v}_{\mathrm{max}}^n|$ where $|\Tilde{v}_{\mathrm{max}}^n|=\max\{|\Tilde{v}^n_i|\}$ for $\forall i\in[1, N]$. Then 
we have 
\begin{equation}\label{he_eq10}
\begin{aligned}
\left|\frac{\theta^n_1}{\hat{\theta}^n_1-\theta^n_1}|\Tilde{v}^n_i|-\theta_{1}^{n}\right|=& \theta_{1}^{n}-\frac{\theta^n_1}{\hat{\theta}^n_1-\theta^n_1}|\Tilde{v}^n_i| \\
=&\left(1-\frac{|\Tilde{v}^n_i|}{\hat{\theta}^n_1-\theta^n_1}\right)\theta_{1}^{n} <\theta_1^n.
\end{aligned}
\end{equation}
When $\hat{\theta}^n_1-\theta^n_1<|\Tilde{v}^n_i|<\hat{\theta}^n_1$, we have 
\begin{equation}\label{he_eq11}
\begin{aligned}
\left|\frac{\theta^n_1}{\hat{\theta}^n_1-\theta^n_1}|\Tilde{v}^n_i|-\theta_{1}^{n}\right|=\left(\frac{|\Tilde{v}^n_i|}{\hat{\theta}^n_1-\theta^n_1}-1\right)\theta_{1}^{n} <\frac{(\theta_1^n)^2}{\hat{\theta}^n_1-\theta^n_1}.
\end{aligned}
\end{equation}

Then, We consider the case of $|\Tilde{v}^n_i|\geq\hat{\theta}^n_1$. In this case, $\Tilde{v}^n_i=v^n_i$, and we obtain from Eq.~\eqref{he_eq16}
\begin{equation}\label{he_eq8}
|v_i^n-x_i^*|
\leq\mu\gamma_1^n\sum_{\substack{j\in \mathbb{S},\\ j\neq i}}\left|x_{j}^{n}-x_{j}^{*}\right|+\left|1-\gamma_1^n\right|\left|x_{i}^{n}-x_{i}^{*}\right|.
\end{equation}

Hereby, we shall calculate $\left\|\mathbf{v}^n - \mathbf{x}^{*}\right\|_1$. Define the sets
\begin{equation}
\begin{aligned}
\mathbb{V}^n_L &= \{i|i\in\mathbb{S},  \hat{\theta}^n_1-\theta^n_1<|\Tilde{v}^n_i|<\hat{\theta}^n_1\}, \\
\mathbb{V}^n_S &= \{i|i\in\mathbb{S},  0\leq|\Tilde{v}^n_i|\leq\hat{\theta}^n_1-\theta^n_1\},
\end{aligned}
\end{equation}
and $|\mathbb{V}_L^n|$ and $|\mathbb{V}_S^n|$ as the cardinality of $\mathbb{V}_L^n$ and $\mathbb{V}_S^n$, respectively. Note that $\hat{\theta}_{1}^{n}=(1+1/\epsilon^n_1)\theta_{1}^{n}$ according to Eq.~(54).
%where $|\Tilde{v}^n_{\mathrm{min}}|=\min\{|\Tilde{v}^n_i|\}$ for $\forall i\in[1, N]$.and define $u_{\mathbf{v}}^n$ as the cardinality of set $U_{\mathbf{v}}^n$, where $U_{\mathbf{v}}^n\triangleq\{i|i\in\mathbb{S}, 0\leq|\Tilde{v}^n_i|<\hat{\theta}^n_1\}$. 
Accumulate all the $\left| v_i^n - x_{i}^{*}\right|$ with all $i\in\mathbb{S}$, we obtain
\begin{align}\label{he_eq14}
&\left\|\mathbf{v}^n - \mathbf{x}^{*}\right\|_1 \nonumber\\
&\quad\leq \sum_{i\in \mathbb{S}}\Bigg(\mu\gamma_1^n\sum_{\substack{j\in \mathbb{S},\\ j\neq i}}\left|x_{j}^{n}-x_{j}^{*}\right|+\left|1-\gamma_1^n\right|\left|x_{i}^{n}-x_{i}^{*}\right|\Bigg) \nonumber \\
&\quad\quad+|\mathbb{V}^n_S|\theta_1^n +\epsilon_1^n|\mathbb{V}^n_L|\theta_1^n\nonumber\\
&\quad\leq\left[\mu\gamma_1^n(|\mathbb{S}|-1)+\left|1-\gamma_1^n\right|\right]\|\mathbf{x}^{n}-\mathbf{x}^{*}\|_{1}\nonumber \\
&\quad\quad+(|\mathbb{V}^n_S|+\epsilon_1^n|\mathbb{V}^n_L|)\theta_1^n \nonumber\\
&\quad\leq\left[\mu\gamma_1^n(|\mathbb{S}|+|\mathbb{V}^n_S|+\epsilon_1^n|\mathbb{V}^n_L|-1)+\left|1-\gamma_1^n\right|\right]\nonumber\\
&\qquad\qquad\qquad\qquad\cdot\sup_{\mathbf{x}^{*}\in\mathcal{X}(B_{\mathbf{x}},\mathbb{S})}\{\|\mathbf{x}^{n}-\mathbf{x}^{*}\|_{1}\}.
\end{align}

$\mathbf{2)~The~upper~bound~for~\|\mathbf{v}^{n+\frac{1}{2}} - \mathbf{x}^{*}\|_1 .}$

Similar to the definition of $\mathbf{K}_{\mathbf{v}^n}$, $\Tilde{\mathbf{v}}^n$ and $\Tilde{v}_i^n$, we also define $\mathbf{K}_{\mathbf{v}^{n+\frac{1}{2}}}$,  $\Tilde{\mathbf{v}}^{n+\frac{1}{2}}$ and $\Tilde{v}_i^{n+\frac{1}{2}}$. 

We first consider the case of  $0\leq|\Tilde{v}^{n+\frac{1}{2}}_i|<\hat{\theta}^n_2$. We have for arbitrary $i\in\mathbb{S}$,
\begin{align}\label{he_eq7}
\Tilde{v}^{n+\frac{1}{2}}_i &=\mathcal{S}_{\theta_{2}^{n}}\left(x_{i}^{n}-\gamma_2^n \sum_{j\in \mathbb{S}}(\mathbf{W}_i)^{T}\mathbf{A}_{j}(v_{j}^{n}-x_{j}^{*})\right)\nonumber\\
&=\mathcal{S}_{\theta_{2}^{n}}\left(x_{i}^{n}-\gamma_2^n \sum_{\substack{j\in \mathbb{S},\\ j\neq i}}(\mathbf{W}_i)^{T}\mathbf{A}_{j}(v_{j}^{n}-x_{j}^{*})-\gamma_2^n(v_{i}^{n}-x_{i}^{*})\right)\nonumber\\
&\in x_{i}^{*} + (x_{i}^{n}-x_{i}^{*})
-\gamma_2^n\sum_{j\in \mathbb{S}, j\neq i}(\mathbf{W}_i)^{T}\mathbf{A}_{j}(v_{j}^{n}-x_{j}^{*})\nonumber\\
&\qquad\qquad\qquad-\gamma_2^n(v_{i}^{n}-x_{i}^{*})-\theta_{2}^{n}\partial \mathit{l}_{1}(\Tilde{v}_{i}^{n+\frac{1}{2}}).
\end{align}
Thus, we obtain
%have from Eq.~\eqref{he_eq6} and \eqref{he_eq7}
\begin{equation}\label{he_eq17}
\begin{aligned}
&v_i^{n+\frac{1}{2}}-x_i^*\\
=&(\mathbf{K}_{\mathbf{v}^{n+\frac{1}{2}}})_i\Tilde{v}^{n+\frac{1}{2}}_i -x_i^* \\
=&(x_{i}^{n}-x_{i}^{*})-\gamma_2^n\sum_{\substack{j\in \mathbb{S},\\ j\neq i}}(\mathbf{W}_i)^{T}\mathbf{A}_{j}(v_{j}^{n}-x_{j}^{*})\\
&-\theta_{2}^{n}\partial \mathit{l}_{1}(\Tilde{v}_{i}^{n+\frac{1}{2}})+\frac{\theta^n_2}{\hat{\theta}^n_2-\theta^n_2}\Tilde{v}_i^{n+\frac{1}{2}}\\
&-\gamma_2^n\Bigg(-\gamma_1^n\sum_{\substack{j\in \mathbb{S},\\ j\neq i}}(\mathbf{W}_i)^{T}\mathbf{A}_{j}(x_{j}^{n}-x_{j}^{*})+(1-\gamma_1^n)(x_{i}^{n}-x_{i}^{*})\\
&-(\mathcal{C}_{\mathbf{v}^n})_i\theta_{1}^{n}\partial \mathit{l}_{1}(\Tilde{v}_{i}^{n})+(\mathcal{C}_{\mathbf{v}^n})_i\frac{\theta^n_1}{\hat{\theta}^n_1-\theta^n_1}\Tilde{v}^n_i\Bigg) \\
=&\left(1-\gamma_2^n+\gamma_1^n\gamma_2^n\right)(x_i^n-x_i^*)-\gamma_2^n\sum_{\substack{j\in \mathbb{S},\\ j\neq i}}(\mathbf{W}_i)^{T}\mathbf{A}_{j}(v_{j}^{n}-x_{j}^{*})\\
&+\gamma_1^n\gamma_2^n\sum_{\substack{j\in \mathbb{S},\\ j\neq i}}(\mathbf{W}_i)^{T}\mathbf{A}_{j}(x_{j}^{n}-x_{j}^{*})-\theta_{2}^{n}\partial \mathit{l}_{1}(\Tilde{v}_{i}^{n+\frac{1}{2}})\\
&+(\mathcal{C}_{\mathbf{v}^n})_i\gamma_2^n\theta_{1}^{n}\partial \mathit{l}_{1}(\Tilde{v}_{i}^{n})+\frac{\theta^n_2}{\hat{\theta}^n_2-\theta^n_2}\Tilde{v}_i^{n+\frac{1}{2}}-(\mathcal{C}_{\mathbf{v}^n})_i\frac{\gamma_2^n\theta^n_1}{\hat{\theta}^n_1-\theta^n_1}\Tilde{v}^n_i.
\end{aligned}
\end{equation}
where $\mathcal{C}_{\mathbf{v}^n}$ is a vector and the $i$th element is defined as
\begin{equation}\label{he_eq12}
(\mathcal{C}_{\mathbf{v}^n})_i=
\begin{cases}
1, & ~~~{\rm if}~ ~i\in(\mathbb{V}^n_L\cup  \mathbb{V}^n_S),\\
0, & ~~~{\rm else},
\end{cases}
\end{equation}
and the second equation holds due to Eq.~\eqref{he_eq6} and \eqref{he_eq8}.
We shall calculate $|v_i^{n+\frac{1}{2}}-x_i^*|$ with the index $i$ that can be divided into two cases. One is $i\in \mathbb{S}$ but $i \not \in \mathrm{supp}(\mathbf{v}^{n+\frac{1}{2}})$, another one is $i\in \mathbb{S}$ and $i\in \mathrm{supp}(\mathbf{v}^{n+\frac{1}{2}})$.
For $i\in \mathbb{S}$ but $i \not \in \mathrm{supp}(\mathbf{v}^{n+\frac{1}{2}})$, we have $\Tilde{v}_{i}^{{n+\frac{1}{2}}}=0$, and
\begin{equation}\label{he_eq9}
\begin{aligned}
|v&_i^{n+\frac{1}{2}}-x_i^*|\\
&\leq\Bigg|\left(1-\gamma_2^n+\gamma_1^n\gamma_2^n\right)(x_i^n-x_i^*)\Bigg|\\
&\qquad+\Bigg|\gamma_1^n\gamma_2^n\sum_{\substack{j\in \mathbb{S},\\ j\neq i}}(\mathbf{W}_i)^{T}\mathbf{A}_{j}(x_{j}^{n}-x_{j}^{*})\Bigg|\\
&\qquad+\Bigg|\gamma_2^n\sum_{\substack{j\in \mathbb{S},\\ j\neq i}}(\mathbf{W}_i)^{T}\mathbf{A}_{j}(v_{j}^{n}-x_{j}^{*})\Bigg|\\
&\qquad+\Bigg|(\mathcal{C}_{\mathbf{v}^n})_i\gamma_2^n\theta_{1}^{n}\bigg(1-\frac{|\Tilde{v}^n_i|}{\hat{\theta}^n_1-\theta^n_1}\bigg)\Bigg|+\theta_{2}^{n} \\
&\leq\Bigg|\left(1-\gamma_2^n+\gamma_1^n\gamma_2^n\right)(x_i^n-x_i^*)\Bigg|+\mu\gamma_1^n\gamma_2^n\sum_{\substack{j\in \mathbb{S},\\ j\neq i}}|x_{j}^{n}-x_{j}^{*}|\\
&\quad+\mu\gamma_2^n\sum_{\substack{j\in \mathbb{S},\\ j\neq i}}|v_{j}^{n}-x_{j}^{*}|+\bigg|1-\frac{|\Tilde{v}^n_i|}{\hat{\theta}^n_1-\theta^n_1}\bigg|(\mathcal{C}_{\mathbf{v}^n})_i\gamma_2^n\theta_1^n+\theta_2^n.
\end{aligned}
\end{equation}
For $i\in \mathbb{S}$ and $i \in \mathrm{supp}(\mathbf{v}^{n+\frac{1}{2}})$, we have $\Tilde{v}_{i}^{{n+\frac{1}{2}}}\not=0$, and
\begin{equation}
\begin{aligned}
|v&_i^{n+\frac{1}{2}}-x_i^*|\\
&\leq\Bigg|\left(1-\gamma_2^n+\gamma_1^n\gamma_2^n\right)(x_i^n-x_i^*)\Bigg|+\mu\gamma_1^n\gamma_2^n\sum_{\substack{j\in \mathbb{S},\\ j\neq i}}|x_{j}^{n}-x_{j}^{*}|\\
&\quad+\mu\gamma_2^n\sum_{\substack{j\in \mathbb{S},\\ j\neq i}}|v_{j}^{n}-x_{j}^{*}|+\bigg|1-\frac{|\Tilde{v}^n_i|}{\hat{\theta}^n_1-\theta^n_1}\bigg|(\mathcal{C}_{\mathbf{v}^n})_i\gamma_2^n\theta_1^n\\
&\quad+\left|\frac{\theta^n_2}{\hat{\theta}^n_2-\theta^n_2}|\Tilde{v}_i^{n+\frac{1}{2}}|-\theta_2^n\right|.
\end{aligned}
\end{equation}
Similar to Eq.~\eqref{he_eq10} and \eqref{he_eq11}, the discussion is based on two cases. When $0<|\Tilde{v}^{n+\frac{1}{2}}_i|\leq\hat{\theta}^n_2-\theta^n_2$,
we obtain 
\begin{equation}
\begin{aligned}
\left|\frac{\theta^n_2}{\hat{\theta}^n_2-\theta^n_2}|\Tilde{v}^{n+\frac{1}{2}}_i|-\theta_{2}^{n}\right|=& \theta_{2}^{n}-\frac{\theta^n_2}{\hat{\theta}^n_2-\theta^n_2}|\Tilde{v}^{n+\frac{1}{2}}_i| \\
=&\left(1-\frac{|\Tilde{v}^{n+\frac{1}{2}}_i|}{\hat{\theta}^n_2-\theta^n_2}\right)\theta_{2}^{n} <\theta_2^n.
\end{aligned}
\end{equation}
When $\hat{\theta}^n_2-\theta^n_2<|\Tilde{v}^{n+\frac{1}{2}}_i|<\hat{\theta}^n_2$, we have 
\begin{equation}
\begin{aligned}
\left|\frac{\theta^n_2}{\hat{\theta}^n_2-\theta^n_2}|\Tilde{v}^{n+\frac{1}{2}}_i|-\theta_{2}^{n}\right|=\left(\frac{|\Tilde{v}^{n+\frac{1}{2}}_i|}{\hat{\theta}^n_2-\theta^n_2}-1\right)\theta_{2}^{n} <\frac{(\theta_2^n)^2}{\hat{\theta}^n_2-\theta^n_2}.
\end{aligned}
\end{equation}

Then, We consider the case of $|\Tilde{v}^{n+\frac{1}{2}}_i|\geq\hat{\theta}^n_2$. In this case, $\Tilde{v}^{n+\frac{1}{2}}_i=v^{n+\frac{1}{2}}_i$, we can easily deduce that
\begin{align}
|v_{i}&^{n+\frac{1}{2}}-x_{i}^{*}| \nonumber\\
&\leq\Bigg|\left(1-\gamma_2^n+\gamma_1^n\gamma_2^n\right)(x_i^n-x_i^*)\Bigg|+\mu\gamma_1^n\gamma_2^n\sum_{\substack{j\in \mathbb{S},\\ j\neq i}}|x_{j}^{n}-x_{j}^{*}|\nonumber\\
&\quad+\mu\gamma_2^n\sum_{\substack{j\in \mathbb{S},\\ j\neq i}}|v_{j}^{n}-x_{j}^{*}|+\bigg|1-\frac{|\Tilde{v}^n_i|}{\hat{\theta}^n_1-\theta^n_1}\bigg|(\mathcal{C}_{\mathbf{v}^n})_i\gamma_2^n\theta_1^n. 
\end{align}

Hereby, we shall calculate $\left\|\mathbf{v}^{n+\frac{1}{2}} - \mathbf{x}^{*}\right\|_1$. Define the sets
\begin{align}
\mathbb{V}^{n+\frac{1}{2}}_L &= \{i|i\in\mathbb{S},  \hat{\theta}^n_2-\theta^n_2<|\Tilde{v}^{n+\frac{1}{2}}_i|<\hat{\theta}^n_2\}, \nonumber\\
\mathbb{V}^{n+\frac{1}{2}}_S &= \{i|i\in\mathbb{S}, 0\leq|\Tilde{v}^{n+\frac{1}{2}}_i|\leq\hat{\theta}^n_2-\theta^n_2\}
\end{align}
and $|\mathbb{V}^{n+\frac{1}{2}}_L|$ and $|\mathbb{V}^{n+\frac{1}{2}}_S|$ as the cardinality of $\mathbb{V}_L^{n+\frac{1}{2}}$ and $\mathbb{V}_S^{n+\frac{1}{2}}$, respectively. Note that $\hat{\theta}_{2}^{n}=(1+1/\epsilon^n_2)\theta_{2}^{n}$ according to Eq.~(54).
Accumulate all the $\left| v_i^{n+\frac{1}{2}} - x_{i}^{*}\right|$ with all $i\in\mathbb{S}$, we obtain
\begin{align}\label{he_eq13}
&\left\|\mathbf{v}^{n+\frac{1}{2}} - \mathbf{x}^{*}\right\|_1 \nonumber\\
&\quad< \sum_{i\in \mathbb{S}}\Bigg(\bigg|\left(1-\gamma_2^n+\gamma_1^n\gamma_2^n\right)(x_i^n-x_i^*)\bigg|+\mu\gamma_1^n\gamma_2^n\sum_{\substack{j\in \mathbb{S},\\ j\neq i}}|x_{j}^{n}-x_{j}^{*}|\Bigg) \nonumber \\
&\qquad+\sum_{i\in \mathbb{S}}\Bigg(\mu\gamma_2^n\sum_{\substack{j\in \mathbb{S},\\ j\neq i}}|v_{j}^{n}-x_{j}^{*}|+\bigg|1-\frac{|\Tilde{v}^n_i|}{\hat{\theta}^n_1-\theta^n_1}\bigg|(\mathcal{C}_{\mathbf{v}^n})_i\gamma_2^n\theta_1^n \Bigg)\nonumber \\
&\quad\quad+|\mathbb{V}^{n+\frac{1}{2}}_S|\theta_2^n +\epsilon_2^n|\mathbb{V}^{n+\frac{1}{2}}_L|\theta_2^n\nonumber\\
&\quad \leq\left|1-\gamma_2^n+\gamma_1^n\gamma_2^n\right|\|\mathbf{x}^n-\mathbf{x}^*\|_1+\mu\gamma_1^n\gamma_2^n(|\mathbb{S}|-1)\|\mathbf{x}^n-\mathbf{x}^*\|_1 \nonumber\\
&\quad\quad+\mu\gamma_2^n(|\mathbb{S}|-1)\|\mathbf{v}^n-\mathbf{x}^*\|_1+\sum_{i\in\mathbb{S}}\bigg|1-\frac{|\Tilde{v}^n_i|}{\hat{\theta}^n_1-\theta^n_1}\bigg|(\mathcal{C}_{\mathbf{v}^n})_i\gamma_2^n\theta_1^n \nonumber \\
&\quad\quad+(|\mathbb{V}_S^{n+\frac{1}{2}}|+\epsilon_2^n|\mathbb{V}_L^{n+\frac{1}{2}}|)\theta_2^n.
\end{align}
According to Eq.~\eqref{he_eq12}, we obtain 
\begin{align}\label{he_eq15}
&\sum_{i\in\mathbb{S}}\bigg|1-\frac{|\Tilde{v}^n_i|}{\hat{\theta}^n_1-\theta^n_1}\bigg|(\mathcal{C}_{\mathbf{v}^n})_i\gamma_2^n\theta_1^n \nonumber\\
&\quad=\sum_{i\in(\mathbb{V}^n_L\cup  \mathbb{V}^n_S)}\bigg|1-\frac{|\Tilde{v}^n_i|}{\hat{\theta}^n_1-\theta^n_1}\bigg|\gamma_2^n\theta_1^n \nonumber\\
&\quad\leq(|\mathbb{V}^n_S|+\epsilon_1^n|\mathbb{V}^n_L|)\gamma_2^n\theta_1^n,
\end{align}
where the last equation holds due to Eq.~\eqref{he_eq10} and \eqref{he_eq11}. Substituting Eq.~\eqref{he_eq14} and \eqref{he_eq15} into \eqref{he_eq13}, we obtain 
\begin{align}\label{he_eq20}
&\left\|\mathbf{v}^{n+\frac{1}{2}} - \mathbf{x}^{*}\right\|_1 \nonumber\\ 
&<\big[\left|1-\gamma_2^n+\gamma_1^n\gamma_2^n\right|+\mu\gamma_1^n\gamma_2^n(|\mathbb{S}|-1)\big]\|\mathbf{x}^n-\mathbf{x}^*\|_1 \nonumber\\ 
&\quad+\mu\gamma_2^n(|\mathbb{S}|-1)\|\mathbf{v}^{n}-\mathbf{x}^{*}\|_{1}\nonumber\\
&\quad+(|\mathbb{V}_S^{n+\frac{1}{2}}|+\epsilon_2^n|\mathbb{V}_L^{n+\frac{1}{2}}|)\mu\gamma_2^n \sup_{\mathbf{x}^{*}\in\mathcal{X}(B_{\mathbf{x}},\mathbb{S})}\{\|\mathbf{v}^{n}-\mathbf{x}^{*}\|_{1}\}\nonumber\\ &\quad+(|\mathbb{V}_S^n|+\epsilon_1^n|\mathbb{V}_L^n|)\mu\gamma_1^n\gamma_2^n\sup_{\mathbf{x}^{*}\in\mathcal{X}(B_{\mathbf{x}},\mathbb{S})}\{\|\mathbf{x}^{n}-\mathbf{x}^{*}\|_{1}\} \nonumber\\
&\leq\big[\left|1-\gamma_2^n+\gamma_1^n\gamma_2^n\right|+\mu\gamma_1^n\gamma_2^n(|\mathbb{S}|+|\mathbb{V}_S^n|+\epsilon_1^n|\mathbb{V}_L^n|-1)\big] \nonumber\\
&\qquad\qquad\qquad\qquad\qquad\qquad\cdot\sup_{\mathbf{x}^{*}\in\mathcal{X}(B_{\mathbf{x}},\mathbb{S})}\{\|\mathbf{x}^{n}-\mathbf{x}^{*}\|_{1}\} \nonumber\\
&\quad+\mu\gamma_2^n(|\mathbb{S}|+|\mathbb{V}_S^{n+\frac{1}{2}}|+\epsilon_2^n|\mathbb{V}_L^{n+\frac{1}{2}}|-1) \nonumber\\ &\qquad\qquad\qquad\qquad\qquad\qquad\cdot\sup_{\mathbf{x}^{*}\in\mathcal{X}(B_{\mathbf{x}},\mathbb{S})}\{\|\mathbf{v}^{n}-\mathbf{x}^{*}\|_{1}\} 
% \nonumber \\
% &\leq\Big\{\left|1-\gamma_2^n+\gamma_1^n\gamma_2^n\right|+\mu\gamma_1^n\gamma_2^n(|\mathbb{S}|+|\mathbb{V}_S^n|+\epsilon_1^n|\mathbb{V}_L^n|-1)    \nonumber \\
% &\quad+\mu\gamma_2^n(|\mathbb{S}|+|\mathbb{V}_S^{n+\frac{1}{2}}|+\epsilon_2^n|\mathbb{V}_L^{n+\frac{1}{2}}|-1) \nonumber \\
% &\qquad\qquad\cdot\big[\mu\gamma_1^n(|\mathbb{S}|+|\mathbb{V}_S^n|+\epsilon_1^n|\mathbb{V}_L^n|-1)+\left|1-\gamma_1^n\right|\big]\Big\}\nonumber \\
% &\qquad\qquad\qquad\qquad\qquad\qquad\quad\cdot\sup_{\mathbf{x}^{*}\in\mathcal{X}(B_{\mathbf{x}},\mathbb{S})}\{\|\mathbf{x}^{n}-\mathbf{x}^{*}\|_{1}\}
\end{align}

$\mathbf{3)~The~upper~bound~for~\|\mathbf{w}^{n}-\mathbf{x}^{*}\|_1 .}$

Similarly, we define $\mathbf{K}_{\mathbf{w}^{n}}$,  $\Tilde{\mathbf{w}}^{n}$ and $\Tilde{w}_i^{n}$.  
When $0\leq|\Tilde{w}^n_i|<\hat{\theta}^n_3$, we have for arbitrary $i\in\mathbb{S}$,
\begin{align}
\Tilde{w}_{i}^{n}=&\mathcal{S}_{\theta_{3}^{n}}\left(u_{i}^{n} -\gamma_3^n \sum_{j\not\in \mathbb{S}}(\mathbf{W}_i)^{T}\mathbf{A}_{j}u_{j}^{n} \nonumber\right.\\
&\left.\qquad\qquad\qquad- \gamma_3^n \sum_{j\in \mathbb{S}}(\mathbf{W}_i)^{T}\mathbf{A}_{j}(u_{j}^{n}-x_{j}^{*})\right)\nonumber\\
=&\mathcal{S}_{\theta_{3}^{n}}\left( u_{i}^{n}-\gamma_3^n\sum_{j\not\in \mathbb{S}}(\mathbf{W}_i)^{T}\mathbf{A}_{j}u_{j}^{n}-\gamma_3^n(u_{i}^{n}-x_{i}^{*}) \right. \nonumber \\
&\left.\qquad\qquad\qquad - \gamma_3^n\sum_{j\in \mathbb{S}, j\neq i}(\mathbf{W}_i)^{T}\mathbf{A}_{j}(u_{j}^{n}-x_{j}^{*})\right)\nonumber\\
\in & x_{i}^{*}-\gamma_3^n\sum_{j\not\in \mathbb{S}}(\mathbf{W}_i)^{T}\mathbf{A}_{j}u_{j}^{n}+(1-\gamma_3^n)(u_{i}^{n}-x_{i}^{*}) \nonumber \\
&-\gamma_3^n \sum_{\substack{j\in \mathbb{S},\\ j\neq i}}(\mathbf{W}_i)^{T}\mathbf{A}_{j}(u_{j}^{n}-x_{j}^{*}) - \theta_{3}^{n}\partial \mathit{l}_{1}(\Tilde{w}_{i}^{n}). 
\end{align}
Thus, we obtain that
\begin{equation}
\begin{aligned}
&w_i^n-x_i^*\\
=&(\mathbf{K}_{\mathbf{w}^n})_i\Tilde{w}^n_i -x_i^* \\
=&-\gamma_3^n\sum_{j\in \mathbb{S}, j\neq i}(\mathbf{W}_i)^{T}\mathbf{A}_{j}(u_{j}^{n}-x_{j}^{*})+(1-\gamma_3^n)(u_{i}^{n}-x_{i}^{*})\\
&-\theta_{3}^{n}\partial \mathit{l}_{1}(\Tilde{w}_{i}^{n})+\frac{\theta^n_3}{\hat{\theta}^n_3-\theta^n_3}\Tilde{w}^n_i-\gamma_3^n\sum_{j\not\in \mathbb{S}}(\mathbf{W}_i)^{T}\mathbf{A}_{j}u_{j}^{n}.
\end{aligned}
\end{equation}
We begin to calculate $|w_i^n-x_i^*|$. For $i\in\mathbb{S}$ and $i\not\in\mathrm{supp}(\mathbf{w}^n)$, we have $\Tilde{w}_i^n=0$, and 
\begin{equation}
|w_i^n-x_i^*|
\leq\mu\gamma_3^n\Bigg(\sum_{j\in \mathbb{S}, j\neq i}|u_{j}^{n}-x_{j}^{*}|+\sum_{j\not\in \mathbb{S}}|u_{j}^{n}|\Bigg)
+|1-\gamma_3^n||u_{i}^{n}-x_{i}^{*}|
+\theta_{3}^{n}.    
\end{equation}
For $i\in\mathbb{S}$ and $i\in\mathrm{supp}(\mathbf{w}^n)$, we have $\Tilde{w}_i^n\not=0$, and 
\begin{multline}
|w_i^n-x_i^*|
\leq\mu\gamma_3^n\Bigg(\sum_{j\in \mathbb{S}, j\neq i}|u_{j}^{n}-x_{j}^{*}|+\sum_{j\not\in \mathbb{S}}|u_{j}^{n}|\Bigg)
\\+|1-\gamma_3^n||u_{i}^{n}-x_{i}^{*}|
+\left|\frac{\theta^n_3}{\hat{\theta}^n_3-\theta^n_3}|\Tilde{w}_i^{n}|-\theta_3^n\right|.    
\end{multline}
Similarly, we obtain
\begin{equation}\label{he_eq18}
\left|\frac{\theta^n_3}{\hat{\theta}^n_3-\theta^n_3}|\Tilde{w}_i^{n}|-\theta_3^n\right|<
\begin{cases}
\theta_3^n, & ~~{\rm if}~ 0<|\Tilde{w}^{n}_i|\leq\hat{\theta}^n_3-\theta^n_3,\\
\frac{(\theta^n_3)^2}{\hat{\theta}^n_3-\theta^n_3}, & ~~{\rm if} ~\hat{\theta}^n_3-\theta^n_3<|\Tilde{w}^{n}_i|<\hat{\theta}^n_3,
\end{cases}
\end{equation}
and 
\begin{equation}
|w_i^n-x_i^*|
\leq\mu\gamma_3^n\Bigg(\sum_{j\in \mathbb{S}, j\neq i}|u_{j}^{n}-x_{j}^{*}|+\sum_{j\not\in \mathbb{S}}|u_{j}^{n}|\Bigg)
+|1-\gamma_3^n||u_{i}^{n}-x_{i}^{*}|  
\end{equation}
when $|\Tilde{w}^{n}_i|\geq\hat{\theta}^n_3$.

Hereby, we shall calculate $\left\|\mathbf{w}^n - \mathbf{x}^{*}\right\|_1$. Define the sets
\begin{equation}
\begin{aligned}
\mathbb{W}^n_L &= \{i|i\in\mathbb{S},  \hat{\theta}^n_3-\theta^n_3<|\Tilde{w}^n_i|<\hat{\theta}^n_3\}, \\
\mathbb{W}^n_S &= \{i|i\in\mathbb{S},  0\leq|\Tilde{w}^n_i|\leq\hat{\theta}^n_3-\theta^n_3\},
\end{aligned}
\end{equation}
and $|\mathbb{W}_L^n|$ and $|\mathbb{W}_S^n|$ as the cardinality of $\mathbb{W}_L^n$ and $\mathbb{W}_S^n$, respectively. Note that $\hat{\theta}_{3}^{n}=(1+1/\epsilon^n_3)\theta_{3}^{n}$ according to Eq.~(54).
Accumulate all the $\left| w_i^n - x_{i}^{*}\right|$ with all $i\in\mathbb{S}$, we obtain
\begin{align}
&\left\|\mathbf{w}^n - \mathbf{x}^{*}\right\|_1 \nonumber\\
&\quad\leq \mu\gamma_3^n\sum_{i\in \mathbb{S}}\Bigg(\sum_{\substack{j\in \mathbb{S},\\ j\neq i}}\left|u_{j}^{n}-x_{j}^{*}\right|+\sum_{j\not\in \mathbb{S}}|u_j^n|\Bigg)\nonumber \\
&\qquad+\sum_{i\in \mathbb{S}}\left|1-\gamma_3^n\right|\left|u_{i}^{n}-x_{i}^{*}\right| +|\mathbb{W}^n_S|\theta_3^n +\epsilon_3^n|\mathbb{W}^n_L|\theta_3^n\nonumber\\
&\quad\leq\left[\mu\gamma_3^n(|\mathbb{S}|-1)+\left|1-\gamma_3^n\right|\right]\|\mathbf{u}^{n}-\mathbf{x}^{*}\|_{1}\nonumber \\
&\quad\quad+(|\mathbb{W}^n_S|+\epsilon_3^n|\mathbb{W}^n_L|)\theta_3^n+\mu\gamma_3^n(N-|\mathbb{S}|)\|\mathbf{u}^n\|_1 \nonumber\\
&\quad\leq\left[\mu\gamma_3^n(|\mathbb{S}|+|\mathbb{W}^n_S|+\epsilon_3^n|\mathbb{W}^n_L|-1)+\left|1-\gamma_3^n\right|\right]\nonumber\\
&\qquad\qquad\qquad\qquad\cdot\sup_{\mathbf{x}^{*}\in\mathcal{X}(B_{\mathbf{x}},\mathbb{S})}\{\|\mathbf{u}^{n}-\mathbf{x}^{*}\|_{1}\}\nonumber\\
&\qquad+\mu\gamma_3^n(N-|\mathbb{S}|)\|\mathbf{u}^n\|_1.
\end{align}

$\mathbf{4)~The~upper~bound~for~\|\mathbf{w}^{n+\frac{1}{2}}-\mathbf{x}^{*}\|_1 .}$

Similarly, we define $\mathbf{K}_{\mathbf{w}^{n+\frac{1}{2}}}$,  $\Tilde{\mathbf{w}}^{n+\frac{1}{2}}$ and $\Tilde{w}_i^{n+\frac{1}{2}}$. The following analysis is similar to the content in $\mathbf{2)}$ when we calculate the upper bound for $\|\mathbf{v}^{n+\frac{1}{2}}-\mathbf{x}^*\|_1$, so we omit some details.

We can easily deduce that when $0\leq|\Tilde{w}^{n+\frac{1}{2}}|<\hat{\theta}_4^n$,
\begin{align}
&\Tilde{w}^{n+\frac{1}{2}}_i \nonumber\\ &=\mathcal{S}_{\theta_{4}^{n}}\left(u_{i}^{n}-\gamma_4^n \sum_{j\in \mathbb{S}}(\mathbf{W}_i)^{T}\mathbf{A}_{j}(w_{j}^{n}-x_{j}^{*})\right)\nonumber\\
&\in x_{i}^{*} + (u_{i}^{n}-x_{i}^{*})
-\gamma_4^n\sum_{j\in \mathbb{S}, j\neq i}(\mathbf{W}_i)^{T}\mathbf{A}_{j}(w_{j}^{n}-x_{j}^{*})\nonumber\\
&\quad-\gamma_4^n(w_{i}^{n}-x_{i}^{*})-\theta_{4}^{n}\partial \mathit{l}_{1}(\Tilde{w}_{i}^{n+\frac{1}{2}})-\gamma_4^n\sum_{j\not\in \mathbb{S}}(\mathbf{W}_i)^{T}\mathbf{A}_{j}w_{j}^{n}.
\end{align}
and 
\begin{equation}
\begin{aligned}
&w_i^{n+\frac{1}{2}}-x_i^*\\
=&(\mathbf{K}_{\mathbf{w}^{n+\frac{1}{2}}})_i\Tilde{w}^{n+\frac{1}{2}}_i -x_i^* \\
=&\left(1-\gamma_4^n+\gamma_3^n\gamma_4^n\right)(u_i^n-x_i^*)-\gamma_4^n\sum_{\substack{j\in \mathbb{S},\\ j\neq i}}(\mathbf{W}_i)^{T}\mathbf{A}_{j}(w_{j}^{n}-x_{j}^{*})\\
&+\gamma_3^n\gamma_4^n\sum_{\substack{j\in \mathbb{S},\\ j\neq i}}(\mathbf{W}_i)^{T}\mathbf{A}_{j}(u_{j}^{n}-x_{j}^{*})-\theta_{4}^{n}\partial \mathit{l}_{1}(\Tilde{w}_{i}^{n+\frac{1}{2}})\\
&+(\mathcal{C}_{\mathbf{w}^n})_i\gamma_4^n\theta_{3}^{n}\partial \mathit{l}_{1}(\Tilde{w}_{i}^{n})+\frac{\theta^n_4}{\hat{\theta}^n_4-\theta^n_4}\Tilde{w}_i^{n+\frac{1}{2}}-\frac{(\mathcal{C}_{\mathbf{w}^n})_i\gamma_4^n\theta^n_3}{\hat{\theta}^n_3-\theta^n_3}\Tilde{w}^n_i\\
&-\gamma_4^n\sum_{j\not\in \mathbb{S}}(\mathbf{W}_i)^{T}\mathbf{A}_{j}w_{j}^{n}+\gamma_3^n\gamma_4^n\sum_{j\not\in \mathbb{S}}(\mathbf{W}_i)^{T}\mathbf{A}_{j}u_{j}^{n}.
\end{aligned}
\end{equation}
where $\mathcal{C}_{\mathbf{w}^n}$ is a vector and the $i$th element is defined as
\begin{equation}
(\mathcal{C}_{\mathbf{w}^n})_i=
\begin{cases}
1, & ~~~{\rm if}~ ~i\in(\mathbb{W}^n_L\cup  \mathbb{W}^n_S),\\
0, & ~~~{\rm else}.
\end{cases}
\end{equation}
The above calculation is similar to Eq.~\eqref{he_eq17}. Next, we shall calculate $|w_i^{n+\frac{1}{2}}-x_i^*|$. For $i\in \mathbb{S}$ but $i \not \in \mathrm{supp}(\mathbf{w}^{n+\frac{1}{2}})$, we have $\Tilde{w}_{i}^{{n+\frac{1}{2}}}=0$, and
\begin{equation}
\begin{aligned}
|w&_i^{n+\frac{1}{2}}-x_i^*|\\
&\leq\Bigg|\left(1-\gamma_4^n+\gamma_3^n\gamma_4^n\right)(u_i^n-x_i^*)\Bigg|+\mu\gamma_3^n\gamma_4^n\sum_{\substack{j\in \mathbb{S},\\ j\neq i}}|u_{j}^{n}-x_{j}^{*}|\\
&\quad+\mu\gamma_4^n\sum_{\substack{j\in \mathbb{S},\\ j\neq i}}|w_{j}^{n}-x_{j}^{*}|+\bigg|1-\frac{|\Tilde{w}^n_i|}{\hat{\theta}^n_3-\theta^n_3}\bigg|(\mathcal{C}_{\mathbf{w}^n})_i\gamma_4^n\theta_3^n+\theta_4^n\\
&\quad+\mu\gamma_4^n\sum_{j\not\in \mathbb{S}}w_{j}^{n}+\mu\gamma_3^n\gamma_4^n\sum_{j\not\in \mathbb{S}}u_{j}^{n}.
\end{aligned}
\end{equation}
For $i\in \mathbb{S}$ and $i \in \mathrm{supp}(\mathbf{w}^{n+\frac{1}{2}})$, we have $\Tilde{w}_{i}^{{n+\frac{1}{2}}}\not=0$, and
\begin{equation}
\begin{aligned}
|w&_i^{n+\frac{1}{2}}-x_i^*|\\
&\leq\Bigg|\left(1-\gamma_4^n+\gamma_3^n\gamma_4^n\right)(u_i^n-x_i^*)\Bigg|+\mu\gamma_3^n\gamma_4^n\sum_{\substack{j\in \mathbb{S},\\ j\neq i}}|u_{j}^{n}-x_{j}^{*}|\\
&\quad+\mu\gamma_4^n\sum_{\substack{j\in \mathbb{S},\\ j\neq i}}|w_{j}^{n}-x_{j}^{*}|+\bigg|1-\frac{|\Tilde{w}^n_i|}{\hat{\theta}^n_3-\theta^n_3}\bigg|(\mathcal{C}_{\mathbf{w}^n})_i\gamma_4^n\theta_3^n\\
&\quad+\left|\frac{\theta^n_4}{\hat{\theta}^n_4-\theta^n_4}|\Tilde{w}_i^{n+\frac{1}{2}}|-\theta_4^n\right|+\mu\gamma_4^n\sum_{j\not\in \mathbb{S}}w_{j}^{n}+\mu\gamma_3^n\gamma_4^n\sum_{j\not\in \mathbb{S}}u_{j}^{n}.
\end{aligned}
\end{equation}
We can easily obtain that
\begin{equation}
\left|\frac{\theta^n_4|\Tilde{w}_i^{n+\frac{1}{2}}|}{\hat{\theta}^n_4-\theta^n_4}-\theta_4^n\right|<
\begin{cases}
\theta_4^n, & ~~{\rm if}~ 0<|\Tilde{w}^{n+\frac{1}{2}}_i|\leq\hat{\theta}^n_4-\theta^n_4,\\
\frac{(\theta^n_4)^2}{\hat{\theta}^n_4-\theta^n_4}, & ~~{\rm if} ~\hat{\theta}^n_4-\theta^n_4<|\Tilde{w}^{n+\frac{1}{2}}_i|<\hat{\theta}^n_4,
\end{cases}
\end{equation}
and 
\begin{equation}
\begin{aligned}
|w&_i^{n+\frac{1}{2}}-x_i^*|\\
&\leq\Bigg|\left(1-\gamma_4^n+\gamma_3^n\gamma_4^n\right)(u_i^n-x_i^*)\Bigg|+\mu\gamma_3^n\gamma_4^n\sum_{\substack{j\in \mathbb{S},\\ j\neq i}}|u_{j}^{n}-x_{j}^{*}|\\
&\quad+\mu\gamma_4^n\sum_{\substack{j\in \mathbb{S},\\ j\neq i}}|w_{j}^{n}-x_{j}^{*}|+\bigg|1-\frac{|\Tilde{w}^n_i|}{\hat{\theta}^n_3-\theta^n_3}\bigg|(\mathcal{C}_{\mathbf{w}^n})_i\gamma_4^n\theta_3^n\\
&\quad+\mu\gamma_4^n\sum_{j\not\in \mathbb{S}}w_{j}^{n}+\mu\gamma_3^n\gamma_4^n\sum_{j\not\in \mathbb{S}}u_{j}^{n}
\end{aligned}
\end{equation}
when $|\Tilde{w}^{n+\frac{1}{2}}|\geq\hat{\theta}_4^n$.

Hereby, we shall calculate $\left\|\mathbf{w}^{n+\frac{1}{2}} - \mathbf{x}^{*}\right\|_1$. Define the sets
\begin{align}
\mathbb{W}^{n+\frac{1}{2}}_L &= \{i|i\in\mathbb{S},  \hat{\theta}^n_4-\theta^n_4<|\Tilde{w}^{n+\frac{1}{2}}_i|<\hat{\theta}^n_4\}, \nonumber\\
\mathbb{W}^{n+\frac{1}{2}}_S &= \{i|i\in\mathbb{S}, 0\leq|\Tilde{w}^{n+\frac{1}{2}}_i|\leq\hat{\theta}^n_4-\theta^n_4\}
\end{align}
and $|\mathbb{W}^{n+\frac{1}{2}}_L|$ and $|\mathbb{W}^{n+\frac{1}{2}}_S|$ as the cardinality of $\mathbb{W}_L^{n+\frac{1}{2}}$ and $\mathbb{W}_S^{n+\frac{1}{2}}$, respectively. Note that $\hat{\theta}_{4}^{n}=(1+1/\epsilon^n_4)\theta_{4}^{n}$ according to Eq.~(54) and 
\begin{align}
&\sum_{i\in\mathbb{S}}\bigg|1-\frac{|\Tilde{w}^n_i|}{\hat{\theta}^n_3-\theta^n_3}\bigg|(\mathcal{C}_{\mathbf{w}^n})_i\gamma_4^n\theta_3^n \nonumber\\
&\quad=\sum_{i\in(\mathbb{W}^n_L\cup  \mathbb{W}^n_S)}\bigg|1-\frac{|\Tilde{w}^n_i|}{\hat{\theta}^n_3-\theta^n_3}\bigg|\gamma_4^n\theta_3^n \nonumber\\
&\quad\leq(|\mathbb{W}^n_S|+\epsilon_3^n|\mathbb{W}^n_L|)\gamma_4^n\theta_3^n,
\end{align}
according to Eq.~\eqref{he_eq18}.
Accumulate all the $\left| w_i^{n+\frac{1}{2}} - x_{i}^{*}\right|$ with all $i\in\mathbb{S}$, we obtain
\begin{align}\label{he_eq19}
&\left\|\mathbf{w}^{n+\frac{1}{2}} - \mathbf{x}^{*}\right\|_1 \nonumber\\
&\quad \leq\left|1-\gamma_4^n+\gamma_3^n\gamma_4^n\right|\|\mathbf{u}^n-\mathbf{x}^*\|_1+\mu\gamma_3^n\gamma_4^n(|\mathbb{S}|-1)\|\mathbf{u}^n-\mathbf{x}^*\|_1 \nonumber\\
&\quad\quad+\mu\gamma_4^n(|\mathbb{S}|-1)\|\mathbf{w}^n-\mathbf{x}^*\|_1+(|\mathbb{W}^n_S|+\epsilon_3^n|\mathbb{W}^n_L|)\gamma_4^n\theta_3^n \nonumber \\
&\quad\quad+(|\mathbb{W}_S^{n+\frac{1}{2}}|+\epsilon_4^n|\mathbb{W}_L^{n+\frac{1}{2}}|)\theta_4^n+\mu\gamma_4^n(N-|\mathbb{S}|)\|\mathbf{w}^{n}\|_1\nonumber \\
&\quad\quad+\mu\gamma_3^n\gamma_4^n(N-|\mathbb{S}|)\|\mathbf{u}^{n}\|_1\nonumber \\
&\quad \leq\big[\left|1-\gamma_4^n+\gamma_3^n\gamma_4^n\right|+\mu\gamma_3^n\gamma_4^n(|\mathbb{S}|+|\mathbb{W}^n_S|+\epsilon_3^n|\mathbb{W}^n_L|-1)\big]\nonumber \\
&\qquad\qquad\qquad\qquad\qquad\cdot\sup_{\mathbf{x}^{*}\in\mathcal{X}(B_{\mathbf{x}},\mathbb{S})}\{\|\mathbf{u}^n-\mathbf{x}^*\|_1 \} \nonumber\\
&\quad\quad+\mu\gamma_4^n(|\mathbb{S}|+|\mathbb{W}^{n+\frac{1}{2}}_S|+\epsilon_4^n|\mathbb{W}^{n+\frac{1}{2}}_L|-1)\nonumber \\
&\qquad\qquad\qquad\qquad\qquad\cdot\sup_{\mathbf{x}^{*}\in\mathcal{X}(B_{\mathbf{x}},\mathbb{S})}\{\|\mathbf{w}^n-\mathbf{x}^*\|_1 \} \nonumber\\
&\quad\quad+\mu\gamma_4^n(N-|\mathbb{S}|)\frac{|\mathbb{S}|+|\mathbb{W}^{n+\frac{1}{2}}_S|+\epsilon_4^n|\mathbb{W}^{n+\frac{1}{2}}_L|-1}{|\mathbb{S}|-1}\|\mathbf{w}^n\|_1 \nonumber\\
&\quad\quad+\mu\gamma_3^n\gamma_4^n(N-|\mathbb{S}|)\frac{|\mathbb{S}|+|\mathbb{W}^{n+\frac{1}{2}}_S|+\epsilon_4^n|\mathbb{W}^{n+\frac{1}{2}}_L|-1}{|\mathbb{S}|-1}\|\mathbf{u}^n\|_1.
\end{align}

$\mathbf{5)~The~upper~bound~for~\|\mathbf{x}^{n+1}-\mathbf{x}^{*}\|_2 .}$

Define 
\begin{equation}
\begin{aligned}
\mathcal{Q}^n_\mathbb{V} &= |\mathbb{S}|+|\mathbb{V}_S^n|+\epsilon_1^n|\mathbb{V}_L^n|-1, \\
\mathcal{Q}^{n+\frac{1}{2}}_\mathbb{V} &= |\mathbb{S}|+|\mathbb{V}_S^{n+\frac{1}{2}}|+\epsilon_2^n|\mathbb{V}_L^{n+\frac{1}{2}}|-1, \\
\mathcal{Q}^n_\mathbb{W} &= |\mathbb{S}|+|\mathbb{W}_S^n|+\epsilon_3^n|\mathbb{W}_L^n|-1,\\ 
\mathcal{Q}^{n+\frac{1}{2}}_\mathbb{W} &= |\mathbb{S}|+|\mathbb{W}_S^{n+\frac{1}{2}}|+\epsilon_4^n|\mathbb{W}_L^{n+\frac{1}{2}}|-1.
\end{aligned}
\end{equation}
and 
\begin{equation}
 \mathcal{Q}^n_* = \max\{\mathcal{Q}^n_\mathbb{V}, \mathcal{Q}^{n+\frac{1}{2}}_\mathbb{V}, \mathcal{Q}^n_\mathbb{W}, \mathcal{Q}^{n+\frac{1}{2}}_\mathbb{W}\}  .
\end{equation}
Combining Eq.~\eqref{he_eq20} and \eqref{he_eq19}, we obtain that 
\begin{align}\label{he_eq24}
&\sup_{\mathbf{x}^{*}\in\mathcal{X}(B_{\mathbf{x}},\mathbb{S})}\|\mathbf{x}^{n+1}-\mathbf{x}^*\|_1  \nonumber\\
&\quad \leq \sup_{\mathbf{x}^*}\left(\alpha^n\|\mathbf{v}^{n+\frac{1}{2}}-\mathbf{x}^*\|_1+(1-\alpha^n)\|\mathbf{w}^{n+\frac{1}{2}}-\mathbf{x}^*\|_1\right) \nonumber\\
&\quad\leq\alpha^n\big(\left|1-\gamma_2^n+\gamma_1^n\gamma_2^n\right|+\mu\gamma_1^n\gamma_2^n\mathcal{Q}^n_\mathbb{V}\big) \nonumber\\
&\qquad\qquad\qquad\qquad\qquad\qquad\cdot\sup_{\mathbf{x}^{*}\in\mathcal{X}(B_{\mathbf{x}},\mathbb{S})}\{\|\mathbf{x}^{n}-\mathbf{x}^{*}\|_{1}\} \nonumber\\
&\quad\quad+\alpha^n\mu\gamma_2^n\mathcal{Q}^{n+\frac{1}{2}}_\mathbb{V}\sup_{\mathbf{x}^{*}\in\mathcal{X}(B_{\mathbf{x}},\mathbb{S})}\{\|\mathbf{v}^{n}-\mathbf{x}^{*}\|_{1}\} \nonumber \\
&\qquad +(1-\alpha^n)\big(\left|1-\gamma_4^n+\gamma_3^n\gamma_4^n\right|+\mu\gamma_3^n\gamma_4^n\mathcal{Q}^n_\mathbb{W}\big)\nonumber \\
&\qquad\qquad\qquad\qquad\qquad\cdot\sup_{\mathbf{x}^{*}\in\mathcal{X}(B_{\mathbf{x}},\mathbb{S})}\{\|\mathbf{u}^n-\mathbf{x}^*\|_1 \} \nonumber\\
&\quad\quad+(1-\alpha^n)\mu\gamma_4^n\mathcal{Q}^{n+\frac{1}{2}}_\mathbb{W}\sup_{\mathbf{x}^{*}\in\mathcal{X}(B_{\mathbf{x}},\mathbb{S})}\{\|\mathbf{w}^n-\mathbf{x}^*\|_1 \} \nonumber\\
&\quad\quad+(1-\alpha^n)\mu\gamma_4^n(N-|\mathbb{S}|)\frac{\mathcal{Q}^{n+\frac{1}{2}}_\mathbb{W}}{|\mathbb{S}|-1}\|\mathbf{w}^n\|_1 \nonumber\\
&\quad\quad+(1-\alpha^n)\mu\gamma_3^n\gamma_4^n(N-|\mathbb{S}|)\frac{\mathcal{Q}^{n+\frac{1}{2}}_\mathbb{W}}{|\mathbb{S}|-1}\|\mathbf{u}^n\|_1
\end{align}

Recalling $\alpha^n$ specified in Eq.~(55), we have 
\begin{align}\label{he_eq22}
(1-\alpha^n)\Bigg(&\mu\gamma_3^n\gamma_4^n\mathcal{Q}^n_\mathbb{W}\sup_{\mathbf{x}^{*}\in\mathcal{X}(B_{\mathbf{x}},\mathbb{S})}\{\|\mathbf{u}^n-\mathbf{x}^*\|_1 \} \nonumber\\
&+\mu\gamma_4^n\mathcal{Q}^{n+\frac{1}{2}}_\mathbb{W}\sup_{\mathbf{x}^{*}\in\mathcal{X}(B_{\mathbf{x}},\mathbb{S})}\{\|\mathbf{w}^n-\mathbf{x}^*\|_1 \} \nonumber\\
&+\mu\gamma_4^n(N-|\mathbb{S}|)\frac{\mathcal{Q}^{n+\frac{1}{2}}_\mathbb{W}}{|\mathbb{S}|-1}\|\mathbf{w}^n\|_1 \nonumber\\
&+\mu\gamma_3^n\gamma_4^n(N-|\mathbb{S}|)\frac{\mathcal{Q}^{n+\frac{1}{2}}_\mathbb{W}}{|\mathbb{S}|-1}\|\mathbf{u}^n\|_1  \Bigg)  \nonumber\\
\leq&\frac{(\gamma_2^n\theta_1^n+\theta_2^n)(\mathcal{Q}^{n}_\mathbb{W}\gamma_4^n\theta_3^n+\mathcal{Q}^{n+\frac{1}{2}}_\mathbb{W}\theta_4^n)}{\gamma_2^n\theta_1^n+\theta_2^n+\gamma_4^n\theta_3^n+\theta_4^n}  \nonumber\\
\leq&\mathcal{Q}^{n}_*(\gamma_2^n\theta_1^n+\theta_2^n)\nonumber\\
\leq&\mathcal{Q}^{n}_*\Bigg(\mu\gamma_1^n\gamma_2^n\sup_{\mathbf{x}^{*}\in\mathcal{X}(B_{\mathbf{x}},\mathbb{S})}\{\|\mathbf{x}^n-\mathbf{x}^*\|_1 \} \nonumber\\
&\qquad+\mu\gamma_2^n\sup_{\mathbf{x}^{*}\in\mathcal{X}(B_{\mathbf{x}},\mathbb{S})}\{\|\mathbf{v}^n-\mathbf{x}^*\|_1 \} \Bigg)
\end{align}
and 
\begin{align}\label{he_eq23}
(1-\alpha^n)&\left|1-\gamma_4^n+\gamma_3^n\gamma_4^n\right|\sup_{\mathbf{x}^{*}\in\mathcal{X}(B_{\mathbf{x}},\mathbb{S})}\{\|\mathbf{u}^n-\mathbf{x}^*\|_1 \} \nonumber\\
&\leq \frac{(\gamma_2^n\theta_1^n+\theta_2^n)\left|1-\gamma_4^n+\gamma_3^n\gamma_4^n\right|}{\gamma_2^n\theta_1^n+\theta_2^n+\gamma_4^n\theta_3^n+\theta_4^n} \frac{\theta_3^n}{\mu\gamma_3^n} \nonumber\\
&\leq\frac{\gamma_1^n\gamma_2^n\left|1-\gamma_4^n+\gamma_3^n\gamma_4^n\right|}{\gamma_3^n}\sup_{\mathbf{x}^{*}\in\mathcal{X}(B_{\mathbf{x}},\mathbb{S})}\{\|\mathbf{x}^n-\mathbf{x}^*\|_1 \} \nonumber\\
&~\quad+\frac{\gamma_2^n\left|1-\gamma_4^n+\gamma_3^n\gamma_4^n\right|}{\gamma_3^n}\sup_{\mathbf{x}^{*}\in\mathcal{X}(B_{\mathbf{x}},\mathbb{S})}\{\|\mathbf{v}^n-\mathbf{x}^*\|_1 \}.
\end{align}
Thus, substituting Eq.~\eqref{he_eq22} and \eqref{he_eq23} into \eqref{he_eq24}, we have
\begin{align}
&\sup_{\mathbf{x}^{*}\in\mathcal{X}(B_{\mathbf{x}},\mathbb{S})}\|\mathbf{x}^{n+1}-\mathbf{x}^*\|_1  \nonumber\\    
&\quad\leq(\alpha^n+1)\mathcal{Q}^n_*\Big(\mu\gamma_1^n\gamma_2^n\sup_{\mathbf{x}^{*}\in\mathcal{X}(B_{\mathbf{x}},\mathbb{S})}\{\|\mathbf{x}^{n}-\mathbf{x}^{*}\|_{1}\} \nonumber\\
&\qquad\qquad\qquad\qquad+\mu\gamma_2^n\sup_{\mathbf{x}^{*}\in\mathcal{X}(B_{\mathbf{x}},\mathbb{S})}\{\|\mathbf{v}^{n}-\mathbf{x}^{*}\|_{1}\}\Big) \nonumber \\
&\qquad+\Big(\alpha^n|1-\gamma_2^n+\gamma_1^n\gamma_2^n|+\frac{\gamma_1^n\gamma_2^n\left|1-\gamma_4^n+\gamma_3^n\gamma_4^n\right|}{\gamma_3^n}\Big)\nonumber\\
&\qquad\qquad\qquad\qquad\qquad\qquad\cdot\sup_{\mathbf{x}^{*}\in\mathcal{X}(B_{\mathbf{x}},\mathbb{S})}\{\|\mathbf{x}^n-\mathbf{x}^*\|_1 \} \nonumber\\
&\qquad+\frac{\gamma_2^n\left|1-\gamma_4^n+\gamma_3^n\gamma_4^n\right|}{\gamma_3^n}\sup_{\mathbf{x}^{*}\in\mathcal{X}(B_{\mathbf{x}},\mathbb{S})}\{\|\mathbf{v}^n-\mathbf{x}^*\|_1 \}.
\end{align}
Substituting Eq.~\eqref{he_eq14} into the above equation, we obtain
\begin{align}
&\sup_{\mathbf{x}^{*}\in\mathcal{X}(B_{\mathbf{x}},\mathbb{S})}\|\mathbf{x}^{n+1}-\mathbf{x}^*\|_1  \nonumber\\  
&\quad \leq \Bigg[2\mathcal{Q}^n_*\mu\gamma_1^n\gamma_2^n\Big(1+\mathcal{Q}^n_*\mu+\frac{|1-\gamma_1^n|}{\gamma_1^n}\Big)+|1-\gamma_2^n+\gamma_1^n\gamma_2^n|\nonumber\\  
&\qquad\quad +\frac{\gamma_1^n\gamma_2^n\left|1-\gamma_4^n+\gamma_3^n\gamma_4^n\right|}{\gamma_3^n}\Big(1+\mathcal{Q}^n_*\mu+\frac{|1-\gamma_1^n|}{\gamma_1^n}\Big)\Bigg] \nonumber\\  
&\qquad\qquad\qquad\qquad\qquad\qquad\cdot\sup_{\mathbf{x}^{*}\in\mathcal{X}(B_{\mathbf{x}},\mathbb{S})}\{\|\mathbf{x}^{n}-\mathbf{x}^{*}\|_{1}\}\nonumber\\  
&\quad=\Bigg[\gamma_1^n\gamma_2^n\Big(1+\mathcal{Q}^n_*\mu+\frac{|1-\gamma_1^n|}{\gamma_1^n}\Big)\Big(2\mathcal{Q}^n_*\mu+\frac{\left|1-\gamma_4^n+\gamma_3^n\gamma_4^n\right|}{\gamma_3^n}\Big)\nonumber\\  
&\qquad\quad +|1-\gamma_2^n+\gamma_1^n\gamma_2^n|\Bigg]\sup_{\mathbf{x}^{*}\in\mathcal{X}(B_{\mathbf{x}},\mathbb{S})}\{\|\mathbf{x}^{n}-\mathbf{x}^{*}\|_{1}\}.
\end{align}
Let us define 
\begin{equation}
\begin{aligned}
c_e^n=-\log\Bigg[&\gamma_1^n\gamma_2^n\Big(1+\mathcal{Q}^n_*\mu+\frac{|1-\gamma_1^n|}{\gamma_1^n}\Big)\\
&\cdot\Big(2\mathcal{Q}^n_*\mu+\frac{\left|1-\gamma_4^n+\gamma_3^n\gamma_4^n\right|}{\gamma_3^n}\Big)\\
&+|1-\gamma_2^n+\gamma_1^n\gamma_2^n|\Bigg]
\end{aligned}
\end{equation}
Then we have
\begin{align}
&\sup_{\mathbf{x}^{*}\in\mathcal{X}(B_{\mathbf{x}},\mathbb{S})}\|\mathbf{x}^{n+1}-\mathbf{x}^*\|_1  \nonumber\\  
&\qquad\leq\exp(-c_e^n)\sup_{\mathbf{x}^{*}\in\mathcal{X}(B_{\mathbf{x}},\mathbb{S})}\|\mathbf{x}^{n}-\mathbf{x}^*\|_1\nonumber\\  
&\qquad\leq\exp(-\sum_{k=0}^{n}c_e^k)\sup_{\mathbf{x}^{*}\in\mathcal{X}(B_{\mathbf{x}},\mathbb{S})}\|\mathbf{x}^{0}-\mathbf{x}^*\|_1\nonumber\\  
&\qquad\leq|\mathbb{S}|B_\mathbf{x}\exp(-\sum_{k=0}^{n}c_e^k).
\end{align}
Since $\|\mathbf{x}\|_2\leq\|\mathbf{x}\|_1$, for arbitrary $n\in\mathbb{N}$,
\begin{equation}
\|\mathbf{x}^n-\mathbf{x}^*\|_2 \leq |\mathbb{S}|B_\mathbf{x}\exp(-\sum_{k=0}^{n-1}c_e^k).
\end{equation}

$\mathbf{6)~The~value~of~c_e^n .}$

To guarantee $c_e^n$>0, the following criterion needs to be satisfied. 
\begin{align}\label{he_eq25}
0<\exp(-c_e^n)<1.
\end{align}

We first show that $0<\gamma_1^n<1$ should be satisfied. Define $\tau_{34}^n=\left|1-\gamma_4^n+\gamma_3^n\gamma_4^n\right|/\gamma_3^n$, we have if $\gamma_1^n\geq 1$
\begin{equation}
\exp(-c_e^n) =\gamma_1^n\gamma_2^n\Big(1+\mathcal{Q}^n_*\mu+\frac{\gamma_1^n-1}{\gamma_1^n}\Big)\Big(2\mathcal{Q}^n_*\mu+\tau_{34}^n\Big)+1+(\gamma_1^n-1)\gamma_2^n>1,    
\end{equation}
which means that Eq~\eqref{he_eq25} cannot hold when  $\gamma_1^n\geq 1$. 

If $0<\gamma_1^n<1$ and $0<\gamma_2^n<1/(1-\gamma_1^n)$,
\begin{align}
&\exp(-c_e^n) \nonumber\\
&=\gamma_1^n\gamma_2^n\Big(1+\mathcal{Q}^n_*\mu+\frac{1-\gamma_1^n}{\gamma_1^n}\Big)\Big(2\mathcal{Q}^n_*\mu+\tau_{34}^n\Big)+1+(\gamma_1^n-1)\gamma_2^n\nonumber\\
&=1+\gamma_2^n\Big((2(\mathcal{Q}^n_*)^2\mu^2+\tau_{34}^n\mathcal{Q}^n_*\mu+1)\gamma_1^n+2\mathcal{Q}^n_*\mu+\tau_{34}^n-1\Big).
\end{align}
Thus, when
\begin{equation}\label{he_eq27}
   0< \gamma_1^n<\frac{1-2\mathcal{Q}^n_*\mu-\tau_{34}^n}{2(\mathcal{Q}^n_*)^2\mu^2+\tau_{34}^n\mathcal{Q}^n_*\mu+1},
\end{equation}
and 
\begin{equation}
   0< \gamma_2^n<
   \min\left\{\frac{1/(1-2\mathcal{Q}^n_*\mu-\tau_{34}^n)}{1-\frac{2(\mathcal{Q}^n_*)^2\mu^2+\tau_{34}^n\mathcal{Q}^n_*\mu+1}{1-2\mathcal{Q}^n_*\mu-\tau_{34}^n}\gamma_1^n}, \frac{1}{1-\gamma_1^n}\right\},
\end{equation}
we have that Eq~\eqref{he_eq25} holds. We can easily obtain that 
\begin{equation}
    \frac{1/(1-2\mathcal{Q}^n_*\mu-\tau_{34}^n)}{1-\frac{2(\mathcal{Q}^n_*)^2\mu^2+\tau_{34}^n\mathcal{Q}^n_*\mu+1}{1-2\mathcal{Q}^n_*\mu-\tau_{34}^n}\gamma_1^n}>\frac{1}{1-\gamma_1^n},
\end{equation}
which means that $0<\gamma_2^n<1/(1-\gamma_1^n)$ suffices to get the conclusion. 
With the assumptions that $0\leq\tau_{34}^n<1$ and
\begin{equation}
    \mathcal{Q}^n_*<\frac{1-\tau_{34}^n}{2\mu},
\end{equation}
there exists such $\gamma_1^n$ and $\gamma_2^n$ to make sure that Eq~\eqref{he_eq25} holds. 

If $0<\gamma_1^n<1$ and $\gamma_2^n\geq 1/(1-\gamma_1^n)$,
\begin{align}
&\exp(-c_e^n) \nonumber\\
&=\gamma_1^n\gamma_2^n\Big(1+\mathcal{Q}^n_*\mu+\frac{1-\gamma_1^n}{\gamma_1^n}\Big)\Big(2\mathcal{Q}^n_*\mu+\tau_{34}^n\Big)+(1-\gamma_1^n)\gamma_2^n-1\nonumber\\
&=\gamma_2^n\Big((2(\mathcal{Q}^n_*)^2\mu^2+\tau_{34}^n\mathcal{Q}^n_*\mu-1)\gamma_1^n+2\mathcal{Q}^n_*\mu+\tau_{34}^n+1\Big)-1.
\end{align}
As
\begin{equation}
\frac{2(\mathcal{Q}^n_*)^2\mu^2+\tau_{34}^n\mathcal{Q}^n_*\mu-1}{2\mathcal{Q}^n_*\mu+\tau_{34}^n+1}\gamma_1^n+1>1-\gamma_1^n\geq\frac{1}{\gamma_2^n},
\end{equation}
we have that $\exp(-c_e^n)>0$ holds. If 
\begin{equation}\label{he_eq26}
\frac{1}{1-\gamma_1^n}\leq\gamma_2^n\\ <\frac{2}{(2(\mathcal{Q}^n_*)^2\mu^2+\tau_{34}^n\mathcal{Q}^n_*\mu-1)\gamma_1^n+2\mathcal{Q}^n_*\mu+\tau_{34}^n+1},
\end{equation}
we have that $\exp(-c_e^n)<1$ holds. One can find that the above inequation holds when $\gamma_1^n$ satisfies Eq~\eqref{he_eq27} with the assumptions that $0\leq \tau_{34}^n<1$ and $\mathcal{Q}^n_*<(1-\tau_{34}^n)/2\mu$.

Finally, we discuss the choices of $\gamma_3^n$ and $\gamma_4^n$ to guarantee $0\leq \tau_{34}^n<1$. Note that $\gamma_3^n\not=0$ and $\gamma_3^n\not=1$. Then we have
\begin{align}\label{he_eq28}
\frac{\left|1-\gamma_4^n+\gamma_3^n\gamma_4^n\right|}{\gamma_3^n}&<1, \nonumber\\
-\gamma_3^n<1-\gamma_4^n+\gamma_3^n\gamma_4^n&<\gamma_3^n.
\end{align}
Eq.~\eqref{he_eq28} means that
\begin{equation}
   (1-\gamma_3^n)(1-\gamma_4^n)<0,~1+\gamma_3^n+\gamma_4^n(\gamma_3^n-1)>0. 
\end{equation}
It is obvious that the above inequation holds when $\gamma_3^n>1$ and $0<\gamma_4^n<1$, or $0<\gamma_3^n<1$ and $1<\gamma_4^n<(1+\gamma_3^n)/(1-\gamma_3^n)$.

As a result, we draw Theorem~8.

\begin{figure*}[!t]
\renewcommand{\baselinestretch}{1.0}
\centering
\subfigure[$N_{\mathcal{W}^{n}}(x)\equiv 0$ vs. Learned $N_{\mathcal{W}^{n}}$]{
\includegraphics[width=0.48\textwidth]{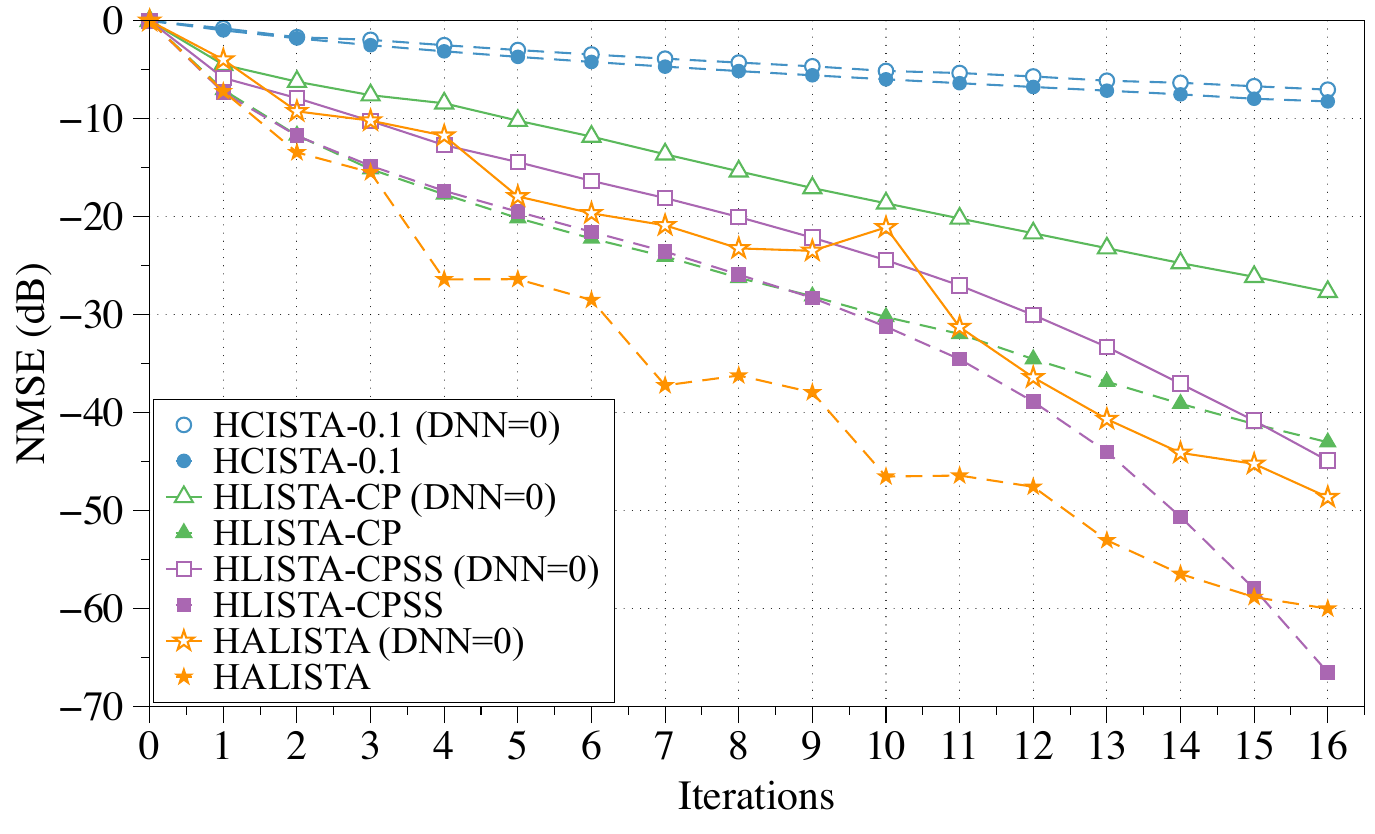}}
\subfigure[Hybrid ISTA with $N_{\mathcal{W}^{n}}(x)\equiv 0$ vs. Baselines]{
\includegraphics[width=0.48\textwidth]{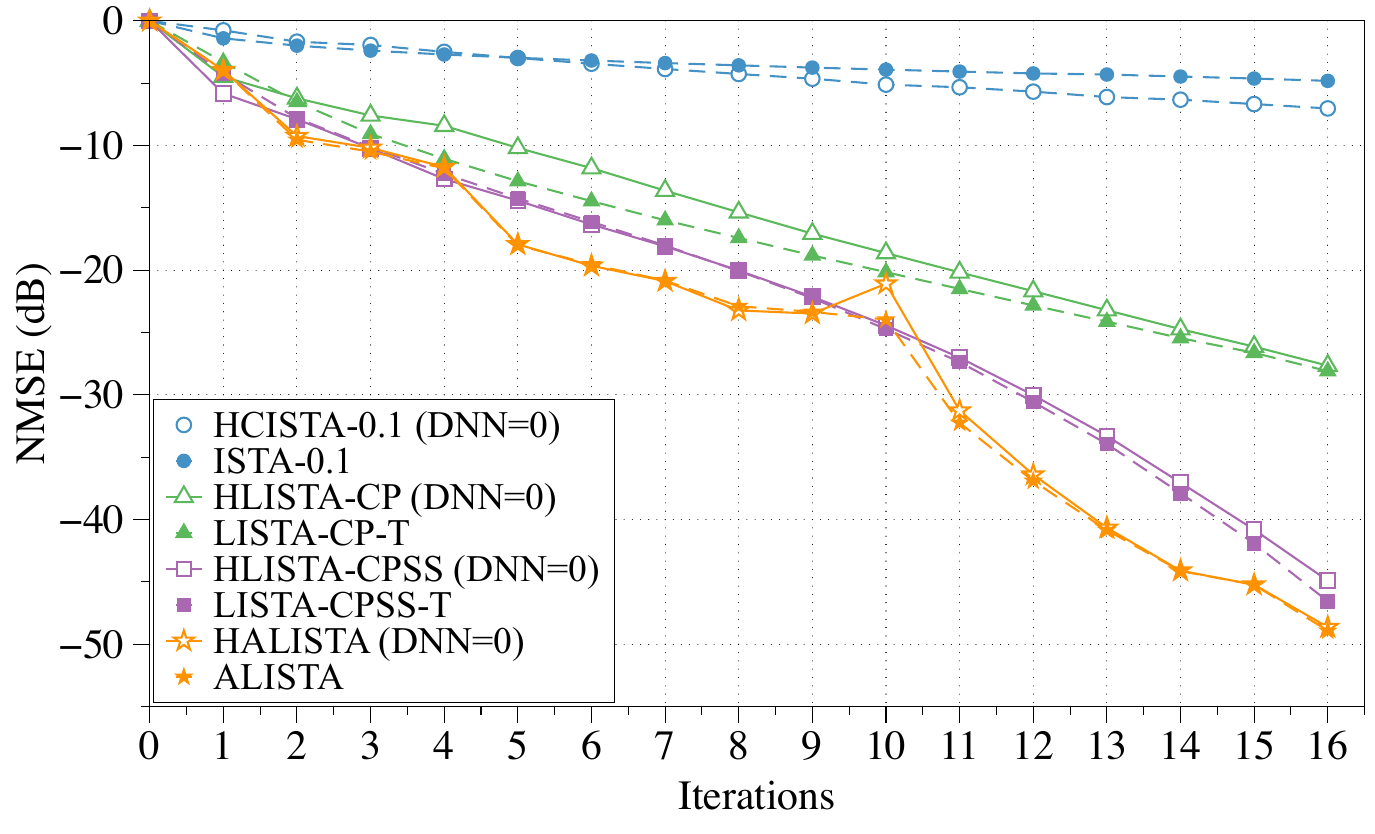}}
\caption{The evaluation of NMSE performance of hybrid ISTA models with $N_{\mathcal{W}^{n}}(\mathbf{x})\equiv 0$ for $\forall \mathbf{x} \in \mathbb{R}^{N}$, where $N_{\mathcal{W}^{n}}$ represents the DNN function as in Eq.~(5), (20), and (31). The left (resp. right) figure elaborates the comparisons between hybrid ISTA models with $N_{\mathcal{W}^{n}}(\mathbf{x})\equiv 0$ and learned $N_{\mathcal{W}^{n}}$ (resp. baselines).} 
\label{rfig1}
\end{figure*}

\section{More Discussions}
\subsection{Comparisons between Hybrid ISTA and Gated LISTA}
As the gain gates and overshoot gates of Gated LISTA seem to be similar to the free-form DNNs and the balancing parameter $\alpha^n$ of hybrid ISTA, we discuss the differences from the perspectives of motivation, formulations (update steps and parameters), theoretical analysis.

\textbf{Motivation.} We first discuss the motivation of two gate mechanisms. Proposition~1 in~[32] found that the components of $\mathbf{x}^n$ must be smaller than or at most equal to those of the $\mathbf{x}^*$, \emph{i.e.}, $|x^n_j|\leq|x^*_j|, ~\forall j\in [1, N]$. To enlarge $\mathbf{x}^n$ to improve the performance, gain gates were proposed to act on the $\mathbf{x}^n$ in the $n$th iteration. Overshoot gates were inspired by the analysis of classical ISTA and were empirically proposed for LISTA to improve the performance. In a word, the two gate mechanisms were proposed to boost the reconstruction performance. By contrast, free-form DNNs are introduced into classical ISTA and LISTA to bring in deep learning technology, relax the restriction on network architectures and improve the performance, while still guaranteeing the convergence. More importantly, we also provide an interesting direction for designing interpretable DNNs used for solving inverse problems.

\textbf{Formulations (Update Steps and Parameters).} We begin with a brief description of the two gate mechanisms in Gated LISTA. When gain gates are introduced into LISTA-CP, the $n$th iteration is formulated as
\begin{equation}
\mathbf{x}^{n+1}= \mathcal{S}_{\theta^{n}}\Big(\mathbf{x}^{n}\odot g_t(\mathbf{x}^{n}, \mathbf{b}|\Lambda_g^n)+(\mathbf{W}^{n})^{T}(\mathbf{b}-\mathbf{A}\mathbf{x}^{n}\odot g_t(\mathbf{x}^{n}, \mathbf{b}|\Lambda_g^n))\Big),
\end{equation}
where the gate function $g_t(\cdot,\cdot|\Lambda_g^n)$ outputs an $N$-dimension vector using a set of its learnable parameters $\Lambda_g^n$ in the $n$th iteration, and $\odot$ represents element-wise multiplication of two vectors. 
When overshoot gates are adopted, the $n$th iteration are
\begin{align}
\mathbf{\bar{x}}^{n+1}&= \mathcal{S}_{\theta^{n}}\left(\mathbf{x}^{n}+(\mathbf{W}^{n})^{T}(\mathbf{b}-\mathbf{Ax}^{n})\right), \nonumber \\
\mathbf{x}^{n+1} &= o_t(\mathbf{x}^{n}, \mathbf{b}|\Lambda_o^n)\odot\mathbf{\bar{x}}^{n+1} +(1-o_t(\mathbf{x}^{n}, \mathbf{b}|\Lambda_o^n)) \odot \mathbf{x}^{n},
\end{align}
where the gate function $o_t(\cdot,\cdot|\Lambda_o^n)$ with the set of learnable parameters $\Lambda_o^n$ produces an $N$-dimension vector. If both gate mechanisms are adopted, the $n$th iteration are formulated as (we write it in the similar form of HLISTA-CP for clear comparison)
\begin{align}\label{r3p2-1}
\mathbf{u}^{n} & = \mathbf{x}^{n}\odot g_t(\mathbf{x}^{n}, \mathbf{b}|\Lambda_g^n), \nonumber\\
\mathbf{w}^{n}&= \mathcal{S}_{\theta^{n}}\left(\mathbf{u}^{n}+(\mathbf{W}^{n})^{T}(\mathbf{b}-\mathbf{Au}^{n})\right), \nonumber \\
\mathbf{x}^{n+1} &= o_t(\mathbf{x}^{n}, \mathbf{b}|\Lambda_o^n)\odot\mathbf{w}^{n} +(1-o_t(\mathbf{x}^{n}, \mathbf{b}|\Lambda_o^n)) \odot \mathbf{x}^{n}.
\end{align}
Recall the steps in the $n$th iteration of HLISTA-CP as follows.
\begin{equation}\label{r3p2-2}
\begin{aligned}
&\mathbf{v}^{n}=\mathcal{S}_{\theta_{1}^{n}}\left(\mathbf{x}^{n}+(\overline{\mathbf{W}}^{n})^{T}(\mathbf{b}-\mathbf{Ax}^{n})\right), \\
&\mathbf{u}^{n}=N_{\mathcal{W}^{n}}(\mathbf{v}^{n}), \\
&\mathbf{w}^{n}=\mathcal{S}_{\theta_{2}^{n}}\left(\mathbf{u}^{n}+(\widehat{\mathbf{W}}^{n})^{T}(\mathbf{b}-\mathbf{Au}^{n})\right),\\
&\mathbf{x}^{n+1}=\alpha^{n}\mathbf{v}^{n}+(1-\alpha^{n})\mathbf{w}^{n}.
\end{aligned}
\end{equation}
Then we compare HLISTA-CP with the Gated LISTA. Though the steps in Eq.~\eqref{r3p2-1} and \eqref{r3p2-2} look alike, they are different. First, two proximal gradient descent steps are adopted in Eq.~\eqref{r3p2-2} but only one in \eqref{r3p2-1}, which makes that Eq.~\eqref{r3p2-1} or \eqref{r3p2-2} cannot be viewed as a special case for the other one. This also leads to many differences in the proofs for convergence. Second, the inserted DNNs are free-form in Eq.~\eqref{r3p2-2} while the gain gate function $g_t(\cdot,\cdot|\Lambda_g^n)$ in Eq.~\eqref{r3p2-1} is restricted. In addition to the limited operation of element-wise multiplication, the $i$the element of $g_t(\mathbf{x}^{n}, \mathbf{b}|\Lambda_g^n)$ needs to satisfy $1\leq g_t(\mathbf{x}^{n}, \mathbf{b}|\Lambda_g^n)_i < 2\theta^n/|x_i^n| + 1$ to guarantee the convergence (see Eq.~(12) and (13) in [32] for more details). Gated ISTA~[32] only provided three choices for gain gate functions (see Eq.~(16) in [32]). By contrast, the inserted DNNs in our methods have no such restriction. Third, the overshoot gate function $o_t(\cdot,\cdot|\Lambda_o^n)$ in Eq.~\eqref{r3p2-1} is totally different from $\alpha^n$ in \eqref{r3p2-2}. The function  $o_t(\cdot,\cdot|\Lambda_o^n)$ outputs a vector where the elements are required to be greater than 1, \emph{i.e.}, $ o_t(\mathbf{x}^{n}, \mathbf{b}|\Lambda_o^n)_i>1$ (see Proposition~2 and Eq.~(18) in [32] for more details). To satisfy this condition, Gated ISTA provided two choices for overshoot gate functions (see Eq.~(18) in [32]). By contrast, $\alpha^n$ is a scalar and should be smaller than or equal to 1, \emph{i.e.}, the ranges are totally different.

\textbf{Theoretical Analysis.} 
Due to the differences of update steps and parameters discussed above, the proofs for convergence are different (see Theorem~4 for HLISTA-CP and Theorem~2 and 3 in~[32] for more details). In addition, the overshoot gates were proposed empirically without theoretical guarantees, while the balancing parameter $\alpha^n$ is indispensable in hybrid ISTA models and the proofs. Here we briefly introduce how the overshoot gates were proposed. K. Wu \emph{et al.}~[32] first improved classical ISTA as follows:
\begin{align}
\mathbf{\bar{x}}^{n+1}&=\mathcal{S}_{\lambda t}\left(\mathbf{x}^{n}-t\mathbf{A}^{T}(\mathbf{Ax}^{n}-\mathbf{b})\right), \nonumber\\
\mathbf{x}^{n+1}&=\eta\mathbf{\bar{x}}^{n+1} +(1-\eta) \mathbf{x}^{n},
\end{align}
where $\eta$ is a scalar for adjusting the output.
They found that $\eta>1$ is a better choice than the common choice $\eta=1$ through theoretical analysis (see Proposition~2 in [32]). Then they empirically extended this improvement to LISTA and changed the scalar $\eta$ to be the overshoot gate function $o_t(\cdot,\cdot|\Lambda_o^n)$ that outputs a vector for a better reconstruction performance. In other words, the convergence of Gated LISTA with overshoot gates has not been proved (see the paragraph above Fig.~2 in [32]).
By contrast, $\alpha^n$ in our models is used to adjust the proportion of $\mathbf{v}^n$ and $\mathbf{w}^n$ and consequently guarantee the convergence. Additionally, the derivation of $\alpha^n$ is not inspired by the analysis in [32] as $\alpha^n\leq 1$ in our models. 

\begin{figure*}[!t]
\renewcommand{\baselinestretch}{1.0}
\centering
\includegraphics[width=0.8\textwidth]{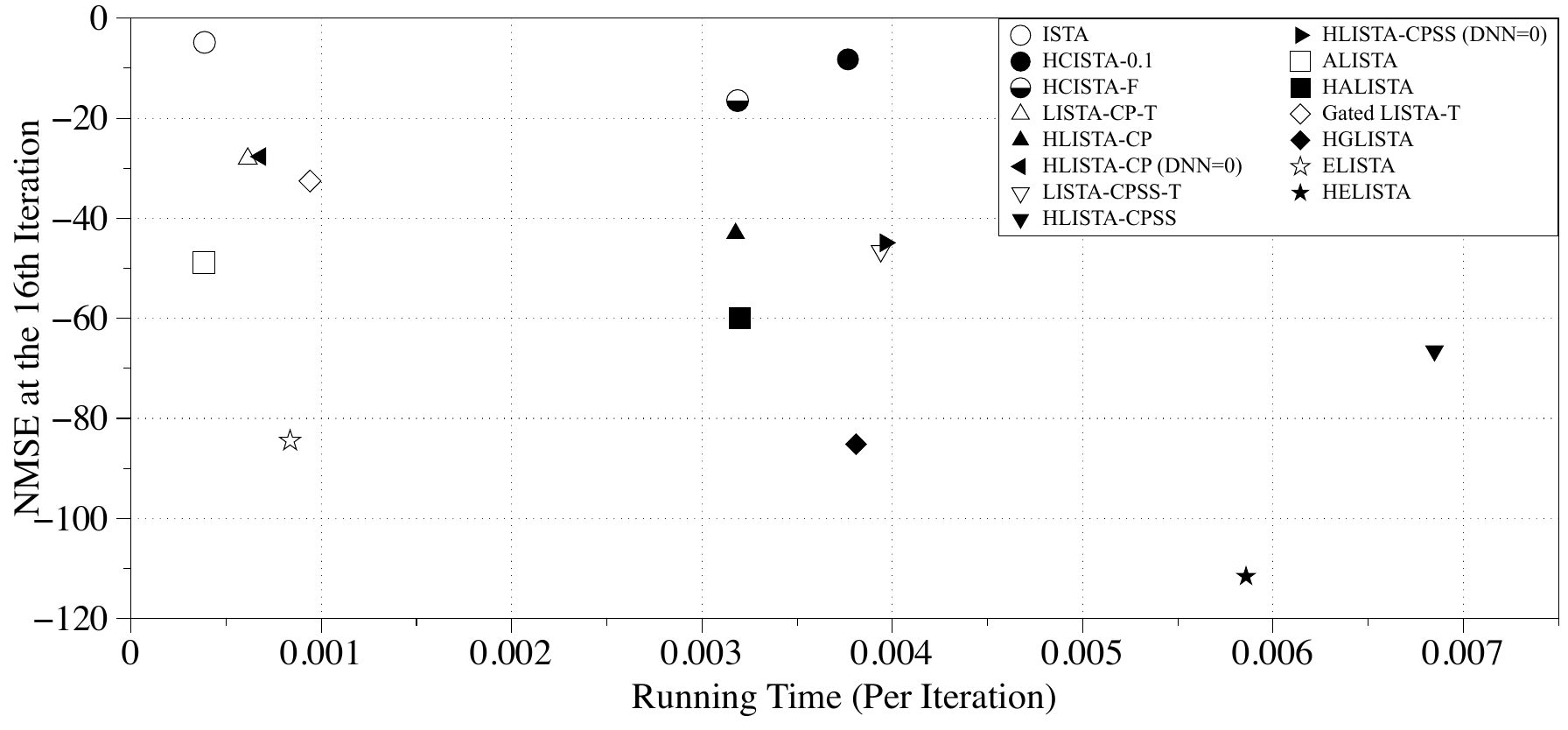}
\caption{Running time per iteration and NMSE at the 16th iteration for hybrid models and the corresponding baselines. HLISTA-CP (DNN=0) / CPSS (DNN=0) mean that the models are equipped with $N_{\mathcal{W}^{n}}(\mathbf{x})\equiv \mathbf{0}$.}
\label{rfig2}
\end{figure*}

\subsection{Functions Expressed by Inserted DNNs}
In this section, we discuss how the free-form DNNs enter the analysis.

The proposed hybrid ISTA methods can support network architectures without constraints and guarantee the convergence. However, it does not mean that any functions of the inserted DNNs are supported. 
Generally speaking, a DNN function is determined by the loss function for training (we adopt MSE function as [24], [30], [31], [32], [33] since we focus on the Lasso problem), rather than the network architecture, unless a network model without learnable parameters is adopted. 
For hybrid ISTA models, implicit requirements on DNN functions are presented to guarantee the convergence, \emph{e.g.}, bounds of $\eta^n$ in Assumption~1, and the choice of $\theta_2^n$ in Theorems~4, 5, 6, and 7. We further validate in the experiments that Assumption~1 can be easily satisfied and proper $\theta_2^n$ can be obtained in a data-driven manner.

However, network architectures do have a considerable impact on the performance, when the networks are required to approximate the desired functions. 
In general, DNNs with feasible and reasonable architectures are more likely to obtain a better approximation, leading to a faster convergence rate in comparison to those unfeasible or unreasonable DNNs.
Thus, we suggest the inserted DNNs to be feasible and reasonable from a practical perspective, but it is not a formal assumption to support our theoretical results. Therefore, we alternatively provide implicit requirements on DNN functions to guarantee the convergence.

We further evaluate the NMSE performance of hybrid ISTA models with $N_{\mathcal{W}^{n}}(\mathbf{x})\equiv \mathbf{0}$ for $\forall \mathbf{x} \in \mathbb{R}^{N}$, where $N_{\mathcal{W}^{n}}$ represents the DNN function as in Eq.~(5), (20), and (31). As shown in Fig.~\ref{rfig1}, the performance of $N_{\mathcal{W}^{n}}(\mathbf{x})\equiv \mathbf{0}$ degenerates in comparison to learned DNN obtained by training, but is extremely close to the baselines, as the DNN makes no progress. This  reveals that learned $N_{\mathcal{W}^{n}}$ via training owns better properties than handcrafted $N_{\mathcal{W}^{n}}$.

\section{Experimental Details and More Experimental Results}
\subsection{Sparse Recovery}
\subsubsection{Running Time}

Fig.~\ref{rfig2} shows the running time per iteration and NMSE at the 16th iteration for hybrid models and the corresponding baselines. HLISTA-CP (DNN=0)/CPSS (DNN=0) mean that the models are equipped with $N_{\mathcal{W}^{n}}(\mathbf{x})\equiv \mathbf{0}$. The proposed models usually take more time to obtain a much higher performance than the baselines. However, the comparisons between LISTA-CP-T/CPSS-T and HLISTA-CP (DNN=0)/CPSS (DNN=0) show that the hybrid models and the corresponding baselines own comparable computational complexity.  
This result suggests that the extra time costs of hybrid ISTA models are almost caused by the inserted DNNs. Though incorporating DNNs greatly improves the performance, reduce time cost of DNN inference is worth noticing. In future, we will explore possible alternatives to further reduce the complexity of inserted DNNs, \emph{e.g.}, network pruning~\cite{han2015learning}, network quantization~\cite{han2015deep}, matrix/tensor factorization, and low-rank approximation~\cite{sainath2013low, jaderberg2014speeding}.

\begin{figure*}[!t]
\renewcommand{\baselinestretch}{1.0}
\centering
\includegraphics[width=6.3in]{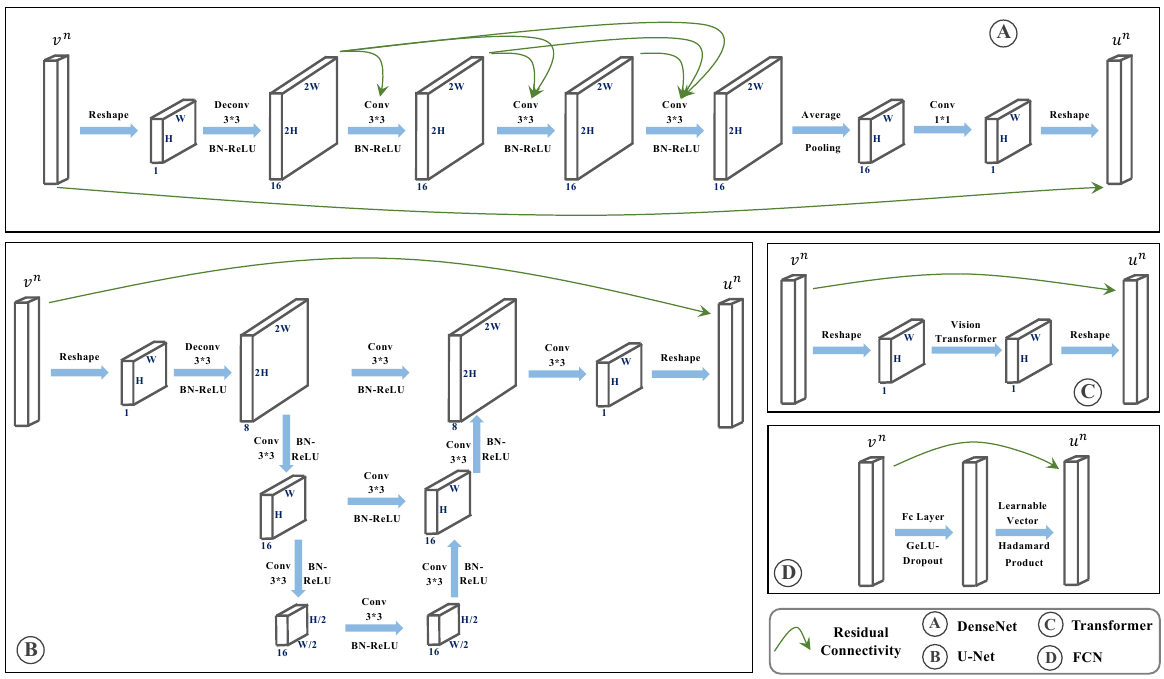}
\caption{Illustration of network architectures for complicated DNNs. To be concrete, the detailed architectures of DenseNet, U-Net, and FCN are illustrated, and one can refer to [53] for a detailed description of Vision Transformer. More details are shown in Appendix~\ref{sec:sc_complicated_DNNs}.
}\label{complicated_DNNs}
\end{figure*}

\begin{figure*}[!t]
\renewcommand{\baselinestretch}{1.0}
\centering
\subfigure[ISTA vs. HCISTA with Complicated DNNs]{
\includegraphics[width=0.48\textwidth]{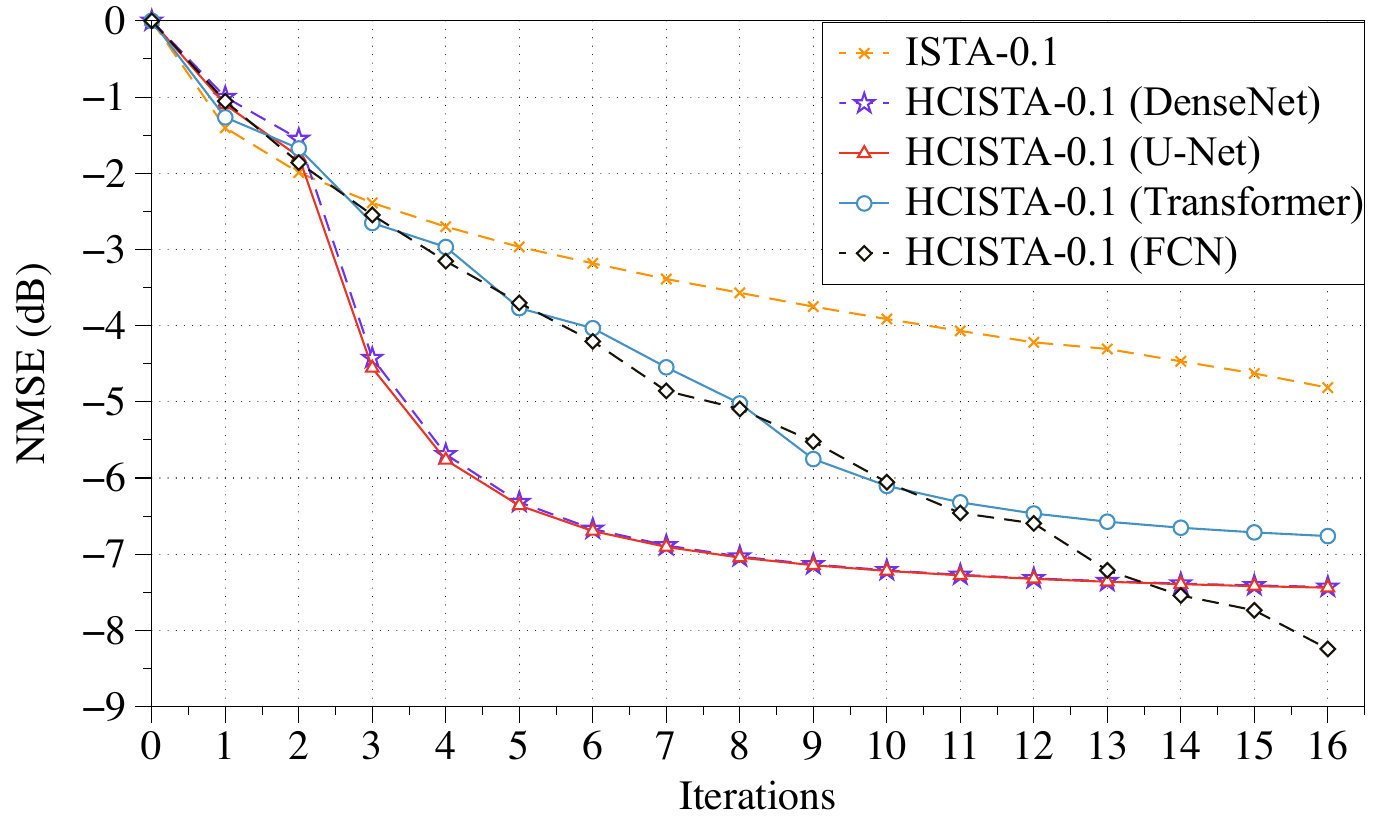}}
\subfigure[LISTA-CP-U vs. HLISTA-CP with Complicated DNNs]{
\includegraphics[width=0.48\textwidth]{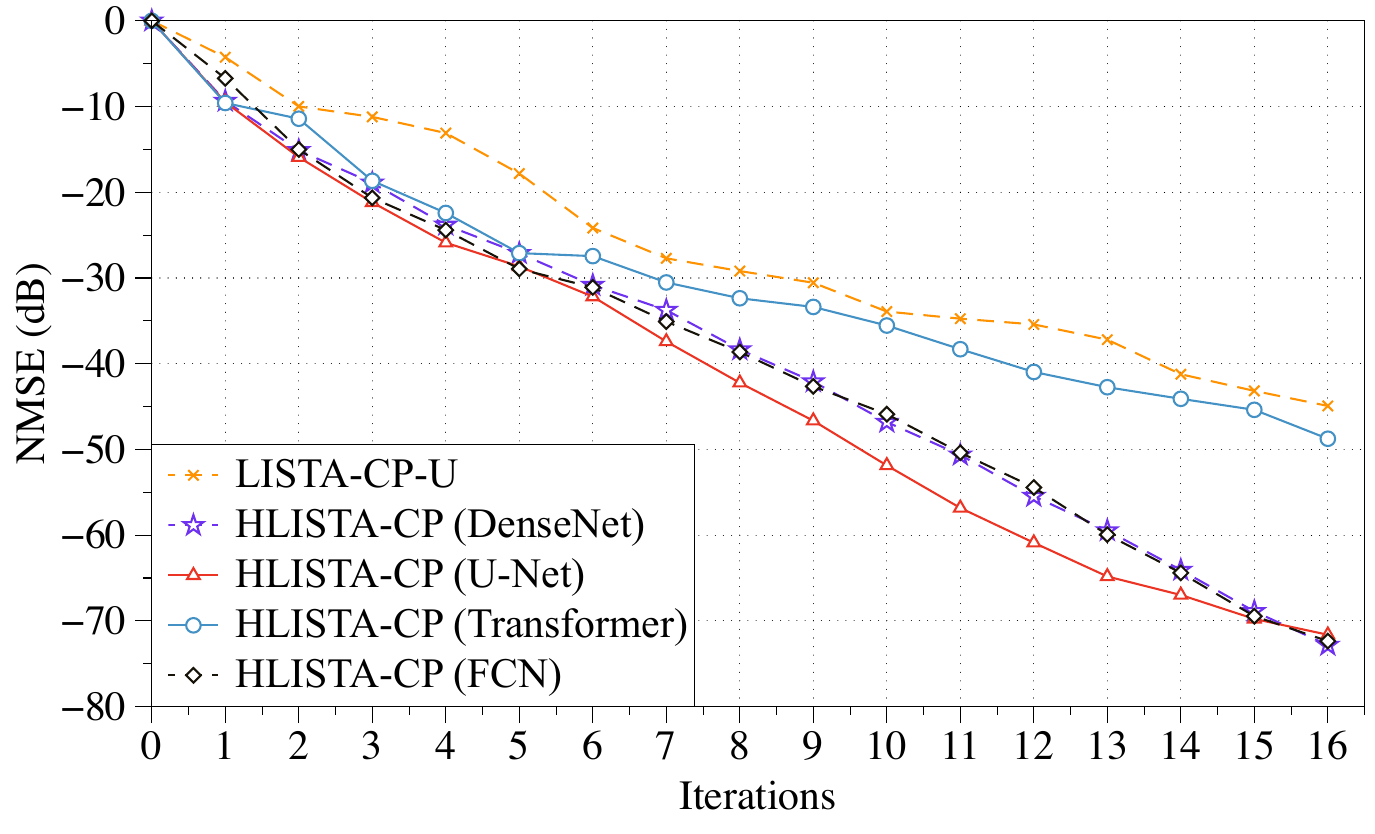}}
\caption{The evaluation of NMSE performance of HCISTA $\&$ HLISTA-CP with complicated DNNs. Specifically, four complicated DNNs, dubbed DenseNet, U-Net, Vision Transformer and FCN, are introduced to hybrid models.} 
\label{complex_sc}
\end{figure*}

\begin{figure*}[!t]
\renewcommand{\baselinestretch}{1.0}
\centering
\subfigure[HCISTA with Trained/Untrained Simple DNNs]{
\includegraphics[width=0.48\textwidth]{figures/FpTp_unorm/eta.pdf}}
\subfigure[HCISTA-0.1 with Complicated DNNs]{
\includegraphics[width=0.48\textwidth]{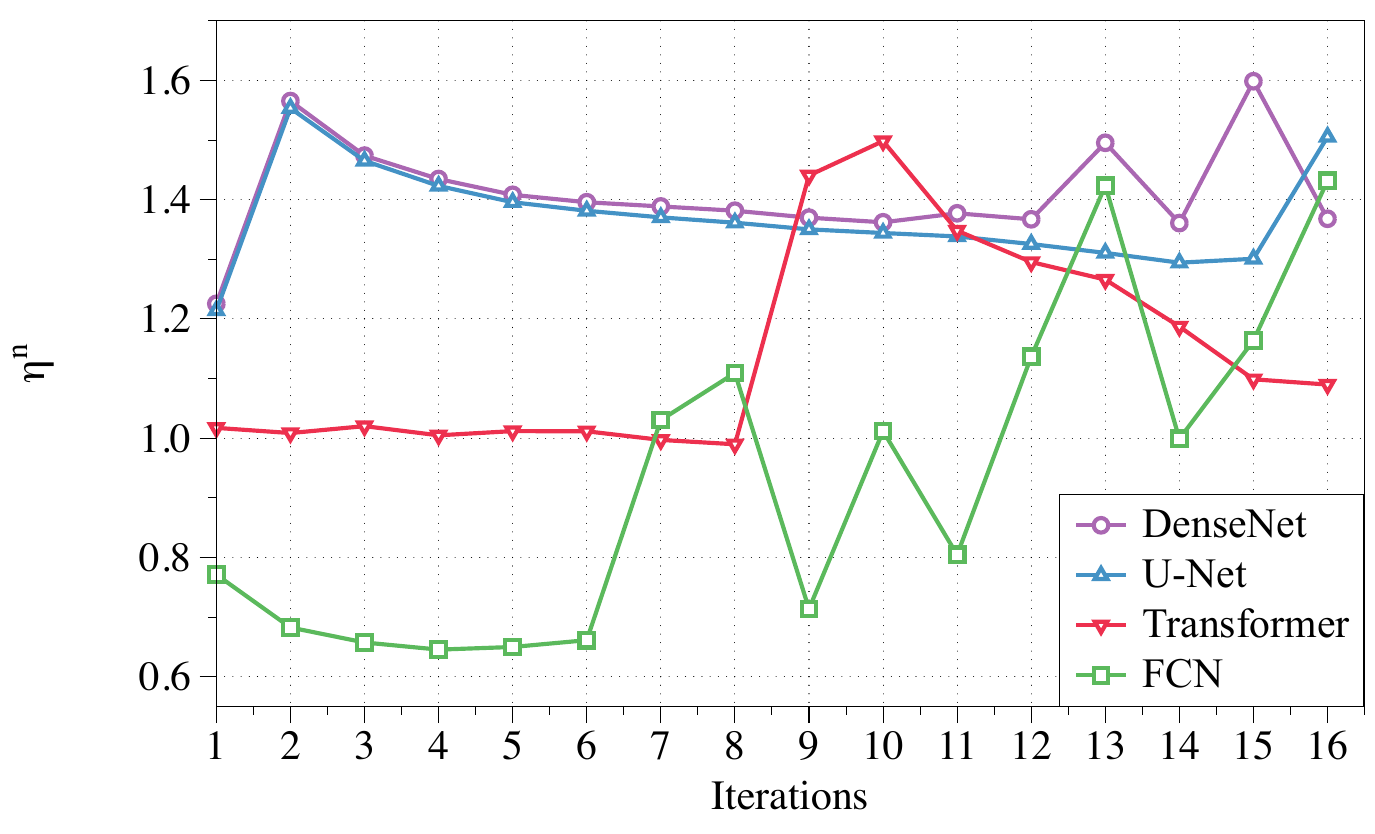}}
\caption{The values of $\eta^n$, $n=1,\cdots,16$ defined in Assumption~1 for HCISTA with trained/untrained simple DNNs and complicated DNNs.} 
\label{complex_assump1}
\end{figure*}

\subsubsection{HCISTA $\&$ HLISTA-CP with Complicated DNNs}
\label{sec:sc_complicated_DNNs}

We evaluate the reconstruction performance with complicated DNNs on the task of sparse recovery. Four complicated DNNs are adopted, including DenseNet~[52], U-Net~[58], Vision Transformer~[53] and fully-connected networks (FCN). To be concrete, we illustrate the network architectures in Fig.~\ref{complicated_DNNs} and elaborate the four DNNs as follows. 

\begin{itemize}[leftmargin=*]
\item 
\textbf{DenseNet.}
Without considering the dimension of batch size, the input vector $\mathbf{v^n}$ is first reshaped as a matrix of size H$\times$W$\times$1 (height$\times$weight$\times$channel). The the matrix is upsampled to the size of 2H$\times$2W$\times$16 via a deconvolution (transposed convolution) operation and then passes through three convolution layers with dense connectivity. Batch normalization and ReLU are utilized after the deconvolution and convolution layers. Then average pooling is utilized for downsampling and a convolution layer is adpoted to reduce the channel numbers. 
 
\item 
\textbf{U-Net.}
Without considering the dimension of batch size, the input vector $\mathbf{v^n}$ is first reshaped as a matrix of size H$\times$W$\times$1. The the matrix is upsampled to the size of 2H$\times$2W$\times$16 via a deconvolution operation and then passes through a neural network shaped like `U'. Finally a convolution layer is adpoted to reduce the channel numbers. 

\item 
\textbf{Transformer.}
We utilize one Vision Transformer network block in each iteration. Vision Transformer includes many DNN components and tricks such as fully connected layers, layer normalization, non-linear activation function like Softmax, attention mechanism and dropout.
As the architecture of Vision Transformer is complicated, we do not elaborate it here. Please refer to~[53] for more details. In addition, we implement this hybrid ISTA model with reference to \textit{https://github.com/emla2805/vision-transformer} and adopt the default parameters of the source code.

\item 
\textbf{FCN.}
The input vector $\mathbf{v^n}$ first passes through a fully-connected layer, followed by GeLU and Dropout. Then the output is multiplied by a learnable vector via the Hadamard product. 
\end{itemize}
Note that $\mathbf{v^n}$ and $\mathbf{u^n}$ are connected with a shortcut in all the four DNNs.

As we adopt complicated DNNs, signals with higher dimensions are utilized for training and testing, \emph{i.e.}, $\mathbf{x}^*\in\mathbb{R}^{1024}$ and $\mathbf{A}\in\mathbb{R}^{512\times 1024}$. We generate $\mathbf{x}^*$ and $\mathbf{A}$ in the same manner as Section~6.1 and [30], [31], [32], as well as the loss function, training strategy and hyperparameters. 

As shown in Fig.~\ref{complex_sc}, all the hybrid ISTA models outperform the corresponding baselines in terms of NMSE. In addition, we find that the architectures of inserted DNNs have impact on the performance to some extent. For example, adopting Transformer seems to obtain a relatively higher NMSE in comparison to the others.

\begin{figure*}[!t]
\renewcommand{\baselinestretch}{1.0}
\centering
\subfigure[]{
\includegraphics[width=0.32\textwidth]{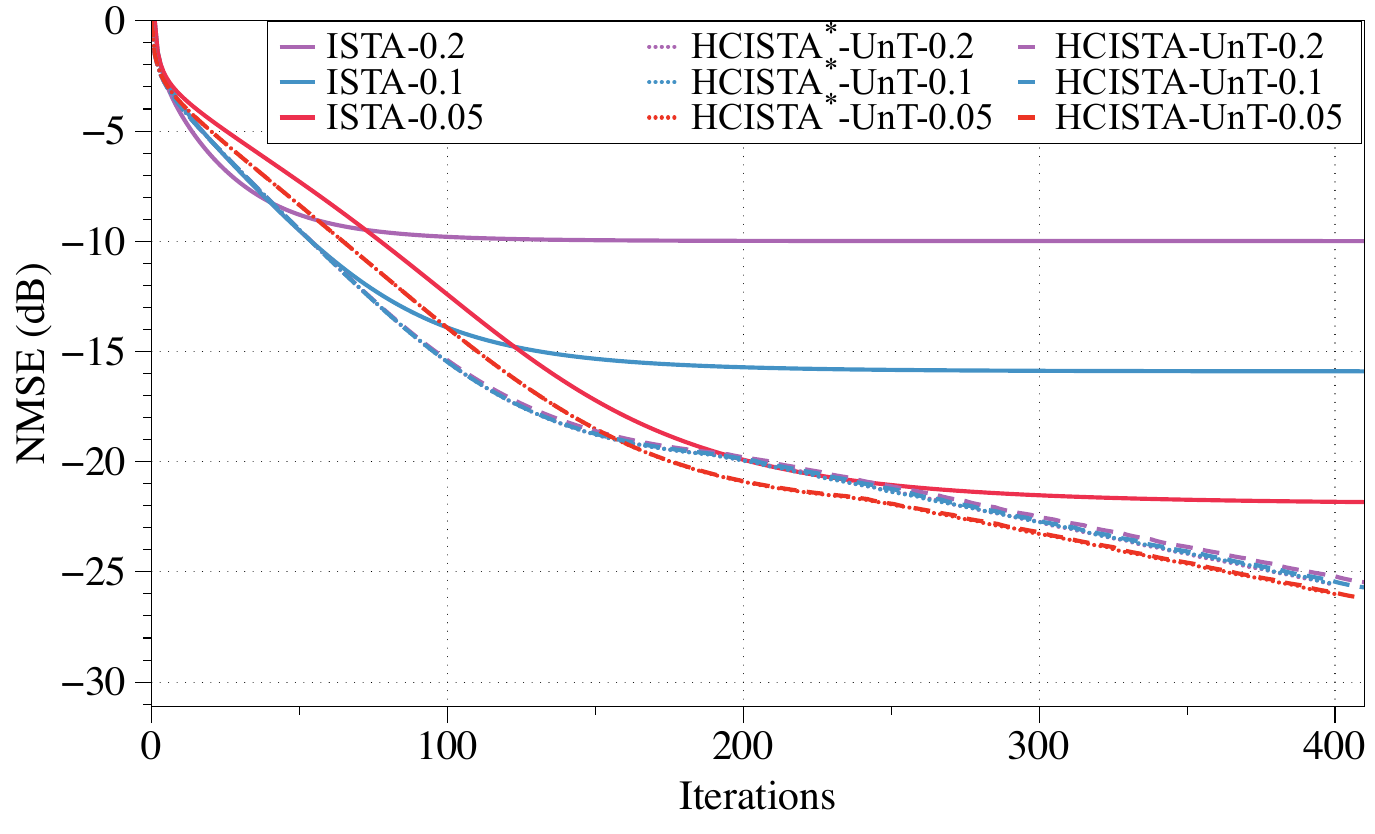}}
\subfigure[]{
\includegraphics[width=0.32\textwidth]{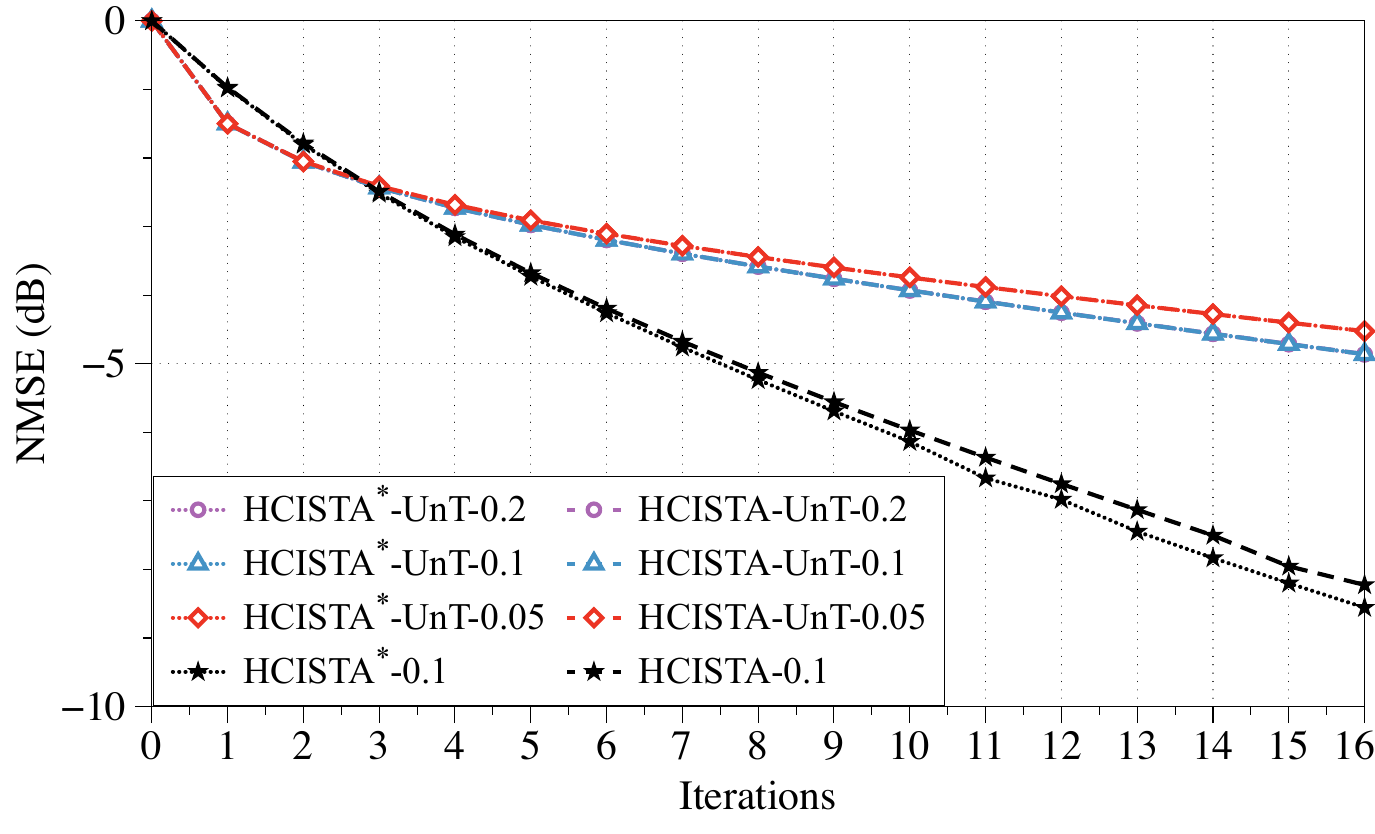}}
\subfigure[]{
\includegraphics[width=0.32\textwidth]{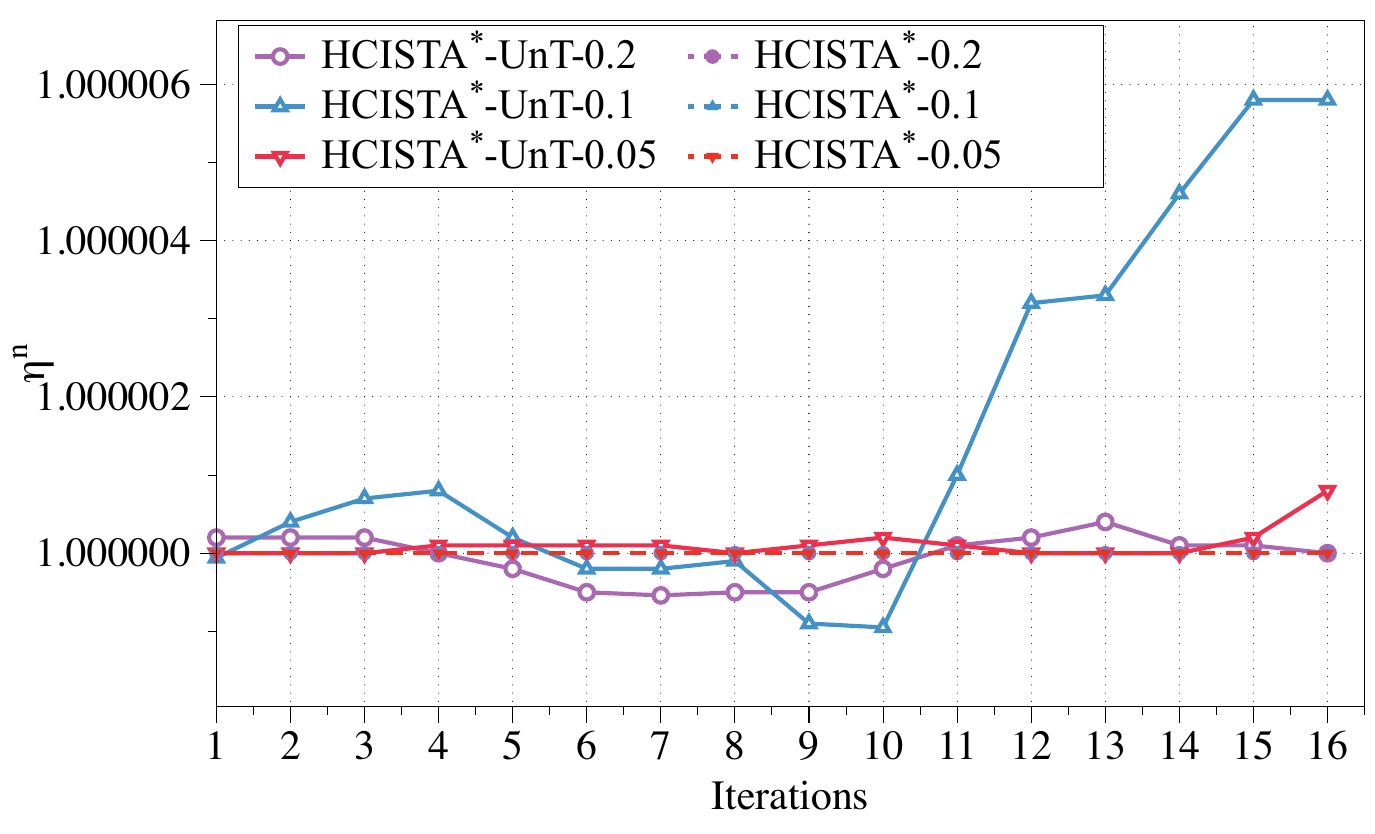}}
\caption{NMSE for ISTA, HCISTA$^*$, and HCISTA and the values of $\eta^n$, $n=1,\cdots,16$ defined in Assumption~1 for HCISTA$^*$ with trained/untrained DNNs as Eq.~\eqref{r2p2-4}.} 
\label{lip_assump1}
\end{figure*}

\subsubsection{Clarification of Assumption~1}

\begin{table*}[!t]
\renewcommand{\baselinestretch}{1.0}
\renewcommand{\arraystretch}{1.0}
\centering
\caption{Comparison of average PSNR (dB) | SSIM on \emph{Set11} obtained at the measurement rates (MRs) of 0.04, 0.10, 0.25 and 0.50 for complicated DNNs.}\label{T1_cs_c}
\begin{tabular}{lccccc}
\toprule
Methods & MR=0.50 & MR=0.25 & MR=0.10 & MR=0.04 \\
\midrule
ISTA (Convergence) & 27.63 | 0.8372 & 20.16 | 0.5874 & 17.29 | 0.4048 & 9.301 | 0.1045 \\
HCISTA (DenseNet) & 33.72 | 0.9403 & 29.17 | 0.8680 & 24.54 | 0.7408 & 21.77 | 0.6193  \\
HCISTA (U-Net) & \textbf{33.84 | 0.9413} & \textbf{29.69 | 0.8775} & \textbf{24.97 | 0.7571} & \textbf{21.92 | 0.6291}  \\
HCISTA (Transformer) & 33.14 | 0.9338 & 28.31 | 0.8508 & 23.91 | 0.7094 & 21.74 | 0.6122  \\
HCISTA (FCN) & 32.40 | 0.9230 & 27.62 | 0.8297 & 23.67 | 0.6973 & 21.43 | 0.5935  \\
\midrule
LISTA-CP-U & 35.03 | 0.9520 & 30.17 | 0.8929 & 25.29 | 0.7735 & 22.14 | 0.6462 \\
HLISTA-CP (DenseNet) & 36.60 | 0.9641 & 31.11 | 0.9038 & 25.64 | 0.7825 & \textbf{22.25} | 0.6512  \\
HLISTA-CP (U-Net) & \textbf{36.88 | 0.9639} & \textbf{31.35 | 0.9085} & \textbf{25.76 | 0.7873} & \textbf{22.25 | 0.6514}  \\
HLISTA-CP (Transformer) & 35.45 | 0.9560 & 30.43 | 0.8960 & 25.55 | 0.7817 &  22.22 | 0.6501  \\
HLISTA-CP (FCN) & 35.24 | 0.9541 & 30.22 | 0.8941 & 25.40 | 0.7775 & 22.24 | 0.6496  \\
\bottomrule
\end{tabular}
\end{table*}

\begin{table*}[!t]
\renewcommand{\baselinestretch}{1.0}
\renewcommand{\arraystretch}{1.0}
\centering
\caption{Comparison of average PSNR (dB) | SSIM on \emph{BSD500} (50 images for test) obtained at the measurement rates (MRs) of 0.04, 0.10, 0.25, and 0.50 for complicated DNNs.}\label{T2_cs_c}
\begin{tabular}{lccccc}
\toprule
Methods & MR=0.50 & MR=0.25 & MR=0.10 & MR=0.04 \\
\midrule
ISTA (Convergence) & 26.56 | 0.7770 & 22.46 | 0.5993 & 17.82 | 0.3874 & 9.661 | 0.1247 \\
HCISTA (DenseNet) & 31.08 | 0.9028 & 27.43 | 0.7931 & 24.20 | 0.6534 & 22.16 | 0.5567 \\
HCISTA (U-Net) & \textbf{31.14 | 0.9034} & \textbf{27.66 | 0.7996} & \textbf{24.39 | 0.6620} & \textbf{22.24 | 0.5606} \\
HCISTA (Transformer) & 30.76 | 0.8970 & 26.86 | 0.7845 & 23.74 | 0.6349 & 22.03 | 0.5459 \\
HCISTA (FCN) & 30.33 | 0.8876 & 26.42 | 0.7682 & 23.57 | 0.6253 & 21.95 | 0.5379 \\
\midrule
LISTA-CP-U & 31.62 | 0.9117 & 27.65 | 0.8059 & 24.46 | 0.6652 & 22.41 | 0.5660 \\
HLISTA-CP (DenseNet) & 32.63 | 0.9249 & 28.19 | 0.8178 & 24.65 | 0.6723 & 22.46 | 0.5681 \\
HLISTA-CP (U-Net) & \textbf{32.86 | 0.9275} & \textbf{28.29 | 0.8203} & \textbf{24.66 | 0.6728} & 22.38 | 0.5677 \\
HLISTA-CP (Transformer) & 32.04 | 0.9165 & 27.87 | 0.8094 & 24.62 | 0.6704 & \textbf{22.49} | \textbf{0.5682} \\
HLISTA-CP (FCN) & 31.84 | 0.9146 & 27.75 | 0.8066 & 24.54 | 0.6688 & \textbf{22.49} | 0.5672 \\
\bottomrule
\end{tabular}
\end{table*}

Though $\eta_c$ is not predefined, we show in Fig.~4 that Assumption~1 can be easily satisfied in the experiments with a relatively small $\eta_c$ when adopting simply constructed DNNs.  We further evaluate $\eta^n$ with complicated DNNs including DenseNet~[52], U-Net~[58], Transformer~[53] and fully connected networks. \figurename~\ref{complex_assump1} plots the values of $\eta^n$ for the 16 iterations for HCISTA with trained/untrained simple DNNs and complicated DNNs. 
For trained/untrained HCISTA with simply constructed DNNs, the values of $\eta^n$ are very close to 1. By contrast, $\eta^n$ fluctuates for complicated DNNs. Nevertheless, one can see that the value is between 0.6 and 1.6. This fact implies that Assumption~1 can be easily satisfied with a relatively small $\eta_c$ even though $\eta_c$ is not predefined when complicated DNNs are adopted.

In addition, Assumption~1 can be satisfied without trial-and-error and predefining $\eta_c$.
We further provide an example of HCISTA with incorporated DNNs that can satisfy Assumption~1 without trial-and-error and predefining $\eta_c$, and show that the efficiency of HCISTA is maintained.
% We further evaluate the performance of HCISTA with inserted DNNs that can satisfy Assumption~1 without trial-and-error and predefining $\eta_c$, and show that the efficiency of HCISTA is maintained. 
Recall that the $n$th iteration of HCISTA is formulated as follows.
\begin{equation}\label{r2p2-1}
\begin{aligned}
&\mathbf{v}^{n}= \mathcal{S}_{\lambda^n t^{n}}(\mathbf{x}^{n}-t^{n}\nabla f(\mathbf{x}^{n})), \\
&\mathbf{u}^{n}=N_{\mathcal{W}^{n}}(\mathbf{v}^{n}), \\
&\mathbf{w}^{n}= \mathcal{S}_{\lambda^n t^{n}}(\mathbf{u}^{n}-t^{n}\nabla f(\mathbf{u}^{n})),\\
&\mathbf{x}^{n+1}= \alpha^{n} \mathbf{v}^{n}+(1-\alpha^{n})\mathbf{w}^{n},
\end{aligned}
\end{equation}
where $\alpha^n, t^n$ and $\lambda^n$ are bounded as Eq.~(7), (8) and (9), respectively. We slightly change the step of generating $\mathbf{u}^{n}$ in Eq.~\eqref{r2p2-1} as follows.
\begin{equation}\label{r2p2-2}
\mathbf{u}^{n}=N_{\mathcal{W}^{n}}(\mathbf{v}^{n}, \mathbf{x}^{n}).
\end{equation}
Note that this change has no influence on the theoretical results and proofs. Furthermore, in this example, we specify Eq.~\eqref{r2p2-2} to be 
\begin{equation}\label{r2p2-4}
\mathbf{u}^{n}=\mathcal{D}_{\mathcal{W}^{n}}(\mathbf{v}^{n})- \mathcal{D}_{\mathcal{W}^{n}}(\mathbf{x}^{n}) + \mathbf{v}^{n},
\end{equation}
where $\mathcal{D}_{\mathcal{W}^{n}}$ is a network with learnable parameters $\mathcal{W}^{n}$. Thus we have for $n\notin\mathcal{T}$
\begin{equation}\label{r2p2-3}
\begin{aligned}
\eta^n &= \frac{\|\mathbf{u}^{n}-\mathbf{x}^{n}\|_2}{\|\mathbf{v}^{n}-\mathbf{x}^{n}\|_2}\\
&= \frac{\|\mathcal{D}_{\mathcal{W}^{n}}(\mathbf{v}^{n})- \mathcal{D}_{\mathcal{W}^{n}}(\mathbf{x}^{n}) + \mathbf{v}^{n}-\mathbf{x}^{n}\|_2}{\|\mathbf{v}^{n}-\mathbf{x}^{n}\|_2} \\
&\leq \frac{\|\mathcal{D}_{\mathcal{W}^{n}}(\mathbf{v}^{n})- \mathcal{D}_{\mathcal{W}^{n}}(\mathbf{x}^{n})\|_2}{\|\mathbf{v}^{n}-\mathbf{x}^{n}\|_2} +1. \\
\end{aligned}
\end{equation}
Recently, lots of researches have investigated the Lipschitz continuity and constants of various DNNs, including fully-connected networks, convolutional neural networks, Transformer~\cite{virmaux2018lipschitz, fazlyab2019efficient, ryu2019plug, zou2019lipschitz, chen2020semialgebraic, kim2021lipschitz}. They proposed efficient methods for estimating or constraining Lipschitz constants of DNNs. Following the analysis of these works, we treat $\mathcal{D}_{\mathcal{W}^{n}}$ as a Lipschitz continuous DNN with a Lipschitz constant $L_D^n$. Then we obtain from Eq.~\eqref{r2p2-3} that $\eta^n\leq L_D^n+1\leq \max\{L_D^n\}_{n=0}^{K}+1$ for HCISTA with $K$ iterations, which means that Assumption~1 is satisfied.  

Experiments are conducted to evaluate the performance and the experimental setups follow Section~6.1. The adopted DNN $\mathcal{D}_{\mathcal{W}^{n}}$ consists of three one-dimensional convolutional layers with the sizes of 9$\times$1$\times$16, 9$\times$16$\times$16, and 9$\times$16$\times$1 and ReLU following the first two convolutional layers. To distinguish from the HCISTA models in Section~6.1, we use HCISTA$^*$ to represent models with DNNs as Eq.~\eqref{r2p2-4}. Fig.~\ref{lip_assump1}(a) evaluates ISTA, HCISTA$^*$ and HCISTA with untrained DNNs for up to 400 iterations under the hyperparameters $\lambda^0=0.2$, $0.1$ and $0.05$. HCISTA$^*$ is comparable to HCISTA and converges faster than ISTA. Fig.~\ref{lip_assump1}(b) further compares HCISTA$^*$ and HCISTA with trained/untrained DNNs in 16 iterations, and HCISTA$^*$ and HCISTA are similar in performance. Fig.~\ref{lip_assump1}(a) and (b) show that the proposed HCISTA$^*$ is efficient to utilize DNNs that satisfy Assumption~1 without trial-and-error and predefining $\eta_c$. %Fig.~\ref{lip_assump1}(a) and (b) show that even we utilize DNNs that can satisfy Assumption~1 without trial-and-error and predefining $\eta_c$, the proposed HCISTA$^*$ is still efficient. 
Fig.~\ref{lip_assump1}(c) further validates that Assumption~1 can be satisfied for HCISTA$^*$.

\begin{figure*}[htb]
\renewcommand{\baselinestretch}{1.0}
\centering
\includegraphics[width=5.6in]{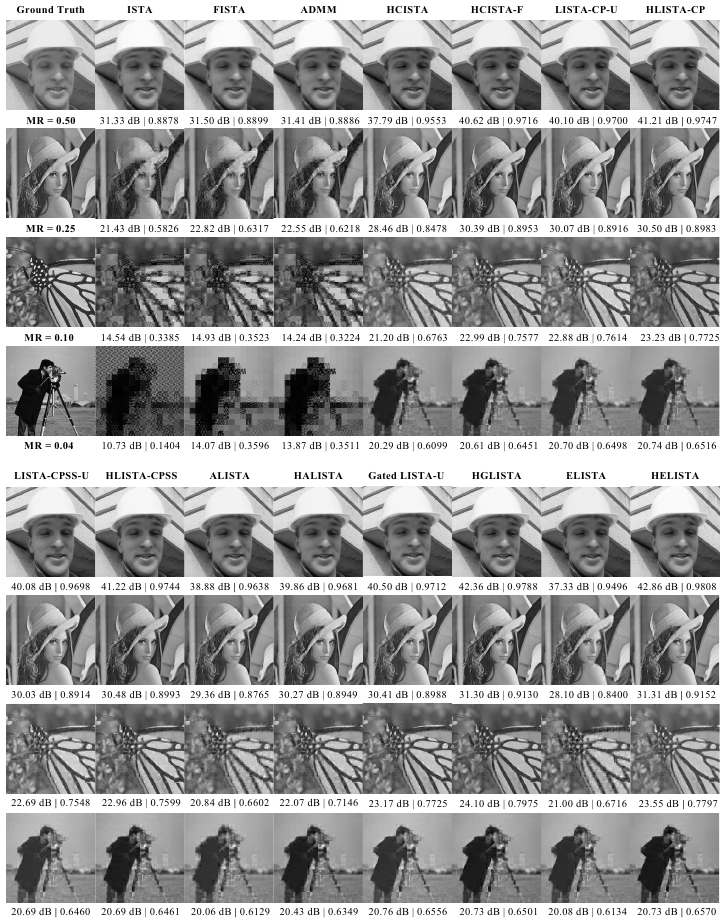}
\caption{Reconstruction of test images \emph{Foreman}, \emph{Lena}, \emph{Monarch}, and \emph{Cameraman} from \emph{Set11} at the MRs of 0.5, 0.25, 0.10, and 0.04, respectively.
}\label{imgs}
\end{figure*}

\subsection{Compressive Sensing}
\subsubsection{Visualization}

\figurename~\ref{imgs} visualizes some reconstructed images of \emph{Set11} at the MRs of 0.5, 0.25, 0.10, and 0.04 when adopting simple DNNs. 
The average PSNR and SSIM on \emph{Set11} are reported in Table~4. 
As shown in \figurename~\ref{imgs}, the edges and textures obtained by hybrid ISTA models are more evident in comparison to the benchmark methods. Although ISTA is severely degraded in terms of visual quality at low MRs of 0.10 and 0.04, HCISTA and HCISTA-F are able to reconstruct the outlines of the ground truth, which proves the effectiveness of hybrid ISTA.

\subsubsection{HCISTA $\&$ HLISTA-CP with Complicated DNNs}

We further evaluate the reconstruction performance with complicated DNNs in the task of compressive sensing. Four complicated DNNs are adopted, including DenseNet~[52], U-Net~[58], Vision Transformer~[53] and fully-connected networks (FCN). The network architectures are illustrated in Fig.~\ref{complicated_DNNs}.

As we utilize complicated DNNs, a large-scale Flickr 30k dataset is adopted for training to evaluate the proposed method. Flickr 30k~[61] contains 31783 images obtained from \textit{Flickr} (the website is \textit{https://www.flickr.com}). We randomly extract 8642160 image patches with size 16$\times$16 from these images for network training. 
As the patch size is 16$\times$16 and is different from the size of training data in sparse recovery experiments, the number of learnable parameters of DNNs is also different to adapt this change when adopting Transformer and FCN. In addition, the loss function, training strategy and hyperparameters are selected in the manner as Section~6.2.

Tables~\ref{T1_cs_c} and \ref{T2_cs_c} report the reconstruction performance in terms of PSNR and SSIM. HCISTA $\&$ HLISTA with complicated DNNs achieve superior performance in comparison with the corresponding baselines. Among the four inserted DNNs, U-Net obtains the best performance, followed by DenseNet. These two network architectures constructed using fully convolutional layers are particularly suitable for handling image tasks, such as compressive sensing. Although HCISTA is not much ahead of classical ISTA in sparse recovery, the recovery performance of HCISTA goes far beyond ISTA in compressive sensing.

\bibliographystyle{IEEEtran}
\bibliography{IEEEabrv,ieee}

\end{document}